\documentclass[11pt,a4paper]{book}  
\input{initial/init.tex}

\begin{document}
\dominitoc 

    \fontsize{11.5}{14}\selectfont

\title{\thesistitle}
\author{Name}
\date{November, 2024}
\DeptName{Department of Electrical and Computer Engineering}
\UniName{Hellenic Mediterranean University}

\Applicant{\textbf{\small Author:} \small }
\CommitteeMembers
{\textbf{\small Supervisor:} \small }
{\textbf{\small Committee Member:} \small }
{\textbf{\small Committee Member:} \small }
{\textbf{\small Committee Member:} \small }
{\textbf{\small Committee Member:} \small }
{\textbf{\small Committee Member:} \small }

\CommitteeChair{\textbf{\small Department Chairman:} \small }

\maketitlepage

\cleardoublepage

\makesigpage

\newpage
\pagestyle{plain}
\thispagestyle{empty}
\mbox{}
\clearpage

    \cleardoublepage

\setstretch{1.15}

\selam

\phantomsection
\addcontentsline{toc}{chapter}{Declaration}
\chapter*{Declaration of Authorship}

\noindent I, Stefanos Gkikas, declare that this thesis titled, 
\textit{\textquotedblleft A Pain Assessment Framework based on multimodal data and Deep Machine Learning methods,\textquotedblright}\space and the work presented in it are my own. I confirm that:

\begin{itemize}[leftmargin=*]
    \item This work was done wholly or mainly while in candidature for a research degree at this University.
    \item Where any part of this thesis has previously been submitted for a degree or any other qualification at this University or any other institution, this has been clearly stated.
    \item Where I have consulted the published work of others, this is always clearly attributed.
    \item Where I have quoted from the work of others, the source is always given. With the exception of such quotations, this thesis is entirely my own work.
    \item I have acknowledged all main sources of help.
    \item Where the thesis is based on work done by myself jointly with others, I have made clear exactly what was done by others and what I have contributed myself.
\end{itemize}

\vspace{1cm} 
\noindent Signed:\\
\noindent\rule[0.20cm]{0.5\textwidth}{0.5pt}

\vspace{-1.7cm} 
\hspace{0.7cm}
\noindent\includegraphics[scale=0.22]{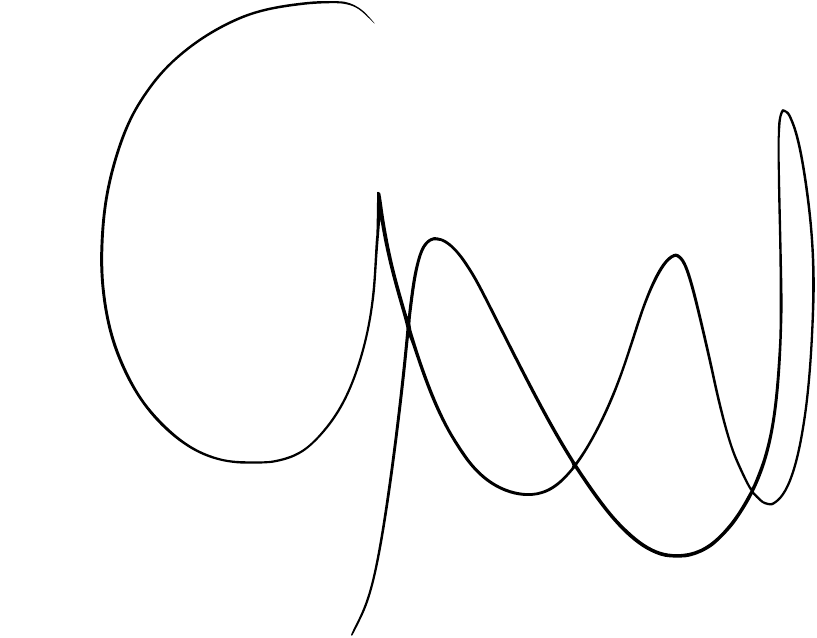}

\vspace{0.5cm} 
\noindent Date: \large \textit{26/10/2024}\\
\noindent\rule[0.20cm]{0.5\textwidth}{0.5pt}

    \cleardoublepage
\pagestyle{plain} \thispagestyle{empty}

\begin{dedication}
\begin{flushright}
\begin{center}


%
%
%
%
%
%

\noindent\begin{minipage}[t]{0.5\textwidth} 
  \raggedright 
  \small 
  \textquotedblleft 
  \textit{Do not go gentle into that good night,\\
  Old age should burn and rave at close of day;\\
  Rage, rage against the dying of the light.}\\[10pt]

  \textit{Though wise men at their end know dark is right,\\
  Because their words had forked no lightning they\\
  Do not go gentle into that good night.}\\[10pt]

  \textit{Good men, the last wave by, crying how bright\\
  Their frail deeds might have danced in a green bay,\\
  Rage, rage against the dying of the light.}\\[10pt]

  \textit{Wild men who caught and sang the sun in flight,\\
  And learn, too late, they grieved it on its way,\\
  Do not go gentle into that good night.}\\[10pt]

  \textit{Grave men, near death, who see with blinding sight\\
  Blind eyes could blaze like meteors and be gay,\\
  Rage, rage against the dying of the light.}\\[10pt]

  \textit{And you, my father, there on the sad height,\\
  Curse, bless, me now with your fierce tears, I pray.\\
  Do not go gentle into that good night.\\
  Rage, rage against the dying of the light.}\textquotedblright \\

   \vspace{10pt}
   -- Dylan Thomas, 1947 
  \normalsize 
\end{minipage}


\end{center}
\end{flushright}
\end{dedication}
\clearpage

\newpage
\pagestyle{empty} \mbox{} \clearpage
  
    \cleardoublepage

\setstretch{1.15}

\selam

\phantomsection
\addcontentsline{toc}{chapter}{Acknowledgments}
\chapter*{Acknowledgments}

First and foremost, I want to express my deepest gratitude to my supervisor, Manolis Tsiknakis, for his invaluable support and guidance throughout my Ph.D. journey. The opportunities he provided, particularly 
my involvement in research projects and international scientific conferences, have profoundly shaped my perspective and worldview. Additionally, he established a supportive and flexible framework that enabled me to work effectively and autonomously, significantly enriching my research experience and personal development
--- I am sincerely thankful.\\

\noindent I want to thank my family for their support throughout my life---my father, Dimitris, my mother, Katerina, and my brother, Alexandros. I never say it to them in person, but I can write it here; maybe they'll read it someday, though probably not.\\

\noindent Lastly, I would like to acknowledge that the success of this Ph.D. project, both in terms of the quantity and quality  (whatever they may be) of the scientific work produced, was made possible by my dedication, ambition, and an almost ascetic lifestyle that allowed me to financially and mentally support myself.\\

\begin{flushright}
To my daim\'{o}nion (\selg daim\'{o}nion\selam)...\\
To Sarah (\selg S\'{a}ra\selam)...\\
To Piki (\selg P\'{i}ki\selam)...
\end{flushright}

    \cleardoublepage

\setstretch{1.15}

\phantomsection
\addcontentsline{toc}{chapter}{Abstract}
\chapter*{Abstract}

Pain is a manifold condition that affects a large portion of the population. Accurate and reliable pain evaluation is essential for creating effective pain management strategies and protocols. In cases where patients cannot communicate their pain, clinicians rely on observing behavioral cues and monitoring vital signs. However, this approach is subjective and deficient in terms of continuous monitoring. Moreover, even in circumstances where patients are able to communicate, numerous challenges persist. Factors such as gender and age significantly influence how pain is perceived and expressed. Individuals with psychological disorders like depression and anxiety exhibit altered pain-related behaviors, further complicating the assessment. Human biases and beliefs, including racial and cultural differences, also shape pain judgment and interpretation. Additionally, social dynamics and personal relationships between the observer and the person experiencing pain can heavily distort the assessment, further intensifying the challenges of accurate pain evaluation.
Automatic pain assessment can alleviate these challenges by providing continuous monitoring through computational systems that utilize designed algorithms to recognize pain indicators from input modalities, including videos and biosignals. In recent years, researchers have increasingly adopted deep learning algorithms to capture and encode the multidimensional aspects of pain into meaningful data attributes.\\

\noindent This thesis initially aims to study the pain assessment process from a clinical-theoretical perspective while exploring and examining existing automatic approaches. Building on this foundation, the primary objective of this Ph.D. project is to develop innovative computational methods for automatic pain assessment that achieve high performance and are applicable in real clinical settings. A primary goal is to thoroughly investigate and assess significant factors, including demographic elements that impact pain perception, as recognized in pain research, through a computational standpoint.  Within the limits of the available data in this research area, our goal was to design, develop, propose, and offer automatic pain assessment pipelines for unimodal and multimodal configurations that are applicable to the specific requirements of different scenarios.
The studies published in this Ph.D. thesis showcased the effectiveness of the proposed methods, achieving state-of-the-art results. Additionally, they paved the way for exploring new approaches in artificial intelligence, foundation models, and generative artificial intelligence.

    \cleardoublepage

\setstretch{1.15}

\phantomsection  

\addcontentsline{toc}{chapter}{Abstract in Greek}

\chapter*{\selg Περίληψη}

{\fontfamily{times}\fontsize{10pt}{12pt}\selectfont  

\selg

Ο πόνος είναι μια πολυεπίπεδη κατάσταση που επηρεάζει μεγάλο μέρος του πληθυσμού. Η ακριβής και αξιόπιστη αξιολόγηση του πόνου είναι ουσιώδης για τη δημιουργία αποτελεσματικών στρατηγικών και πρωτοκόλλων διαχείρισης του πόνου. Σε περιπτώσεις που οι ασθενείς δεν μπορούν να επικοινωνήσουν τον πόνο τους, οι κλινικοί βασίζονται στην παρατήρηση συμπεριφορικών ενδείξεων και στην παρακολούθηση ζωτικών σημείων. Ωστόσο, αυτή η προσέγγιση είναι υποκειμενική και ελλιπής όσον αφορά τη συνεχή παρακολούθηση. Επιπλέον, ακόμη και σε περιπτώσεις όπου οι ασθενείς μπορούν να επικοινωνήσουν, υπάρχουν πολλές προκλήσεις. Παράγοντες όπως το φύλο και η ηλικία επηρεάζουν σημαντικά τον τρόπο που ο πόνος αντιλαμβάνεται και εκφράζεται. Άτομα με ψυχολογικές διαταραχές όπως η κατάθλιψη και η αγχώδης διαταραχή εκδηλώνουν τροποποιημένες συμπεριφορές σχετικές με τον πόνο, δυσχεραίνοντας περαιτέρω την αξιολόγηση. Οι ανθρώπινες προκαταλήψεις και πεποιθήσεις, περιλαμβανομένων των φυλετικών και πολιτισμικών διαφορών, επίσης διαμορφώνουν την κρίση και την ερμηνεία του πόνου. Επιπλέον, οι κοινωνικές δυναμικές και οι προσωπικές σχέσεις μεταξύ του παρατηρητή και του ατόμου που βιώνει τον πόνο μπορούν να διαστρεβλώσουν σημαντικά την αξιολόγηση, εντείνοντας περαιτέρω τις προκλήσεις της ακριβούς αξιολόγησης του πόνου. Η αυτόματη αξιολόγηση του πόνου μπορεί να αμβλύνει αυτές τις προκλήσεις παρέχοντας συνεχή παρακολούθηση μέσω υπολογιστικών συστημάτων που χρησιμοποιούν ειδικά σχεδιασμένους αλγόριθμους για να αναγνωρίσουν ενδείξεις πόνου από διαφόρους τύπους δεδομένων, συμπεριλαμβανομένων των βίντεο και των βιοσημάτων. Τα τελευταία χρόνια, οι ερευνητές έχουν υιοθετήσει ολοένα και περισσότερο αλγόριθμους βαθιάς μάθησης για να αποτυπώσουν και να κωδικοποιήσουν τις πολυδιάστατες πτυχές του πόνου σε χρήσιμα δεδομενικά χαρακτηριστικά.\\

\noindent Η παρούσα διδακτορική διατριβή αρχικά στοχεύει να μελετήσει τη διαδικασία αξιολόγησης του πόνου από μια κλινικο-θεωρητική προοπτική ενώ διερευνά και εξετάζει τις υπάρχουσες αυτόματες προσεγγίσεις. Στηριζόμενη σε αυτή τη βάση, ο κύριος στόχος αυτού του διδακτορικού έργου είναι να αναπτύξει καινοτόμες υπολογιστικές μεθόδους για αυτόματη αξιολόγηση του πόνου που επιτυγχάνουν υψηλή απόδοση και είναι εφαρμόσιμες σε πραγματικά κλινικά περιβάλλοντα. Ένας πρωταρχικός στόχος είναι να διερευνήσει και να αξιολογήσει σε βάθος σημαντικούς παράγοντες, συμπεριλαμβανομένων των δημογραφικών στοιχείων που επηρεάζουν την αντίληψη του πόνου, όπως αναγνωρίζονται στην έρευνα πόνου, από μια υπολογιστική σκοπιά. Εντός των ορίων των διαθέσιμων δεδομένων σε αυτό τον τομέα έρευνας, ο στόχος μας ήταν να σχεδιάσουμε, να αναπτύξουμε, να προτείνουμε και να προσφέρουμε αυτόματες αλυσίδες επεξεργασίας αξιολόγησης του πόνου για μονομορφικές και πολυμορφικές διαμορφώσεις που είναι εφαρμόσιμες στις συγκεκριμένες απαιτήσεις διαφορετικών σεναρίων. Οι μελέτες που δημοσιεύτηκαν σε αυτή τη διδακτορική διατριβή παρουσίασαν την αποδοτικότητα των προτεινόμενων μεθόδων, επιτυγχάνοντας πρωτοπόρα επιτεύγματα. Επιπλέον, καινοτόμησαν στην εξερεύνηση νέων προσεγγίσεων στην τεχνητή νοημοσύνη, τα μοντέλα θεμελίωσης και την γενετική τεχνητή νοημοσύνη.

}

\selam

    \setstretch{1.15}

\tableofcontents

\listoffigures

\listoftables

\mainmatter
\StartBody
\mainmatter

	\setcounter{mtc}{7}
    \setstretch{1.15}

\chapter{Introduction}
\label{chapter:intro}
\minitoc  

\section{Context and Motivation} 
\textbf{\textit{Pain}} is a complex and deeply personal experience that is subjective by nature. Traditionally, it has been described in terms of its sensory dimension \cite{merskey_1979_2}. However, extensive research has highlighted the importance of affective, cognitive and social aspects in shaping this experience \cite{williams_craig_2016}. Studies have explored physiological, psychological, and socio-environmental factors that contribute to the experience of pain. It is understood as a result of biological evolution and as influenced by psychological and social factors.
As Ridell \textit{et al.} \cite{riddell_craig_2018} noted, \textit{\textquotedblleft Pain is a synthesis--a sum that is greater than its parts.\textquotedblright}\space The brain's ability to alter the perception of sensory inputs through the interplay of emotion, cognition, and social processes is significant. 
Although natural systems establish the initial biological framework for pain perception, this structure is highly adaptable, particularly in humans. Throughout a person's life, both biological developments and personal experiences significantly reshape this framework.

A key question driving pain research across biological, psychological, and computational fields is why this topic of pain is meaningful and important. This question also forms the basis for initiating this thesis, highlighting the broader relevance of studying pain.
Williams and Kappesser \cite{williams_kappesser_2018} provide a compelling explanation, stating, \textit{\textquotedblleft We care because we are wired to care: to attend to other people's expression of pain and to understand its meaning; to feel distress in relation to their distress; and to be motivated to reduce their distress, and ours, if we are able to do so.\textquotedblright}\space  This highlights the intrinsic human response to empathize and alleviate pain, underlining the fundamental importance of this research area.
Indeed, from a Darwinian perspective, pain serves a crucial role. The manifestation of pain in humans and the reactions it elicits are examined through an evolutionary lens. Pain facilitates recovery by promoting responses to harmful stimuli and behaviors that demonstrate the adverse nature of painful experiences, common among animals. Specifically, the facial expression of pain, which communicates discomfort directly to those nearby, is universally recognized across different ages, ethnicities, roles, and relationships. Evidence from healed major fractures \cite{mithen_1996, redfern_2010} suggests that injured members of hominid groups were not left to fend for themselves but were supported through their recovery, indicating the fundamental importance of pain expression in our evolutionary history.

Pain is a widespread health concern globally, affecting up to $30\%$ of the adult population \cite{cohen_vasem_hooten_2021} and between $83\%$ and $93\%$ of elderly adults in residential care \cite{abdulla_adams_2013}. The Global Burden of Disease (GBD) study identifies pain as the primary cause of years lived with disability (YLD) \cite{james_abate_2018}, with major contributors including chronic back pain, musculoskeletal disorders, and neck pain \cite{usa_bdc_2013}. Pain impacts individuals and poses significant clinical, economic, and social challenges. In the United States, the economic and healthcare costs related to pain due to reduced work productivity range from $\$560$ to $\$635$ billion annually, surpassing the costs associated with heart disease, cancer, and diabetes combined \cite{gaskin_richard_2012}. In Europe, chronic pain's direct healthcare costs and indirect socioeconomic impacts account for $3\%$ to $10\%$ of the GDP \cite{breivik_eisenberg_2013}. In Australia, the average annual cost for individuals among the $15.4\%$ living with chronic pain ranges from AU$\$ 22,588$ to AU$\$42,979$, including non-financial costs \cite{deloitte_australia}. Beyond direct effects on health, pain contributes to a range of adverse outcomes, such as opioid dependency, drug overuse, addiction, declining social relationships, and psychological disorders \cite{dinakar_stillman_2016}. In the last two decades, prescription opioid use has surged in the United States, where overdose deaths have increased more than fourfold from $1999$ to $2016$ \cite{seth_rudd_2018}. Additionally, side effects from these opioids, like lethargy, depression, anxiety, and nausea, severely impact workforce productivity and overall life quality \cite{benyamin_trescot_2008}.

Accurate pain assessment is crucial for early diagnosis, disease progression monitoring, and treatment effectiveness evaluation, particularly in managing chronic pain \cite{gkikas_tsiknakis_slr_2023}. This critical role has resulted in pain being recognized as \textit{\textquotedblleft the fifth vital sign\textquotedblright} in nursing literature \cite{joel_lucille_1999}. Pain assessment is also fundamental in physiotherapy, where therapists apply external stimuli and need to gauge the patient's pain levels accurately \cite{badura_2021}.
Objective evaluation of pain is essential to provide appropriate care, especially for vulnerable populations who may not be able to communicate their pain effectively, such as infants, young children, individuals with mental health issues, and the elderly. Various methods are used for pain assessment, with self-reporting--where individuals describe their pain experiences--considered the gold standard \cite{dvader_Bostick_2021}. Pain evaluation methods in clinical environments include quantifiable measures like the Numeric Pain Rating Scale (NPRS), Visual Analogue Scale (VAS), and quantitative sensory testing techniques such as the pressure pain detection threshold (PPDT) \cite{straatman_lukacs_2022}. Behavioral indicators are also crucial and include facial expressions (\textit{e.g.}, grimacing, open mouth, lifted eyebrows), vocalizations (like crying, moaning, or screaming), and movements of the body and head \cite{rojas_brown_2023}. Physiological measures such as electrocardiography (ECG), electromyography (EMG), galvanic skin responses (GSR), and respiration rates further contribute to understanding pain's physiological aspects \cite{gkikas_tsiknakis_slr_2023}. Additionally, brain monitoring techniques like near-infrared spectroscopy (fNIRS) have effectively detected changes in hemodynamic activity associated with pain stimuli \cite{rojas_liao_2019}.

Caregivers and family members often determine the presence or absence of pain in patients by observing their behavioral or physiological responses \cite{gkikas_tsiknakis_slr_2023}. However, accurately assessing pain poses a significant challenge for clinicians \cite{aqajari_cao_2021}, especially with nonverbal patients such as the elderly, who may have reduced expressive abilities or may be reluctant to communicate pain \cite{yong_gibson_2001}. Extensive research indicates that pain manifestations vary significantly across different genders and ages, adding to the complexity of its assessment \cite{bartley_fillingim_2013}. Further complicating the assessment process are the heightened workload and fatigue experienced by nursing staff due to the demands of patient monitoring \cite{roue_morag_2021}. 
Technological solutions are necessary for continuous patient monitoring. Nevertheless, concerns remain about the objectivity and accuracy of these observations, as inadequately trained or biased observers may struggle to assess pain \cite{dekel_gori_2016} accurately. Even among trained observers, interpretations of behaviors can vary \cite{rojas_brown_2023}, and social and interpersonal dynamics can significantly affect the pain assessment process, influencing both the evaluators' judgments and the patients' expressions of pain \cite{hoffman_trawalter_axt_2016}. Additionally, the presence of an observer can lead patients to modify their behavior \cite{keefe_sommers_2011}, and expressing pain through scales and measurements can be challenging \cite{miglio_stanier_2022}. 
While self-reporting is used because pain is inherently subjective, relying solely on a one-dimensional pain score fails to capture this complex phenomenon, often leading to inadequate pain management \cite{leroux_lynn_2021}.

Given the challenges described above, scientific computing (SC) researchers have focused on developing models and algorithms to enhance automatic pain recognition systems over the last two decades. Their goal is to accurately determine the presence and intensity of pain by analyzing physiological and behavioral indicators. Adopting deep learning and artificial intelligence (AI) techniques has expanded these automatic methods, designed to interpret the complex and varied nature of pain \cite{gkikas_tsiknakis_slr_2023}. 
Numerous studies have underscored the effectiveness of automated systems that utilize behavioral or physiological modalities for pain assessment \cite{werner__martinez_2019}. Sario \textit{et al}. \cite{sario_haider_2023} have shown the capability of these systems to accurately recognize pain through facial expressions, proving their utility in clinical environments. Multimodal sensing has shown particular promise, offering enhanced accuracy in pain detection systems \cite{rojas_brown_2023}. Furthermore, including temporal aspects in these modalities has proven to significantly improve the accuracy of pain assessments \cite{gkikas_tsiknakis_slr_2023}.

\section{Scope and Challenges} 
Although considerable research has been conducted on automatic pain assessment, studies have yet to explore factors like demographics and social aspects from a computational angle. Furthermore, despite the existence of deep learning-based methods, the approaches we observe are often outdated and repeatedly recycled.
For these reasons, we aimed to address two issues by \textit{(i)} attempting to evaluate the social or demographic context, which significantly impacts and influences pain sensation and perception, and \textit{(ii)} introducing innovative deep learning methods inspired by the latest developments in AI and generative AI literature. We believe these approaches can forge new paths in pain research, enhance the accuracy of recognizing this complex phenomenon, and, ultimately, be adopted in real-world scenarios to assist those in need.
Additionally, \textit{(iii)} recognizing the skepticism towards new technologies among clinicians and the general public, especially regarding the limited understanding of how deep learning models function, we have devoted a portion of our research to interpreting these models to offer some level of explanation and help the adoption process of them in clinical settings.

Nevertheless, this thesis initially faced challenges related to our objectives and goals as the research progressed.
The availability of pain datasets (to be discussed in the next chapter) is limited. Only a few datasets are available, and crucially, they are limited in size. This restriction poses a significant challenge for developing deep learning models, which typically require a large volume of data.
In automatic pain assessment, researchers who develop deep learning methods typically confront a decision: either train their models from scratch, which can introduce performance limitations, or employ pre-trained models. These pre-trained models are generally trained on broadly available image datasets that include a variety of subjects like animals and objects, or they rely on older architectures that were trained explicitly on facial datasets. In this thesis, we addressed these issues by independently pre-training our deep-learning models using diverse datasets related explicitly to human facial images and biosignals. This strategy allowed us to design specific architectures to meet our unique needs for each scenario, free from the constraints of relying on models developed and trained by others. Furthermore, we explored and evaluated several pre-training techniques to assess their effectiveness in pain assessment applications. 

Regarding, our objective to explore methods that utilize various modalities individually and in combination in a multimodal manner further constrains our dataset options. Moreover, as previously outlined, our interest in the sociodemographic aspects of pain necessitates datasets that include this information type, intensifying our challenges. For these reasons, this thesis focuses specifically on examining the impact of age and gender on pain. 
In addition, led us to utilize two pain datasets that most closely match the characteristics necessary for our research, particularly in terms of demographic elements and multimodality.



\section{Contributions -- Peer-review Publications} 
This section outlines the publications and projects produced during the Ph.D. research on automatic pain assessment, where I was the first author.

\begin{enumerate}
\item[\bfseries 1.]\textbf{\textit{Automatic assessment of pain based on deep learning methods: A systematic review} \cite{gkikas_tsiknakis_slr_2023}}\\
This systematic literature review (SLR) was conducted at the start of this Ph.D. research. 
This paper aims to explore the surge in recent years of deep learning algorithms adopted by researchers to encode the multidimensional nature of pain into meaningful features. Specifically, this systematic review examines the models, methods, and data types used to establish the foundation for deep learning-based automatic pain assessment systems.
It identified relevant original studies from digital libraries such as \textit{Scopus}, \textit{IEEE Xplore}, and \textit{ACM Digital Library}, following defined inclusion and exclusion criteria for studies published until December $2021$. The findings highlight the critical role of multimodal approaches in automatic pain estimation, particularly in clinical environments, and emphasize the substantial gains observed with the inclusion of temporal exploitation of modalities. The review also recommends selecting high-performing deep learning architectures and methods, encouraging the adoption of robust evaluation protocols and interpretability techniques to deliver reliable and understandable outcomes. Additionally, it underscores the current limitations of existing pain databases in adequately supporting the development, validation, and practical application of deep learning models as decision-support tools in real-world settings.
Furthermore, we believe this paper is valuable not only for this Ph.D. project but also for other practitioners and researchers in the field.

\item[\bfseries 2.]\textbf{\textit{Automatic Pain Intensity Estimation based on Electrocardiogram and Demographic Factors} \cite{gkikas_chatzaki_2022}}\\
This study investigated the relationship between gender, age, and pain sensation and their effects on the automatic pain assessment process. By analyzing physiological signals, particularly electrocardiography (ECG), we estimated pain intensity and examined the influence of these demographic factors. Utilizing the Pan-Tompkins algorithm for feature extraction and applying well-established classification methods, we explored the correlation between gender, age, and pain manifestation.

\item[\bfseries 3.]\textbf{\textit{Multi-task Neural Networks for Pain Intensity Estimation Using Electrocardiogram and Demographic Factors} \cite{gkikas_chatzaki_2023}}\\
Inspired by the previous study, this research further explored the influence of gender and age on pain perception. 
In this work, we analyze electrocardiography signals to uncover variations in pain perception across different demographic groups. We leveraged these insights by developing a novel multi-task neural network for automatic pain estimation, incorporating age and gender data for each individual. The study demonstrated the advantages of this approach compared to other existing methods.

\item[\bfseries 4.]\textbf{\textit{A Full Transformer-based Framework for Automatic Pain Estimation
using Videos} \cite{gkikas_tsiknakis_embc}}\\
This study introduced an innovative full transformer-based framework featuring a \textit{Transformer in Transformer (TNT)} model combined with cross-attention and self-attention blocks. We achieved state-of-the-art performance using video data from the \textit{BioVid} database, demonstrating the model's effectiveness, efficiency, and strong generalization across primary pain estimation tasks.

\item[\bfseries 5.]\textbf{\textit{Multimodal automatic assessment of acute pain through facial videos and heart rate signals utilizing transformer-based architectures} \cite{gkikas_tachos_2024}}\\
This study presented a multimodal automatic acute pain assessment framework, integrating video and heart rate signals. The framework consists of four key modules: the \textit{Spatial Module}, which extracts embeddings from videos; the \textit{Heart Rate Encoder}, which maps heart rate signals into a higher-dimensional space; the \textit{AugmNet}, which generates learning-based augmentations in the latent space; and the \textit{Temporal Module}, which leverages the video and heart rate embeddings for the final assessment. The \textit{Spatial Module} undergoes a two-stage pre-training process: first, it learns universal facial features through face recognition, followed by emotion recognition in a multitask learning approach, enabling high-quality embeddings for pain assessment. Experiments with facial videos and heart rate data extracted from electrocardiograms in the \textit{BioVid} database, alongside direct comparisons to $29$ studies, demonstrate state-of-the-art performance in unimodal and multimodal settings while maintaining high efficiency. In the multimodal setting, the framework achieved $82.74\%$ accuracy for binary pain classification and $39.77\%$ for multi-level pain classification, using only $9.62$ million parameters across the entire framework.

\item[\bfseries 6.]\textbf{\textit{Synthetic Thermal and RGB Videos for Automatic Pain Assessment utilizing a Vision-MLP Architecture} \cite{gkikas_tsiknakis_thermal_2024}}\\
This paper introduced synthetic thermal videos generated by \textit{Generative Adversarial Networks}, which are integrated into the pain recognition process to assess their effectiveness. The framework employs a \textit{Vision-MLP} and \textit{Transformer}-based module, leveraging RBG and synthetic thermal videos in unimodal and multimodal settings. Experiments conducted using facial videos from the \textit{BioVid} database highlighted synthetic thermal videos' effectiveness and showcased their potential benefits in pain recognition tasks.

\item[\bfseries 7.]\textbf{\textit{Twins-PainViT: Towards a Modality-Agnostic Vision Transformer Framework for Multimodal Automatic Pain Assessment using Facial Videos and fNIRS} \cite{gkikas_tsiknakis_painvit_2024}}\\
This study was submitted to the \textit{First Multimodal Sensing Grand Challenge for Next-Gen Pain Assessment (AI4PAIN)}. The proposed multimodal framework leverages facial videos and fNIRS, offering a modality-agnostic approach that eliminates the need for domain-specific models. Utilizing a dual ViT configuration and waveform representations for both fNIRS and the extracted embeddings from the two modalities, the method demonstrates its effectiveness, achieving an accuracy of $46.76\%$ in the multilevel pain assessment task.

\item[\bfseries 8.]\textbf{\textit{PainFormer: a Vision Foundation Model for Automatic Pain Assessment}\cite{gkikas_rojas_foundation_2024}\footnote{Under Review}}\\
This study introduces \textit{PainFormer}, a vision foundation model built on multi-task learning principles and trained across $14$ distinct tasks and datasets comprising $10.9$ million samples. As an embedding extractor for various input modalities, \textit{PainFormer} provides feature representations to the \textit{Embedding-Mixer}, a transformer-based module responsible for conducting the final pain assessment. Extensive experimentation using both behavioral modalities--including RGB, synthetic thermal, and estimated depth videos--and physiological modalities like ECG, EMG, GSR, and fNIRS revealed \textit{PainFormer}'s ability to extract high-quality embeddings from diverse inputs. Tested on the \textit{BioVid} and \textit{AI4Pain} datasets and compared to more than  $60$ existing methods, the framework demonstrated state-of-the-art performance in unimodal and multimodal settings, positioning itself as a step toward developing general-purpose models for automated pain evaluation.

\end{enumerate}

\section{Thesis Outline} 
The dissertation is organized into the following chapters:\\
\textbf{Chapter 2} introduces the foundational concepts of pain from biological, psychological, and clinical perspectives.\\
\textbf{Chapter 3} reviews existing literature on automatic pain assessment using deep learning methods and details the pain datasets used.\\
\textbf{Chapter 4} outlines and proposes methods for evaluating demographic variables, their utilization, and their integration into an automatic pain assessment framework.\\
\textbf{Chapter 5} discusses methods that utilize video and wearable device data, exploring the trade-offs between efficiency and accuracy. It also proposes efficient, fast, effective models suitable for real-world applications.\\
\textbf{Chapter 6} explores synthetic data in pain assessment and introduces synthetic thermal imagery techniques to enhance performance in automatic pain recognition.\\
\textbf{Chapter 7} discusses general-purpose models, introduces a modality-agnostic framework, and presents the first foundation model used in automatic pain assessment.\\
\textbf{Chapter 8} concludes the thesis with a final discussion, offering perspectives and ideas for future research in automatic pain assessment.

    \chapter{Clinical Pain Assessment}
\label{clinical_pain_assessment}
\minitoc  

\section{Chapter Overview}
This chapter provides an anatomical and physiological overview of pain, focusing on the mechanisms responsible for generating, transmitting, processing, and interpreting pain signals. It examines the various types of pain and explores the actions and expressions typically associated with pain. Additionally, it reviews current pain assessment methods used in clinical settings for adults, children, and newborns. The chapter also discusses developing and validating existing clinical pain assessment tools. This foundational knowledge is essential for understanding the development and validation of computer-assisted pain assessment methods discussed in later chapters. Finally, it highlights the challenges faced in clinical pain assessment and underscores the need for automated pain assessment techniques.

\section{Biology of Pain}
Pain, according to the International Association for the Study of Pain (IASP) \cite{iasp_2020}, is \textit{\textquotedblleft an unpleasant sensory and emotional experience associated with actual or potential tissue damage, or described in terms of such damage\textquotedblright.} Biologically, pain is an undesirable sensation originating from the peripheral nervous system. Its fundamental function is to engage sensory neurons, notifying the organism of potential harm and playing a vital role in recognizing and responding to threats \cite{khalid_tubbs_2017}.

The transmission of a noxious stimulus from the periphery to the central nervous system involves a complex pathway through the spinal cord, resulting in the physical sensation of pain and a corresponding emotional response and memory. This process culminates in the perception of pain. The initial stage of pain processing occurs when a stimulus at nociceptive sensory fibers in the periphery is converted into an action potential. A nerve signal is generated if the stimulus is strong enough to surpass the action potential threshold \cite{garland_2012}. This signal travels along the primary afferent fiber toward the central nervous system. As the stimulus intensity grows, more nerve fibers and areas of the nervous system are engaged \cite{garland_2012}. Due to their branching nature, primary afferent fibers typically relay information from several pain receptors. These fibers and their receptors comprise a sensory unit, which gathers data from a specific receptive field \cite{garland_2012}. When receptive fields are larger and overlap with nearby fields, it becomes more challenging for the sensory system to locate the source of pain accurately. The primary afferent neuron is a pseudounipolar neuron that splits into a peripheral and central axon. The cell bodies of these neurons are located in the peripheral nervous system, within the posterior or cranial root ganglia. The peripheral axon extends to the skin, muscles, tendons, or joints, branching into terminal fibers that connect with somatosensory receptors. In contrast, the central axon leads to the central nervous system \cite{almeida_roizenblatt_2004}.

\begin{figure}
\begin{center}
\includegraphics[scale=0.65]{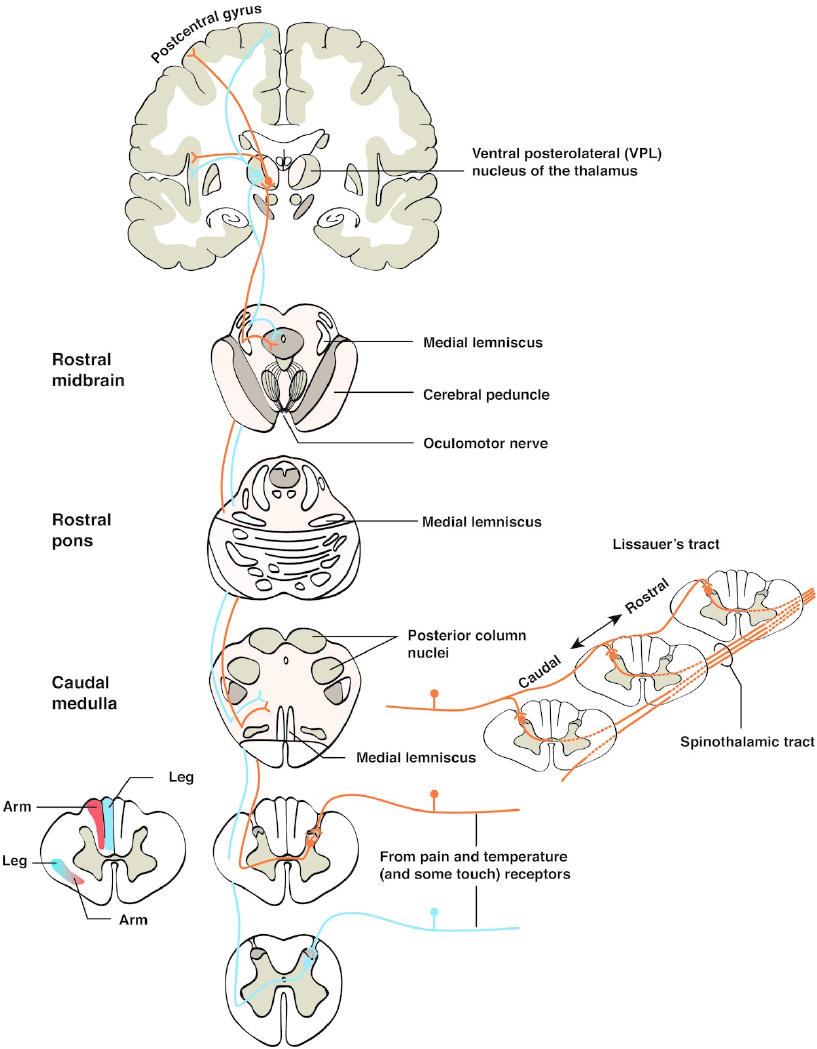} 
\end{center}
\caption{The spinothalamic tract (STT) \cite{khalid_tubbs_2017}. Pain, temperature, and some touch afferents end in the posterior horn, where second-order fibers cross the midline to form the spinothalamic tract, ascending to the thalamus and projecting to various cortical areas. Along the way, collaterals connect to the reticular formation. Due to the rostral inclination of fibers in Lissauer's tract, cordotomy must be performed several segments above the pain level for effective relief.}
\label{anatomy_of_pain}
\end{figure}

Peripheral somatosensory fibers are categorized into three main groups. The first group includes $A-\alpha$, $A-\beta$, $A-\gamma$ fibers, large, myelinated fibers that rapidly conduct signals \cite{apkarian_bushnell_2005}. These fibers involve touch and proprioception but are not associated with pain perception. The second group consists of $A-\delta$ fibers, which are smaller and slower conducting. Certain $A-\delta$ fibers play a key role in pain sensation, with some responding only to intense mechanical stimuli and others reacting to noxious and non-noxious heat. The third group comprises $C$ fibers, which are small, unmyelinated, and conduct signals very slowly. Most $C$ fibers are polymodal for pain perception, responding to various noxious mechanical, thermal, and chemical stimuli. These fibers are mainly linked to burning pain sensations \cite{khalid_tubbs_2017}.
The sensation of pain, known as nociception, is primarily facilitated by various intracellular and extracellular molecular messengers. When activated by a specific stimulus, nociceptors relay information through glutamate, an excitatory neurotransmitter. Additionally, inflammatory mediators are released at the injury site, further stimulating nociceptor activation by releasing chemicals such as neurotransmitters (\textit{e.g.}, serotonin), lipids (\textit{e.g.}, prostaglandins), peptides (\textit{e.g.}, bradykinin), and neurotrophins (\textit{e.g.}, nerve growth factor) \cite{apkarian_bushnell_2005}.
There are ascending tracts responsible for transmitting sensory information from the periphery to the central nervous system. Fibers that convey two-point discrimination, tactile information, pressure, vibration, and proprioception ascend via the dorsal column of the spinal cord, forming the gracile and cuneate fasciculi. Fibers transmitting pain, temperature, and crude touch from somatic and visceral structures travel through the lateral spinothalamic tract. The anterior spinothalamic tract also transmits pain, temperature, and touch information to the brainstem and diencephalon (Figure \ref{anatomy_of_pain})\cite{babos_grady_2013}.

\section{Classification and Characteristics of Pain}
\begin{figure}
\begin{center}
\includegraphics[scale=0.15]{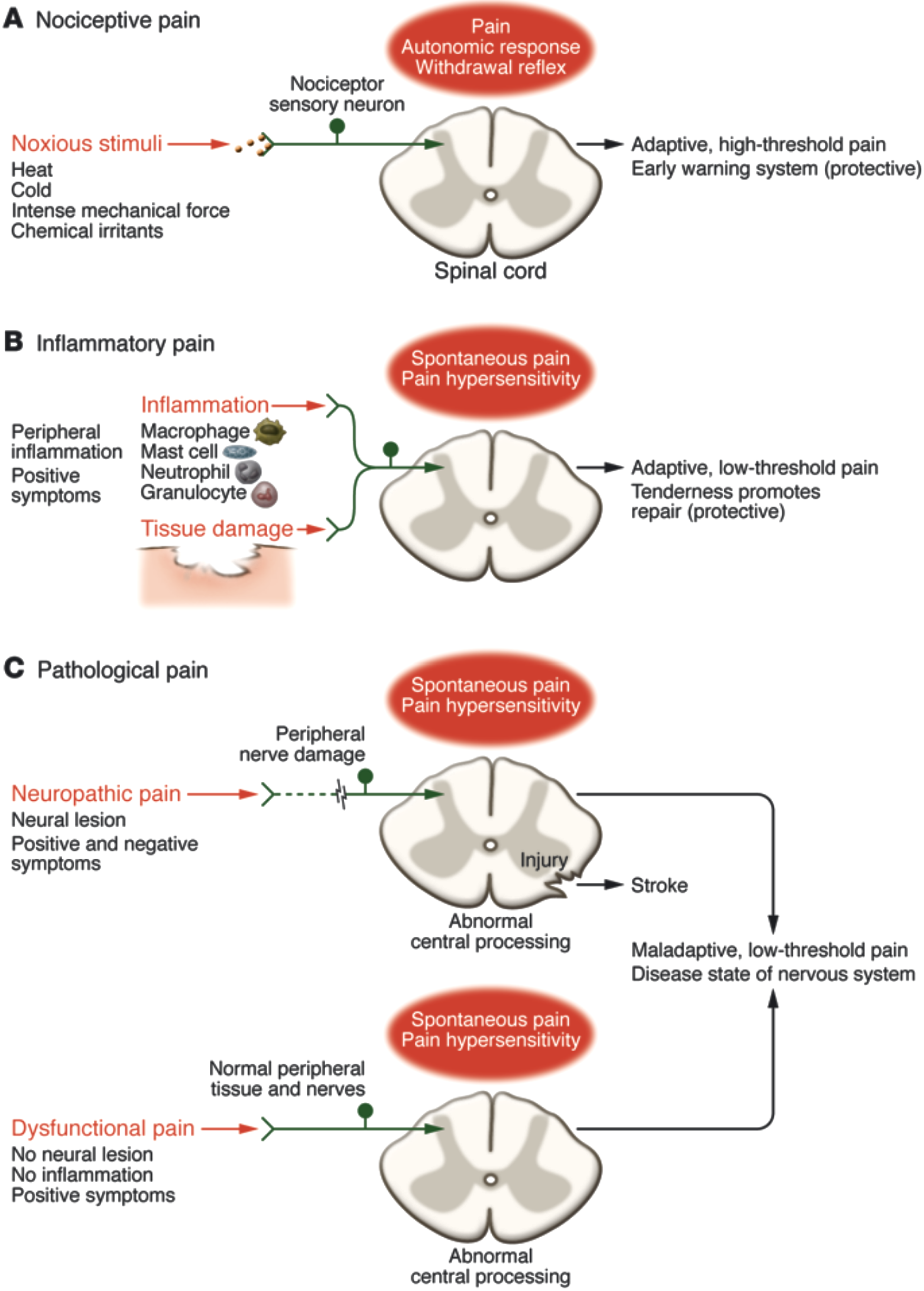} 
\end{center}
\caption{Pain classification \cite{wolf_2010}: \textbf{(A)} \textit{Nociceptive pain}, which results from detecting potentially harmful stimuli and serves a protective function. \textbf{(B)} \textit{Inflammatory pain} is linked to tissue damage and immune cell infiltration, increasing pain sensitivity during healing. \textbf{(C)} \textit{Pathological pain} is a disease state caused by either nervous system damage (neuropathic) or abnormal nervous system function (dysfunctional).}
\label{classes_of_pain}
\end{figure}

According to neurobiologist Clifford Woolf \cite{wolf_2010}, pain can be classified into three categories based on its function and characteristics: \textit{nociceptive}, \textit{inflammatory}, and \textit{pathological} pain. These classes and their respective functions are illustrated in Figure \ref{classes_of_pain}.

\textit{Nociceptive pain} (refer to \hyperref[classes_of_pain]{Figure. 2.2(A)}), arising from tissue damage, is a high-threshold pain that activates only in response to intense stimuli \cite{basbaum_bautista_2009}, serving as a vital warning signal to the body. The neurobiological system responsible for nociceptive pain evolved from the ability of even the most primitive nervous systems to detect impending or actual tissue damage caused by external stimuli. Its protective role requires immediate attention and action, achieved through the withdrawal reflex it initiates, the unpleasant sensation it produces, and the emotional distress it triggers. Nociceptive pain demands avoidance in the present moment, and when activated, it overrides most other neural processes \cite{wolf_2010}.

\textit{Inflammatory pain} (refer to \hyperref[classes_of_pain]{Figure. 2.2(B)}) is also protective and adaptive, increasing sensory sensitivity following tissue damage to aid healing by discouraging movement and contact with the injured area. This heightened sensitivity, or tenderness, helps prevent further harm and supports recovery, as seen after surgical wounds or inflamed joints where normally non-painful stimuli now cause pain. It is triggered by immune system activation in response to tissue injury or infection. Despite its adaptive role, this pain often needs to be alleviated in patients with persistent inflammation, such as in rheumatoid arthritis or severe injuries \cite{wolf_2010}.

\textit{Pathological pain} (\hyperref[classes_of_pain]{Figure. 2.2(C)}) is maladaptive, arising from abnormal nervous system functioning and not serving a protective role. Unlike nociceptive and inflammatory pain, pathological pain is a disease state of the nervous system itself. It may occur following nerve damage (neuropathic pain) or in conditions without apparent damage or inflammation (dysfunction l pain). Examples of dysfunctional pain include fibromyalgia, irritable bowel syndrome, tension headaches, and temporomandibular joint disease, where significant pain exists without an apparent noxious stimulus or peripheral pathology. Pathological pain, a low-threshold pain primarily driven by amplified sensory signals in the central nervous system, is the clinical pain syndrome with the greatest unmet need. To analogize, while nociceptive pain acts as a fire alarm for intense heat, and inflammatory pain reacts to warm temperatures, pathological pain is a false alarm triggered by a system malfunction. Thus, treatment must specifically target the underlying mechanisms causing each type of pain \cite{wolf_2010}.

Pain from a time-duration perspective can be categorized by duration into \textit{acute} and \textit{chronic}, with \textit{chronic} pain persisting or recurring for more than three months \cite{treede_rief_barke_2015}. \textit{Acute} pain is typically related to identifiable physiological damage from injury, surgery, illness, trauma, or medical procedures and generally subsides once the underlying cause is resolved. However, if untreated, it may develop into \textit{chronic} pain. 
\textit{Acute} pain is further classified into \textit{procedural} pain, caused by medical interventions such as muscular injections \cite{egede_2019_thesis}, and \textit{postoperative} pain, which occurs after surgery and is a significant concern for both patients and healthcare providers. 
Effective management is crucial to aid recovery and prevent the transition to chronic pain \cite{sinatra_2010}. \textit{Chronic} pain manifests in various forms, including \textit{chronic-recurrent} pain, like migraine headaches, and \textit{chronic-continuous} pain, such as persistent low back pain \cite{ruddere_tait_2018}.

\section{Pain Indicators}
Pain can manifest in numerous ways and is often shaped by individual characteristics and environmental influences. Various human expressions, actions, and bodily responses have been linked to pain, serving both communicative and coping purposes. These pain indicators are generally categorized into three primary groups: \textit{(i)} behavioral, \textit{(ii)} physiological, and \textit{(iii)} biochemical. While these indicators are universally present, certain expressions are more prominent in specific groups. For instance, crying is a common pain response across all age groups but is more frequently observed in younger infants. This may be due to contextual factors---such as culture, social status, age, and ego---influencing how pain is expressed over time. Adults, for example, may suppress crying in favor of other vocalizations, such as groans and moans, as crying could be perceived as inappropriate in certain contexts. These mediating factors are often considered when interpreting pain indicators. The following sections will delve into each of these three categories \cite{egede_2019_thesis}.

\subsection{Behavioral Indicators}
Behavioral indicators such as facial expressions (\textit{e.g.}, grimacing, open mouth, raised eyebrows), vocalizations (\textit{e.g.}, crying, moaning, screaming), and various bodily movements (\textit{e.g.}, changes in posture, signs of tension) are vital markers used in assessing pain \cite{rojas_brown_2023}.
Facial expressions and limb movements in response to acute pain are typically rapid and involuntary. Facial reactions include brow bulging, eye squeezing, nasolabial furrow formation \cite{oliveira_jesus_2011}, grimacing, clenched teeth, jaw-dropping, and tightened lips \cite{feldh_2000}. Body movements associated with pain include bracing (gripping an object or the affected area during movement), rubbing (massaging the painful area), restlessness (constant shifting of position) \cite{feldh_2000}, and knee flexion \cite{evans_vogelpohl_1997}. Non-verbal vocalizations such as groaning, moaning, sighing, crying, and gasping \cite{abu_bours_1998} also indicate pain. Verbal expressions like \textit{\textquotedblleft ouch\textquotedblright}, \textit{\textquotedblleft stop\textquotedblright}, \textit{\textquotedblleft that hurts\textquotedblright}, \textit{\textquotedblleft that is enough\textquotedblright}, and even cursing \cite{feldh_2000} also serve as pain indicators. Interestingly, swearing has been found to significantly alleviate pain, although its effect diminishes with frequent use over a short period \cite{stephens_atkins_2009, stephens_umland_2011}.

\subsection{Physiological Indicators}
Vital signs can reflect the state of the central nervous system, and since pain is mediated through this system, trends in vital signs can provide insights into pain levels. Clinical studies \cite{greisen_juhl_2001, stevens_johnston_1994} have examined physiological changes in response to pain and established empirical solid evidence linking pain to vital sign alterations. However, as vital signs can also change due to other non-pain-related pathological conditions, it is recommended that they be assessed alongside behavioral pain indicators for accuracy. Physiological pain responses are considered more reliable than behavioral signals, as they cannot be consciously controlled or altered. Physiological measurements such as electrocardiography (ECG), electromyography (EMG), galvanic skin responses (GSR), and respiration rate provide critical insights into the body's reaction to pain \cite{gkikas_tsiknakis_slr_2023}. In addition, brain monitoring techniques like near-infrared spectroscopy (fNIRS) have demonstrated the ability to detect pain-related hemodynamic changes \cite{rojas_liao_2019}. At the same time, functional magnetic resonance imaging (fMRI) has been explored for assessing pain in both normal and pathological conditions \cite{hoffman_2004}.

\subsection{Biochemical Indicators}
Compared to other pain indicators, biochemical changes are the most precise and sensitive reactions to pain. However, their routine use in pain assessment is restricted due to the invasive nature of measurement techniques \cite{mathew_mathew_2003}. These biochemical responses are most evident during surgical procedures with limited anesthesia, leading to increased levels of endorphins, norepinephrine, cortisol, growth hormones, renin, glucagon, aldosterone, and catecholamines, along with a decrease in insulin levels \cite{greisen_juhl_2001}.

\section{Sociodemographic and Psychological Variables}
In $1965$, Melzack and Wall \cite{melzack_wall_1965} introduced the \textit{\textquotedblleft Gate Control Theory\textquotedblright}, which interprets pain from two perspectives. The first involves the mechanisms of nociceptive signal transmission and modulation, while the second emphasizes pain as a psychophysiological phenomenon arising from the interaction between physiological and psychological factors \cite{ruddere_tait_2018}. Observations, empirical research, and theoretical models increasingly suggest that a comprehensive understanding of pain requires a biopsychological approach. It is also becoming apparent that, although pain is often regarded as private and subjective, it is also fundamentally a social experience \cite{ruddere_tait_2018}. Pain is not solely explained by biomedical components (\textit{e.g.}, muscle damage) but also involves psychological (\textit{e.g.}, cognitive, affective) and social factors (\textit{e.g.}, friends, family, health professionals), leading to what is known as a biopsychosocial sensation \cite{hadjistavropoulos_craig_2011}. Numerous factors contribute to how painful experiences are expressed and perceived, varying wildly due to social and personal biases.
These factors prompted Williams and Craig \cite{williams_craig_2016} to define pain as \textit{\textquotedblleft a distressing experience associated with actual or potential
tissue damage with sensory, emotional, cognitive, and social
components.\textquotedblright}

\subsection{Sex and Gender}
Several studies have explored the relationship between gender and pain expressiveness, as well as variations in pain reporting. Research indicates that women generally exhibit a lower pain threshold compared to men. A meta-analysis by Boerner \textit{et al.} \cite{Boerner_birnie_2014} on gender differences in children and adolescents found that girls over the age of $12$ reported higher pain intensity in response to cold-induced pain than boys. Furthermore, multiple studies suggest that women tend to describe a greater degree of pain compared to men. In addition to biological differences, psychological aspects linked to gender also play a role. For instance, individuals with a masculine identity may be less inclined to express or report their pain or seek assistance \cite{Keogh_2015}. 

Moreover, the manifestation of pain is not only influenced by the individual's gender but also by dyadic interactions between people of different sexes. Levine and Desimone \cite{Levine_simone_1991} conducted one of the initial studies on this phenomenon, showing that male participants in a cold pressure experiment reported lower pain intensity when a female experimenter was present. Similarly, McClelland and McCubbin \cite{McClelland_mccubibin_2008} found that female participants expressed and reported higher pain levels when accompanied by a female friend. This dynamic also extends to patient-healthcare provider interactions. In studying health records, Vigil and Alcock \cite{vigil_alcock_2014} discovered that when the pain intensity was reported as high, the patients (\textit{i.e.}, men and women) were examined by a female doctor or nurse.
Additionally, studies examining gender differences among physicians in pain treatment options revealed that female patients were more likely to receive prescriptions for more potent drugs, such as analgesics, and female physicians were more likely to prescribe medications. 
Extensive research has also shown that both lay observers and healthcare professionals tend to estimate higher pain levels for female patients compared to male patients \cite{Raftery_coggins_1995}. 
Hooper \textit{et al.} \cite{Hooper_comstock_1982} further noted that clinicians communicate more effectively with female patients, often displaying greater empathy. Gender roles, beliefs, and expectations play a significant role in understanding how social factors influence the differences in pain perception and experience between men and women \cite{Keogh_book_chapter_2018}.

\subsection{Age}
Age plays a crucial role in pain assessment and management. At the same time, there are significant challenges, limitations, and biases related to the patient's age group. Two of the most vulnerable groups, albeit for different reasons, are the elderly and infants.

Pain recognition and interpretation among the elderly, particularly by caregivers, often present unique challenges. Older adults frequently exhibit stoicism and reluctance to express their pain, while healthcare providers struggle to accurately assess the patient's pain, leading to inappropriate pain management decisions \cite{Hadjistavropoulos_gallant_2018}. McPherson \textit{et al.} \cite{McPherson_Hadjistavropoulos_Devereaux_Lobchuk_2014} noted that despite caregivers' accommodating and empathetic relationships with elderly patients, conflicts still arise. Older patients may resist acknowledging their weaknesses and accepting help, which can cause them to conceal their pain. The situation becomes even more complex when dealing with dementia, a disorder encompassing a range of conditions (\textit{e.g.}, Parkinson's, Alzheimer's, Vascular dementia), characterized by abnormal brain changes that impair cognitive and linguistic abilities. A person with dementia may find it challenging to communicate their pain verbally. However, non-verbal pain expressions remain intact even in moderate dementia, although such reactions can be exaggerated \cite{Hadjistavropoulos_lachapelle_macleord_2000}. However, aggressive behavior and disturbances in dementia patients, often caused by pain, are frequently misinterpreted as psychiatric symptoms, leading to improper medication that can have life-threatening consequences \cite{ballard_hanney_2009}. 
Caregivers of dementia patients face additional challenges, not only related to pain management but also in addressing dementia's impact on language and memory. Particularly in the later stages of dementia, patients encounter severe pain communication difficulties due to cognitive decline, necessitating that caregivers recognize behavioral and contextual indicators of pain \cite{Hadjistavropoulos_gallant_2018}.
Age is also known to cause changes in skin characteristics, such as texture, rigidity, and elasticity, which impact the performance of emotional face recognition tasks \cite{ochi_midorikawa_2021}.

Infants represent another vulnerable age group where pain assessment requires specialized attention, particularly when they experience painful events. The first challenge is obviously their limited reporting ability to express their pain through language. Although crying might appear to signal pain, this is an oversimplified and unreliable method, as crying can indicate a variety of situations, such as discomfort, hunger, or pain. Accurately discerning the type of cry is only one part of the challenge; assessing pain in infants is far more complex and influenced by numerous factors, including the interpersonal relationships within their environment. Riddell and Racine \cite{riddell_racine_2009} found that through various distressing experiences, infants can learn that specific signaling behaviors can prompt their caregiver's proximity. This attachment dynamic suggests that, to some extent, infants may consciously utilize pain-related behaviors to elicit responses from their caregivers. Similarly, the context affects older children as well; for example, self-reports of pain tend to be significantly lower when a parent is present compared to when the child is alone \cite{vervoort_caes_2011}.

\subsection{Psychological Factors}
Multiple studies have revealed that several psychological factors are consistently linked with pain-related behavior, including depression, pain-related fear, and catastrophizing. Research focusing on the impact of depression and anxiety on pain-related behavior has been conducted mainly on patient populations. These studies have shown that depressed individuals exhibit more pronounced protective and communicative behaviors compared to non-depressed patients \cite{keefe_wilkins_1986}. Similarly, numerous studies suggest that patients with higher levels of anxiety demonstrate more pain-related behaviors than those with lower anxiety levels \cite{burns_quartana_2008}. Despite the frequent coexistence of pain with psychological conditions, research indicates that these patients often experience underestimation of their pain. For instance, De Ruddere \textit{et al.} \cite{Ruddere_goubert_2014} found that patients dealing with psychological stressors such as anxiety, depression, and daily life challenges are often perceived by physiotherapists as experiencing less severe pain, illustrating the influence of psychosocial factors on the patient's pain experience.

\subsection{Race and Culture}
Pain expression is generally understood across ethnicities and cultures, though differences exist in how it is conveyed \cite{williams_kappesser_2018}. However, cultural variations and the nuances of facial expressions related to emotion are complex and necessitate deeper study. Additionally, racial and cultural biases significantly influence pain assessment, judgment, and interpretation. Extensive research highlights the impact of a patient's race as a sociodemographic factor on observer responses. The most examined topic relates to the different responses toward Caucasian versus non-Caucasian individuals, particularly African Americans, who are more likely to have their pain underestimated and undertreated by healthcare providers \cite{staton_panda_2007}.

Ethnocultural factors are crucial in shaping how individuals perceive and express pain. For example, Western cultures often emphasize conservative expressions and self-control, leading to restrained responses in personal pain experiences and in perceiving others' pain \cite{riddell_craig_2018}. Differences also arise in coping mechanisms; African Americans, for instance, are more prone to catastrophizing pain events compared to European Americans \cite{meints_stout_2017}. Furthermore, evidence shows racial biases in pain treatment across various racial groups, with certain groups being more sensitive to pain but receiving lower-quality treatment \cite{drwecki_2018}. For example, Cleeland \textit{et al.} \cite{cleeland_gonin_1994} found that minority cancer patients, mainly Black and Hispanic individuals, were more likely to experience inadequate analgesia compared to non-minority patients.

\subsection{Observer's Impact on Pain}
The variability in pain management stems from the interplay of various elements, including sociocultural, biomedical, and psychosocial factors, especially in cases of chronic pain \cite{DeRuddere_tait_2018}. When it comes to the observer responsible for assessing a patient's pain, several characteristics directly influence the objectivity of their evaluation.
The first and perhaps most critical factor is the observer's experience level. One would expect that more experience leads to better and more accurate assessments, but studies show that even experienced healthcare providers consistently underestimate pain, much like laypersons \cite{dekel_gori_2016}. The greater the experience, the more pronounced the underestimation tends to be. This may be due to desensitization caused by repeated exposure to pain events, as seen in the differences between internists and surgeons in their evaluation of postoperative pain, with surgeons often encountering severe pain more regularly \cite{tait_chibnall_2010}. Another significant factor is the observer's knowledge and beliefs about pain. For example, \cite{Ruddere_goubert_2014} found that laypersons and healthcare professionals without physical signs of pain might view the patient's complaints less seriously. Proper training is also essential for adequate pain assessment, which is why the Department of Health and Human Services (DHHS) initiated a strategic program to improve healthcare providers' education and knowledge regarding pain management, following evidence of inadequate training in the field \cite{national_academies_press}.

\section{Impact of Inadequate Pain Management}
The experience of pain, particularly persistent pain, can have detrimental effects on the individual and their surrounding environment. Thoughts about severe pain often lead to grief and fear, causing individuals to perceive pain as a threat and feel incapable of managing it. This can prompt avoidance behaviors aimed at escaping perceived harm \cite{caes_goubert_2018}. Studies have shown that children with a catastrophizing mindset about pain struggle with daily activities, while adolescents with chronic pain tend to have fewer friends and may miss out on social and entertainment opportunities, putting them at greater risk of victimization \cite{Crombez_bijttebier_2003}. These adolescents often feel isolated and lonely compared to their healthy peers, and they may experience anxiety in social interactions \cite{Forgeron_mcgrath_stevens_2011}. Parental reactions to their children's pain can further complicate the situation, as parents with catastrophic tendencies tend to engage in overprotective behaviors that hinder the child's functioning and psychosocial development \cite{Goubert_simons_2013}. Additionally, the family's overall dynamic is affected, with the patient's sadness, sleep disorders, and changes in leisure activities impacting the household \cite{Hadjistavropoulos_gallant_2018}.

On a biological level, pain, particularly when experienced early and severely, can alter the brain and nervous system. These early pain experiences can disrupt neurobiological development and affect how pain is processed later in life \cite{Fitzgerald_walker_2009}. A growing body of research links chronic pain to changes in the medial prefrontal cortex, a region crucial to emotional processing. Chronic pain is associated with structural and biochemical alterations in this brain area, suggesting that these changes play a role in the pathophysiology of chronic pain \cite{kang_shariati_2021}.

\section{Pain Measurement Scales and Metrics}
In clinical settings, self-reporting remains the gold standard for assessing pain, allowing individuals to describe their pain's intensity and location. Various self-report scales have been developed for different age groups, such as the visual analog scale (VAS) \cite{delgado_lambert_2018} and the verbal rating scale (VRS) \cite{haefeli_2006}. Additionally, observation-based scales, where a third party evaluates the pain's severity, include tools like the Prkachin and Solomon pain intensity scale (PSPI) \cite{Prkachin_solomon_2008} and the neonatal/infant pain scale (NIPS) \cite{lawrence_1993}. However, some studies suggest that patients may exaggerate their pain severity to prompt more aggressive treatment interventions \cite{Weissman_haddox_1989}, raising concerns about the accuracy of self-reported symptoms. Therefore, objective pain measurement remains clinically crucial.

    \chapter{Automatic Pain Assessment--A Literature Review}
\label{slr}
\minitoc  

\section{Chapter Overview}
This chapter corresponds to the publication \cite{gkikas_tsiknakis_slr_2023},
a systematic literature review (SLR) conducted at the start of this Ph.D. research. This review facilitated an understanding of automatic pain assessment methods, particularly those based on deep learning, and the techniques and strategies employed. It enabled the identification and proposal of new approaches that could enhance the effectiveness of pain recognition. 

Additionally, it allowed for identifying gaps in the literature from other reviews conducted on this specific research topic.
Every existing systematic review on pain assessment was identified and assessed, revealing several insights. The first review on automatic pain assessment, published by Prkachin in 2009 \cite{prkachin_2009}, did not cover papers on deep learning, as the practical implementations of deep architectures only began around 2012. Zamzmi \textit{et al.} \cite{zamzmi_2018} focused their review exclusively on infants, omitting deep learning methods. In 2018, Chen \textit{et al.} \cite{chen_2018} reviewed automated pain detection methods using the Facial Action Coding System (FACS), noting only three publications that employed deep learning techniques.
In 2019, Hassan \textit{et al.} \cite{hassan_2019} included only seven papers that used deep learning methods in their review. Similarly, Werner \textit{et al.} \cite{werner_2019}, also in 2019, discussed pain assessment without restrictions on modalities or age groups, finding fewer than ten papers that reported on deep learning methods. In 2020, Al-Eidan \textit{et al.} \cite{eidan_2020} published the first systematic literature review titled \textquotedblleft Deep-Learning-Based Models for Pain Recognition: A Systematic Review\textquotedblright , which included fifteen papers but was critiqued for having significant limitations and incorrect information. 
It was noted that some papers analyzed might not be relevant, and there was confusion between \textquotedblleft neural networks\textquotedblright\space and \textquotedblleft deep learning\textquotedblright . For instance, while study \cite{hassan_2019} mentioned using neural network approaches, they did not provide evidence of using deep learning methods. Moreover, in the study \cite{chen_2018}, the authors developed a neural network with only two layers combined with handcrafted features, which does not qualify as a deep learning method. Additionally, studies \cite{zamzmi_2018,eidan_2020} focused on detecting protective movement behaviors in chronic pain patients, which deviates from the central topic of automatic pain assessment.
Several reviews and SLRs on automatic pain assessment have been published, but none exclusively or adequately focus on deep learning methods. This SLR aims to bridge this gap by thoroughly reviewing deep learning techniques used for automatic pain assessment.


\section{Modalities and Hardware for Automatic Pain Assessment}
Creating an automatic pain assessment system hinges on capturing the necessary input data through various information channels, referred to as modalities. These modalities are categorized into behavioral and physiological types. A system utilizing only one modality is termed unimodal, whereas a multimodal system incorporates multiple modalities.

Key behavioral modalities encompass facial expressions, body movements, gestures, and auditory signals. Researchers use a range of optical and light sensors to record images or video sequences of facial and body movements. Commonly, researchers employ color RGB cameras, but depth and thermal sensors are also used to enhance visual data. Motion capture sensors are also employed to track movements, and microphones are frequently employed to capture sound. On the physiological front, modalities often involve biosignals that detect electrical activities from various tissues and organs. Techniques such as electrocardiography (ECG), electromyography (EMG), electrodermal activity (EDA), photoplethysmography (PPG), blood oxygen saturation (SpO2), near-infrared spectroscopy (NIRS), respiration rate, and skin temperature are commonly used to gauge pain. Multiple sensors can measure several modalities simultaneously --- for instance, strain sensors and cameras can track respiration rates.

Besides the sensors that gather input data, the computational hardware is crucial. Deep learning-based systems operate in two phases: training and inference. The training phase is particularly resource-intensive, necessitating a graphics processing unit (GPU). The trained model makes predictions on new data during inference, typically processed on a central processing unit (CPU). The choice of hardware depends on various factors, especially in real-time scenarios where low latency is crucial, compared to offline settings where data processing can be deferred. Additionally, characteristics of the model, such as floating point operations per second (FLOPS) and total computational operations, are significant considerations.

\section{Pain Databases}
\begin{figure}
\centering
\includegraphics[scale=0.60]{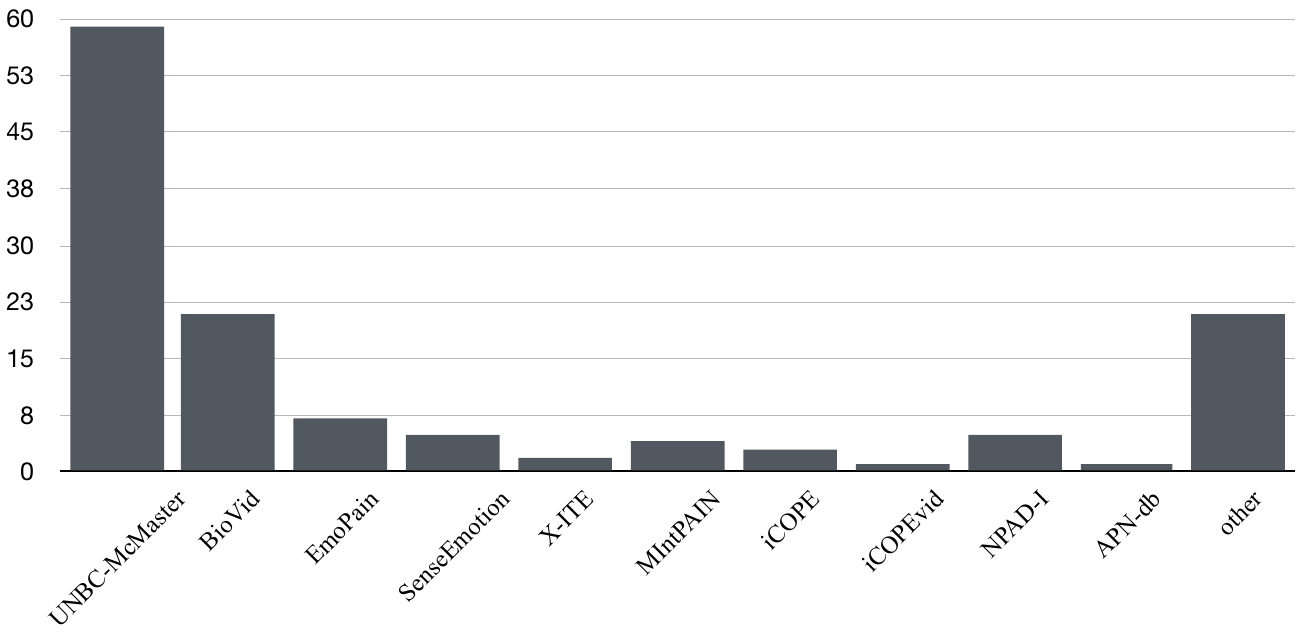}
\caption{The number of studies utilizing these specific datasets. Note that various studies used multiple datasets to conduct their experiments.}
\label{fig:pain_datasets}
\end{figure}

Access to data is crucial for evaluating methods and algorithms in automatic pain assessment. 
However, only a few databases have explicitly been developed for automatic pain recognition based on human behavioral and physiological changes. Unlike the extensive data found in most facial expression databases, publicly accessible pain datasets often offer limited samples and suffer from significant class imbalance. This primarily stems from the ethical concerns associated with collecting pain data.
Table \ref{table:pain_databases} lists the principal databases reviewed in the studies. Figure \ref{fig:pain_datasets} shows how frequently each database was used. Most research utilized publicly available datasets, with some studies exploring multiple datasets. Few studies used private datasets, mainly those aimed at detecting pain in neonates.
The \textit{UNBC-McMaster Shoulder Pain Archive Database} \cite{unbc_2011} is the most utilized, followed by The \textit{BioVid Heat Pain Database} \cite{biovid_2013}. The former contains $200$ facial videos of $25$ individuals with shoulder pain. At the same time, the latter includes facial videos and biopotentials of $90$ healthy participants subjected to experimentally induced heat pain at four intensity levels. 
The following subsections provide a brief description of some of these datasets.

\renewcommand{\arraystretch}{1.5}
\begin{table*}[htb] 
\myfontsize
\begin{threeparttable}
\caption{Most commonly utilized pain databases.}
\label{table:pain_databases}

\begin{tabular}{ p{2.2cm} p{3.2cm} p{3.5cm} p{1.9cm} p{2.4cm}}
\hline
\toprule
\textbf{Database} & \textbf{Modality}  & \textbf{Population} &\textbf{Annotation \newline Granularity} &\textbf{Annotation Labels}\\
\hline
\hline

\textbf{UNBC-McMaster} \newline Shoulder Pain$^{A}$ \cite{unbc_2011} & RGB video of face &25 adults with shoulder pain &\begin{tabular}{@{}l@{}}Frame level\\Sequence level\end{tabular} &\begin{tabular}{@{}l@{}}FACS\\VAS, OPI\end{tabular}\\

\textbf{BioVid$^{A}$} \cite{biovid_2013} & RGB video of face, EDA, ECG, EMG &87 healthy adults  &Sequence level &stimulus \newline(calibrated per person)\\

\textbf{MIntPAIN$^{A}$} \cite{haque_bautista_2018} & RGB-Depth-Thermal video of face & 20 healthy adults &Sequence level &stimulus \newline(calibrated per person), VAS\\

\textbf{iCOPE$^{A}$} \cite{brahnam_chuang_2006} &RGB photographs of face &26 healthy neonates &Frame level &pain, cry, rest, air puff, friction\\

\textbf{iCOPEvid$^{A}$} \cite{brahnam_nanni_2019}  & Grayscale video of face & 49 neonates &Sequence level &pain, no pain\\

\textbf{NPAD-I$^{A}$} \cite{zamzmi_2019}  & RGB video of face \& body, HR, SpO2, BP, NIRS  & 36 healthy neonates \& 9 neonates \newline with tissue injured by surgery &Sequence level &NIPS, N-PASS\\

\textbf{APN-db$^{A}$} \cite{egede_valstar_2019_a}  & RGB video of face & 112 healthy neonates &Sequence level &NFLAPS, NIPS, NFCS\\
\hline

\textbf{EmoPain$^{N}$} \cite{emopain_dataset_2016} &video, audio, EMG, MoCap &22 adults with chronic pack pain \& 28 healthy adults &Sequence level\ &self-report, naive OPI\\

\textbf{SenseEmotion$^{N}$} \cite{senseemotion_database_2017} &video of face, audio, EDA, ECG, EMG, RSP& 45 healthy adults &Sequence level &stimulus \newline(calibrated per person)\\
\textbf{X-ITE$^{N}$} \cite{x-ite_databaase_2019} & RGB-Thermal video of face, \newline RGB-Depth video of body, audio, EDA, ECG, EMG & 134 healthy adults &Sequence level &stimulus \newline(calibrated per person)
 
\\\bottomrule
\end{tabular}
\begin{tablenotes}
\myfontsize
\item ${A}$: Publicly available by request, complete or part of the dataset 
${N}$: Not yet available
\textbf{Modality:} HR: heart rate SpO2: oxygen saturation rate BP: blood pressure NIRS: near-infrared spectroscopy MoCap: motion capture RSP: respiration rate EDA: electrodermal activity ECG: electrocardiogram  EMG: electromyogram 
\textbf{Annotation Labels:} FACS: Facial Action Coding System VAS: visual analogue scale OPI: observer pain intensity NIPS: neonatal infant scale  N-PASS: neonatal pain, agitation and sedation scale NFLAPS: neonatal face and limb acute pain scale NFCS: neonatal facial coding system 
          \end{tablenotes}
          \end{threeparttable}

\end{table*}

\subsection{The UNBC-McMaster Shoulder Pain Expression Archive Database}
The \textit{UNBC McMaster Shoulder Pain Database} \cite{unbc_2011} comprises $200$ video sequences showing the facial expressions of $25$ subjects undergoing motion tests, including arm abduction and external and internal rotations. The data collection utilized both active and passive approaches: in the active mode, subjects moved their affected arms to their bearable limit, while in the passive mode, a physiotherapist moved the subjects' arms. Each video sequence contains about $60$ to $700$ frames, totaling $48,398$, with $82.71\%$ of frames scoring a pain rating of zero, indicating a significant imbalance in the data.
All frames are FACS-coded for pain-related action units (AUs)---AU4, AU6, AU7, AU9, AU10, AU12, AU20, AU25, AU26, AU27, and AU43---with each AU coded for intensity from A to E, $0$, or $5$, except for AU43 (closed eyes), which is coded as either present or absent. Pain scores are assigned using the PSPI metric based on the intensity of the AUs present. Additionally, the database includes $66$ facial landmarks per frame, determined by an active appearance model. Pain assessments also include self-reports using two Likert scales with $15$ options each and a visual analog scale (VAS) from $1$ (no pain) to $10$ (extreme pain). One scale measures the sensory intensity from \textquotedblleft extremely weak\textquotedblright\space to \textquotedblleft extremely intense\textquotedblright\space , while the other assesses the affective-motivation aspect of pain from \textquotedblleft bearable\textquotedblright\space to \textquotedblleft extremely excruciating\textquotedblright . Independent observer pain intensity (OPI) ratings use a $6$-point scale from $0$ (no pain) to $5$ (intense pain). The \textit{UNBC} database is currently the most extensively utilized dataset for automatic pain recognition among publicly available resources.

\subsection{The Biopotential and Video (BioVid) Heat Pain Database}
The \textit{BioVid} dataset \cite{biovid_2013} is a prominent resource in pain research, comprising facial videos, electrocardiograms, electromyograms, and galvanic skin response data from eighty-seven $(n\text{=}87)$ healthy participants ($44$ males and $43$ females, aged $20$ to $65$). The pain was induced using a thermode on the participants' right arm, with pain and tolerance thresholds established before data collection. These thresholds defined the range of pain from No Pain (NP) to Very Severe Pain (P\textsubscript{4}), encompassing five levels of pain intensity. The temperatures for the pain inductions ranged from P\textsubscript{1} to P\textsubscript{4} and did not exceed $50.5^\circ\text{C}$. Each participant underwent $20$ inductions at each of four pain levels, with each induction lasting $4s$ followed by a recovery period of $8$ to $12s$.
In addition, $20$ baseline measurements were taken at $32^\circ\text{C}$ (NP), totaling $100$ stimulations per participant, randomly administered. Data processing segmented these into $5.5s$ durations starting $1s$ after the target temperature was reached, resulting in $8,700$ samples across the five pain intensity classes, equally distributed among all modalities for each participant.
Video recordings were made at a frame rate of $25$ FPS, and biosignal (ECG, EMG, GSR) recordings were sampled at $512$ Hz.

\subsection{The EmoPain Database}
The \textit{EmoPain} \cite{emopain_dataset_2016} dataset encompasses various pain indicators, including body movements, audio, biosignals, and postural and facial expressions. It features video and audio recordings of $22$ patients ($7$ male, $15$ female) exhibiting natural pain expressions while engaging in physiotherapy-like exercises. These exercises, performed at regular and challenging levels, include a sitting-standing sequence, balancing on one leg for five minutes, and reaching forward while standing.
The video signals are captured in high resolution ($1024\times1024$ pixels) using eight cameras positioned at various angles, enhanced by specialized lighting conditions. Audio is recorded with two microphones: an \textit{AKG C-1000S MKIII} placed near the cameras and an \textit{AKG HC 577 L} worn by the patients, both operating at a $48$kHz sampling rate with bit Pulse Code Modulation. Body movements and postures are tracked using a motion capture suit with $18$ sensors distributed across the body. Biosignals are monitored with four sEMG sensors attached to the trapezius and lumbar para-spinal muscles.
Additionally, the dataset provides continuous frame-wise pain ratings for facial expressions by eight naive annotators and binary frame-wise annotations for protective behaviors by four experts, along with coordinates from $26$ body nodes. Six annotated protective behaviors include stiffness, bracing, hesitation, limping, rubbing, and abrupt actions. Audio and EMG signals from the eight activities per subject also contribute to multimodal pain recognition.
Like the \textit{UNBC} database, \textit{EmoPain} faces significant challenges due to data sparsity and imbalance---only $11.4\%$ of frames show facial expressions of pain, and $8.6\%$ show protective behaviors. This scarcity complicates pain recognition research, necessitating the development of methods that efficiently utilize limited data to achieve optimal performance.

\subsection{The Experimentally Induced Thermal and Electrical (X-ITE) Pain Database}
The \textit{X-ITE} \cite{x-ite_databaase_2019} dataset is one of the largest pain datasets but is not publicly available. It involved $134$ healthy adults ($67$ men and $67$ women) aged between $18$ and $50$. The average age was $31.4$ years (SD = $9.7$), with men averaging $33.4$ years (SD = $9.3$) and women $32.9$ years (SD = $10.2$). Participants had no chronic pain, depression, psychiatric disorders, neurological conditions, headache syndromes, or cardiovascular disease, nor had they taken pain medication or painkillers before the experiment.
Pain stimuli were stimulated using the \textit{Medoc PATHWAY Model ATS} for heat pain on the forearm and the \textit{Digitimer DS7A} for electrical pain on the index and middle fingers. Both modalities featured phasic stimuli (short, $5$ seconds) and tonic stimuli (long, $60$ seconds), each in three intensities. After calibration, participants underwent a $90$-minute stimulation phase where phasic stimuli were repeated $30$ times in a randomized sequence with $8$-$12$-second pauses. The tonic stimuli were applied once per intensity, totaling six per participant, each followed by a five-minute pause. The highest intensity tonic stimuli for heat and electrical pain were induced at the experiment's ending, with the other stimuli randomly interspersed during the phasic period.
Simultaneous to the pain stimulation, various sensors collected multimodal pain response data: frontal and side view RGB videos for facial expression and head pose analysis, audio for paralinguistic response analysis, electrocardiogram (ECG) to monitor heart rate variability, surface electromyography (EMG) to assess muscle activity in the trapezius, corrugator supercilii, and zygomaticus major, electrodermal activity (EDA) to measure sweating, video for body movement analysis, and thermal video for facial temperature changes.

\subsection{The AI4Pain Database}
The \textit{AI4Pain Grand Challenge 2024} \cite{ai4pain} dataset is a recent contribution to the pain research field, tailored for sophisticated pain recognition tasks using fNIRS and facial video data. This dataset involves sixty-five volunteers $(n\text{=}65)$, including $23$ females, with ages ranging from $17$ to $52$ years (mean age of $29.06$ years and a standard deviation of $8.28$ years). Although it captures physiological signals such as photoplethysmography (PPG), electrodermal activity (EDA), and respiration (RESP), these signals are not publicly available yet. The dataset is segmented into three parts: training ($41$ volunteers), validation ($12$ volunteers), and testing ($12$ volunteers).
The experimental setup includes fNIRS data recorded with an \textit{Artinis} device, measuring changes in oxygenated and deoxygenated haemoglobin concentrations across $24$ channels targeting the prefrontal cortex. The optodes configuration includes $10$ sources and $8$ detectors spaced $30$ mm apart, using near-infrared light at $760$ nm and $840$ nm, sampled at $50$ Hz. Additionally, facial movements are captured by a \textit{Logitech StreamCam} at $30$ FPS.
The \textit{AI4Pain} dataset categorizes pain into three levels: \emph{No Pain}, \emph{Low Pain}, and \emph{High Pain}. It features $65$ instances of \emph{No Pain} (each lasting $60$s), $780$ instances of \emph{Low Pain} (each lasting $10$s), and $780$ instances of \emph{High Pain} (each lasting $10$s). The \emph{No Pain} instances, recorded during baseline, serve as control data. The \emph{Low Pain} instances reflect mild pain responses, and the \emph{High Pain} instances capture significant pain, both derived from a pain tolerance test and reflected in the corresponding neurological and behavioral data recorded.

\section{Unimodal studies}
\label{sec:unimodal}
This section presents the studies that utilized only one information channel to estimate the subject's pain condition.

\subsection{Vision-based: Static Analysis}
\label{sec:non_temporal}
The first publicly available pain database that significantly contributed to the development of automatic pain assessment methods was the \textit{UNBC-McMaster Shoulder Pain Database}. Numerous studies have employed this dataset. Pedersen \cite{pedersen_2015} implemented the first deep learning approach in 2015 to address the pain assessment problem, utilizing a $4$-layer contractive autoencoder. He combined the encoded representations with a support vector machine (SVM), achieving high performance in frame-level pain detection.
A significant advancement in vision-based pain recognition methods was the EmoPain challenge in 2020, which became the first international competition to compare machine learning methods for chronic pain assessment. Egede \textit{et al.} \cite{egede_song_olugbade_2020} presented the \textit{EMOPAIN 2020 Challenge}, utilizing a dataset composed of features extracted via both handcrafted methods and deep-learned models. They utilized facial landmarks, histogram of oriented gradients (HOG), and deep vectors from \textit{VGG-16} \cite{simonyan_karen_2015} and \textit{ResNet-50} \cite{he_zhang_2016}, both pre-trained on the \textit{Aff-Wild} dataset\footnote{\url{https://ibug.doc.ic.ac.uk/resources/first-affect-wild-challenge}}. The authors report that combining hand-engineered features with deep learning cues led to the best performance. Similarly, Yang \textit{et al.} \cite{yang_hong_2018} extracted both low- and high-level features from local descriptors and the pre-trained \textit{VGG-16} CNN, combining them through weighted coefficients. Semwal and Londhe \cite{semwal_londhe_august_2021} demonstrated that fusing deep-learned features with facial landmarks is beneficial for multi-class pain estimation. Lakshminarayan \textit{et al.} \cite{lakshminarayan_hinduja_2020} combined deep-learned features with handcrafted ones---namely features from \textit{VGG-16} \cite{simonyan_karen_2015} and \textit{ResNet-50} \cite{he_zhang_2016}, HOG, action unit occurrence and intensity, facial landmarks, and head pose---through a fully connected network. Their study found that combining \textit{VGG-16} with handcrafted features lowered regression error, whereas \cite{huynh_yang_2020} achieved maximal performance using only \textit{VGG-16} features with a fully connected network.

Conversely, Semwal and Londhe \cite{semwal_londhe_2018} noted the limitations of traditional handcrafted feature engineering and the computational expense of deep neural networks. As a solution, they proposed a relatively shallow $4$-layer CNN, which reduces computational costs due to fewer parameters while achieving performance comparable to deeper models. A different approach came from \cite{tavakolian_hadid_2018_b}, where the authors focused on representing facial expressions as compact binary codes for pain intensity classification. Feature extraction was conducted using a pre-trained model \cite{yi_lei_2014}, with a fully connected network used to generate the binary codes.

Several studies utilized CNN ensemble designs with varying architectures to exploit feature diversity. Semwal and Londhe \cite{semwal_londhe_2021} combined predictions from three compact CNNs---\textit{VGG-16}, \textit{M-MobileNet} \cite{lawhern_solon_2018}, and \textit{GoogleNet} \cite{szegedy_2017}---using the average ensemble rule, resulting in improved classification performance. Kharghanian \textit{et al.} \cite{kharghanian_peiravi_2016} developed a convolutional deep belief network (CDBN) using unsupervised feature learning. An SVM used the extracted features to differentiate between two states for binary pain classification (\textit{i.e.}, pain vs. no pain). Later, \cite{kharghanian_peiravi_moradi_2021} added two layers to the CDBN, though the results were not directly comparable due to differing evaluation methods.

Several papers suggest that because pain is predominantly expressed in specific facial regions, focusing on these areas rather than the whole face could improve model accuracy by reducing noise. Huang \textit{et al.} \cite{huang_xia_li_2019} initially identified the left eye, right eye, nose, and mouth as key regions and utilized a multi-stream CNN for feature extraction, assigning learned weights to enhance attention on these regions. Xin \textit{et al.} \cite{xin_lin_2020} employed a $9$-layer CNN with an attention mechanism to assign different weights to face regions, resulting in more accurate attention face maps and boosting prediction accuracy by up to $19\%$. Cui and Huang \cite{cui_huang_2021} introduced a multi-scale regional attention network (MSRAN), which uses multiple cropping regions from video frames. The framework includes self-attention and relation-attention modules to highlight pain-relevant regions and explore interrelationships. Li \textit{et al.} \cite{li_zhu_2018} extended this concept by integrating contrastive and multi-task training through an autoencoder, building on the work of \cite{schroff_kalenichenko_2015}.

One challenge in pain intensity estimation is that individual facial features, such as face shape, can introduce significant variability in how pain is expressed. This makes it difficult to distinguish between adjacent intensity levels. To address this, Peng \textit{et al.} \cite{peng_huang_zhang_2020} examined facial shape information and developed a deep multi-task network to account for the relationship between pain recognition and shape, which improved pain estimation performance. Similarly, Xin \textit{et al.} \cite{xin_li_yang_2021} proposed a novel multi-task framework that combines a CNN feature learning module with an autoencoder attention component, also estimating subject identity, as individual differences in pain manifestation are key. Their experiments achieved state-of-the-art results on publicly available datasets.

Most studies report results obtained from controlled laboratory settings, which typically feature proper lighting, minimal head pose variability, and no occlusions. However, such conditions do not represent typical hospital environments. Semwal and Londhe \cite{semwal_londhe_b_2021} addressed this by focusing on pain assessment in uncontrolled settings, developing a shallow CNN with three convolutional layers that performed comparably to deeper pre-trained models. In a subsequent study \cite{semwal_londhe_c_2021}, they introduced a more complex framework comprising three modules that leveraged high-level spatial descriptors with both local and global geometric cues, achieving results comparable to models like \textit{GoogleNet} \cite{szegedy_liu_2015} and \textit{VGG} \cite{simonyan_karen_2015}. Lee and Wang \cite{lee_wang_2019} explored pain assessment in intensive care unit (ICU) settings, where partially occluded faces frequently complicate facial analysis. They developed a 4-layer CNN combined with an extreme learning machine (ELM) for final estimation. Virrey and Caesarendra \cite{virrey_caesarendra_2019} used CNNs to classify sections of frames where pain was triggered, peaked, and subsided. Nugroho \textit{et al.} \cite{nugroho_harmanto_2018} tackled pain detection in smart home-care settings, particularly for elderly patients, using relatively low-power mobile devices. They modified the \textit{OpenFace}\footnote{\url{http://cmusatyalab.github.io/openface}} library, based on pre-trained \textit{FaceNets} \cite{facenet_schroll_2015}, and showed that transfer learning could enable real-time binary classification (\textit{pain} vs. \textit{no pain}), even on low-powered hardware.

Researchers like Dai \textit{et al.} \cite{dai_broekens_2019} and Menchetti \textit{et al.} \cite{menchetti_chen_2019} have noted that most models, whether deep or shallow, are trained on dataset-specific features rather than actual pain-related features. Moreover, most studies employ validation methods using the same dataset, while cross-dataset performance is rarely addressed, limiting real-world applicability. To tackle these issues, Dai \textit{et al.} \cite{dai_broekens_2019} combined pain and emotion detection datasets to develop a real-time pain assessment system with better generalization capabilities. They emphasized the importance of cross-corpus evaluation, real-time testing, and the need for well-balanced, ecologically valid pain datasets \cite{andrade_2018}.

Several studies have explored combining pain scales to improve prediction objectivity and reliability. Liu \textit{et al.} \cite{liu_peng_2017} developed a two-stage personalized model trained using active appearance model (AAM) facial landmarks and multi-task learning, with visual analog scale (VAS) and observed pain index (OPI) as ground truth. Xu \textit{et al.} \cite{xu_huang_sa_2020} similarly reduced mean square error (MSE) by incorporating various pain scales with the \textit{VGG-Face} model. However, Casti \textit{et al.} \cite{casti_mencattini_2019} pointed out the limitations of original ground truth data due to subjectivity and annotation inconsistencies. To address this, they re-annotated their dataset with judgments from multiple experts, using multidimensional scaling to map frames to illumination-invariant 3D space, which they then fed into a pre-trained \textit{AlexNet} \cite{krizhevsky_sutskever_hinton_2012}.

Celona and Manoni \cite{celona_manoni_2017} investigated neonatal facial expressions to detect pain, achieving the highest accuracy when utilizing two pre-trained models: \textit{VGG-Face} \cite{parkhi_vedaldi_2015} and mapped LBP+CNN (MBPCNN) \cite{levi_hassner_2015}. Similarly, Lu and Hao \cite{lu_hao_2018} found that pre-trained models were crucial for small datasets like neonates, as training from scratch led to overfitting. They achieved optimal classification performance by fine-tuning the entire \textit{VGG-16} model \cite{he_zhang_2016}. However, Zamzmi \textit{et al.} \cite{zamzmi_paul_salekin_2019} argue that most face recognition methods are tailored for adults and thus less applicable to infants. They developed a lightweight 2D CNN trained end-to-end and achieved high pain detection accuracy, but external validation on a different neonatal dataset revealed challenges with generalizability. In 2019, Brahnam \textit{et al.} \cite{brahnam_nanni_2019} introduced the \textit{iCOPEvid} neonatal video dataset, a significant contribution since the only publicly available neonatal pain dataset \cite{brahnam_chuang_2006} previously contained only static images. Their experiments showed that local descriptors based on the bag-of-features (BoF) approach outperformed deep learning models like \textit{VGG-Face} and \textit{ResNet}. Combining handcrafted and deep-learned features offered only a marginal improvement in performance. In contrast, Zamzmi \textit{et al.} \cite{zamzmi_goldgof_2018} found that the most effective approach for binary classification (pain vs. no pain) was the fusion of high-level features from \textit{VGG} \cite{chatfield_simonyan_2014} and optical flow strains, with naive Bayes serving as the classifier. Celona and Brahnam \cite{celona_brahnam_2019} applied a Wasserstein generative adversarial network with gradient penalty (WGAN-GP) \cite {arjovsky_2017}, demonstrating that training set augmentation with synthetic samples improved classification performance.
Table \ref{table:camera_based_studies_non_temporal} summarizes the vision-based studies focusing exclusively on the spatial dimension.

\begin{landscape}
\renewcommand{\arraystretch}{1.5}
\begin{table}
\myfontsize
\begin{threeparttable}
\caption{Vision-based studies with static analysis.}
\label{table:camera_based_studies_non_temporal}
\centering
\begin{tabular}{ p{0.9cm} p{1.0cm} p{1.2cm} p{0.7cm} p{1.5cm} p{1.0cm} p{0.9cm} p{1.0cm} p{1.2cm} p{0.6cm} p{1.0cm} p{1.0cm} p{1.7cm} p{1.5cm}}
\hline
\toprule
\multirow{2}[3]{*}{Paper} &\multicolumn{3}{c}{Input}  &\multicolumn{4}{c}{Processing} & \multicolumn{6}{c}{Evaluation}\\
\cmidrule(lr){2-4} \cmidrule(lr){5-8} \cmidrule(lr){9-14}

\ & Modality  &Non deep \newline features &Fusion M/E &Deep model  &Non deep model &Learning Method &Classific. $/$Regres.  &Objective &GT &Number subjects &Validation Method &Dataset &Metrics\\
\hline\hline

\rowcolor{mygray} \ \textquotesingle 19 \cite{brahnam_nanni_2019} &F (RGB) &texture \newline descriptors &- FF &2D CNN$^{+}$ &SVM &SL &C &P &O &49 &k-fold &iCOPEvid &79.80 AUC\\

\ \textquotesingle 15 \cite{pedersen_2015} &F (RGB)  &- &- \space - &AE &SVM &SeSL, SL  &C &P &PS &25 &LOSO &UNBC &86.10 ACC, \newline 96.50 AUC\\

\rowcolor{mygray}\ \textquotesingle 20 \cite{egede_song_olugbade_2020} &F (RGB) &- &- FF &2D CNN$^{+}$ &NN &SL &R &IC &O &36 &hold-out &EmoPain &0.91 MAE$^\ddagger$\\

\ \textquotesingle 18 \cite{yang_hong_2018}  &F (RGB) &HOG, \newline statistics &- FF &2D CNN$^{+}$ &SVR &SL &R &IC &PS &25 &LOSO &UNBC &1.44 MSE$^\ddagger$\\

\rowcolor{mygray}\ \textquotesingle 21 \cite{semwal_londhe_2021} &F (RGB) &- &- DF  &2D CNN$^{+}$ &- &SL &C &ID &PS &25 &k-fold &UNBC 
&93.87 ACC$^\ddagger$\\

\ \textquotesingle 16 \cite{kharghanian_peiravi_2016} &F (RGB) &- &- \space - &CDBN  &SVM &UL 
&C &P &PS &25 &LOSO$^\dagger$ &UNBC &87.20 ACC$^\ddagger$\\

\rowcolor{mygray}\ \textquotesingle 21 \cite{kharghanian_peiravi_moradi_2021} &F (RGB) &- &- \space - &CDBN &SVM &SL &C &P &PS &25 &LOSO &UNBC &93.16 AUC\\

\ \textquotesingle 19 \cite{huang_xia_li_2019} &F (RGB) &- &- FF &2D CNN &- &SL &C &ID$^{1}$, IC &PS &25 &LOSO &UNBC 
&88.19$^{1}$ ACC\\

\rowcolor{mygray}\ \textquotesingle 20 \cite{xin_lin_2020} &F (RGB) &- &- \space -  &2D CNN &- &SL &C &ID &PS &25 &hold-out &UNBC 
&51.10 ACC$^\ddagger$\\

\ \textquotesingle 20 \cite{peng_huang_zhang_2020} &F (RGB) &- &- FF  &2D CNN$^{+}$ &- &SL &R &ID &S &25 &? &UNBC &79.94 ACC$^\ddagger$\\

\rowcolor{mygray}\ \textquotesingle 21 \cite{semwal_londhe_b_2021} &F (RGB) &- &- \space - &2D CNN &- &SL 
&C &ID &O &8 &k-fold &other &97.48 ACC$^\ddagger$\\

\ \textquotesingle 19 \cite{lee_wang_2019}  &F (RGB) &- &- \space -  &2D CNN &ELM &SL &R &IC &PS &25 &k-fold &UNBC$^\bullet$ &1.22 MSE$^\ddagger$\\

\rowcolor{mygray}\ \textquotesingle 19 \cite{virrey_caesarendra_2019} &F (RGB) &- &- \space - &2D CNN &- &SL &C &TR, CL, DI &PS &25 &k-fold &UNBC &60.00 ACC\\

\ \textquotesingle 19 \cite{menchetti_chen_2019} &F (RGB) &- &- \space - &2D CNN$^{+}$ &- &SL 
&C &AUs-D &PS & 25, 43 &k-fold &UNBC \& CK+$^{1}$, Wilkie &97.70$^{1}$ ACC$^\ddagger$\\

\rowcolor{mygray}\ \textquotesingle 17 \cite{liu_peng_2017} &F (RGB)  &statistics &- \space - &NN &GPM &WSL &R &IC &O, S &25 &k-fold &UNBC 
&2.18 MAE\\

\ \textquotesingle 20 \cite{xu_huang_sa_2020} &F (RGB) &statistics &- FF &2D CNN$^{+}$ &NN &SL &R &IC &S &25 &k-fold &UNBC
&1.95 MAE$^\ddagger$\\

\rowcolor{mygray}\ \textquotesingle 19 \cite{casti_mencattini_2019} &F (RGB) &LBP, MDS &- \space - &2D CNN$^{+}$ &- &SL &C &ID &O &25 &hold-out &UNBC &80.00 ACC\\

\ \textquotesingle 18 \cite{lu_hao_2018}  &F (RGB) &- &- \space - &2D CNN$^{+}$ &- &SL 
&C &ID &O &? &hold-out &other &78.30 ACC\\

\bottomrule
\end{tabular}
\begin{tablenotes}
\myfontsize

\item ${+}$: Pre-trained model -:Not exist 
\&: in Dataset indicates the utilization of cross-database training/validation 
?: Not found 
$\dagger$: The authors provide additional experiments with other validation methods
$\bullet$: The authors utilized occluded facial images  
$\ddagger$: The authors provide additional metrics  
\textbf{Modality:} F: face region
\textbf{Non deep features:} LBP: local binary pattern MDS: multidimensional scaling
\textbf{Fusion:} M: fusion of modalities E: fusion of deep learned features or hand-crafted features 
\textbf{Deep models:} AE: autoencoder RCNN: recurrent convolutional neural network CDBN: convolutional deep belief network CNN: convolutional neural network NN: neural network WGAN-GP: Wasserstein generative adversarial model with gradient penalty 
\textbf{Non deep model:} SVM: support vector machine GPM: Gaussian process regression model
kNN: k-nearest neighbors NB: naive Bayes ELM: extreme learning machine
\textbf{Learning Method:} SL: supervised learning SeSL: semi-supervised learning UL: unsupervised learning 
WSL: weakly supervised learning 
\textbf{Classific./Regres.:} C: classification R: regression
\textbf{Objective:} P: presence of pain ID: intensity in discrete scale IC: intensity in continuous scale TR: trigger CL: climax DI: diminishing AUs-D: Action Units detection   
\textbf{GT}: ground truth PS: Prkachin and Solomon S: self-report O: observer rating ST: stimulus 
\textbf{Validation Method:} LOSO: leave one subject out  
\textbf{Metrics:} AUC: Area Under the ROC Curve ACC: accuracy PPV: precision MSE: mean squared error 
MAE: mean absolute error  
   
\end{tablenotes}
\end{threeparttable}
\end{table}
\end{landscape}

\begin{landscape}
\renewcommand{\arraystretch}{1.5}
\begin{table}
\myfontsize
\begin{threeparttable}
\ContinuedFloat

\caption{Vision-based studies with static analysis (continued).}
\label{table:camera_based_studies_non_temporal_2}
\centering
\begin{tabular}{ p{0.9cm} p{1.1cm} p{1.8cm} p{0.7cm} p{1.7cm} p{1.0cm} p{0.9cm} p{1.0cm} p{1.2cm} p{0.6cm} p{1.0cm} p{0.9cm} p{1.3cm} p{1.6cm}}
\hline
\toprule
\multirow{2}[3]{*}{Paper} &\multicolumn{3}{c}{Input}  &\multicolumn{4}{c}{Processing} & \multicolumn{6}{c}{Evaluation}\\
\cmidrule(lr){2-4} \cmidrule(lr){5-8} \cmidrule(lr){9-14}

\ & Modality  &Non deep \newline features &Fusion M/E  &Deep model  &Non deep model &Learning Method &Classific. $/$Regres.  &Objective &GT &Number subjects &Validation Method &Dataset &Metrics\\
\hline\hline

\rowcolor{mygray}\ \textquotesingle 21 \cite{semwal_londhe_august_2021} &F (RGB) &facial landmarks &- FF  &2D CNN &NN &SL &C, R &ID, IC$^{1}$ &P &25 &LOSO$^\dagger$ &UNBC &0.17$^{1}$ MSE$^\ddagger$\\

\ \textquotesingle 20 \cite{lakshminarayan_hinduja_2020} &F (RGB) &HOG, head pose, AUs intensity/ occurrence, facial landmarks  &FF - &2D CNN$^{+}$ &NN &SL &R &IC &O &36 &hold-out &EmoPain &5.48 RMSE$^\ddagger$\\

\rowcolor{mygray}\ \textquotesingle 20 \cite{huynh_yang_2020} &F (RGB) &- &- \space - &2D CNN$^{+}$ &NN &SL &R &IC &O &36 &hold-out &EmoPain &1.49 RMSE$^\ddagger$\\

\ \textquotesingle 18 \cite{semwal_londhe_2018} &F (RGB) &- &- \space - &2D CNN &- &SL 
&C &ID &PS &25 &hold-out &UNBC &92.00 ACC$^\ddagger$\\

\rowcolor{mygray}\ \textquotesingle 18 \cite{tavakolian_hadid_2018_b} &F (RGB) &statistics, distance metrics &-\,FF 
&2D CNN$^{+}$ &- &SL &C, R &ID, IC &PS &25 &LOSO &UNBC &0.81 PCC, \newline 0.69 MSE\\

\ \textquotesingle 21 \cite{cui_huang_2021} &F (RGB) &- &- FF &2D CNN$^{+}$ &- &SL &C, R &ID, IC &P &25 &LOSO &UNBC &91.13 ACC, \newline 0.78 PCC, \newline 0.46 MSE\\

\rowcolor{mygray}\ \textquotesingle 18 \cite{li_zhu_2018} &F (RGB) &- &- \space - &AE$^{+}$ &- &SL 
&R &IC &PS &25 &k-fold &UNBC &0.33 MAE$^\ddagger$\\

\ \textquotesingle 21 \cite{xin_li_yang_2021} &F (RGB) &- &- FF  &[AE, 2D CNN]$^\cup$ &- &SL &C, R &ID$^{1}$, IC$^{2}$, P$^{3}$ &P, ST &25, 87 &LOSO &UNBC$^{1}$, BioVid (A)$^{2}$ &89.17$^{11}$ ACC, \newline 0.81$^{21}$ PCC, \newline 85.65$^{32}$ ACC, \newline 40.40$^{12}$ ACC\\

\rowcolor{mygray}\ \textquotesingle 21 \cite{semwal_londhe_c_2021} &F (RGB) &entropy texture \newline descriptors &- \space - &2D CNN$^{+}$ &- &SL &C &ID &O &8  &k-fold &other &0.92 PPV$^\ddagger$\\

\ \textquotesingle 18 \cite{nugroho_harmanto_2018} &F (RGB) &- &- \space - &2D CNN$^{+}$ &- &SL 
&C &P &PS &14 &k-fold &UNBC &93.00 ACC\\

\rowcolor{mygray}\ \textquotesingle 19 \cite{dai_broekens_2019} &F (RGB) &- &- \space - &2D CNN &- &SL 
&C &P &PS &25, 20 &k-fold &UNBC \& BioVid (A)$^\diamond$ &56.75 ACC\\

\ \textquotesingle 17 \cite{celona_manoni_2017} &F (RGB) &HOG, LBP &- FF  &2D CNN$^{+}$ &SVM &SL &C &P &O &26 &LOSO &iCOPE &73.78 ACC\\

\rowcolor{mygray}\ \textquotesingle 19 \cite{zamzmi_paul_salekin_2019} &F (RGB) &- &- \space - &2D CNN &- &SL &C &P &O &31 &LOSO 
&NPAD$^{1}$, iCOPE$^{2}$ &96.98$^{1}$ ACC$^\ddagger$, \newline 89.80$^{2}$ ACC\\

\ \textquotesingle 21 \cite{rudovic_tobis_2021} &F (RGB) &- &- \space -  &2D CNN$^{+}$ &- &FL 
&C &P &PS &25 &LOSO &UNBC &76.00 ACC$^\ddagger$\\

\rowcolor{mygray}\ \textquotesingle 21 \cite{pikulkaew_boonchieng_2021} &F (RGB) &- &- \space -  &2D CNN$^{+}$ &- &SL &C &P &O &25 
&hold-out &UNBC &75.49 ACC\\

\ \textquotesingle 21 \cite{elmorabit_rivenq_2021} &F (RGB) &- &- \space - &2D CNN$^{+}$ &SVR &SL 
&R &IC &P &25 &LOSO &UNBC &0.34 MSE\\

\rowcolor{mygray}\ \textquotesingle 21 \cite{li_pourtaheian_2021} &F (RGB) &- &- \space - &2D R-CNN &- &SL &C &P &O &? &hold-out &other &87.80 PPV\\

\bottomrule
\end{tabular}
 \begin{tablenotes}
\myfontsize
\item $\cup$: The authors combined the deep models into a unified framework $\diamond$: The authors experimented with additional datasets combinations
\textbf{Non deep features:} AUs: actions units HOG: histogram of oriented gradients
\textbf{Non deep model:} SVR: support vector regression 
\textbf{Learning Method:} FL: federated learning
\textbf{Metrics:} RMSE: root mean squared error      
  \end{tablenotes}
\end{threeparttable}
\end{table}
\end{landscape}

\begin{landscape}
\renewcommand{\arraystretch}{1.5}
\begin{table}
\myfontsize
\begin{threeparttable}
\ContinuedFloat

\caption{Vision-based studies with static analysis (continued).}
\label{table:camera_based_studies_non_temporal_3}
\centering
\begin{tabular}{ p{0.9cm} p{1.1cm} p{1.5cm} p{0.7cm} p{1.7cm} p{1.0cm} p{0.9cm} p{1.0cm} p{1.2cm} p{0.6cm} p{1.0cm} p{0.9cm} p{1.3cm} p{1.6cm}}
\hline
\toprule
\multirow{2}[3]{*}{Paper} &\multicolumn{3}{c}{Input}  &\multicolumn{4}{c}{Processing} & \multicolumn{6}{c}{Evaluation}\\
\cmidrule(lr){2-4} \cmidrule(lr){5-8} \cmidrule(lr){9-14}

\ & Modality  &Non deep \newline features &Fusion M/E  &Deep model  &Non deep model &Learning Method &Classific. $/$Regres.  &Objective &GT &Number subjects &Validation Method &Dataset &Metrics\\
\hline\hline

\rowcolor{mygray}\ \textquotesingle 18 \cite{zamzmi_goldgof_2018} &F (RGB) &optical flow &- FF &2D CNN$^{+}$ &SVM, kNN, NB &SL &C &P &O &31 &k-fold &other &92.71 ACC, \newline 94.80 AUR\\

\ \textquotesingle 19 \cite{celona_brahnam_2019} &F (RGB) &- &- \space - &WGAN-GP &- &SL 
&C &P &O &26 &LOSO &iCOPE &93.38 ACC\\

\rowcolor{mygray}\ \textquotesingle 17 \cite{wang_xiang_2017} &F (RGB)  &- &- \space    - &2D CNN$^{+}$ &- &SL &R &IC &PS &25 &LOSO &UNBC &0.99 MAE$^\ddagger$\\

\ \textquotesingle 20 \cite{dragomir_florea_2020} &F (RGB) &- &- \space    - &2D CNN &- &SL 
&C &ID &ST &87 &hold-out &BioVid (A) &36.60 ACC\\

\rowcolor{mygray}\ \textquotesingle 20 \cite{semwal_londhe_2020} &F (RGB) &- &- \space - &2D CNN &- &SL 
&C &P &PS &25 &hold-out &UNBC &97.00 PPV$^\ddagger$\\

\ \textquotesingle 21 \cite{rathee_pahal_2021} &F (RGB) &- &- \space - &2D CNN &- &SL &C &ID &P &28 &LOSO$^\dagger$ &UNBC &90.30 ACC\\

\rowcolor{mygray}\ \textquotesingle 19 \cite{zamzmi_paul_2019} &F (RGB) &- &- \space - &2D CNN &- &SL &C &P &O &31 &hold-out
&NPAD$^{1}$, iCOPE$^{2}$ &91.00$^{1}$ ACC$^\ddagger$, \newline 84.50$^{2}$ ACC$^\ddagger$\\

\ \textquotesingle 21 \cite{carlini_ferreira_2021} &F (RGB) &- &- \space - &2D CNN$^{+}$ &- &SL 
&C &P &O &26, 30 &hold-out &iCOPE \& UNIFESP &89.90 ACC$^\ddagger$\\

\rowcolor{mygray}\ \textquotesingle 21 \cite{nerella_cupka_2021} &F (RGB) &- &- \space - &2D CNN &- &SL 
&C &AUs-D &P &10 &hold-out &Pain-ICU &77.00 ACC$^\ddagger$\\

\bottomrule
\end{tabular}
\end{threeparttable}
\end{table}
\end{landscape}

\subsection{Vision-based: Temporal Utilization (Non-ML Approach)}
\label{sec:temporal_non_learned}
Pain assessment is particularly challenging due to its complex and dynamic nature. 
Relying on static, individual frames to assess pain fails to capture the phenomenon's temporal progression and often leads to inaccurate estimations. Additionally, many studies highlight the difficulties of applying deep learning techniques to small datasets, with one proposed solution being the combination of deep learning and traditional feature extraction methods.
Egede \textit{et al.} \cite{egede_valstar_2017_b} addressed this by extracting deep features from a pre-trained CNN, explicitly targeting the eyes and mouth regions. Using a relevance vector regressor (RVR), they demonstrated that combining deep and hand-crafted features led to optimal performance. Despite the valuable insights the \textit{UNBC-McMaster} database provides, its imbalanced sample distribution---particularly the limited number of frames showing pain---poses a significant challenge for deep learning models. In response, Egede and Valstar \cite{egede_valstar_2017} devised a method based on the observation that neighboring pain level classes share many common features. This approach allowed them to avoid extracting all possible features for classes with fewer samples, as certain features had already been utilized from other related classes. The study also showed that combining deep and hand-crafted features improved performance. However, in a later study \cite{jaiswal_egede_2018}, the authors applied a similar approach, using only deep-learned features to address data imbalance, but could not replicate the same high-performance levels.

Tavakolian \textit{et al.} \cite{tavakolian_cruces_2019} took a different approach, focusing on the detection of genuine versus acted pain through facial expressions, a technique with important applications in both medical and forensic contexts. They developed a residual GAN (R-GAN) to capture subtle facial changes and the dynamic nature of expressions, using a weighted spatio-temporal pooling (WSP) method. In a subsequent study \cite{tavakolian_bordallo_liu_2020}, the authors suggested that self-supervised learning could reduce the time and effort needed for data labeling, as it does not require complete dataset annotation. They introduced a new similarity function for learning generalized representations with a Siamese network. They also employed statistical spatio-temporal distillation (SSD) based on the Gaussian scale mixture (GSM) to improve computational efficiency. This technique encodes spatiotemporal variations in facial videos into a single RGB image, simplifying the model while maintaining effectiveness.

Other studies also aim to capture the dynamic aspects of pain. For instance, \cite{othman_werner_saxen_2021} combined a random forest classifier with the pre-trained \textit{MobileNetV2} model \cite{sandler_howard_2018}, encoding videos by selecting and merging three frames from different time points into a single image. Othman \textit{et al.} \cite{othman_werner_2019} emphasized the importance of using diverse datasets---including varying age, gender, pose, occlusion, and lighting conditions---to improve model generalization. They used multiple data combinations and a reduced version of \textit{MobileNetV2}, showing that cross-dataset training is essential for achieving better generalizability.

\subsection{Vision-based: Implicit Temporal Utilization}
\label{sec:temporal_implicit}
Several studies have explored the application of 3D CNNs for pain assessment. Tavakolian and Hadid \cite{tavakolian_hadid_2018} developed a 3D CNN to capture dynamic facial representations from videos. They noted that researchers often use fixed temporal kernel depths when employing 3D convolution techniques, which limits the ability to capture short, mid, and long temporal ranges simultaneously. To address this, they designed a model with parallel 3D convolutional layers featuring variable temporal depths, allowing the capture of temporal dependencies from $32$ consecutive frames.
Similarly, Wang and Sun \cite{wang_sun_2018} applied 3D convolutions based on the architecture proposed in \cite{tran_bourdev_2015}, consisting of $8$ convolutional layers with $3\times3\times3$ filters. While they reported high performance, the authors acknowledged that extracting deep features from small datasets posed a challenge for model generalization. In a related study, Huang \textit{et al.} \cite{huang_qing_xu_2021} developed a framework that integrated 3D, 2D, and 1D CNNs to extract spatio-temporal, spatial, and geometric features. For the 3D CNN component, they modified the architecture from \cite{xie_sun_2018} by using discrete kernels of $1\times3\times3$ and $3\times1\times1$ rather than the traditional $3\times3\times3$ kernel.
Other researchers have also proposed 3D deep CNNs with varying temporal depths to capture short, mid, and long-range facial expression variations \cite{tavakolian_hadid_2019}. Recognizing the difficulty and time consumption involved in training a deep 3D CNN from scratch, they introduced a cross-architecture knowledge transfer learning technique, utilizing a pre-trained 2D CNN to assist in the training of the 3D CNN. In studies by Praveen \textit{et al.} \cite{praveen_granger_cardinal_2020} and \cite{gnana_praveen_granger_2020}, the authors employed weakly-supervised domain adaptation, where the source domain focused on human affective expressions and the target domain was explicitly related to pain expressions. Their framework featured an inflated 3D-CNN (I3D) \cite{carreira_zisserman_2018}, incorporating $3$ convolutional layers and $3$ inception modules \cite{szegedy_2017} to capture both spatial and temporal information from video data.

Bargshady \textit{et al.} \cite{bargshady_zhou_deo_2020} opted to use the HSV color space instead of RGB, arguing that it better reflects human visual perception for tasks such as skin pixel detection and multi-face detection. They employed the pre-trained \textit{VGG-Face} \cite{parkhi_vedaldi_2015} for feature extraction, followed by a temporal convolutional network (TCN) using dilated causal convolutional operations to leverage temporal dependencies. Rezaei \textit{et al.} \cite{rezaei_moturu_2020} tackled the challenge of pain detection in people with dementia, a difficult task due to insufficient pain-related images or videos of elderly subjects in existing datasets. They developed a $10$-layer 2D CNN that processed pairs of pain and no-pain images, analyzing frame-to-frame changes and employing contrastive training methods \cite{hinton_2002}. The model demonstrated high performance in both healthy individuals and people with dementia.
In another study, Pandit and Schmitt \cite{pandit_schmitt_2020} explored the potential of using shallow 1D CNN architectures for real-time pain recognition. They extracted facial action units from each frame using the \textit{OpenFace 2.0}\footnote{\url{https://github.com/TadasBaltrusaitis/OpenFace}} toolkit, with promising results for pain detection in real-time settings.

\subsection{Vision-based: Explicit Temporal Utilization}
\label{sec:temporal_explicit}
Several efforts have focused on addressing the limitations of static frames by developing dedicated temporal modules. Zhou \textit{et al.} \cite{zhou_hong_2016} tackled this issue using a regression framework based on a $4$-layer recurrent convolutional neural network (RCNN), each with a sequence length of $3$ time steps. Rodriguez \textit{et al.} \cite{rodriquez_cucurull_2017} leveraged dynamic information by designing an LSTM model fed with feature vectors extracted from \textit{VGG-16} \cite{he_zhang_2016}. Similarly, Bellantonio \textit{et al.} \cite{bellantonio_hague_2017} emphasized that facial expressions evolve, making it essential to analyze the spatio-temporal dimension of pain. They improved estimation performance using a fine-tuned 16-layer CNN model \cite{parkhi_vedaldi_2015}, an LSTM processing $16$ frames as a time window, and super-resolution techniques.
In another study, Bargshady \textit{et al.} \cite{bargshady_soar_zhou_2019} combined the \textit{VGG-Face} CNN \cite{parkhi_vedaldi_2015} with a $3$-layer LSTM to extract spatio-temporal features from grayscale images, applying zero-phase component analysis (ZCA). In \cite{bargshady_zhu_deo_2020_c}, principal component analysis (PCA) was used to reduce dimensionality. Mauricio \textit{et al.} \cite{mauricio_cappabianco_2019} also employed \textit{VGG-Face} but replaced LSTM with a $2$-layer gated recurrent unit (GRU) to capture temporal dependencies.
Thuseethan \textit{et al.} \cite{thuseethan_rajasegarar_2019} used a conventional 2D CNN and two RCNNs to extract temporal features from previous and subsequent frames, enhancing the time dimension of expression analysis. 

A similar approach was followed by Bargshady \textit{et al.} \cite{bargshady_zhou_deo_2020_b}, who employed ensemble learning with three distinct CNN-biLSTM modules, merging their outputs for the final prediction. Salekin \textit{et al.} \cite{salekin_zamzmi_goldgof_2020} used a bilinear CNN (B-CNN) based on the \textit{VGG} architecture \cite{simonyan_karen_2015}, pre-trained on \textit{VGGFace2}\footnote{\url{https://www.robots.ox.ac.uk/~vgg/data/vgg_face}} and \textit{ImageNet}\footnote{\url{https://www.image-net.org}} datasets, along with an LSTM to capture temporal dependencies in image sequences.
Kalischek \textit{et al.} \cite{kalischek_thiam_2019} explored deep domain adaptation for facial expression and pain detection, utilizing the self-ensembling approach \cite{french_mackiewicz_2028} with a long-term recurrent convolutional network (LRCN). While they achieved state-of-the-art results for facial expression recognition, performance was lower for pain detection, likely due to the subtle nature of pain-related expressions.

Despite the availability of additional information in pain datasets, multi-task approaches remain limited. Martinez \textit{et al.} \cite{martinez_rudovic_2017} proposed a personalized multi-task learning method based on individual physiological and behavioral pain responses. They extracted AAM facial landmarks, processed them through a biLSTM to produce PSPI scores, and predicted the final VAS score. Erekat \textit{et al.} \cite{erekat_hammal_2020} combined \textit{AlexNet} \cite{krizhevsky_sutskever_hinton_2012} with 2 GRU layers to capture temporal dependencies, using both self and observer-reported pain intensity as ground truth. Vu \textit{et al.} \cite{vu_aimer_2021} developed a multi-task framework to estimate pain levels while reconstructing heatmaps of action unit locations, improving model generalization with a CNN-LSTM combination to capture micro facial movements.

Huang \textit{et al.} \cite{huang_xia_2020} noted that specific frames within a video sequence exhibit more pronounced pain expressions, requiring special handling. They developed a novel framework using attention saliency maps with a \textit{VGG-16} model, GRUs and learned weights for each frame's contribution to pain intensity estimation. The study demonstrated that dynamic and salient features can significantly improve performance. Similarly, Yu \textit{et al.} \cite{yu_kurihara_2019} used \textit{VGG-11 (configuration A)} and an LSTM to create an attention mechanism, predicting pain intensity from $16$ consecutive frames. Xu and Liu \cite{hu_liu_2021} adopted a \textit{ResNet-50} model with an attention mechanism to extract spatial features, followed by a transformer encoder to capture temporal sequences, achieving promising results.

In other studies, Ragolta \textit{et al.} \cite{ragolta_liu_2020} used extracted action units to train a $2$-layer LSTM predicting pain on an 11-point scale, employing curriculum learning. Guo \textit{et al.} \cite{guo_wang_2021} developed a convolutional LSTM (C-LSTM) to extract both spatial and temporal features from videos, showing that temporal models outperform non-temporal models for pain estimation accuracy. Rasipuram \textit{et al.} \cite{rasipuram_sai_2020} utilized in-the-wild video data for pain detection, generating a 3D morphable model without relying on facial landmarks and combining it with an LSTM.
Zhi and Wan \cite{zhi_wan_2019} introduced sparse coding with LSTM (SLTM), using the iterative hard thresholding algorithm (ISTA) \cite{blumensast_davies_2008} to capture dynamic facial expressions. Although SLTM did not achieve high performance, it offers speed and efficiency for specific applications. Finally, Thiam \textit{et al.} \cite{thiam_kestler_schenker_2020} developed a method combining motion history and optical flow images with a $10$-layer CNN and $2$-layer biLSTM, showing that weighted score aggregation improves performance. 
Table \ref{table:camera_based_studies_temporal} summarizes studies incorporating the modalities' temporal dimensions.

\begin{landscape}
\renewcommand{\arraystretch}{1.5}
\begin{table}
\myfontsize
\begin{threeparttable}
\caption{Vision-based studies with temporal utilization.}
\label{table:camera_based_studies_temporal}
\centering
\begin{tabular}{ p{1.0cm} p{1.0cm} p{1.0cm} p{0.6cm} p{1.1cm} p{2.2cm} p{1.0cm} p{0.7cm} p{0.9cm} p{1.0cm} p{0.4cm} p{0.8cm} p{0.9cm} p{1.3cm} p{1.3cm}}
\hline
\toprule
\multirow{2}[3]{*}{Paper} &\multicolumn{4}{c}{Input}  &\multicolumn{4}{c}{Processing} & \multicolumn{6}{c}{Evaluation}\\
\cmidrule(lr){2-5} \cmidrule(lr){6-9} \cmidrule(lr){10-15}

\ & Modality  &Non deep \newline features &Fusion M/E &Temporal \newline Exploitation &Deep model  &Non deep \newline model &Learning Method &Classific. $/$Regres.  &Objective &GT &Number subjects &Validation Method &Dataset &Metrics\\
\hline\hline

\rowcolor{mygray}\ \textquotesingle 17 \cite{egede_valstar_2017_b} &F (RGB) &HOG, \newline distance metrics &- \,DF &NL &2D CNN$^{+}$ &RVR &SL &R &IC &PS &25 &LOSO &UNBC &0.99 RMSE, \newline 0.67 PCC\\

\ \textquotesingle 17 \cite{egede_valstar_2017} &F (RGB) &HOG, \newline distance metrics  &- \,DF &NL &2D CNN$^{+}$ &RVM &SL &R &IC &PS &25 &LOSO &UNBC &1.04 RMSE, \newline 0.64 PCC\\

\rowcolor{mygray}\ '18 \cite{jaiswal_egede_2018} &F (RGB) &- &- \space - &NL &2D CNN &- &SL &R &IC &PS &25 &LOSO &UNBC &1.20 RMSE, \newline 0.47 PCC\\

\ \textquotesingle 18 \cite{tavakolian_hadid_2018} &F (RGB) &- &- \space - &I &3D CNN  &- &SL 
&R &IC &PS &25 &LOSO &UNBC &0.53 MSE, \newline 0.84 PCC$^\ddagger$\\

\rowcolor{mygray}\ \textquotesingle 18 \cite{wang_sun_2018} &F (RGB) &HOG, \newline geometric difference &- \,DF &I &3D CNN &SVR &SL &R &IC &PS &25 &LOSO &UNBC &0.94 RMSE, \newline 0.67 PCC\\

\ \textquotesingle 20 \cite{gnana_praveen_granger_2020} &F (RGB) &- &- \space - &I &3D CNN$^{+}$ &- &WSL &R &IC &PS &24, ? &LOSO &UNBC \newline \& RECOLA &0.64 MAE, \newline 0.82 PCC$^\ddagger$\\

\rowcolor{mygray}\ \textquotesingle 16 \cite{zhou_hong_2016} &F (RGB) &- &- \,FF &E  &RCNN  &- &SL &R &IC &PS &25 &LOSO &UNBC &1.54 MSE, \newline 0.65 PCC\\

\ \textquotesingle 17 \cite{rodriquez_cucurull_2017} &F (RGB) &- &- \,FF &E &[2D CNN$^{+}$, LSTM]$^\cup$ &- &SL &C, R &P, IC$^{1}$ &PS &25 &LOSO &UNBC &0.74$^{1}$ MSE, \newline 0.78$^{1}$ PCC$^\ddagger$\\

\rowcolor{mygray}\  \textquotesingle 17 \cite{bellantonio_hague_2017} &F (RGB) &- &- FF &E &[2D CNN$^{+}$, LSTM]$^\cup$ &- &SL &C &ID &PS &25 &LOSO &UNBC &61.90 ACC\\

\ \textquotesingle 19 \cite{bargshady_soar_zhou_2019} &F (RGB) &- &- FF &E &[2D CNN$^{+}$, LSTM]$^\cup$ &- &SL &C &ID &PS &25 &LOSO &UNBC &75.20 ACC\\

\rowcolor{mygray}\ \textquotesingle 20 \cite{bargshady_zhu_deo_2020_c} &F (RGB) &PCA &- DF &E &[2D CNN$^{+}$, \newline 1D CNN, biLSTM]$^\cup$ &- &SL  &C &ID &PS &25 &LOSO$^\dagger$ &UNBC &85.00 ACC$^\ddagger$\\

\ \textquotesingle 19 \cite{mauricio_cappabianco_2019} &F (RGB) &- &- \space - &E &[2D CNN$^{+}$, GRU]$^\cup$ &- &SL &C &ID, IC &PS &25 &LOSO &UNBC &85.40 ACC, \newline 0.62 MSE$^\ddagger$\\

\rowcolor{mygray}\ \textquotesingle 19 \cite{thuseethan_rajasegarar_2019} &F (RGB) &- &- FF &E &[2D CNN, RCNN]$^\cup$ &- &SL &R &IC &PS &25 &LOSO &UNBC &1.29 MSE, \newline 0.73 PCC\\

\ \textquotesingle 17 \cite{martinez_rudovic_2017} &F (RGB) &- &- FF &E &biLSTM &HCRF, FC &SL &C &IC &O, S &25 &hold-out &UNBC &2.46 MAE$^\ddagger$\\

\bottomrule
\end{tabular}
\begin{tablenotes}
\myfontsize
\item 
\textbf{Non deep features:} PCA: principal component analysis  
\textbf{Temporal Exploitation:} NL: non-machine learning method I: implicit method E: explicit method
\textbf{Deep models:} RCNN: recurrent convolutional neural network LSTM: long short memory networks biLSTM: bidirectional neural network GRU: gated recurrent unit \textbf{Non deep models:} SVM: support vector machine RVM: relevance vector machine GPM: Gaussian process regression model HCRF: hidden conditional random fields FC: fully connected SVR: support vector regression \textbf{Objective:} I2: intensity in binary pairs
\textbf{Metrics:} PCC: Pearson correlation coefficient   
\end{tablenotes}
\end{threeparttable}
\end{table}
\end{landscape}

\begin{landscape}
\renewcommand{\arraystretch}{1.5}
\begin{table}
\myfontsize
\begin{threeparttable}
\ContinuedFloat
\caption{Vision-based studies with temporal utilization (continued).}
\label{table:camera_based_studies_temporal_2}
\centering
\begin{tabular}{ p{0.9cm} p{1.0cm} p{1.0cm} p{0.5cm} p{1.0cm} p{2.2cm} p{1.0cm} p{0.7cm} p{0.9cm} p{1.0cm} p{0.4cm} p{0.8cm} p{0.9cm} p{1.5cm} p{1.3cm}}
\hline
\toprule
\multirow{2}[3]{*}{Paper} &\multicolumn{4}{c}{Input}  &\multicolumn{4}{c}{Processing} & \multicolumn{6}{c}{Evaluation}\\
\cmidrule(lr){2-5} \cmidrule(lr){6-9} \cmidrule(lr){10-15}

\ & Modality  &Non deep \newline features &Fusion M/E &Temporal \newline Exploitation &Deep model  &Non deep \newline model &Learning Method &Classific. $/$Regres.  &Objective &GT &Number subjects &Validation Method &Dataset &Metrics\\
\hline\hline

\rowcolor{mygray}\ \textquotesingle 19 \cite{egede_valstar_2019_a} &F (RGB) &HOG, \newline distance metrics &- \,DF &NL &2D CNN & RVR &SL &R &IC &O &13 &LOSO &APN-DB &1.71 MAE$^\ddagger$\\ 

\ \textquotesingle 19 \cite{tavakolian_cruces_2019} &F (RGB) &- &- \space - &NL &R-GAN &- &UL &C &genuine \newline vs posed &PS, ST &25, \newline 34, \newline 87, \newline 87 &? &UNBC \newline \& STOIC \newline \& BioVid (A) \newline \& BioVid (D) &90.97 ACC\\

\rowcolor{mygray}\ \textquotesingle 20 \cite{tavakolian_bordallo_liu_2020} &F (RGB) &- &- FF &NL  &2D CNN$^{+}$ &- &SSL &C &IC &P, ST &25 \newline 87 &LOSO &UNBC$^{1}$, \newline BioVid (A)$^{2}$ $^\oplus$ &0.78$^{1}$ PCC$^\ddagger$, \newline $71.02^{2}$AUC$^\ddagger$\\

\ \textquotesingle 21 \cite{othman_werner_saxen_2021} &F (RGB) &AUs \newline intensity &- H &NL &2D CNN$^{+}$ &RF &SL &C &ID &ST &127 &k-fold &X-ITE &25.00 ACC\\

\rowcolor{mygray}\ \textquotesingle 19 \cite{othman_werner_2019}  &F (RGB) &- &- \space - &NL &2D CNN &- &SL &C &P &ST &87 \newline 134 &k-fold 
&BioVid (A) \newline \& X-ITE$^\oplus$ &67.90 ACC\\

\ \textquotesingle 20 \cite{erekat_hammal_2020} &F (RGB) &- &- FF &E &[2D CNN$^{+}$, GRU]$^\cup$ &- &SL 
&R &IC &O, S &25 &k-fold &UNBC &2.34 MAE\\

\rowcolor{mygray}\ \textquotesingle 20 \cite{huang_xia_2020} &F (RGB) &- &- FF &E &[2D CNN$^{+}$, GRU]$^\cup$ &- &SL &R &IC &PS &19 &LOSO &UNBC &0.21 MSE, \newline 0.89 PCC\\

\ \textquotesingle 19 \cite{yu_kurihara_2019} &F (RGB) &- &- FF &E &[2D CNN, LSTM]$^\cup$ &- &SL &R &IC &PS &24 &LOSO &UNBC &1.22 MSE$^\ddagger$, \newline 0.40 PCC$^\ddagger$\\

\rowcolor{mygray}\ \textquotesingle 20 \cite{ragolta_liu_2020} &F (RGB) &AUs \newline intensity &- \space - &E &LSTM &- &SL &R &IC &O &36 &hold-out &EmoPain &2.12 RMSE, \newline 1.60 MAE$^\ddagger$\\

\ \textquotesingle 20 \cite{rasipuram_sai_2020} &F (RGB) &- &- FF &E &[2D CNN$^{+}$, LSTM]$^\cup$ &- &SL &C &P &O &? &k-fold &UNBC &78.20 ACC$^\ddagger$\\

\rowcolor{mygray}\ \textquotesingle 20 \cite{thiam_kestler_schenker_2020} &F (RGB) &- &- DF &E &[2D CNN, biLSTM, NN]$^\cup$ &- &SL &C &P &ST &87 \newline 40 &LOSO &BioVid (A)$^{1}$, \newline SenseEmotion$^{2}$ &69.25$^{1}$ ACC, \newline 64.35$^{2}$ ACC\\

\ \textquotesingle 20 \cite{mauricio_pena_2020} &F (RGB) &- &- FF &E &[2D CNN$^{+}$, GRU]$^\cup$ &- &SL &C &ID, IC &PS &25 &LOSO &UNBC &0.84 ACC, \newline 0.69 PCC$^\ddagger$\\

\bottomrule
\end{tabular}

\begin{tablenotes}
\myfontsize
\item $\oplus$: The authors provide experiments with cross-dataset settings  
\textbf{Fusion:} H: hybrid  
\textbf{Non deep models:} RF: random forest classifier  
\end{tablenotes}
\end{threeparttable}
\end{table}
\end{landscape}

\begin{landscape}
\renewcommand{\arraystretch}{1.5}
\begin{table}
\myfontsize
\begin{threeparttable}
\ContinuedFloat
\caption{Vision-based studies with temporal utilization (continued).}
\label{table:camera_based_studies_temporal_3}
\centering
\begin{tabular}{ p{0.9cm} p{0.9cm} p{1.0cm} p{0.6cm} p{1.1cm} p{1.8cm} p{1.0cm} p{0.7cm} p{0.9cm} p{1.0cm} p{0.4cm} p{0.8cm} p{0.8cm} p{1.8cm} p{1.5cm}}
\hline
\toprule
\multirow{2}[3]{*}{Paper} &\multicolumn{4}{c}{Input}  &\multicolumn{4}{c}{Processing} & \multicolumn{6}{c}{Evaluation}\\
\cmidrule(lr){2-5} \cmidrule(lr){6-9} \cmidrule(lr){10-15}

\ & Modality  &Non deep \newline features &Fusion M/E &Temporal \newline Exploitation &Deep model  &Non deep \newline model &Learning Method &Classific. $/$Regres.  &Objective &GT &Number subjects &Validation Method &Dataset &Metrics\\
\hline\hline

\rowcolor{mygray}\ \textquotesingle 21 \cite{huang_qing_xu_2021} &F (RGB) &facial \newline landmarks &- DF &I &[3D CNN$^{+}$, \newline 2D CNN$^{+}$, \newline 1D CNN, FC]$^\cup$ &- &SL &R &IC &PS &25 &LOSO &UNBC &0.76 MSE, \newline 0.82 PCC$^\ddagger$\\

\ \textquotesingle 19 \cite{tavakolian_hadid_2019} &F (RGB) &- &- \space - &I &[2D CNN$^{+}$, \newline 3D CNN]$^\cup$ &- &UL, SL &C, R &IC$^{1}$, P$^{2}$ &P, ST &25, 87 &LOSO &UNBC$^{1}$, \newline BioVid (A)$^{2}$ &0.92$^{11}$ PCC$^\ddagger$, \newline 86.02$^{22}$ AUC\\

\rowcolor{mygray}\ \textquotesingle 20 \cite{praveen_granger_cardinal_2020} &F (RGB) &- &- \space - &I &3D CNN$^{+}$ &- &WSL &R &IC &PS &24,?, 87, 18 &LOSO &UNBC$^{1}$ \newline \& RECOLA \newline \& BioVid (A)$^{2}$ $^\oplus$ &0.74$^{1}$ PCC, \newline 0.34$^{2}$ PCC\\

\ \textquotesingle 20 \cite{bargshady_zhou_deo_2020} &F (RGB) &PCA &- FF &I &[2D CNN$^{+}$, \newline TCN]$^\cup$ &- &SL &C &ID &P, ST &25, 20 &LOSO$^\dagger$ &UNBC$^{1}$, \newline MIntPAIN$^{2}$ &92.44$^{1}$ ACC$^\ddagger$, \newline 89.00$^{2}$ ACC$^\ddagger$\\

\rowcolor{mygray}\ \textquotesingle 20 \cite{rezaei_moturu_2020}&F (RGB) &- &- \space - &I &2D CNN &- &SL &C, R &IC, P$^{1}$ &P &95, 25 &k-fold &UofR \& UNBC$^{1}$ &82.00$^{11}$ PCC$^\ddagger$\\

\ \textquotesingle 20 \cite{pandit_schmitt_2020} &F (RGB) &AUs \newline occurrence &- FF &I &1D CNN &- &SL &R &IC &P &24, 87 &hold-out &UNBC$^{1}$, \newline BioVid (A) &0.80$^{1}$ CCC\\

\rowcolor{mygray}\ \textquotesingle 20 \cite{bargshady_zhou_deo_2020_b} &F (RGB) &PCA &- \,DF &E &[2D CNN$^{+}$, 1D CNN, biLSTM]$^\cup$ &- &SL &C &ID &PS, ST &25, 20 &k-fold &UNBC$^{1}$, \newline MIntPAIN$^{2}$ &86.00$^{1}$ ACC$^\ddagger$  \newline 92.26$^{2}$ ACC$^\ddagger$\\

\ \textquotesingle 20 \cite{salekin_zamzmi_goldgof_2020} &F (RGB) &- &- FF &E &[2D CNN$^{+}$, \newline LSTM]$^\cup$ &- &SL &R &P, IC$^{1}$ &O &45 &LOSO &NPAD &3.99$^{1}$ MSE, \newline 1.55$^{2}$ MAE\\

\rowcolor{mygray}\ \textquotesingle 19 \cite{kalischek_thiam_2019} &F (RGB) &- &- FF &E &[2D CNN$^{+}$, \newline LSTM]$^\cup$ &- &UL &C &P &ST &40 &LOSO &SenseEmotion &60.61 ACC\\

\ \textquotesingle 21 \cite{vu_aimer_2021} &F (RGB) &- &- \space - &E &[2D CNN$^{+}$, \newline LSTM]$^\cup$ &- &SL &R &IC &P &25, 27 &LOSO &UNBC$^{1}$, \newline DISFA$^\oplus$ &0.60$^{+}$ MSE, \newline 0.82$^{+}$ PCC$^\ddagger$\\

\rowcolor{mygray}\ \textquotesingle 21 \cite{hu_liu_2021} &F (RGB) &- &- \space - &E &[2D CNN$^{+}$, \newline Transformer]$^\cup$ &- &SL &R &IC &P &25 &LOSO &UNBC &0.40 MSE, \newline 0.76 PCC$^\ddagger$\\

\ \textquotesingle 21 \cite{guo_wang_2021} &F (RGB) &- &- \space - &E &2D C-LSTM &- &SL &C &ID &S &29 &hold-out &other &69.58 F1\\

\rowcolor{mygray}\ \textquotesingle 19 \cite{zhi_wan_2019} &F (RGB) &- &- FF &E &SLSTM &- &SL &C &P$^{1}$, ID$^{2}$ &ST & 85 &LOSO &BioVid (A) &61.70$^{1}$ ACC \newline 29.70$^{2}$ ACC\\

\ \textquotesingle 21 \cite{ting_yang_2021} &F (RGB) &- &- \space - &I &3D CNN$^{+}$ &- &SL &R 
&IC &S &25 &k-fold &UNBC &0.66 ICC$^\ddagger$\\

\bottomrule
\end{tabular}

\begin{tablenotes}
\myfontsize
\item
\textbf{Fusion:} H: hybrid \textbf{Deep models:} TCN: temporal convolutional neural network C-LSTM: convolutional-LSTM SLTM: sparse long short memory network
\textbf{Learning Method:} SSL: self-supervised learning 
\textbf{Metrics:} F1: F1 score CCC: concordance
correlation coefficient 
 
\end{tablenotes}
\end{threeparttable}
\end{table}
\end{landscape}

\subsection{Touch sensor-based}
\label{sec:contact_sensor_based}
Touch (contact) sensors provide a viable alternative for pain assessment, often outperforming vision-based methods. Table \ref{table:contact_sensor_studies} highlights studies that utilized contact sensor data to evaluate pain. Yu \textit{et al.} \cite{yu_sun_zhu_2020} analyzed three categories of pain-no pain, moderate pain, and severe pain---using EEG signals. They extracted several bands from the biosignals, including alpha, beta, and gamma, and applied a convolutional module. The study found that combining these bands yielded better results than evaluating them independently. Similarly, \cite{wang_wei_2020} used EEG potentials with an autoencoder to compress the raw data and applied a logistic regressor for classification.

Other researchers, such as Rojas \textit{et al.} \cite{rojas_romero_2021}, utilized functional near-infrared spectroscopy (fNIRS) for pain detection. They developed three models---multilayer perceptron (MLP), LSTM, and biLSTM---with biLSTM demonstrating superior accuracy. Additionally, \cite{lim_kim_2019} focused on PPG signals, extracting hand-crafted features from the time and frequency domains, which were then combined with a deep belief network (DBN) to achieve over $65\%$ accuracy in a $4$-class pain assessment task. Hu \textit{et al.} \cite{hu_kim_2018} used kinematic data to compare healthy individuals with those suffering from low back pain (LBP). Their approach, which employed two stacked LSTM layers, reached over $97\%$ accuracy in binary classification using raw motion data. Lastly, Mamontov \textit{et al.} \cite{mamontov_polonskaia_2019} were the first to apply evolutionary algorithms in the design of an optimized recurrent neural network (RNN) for pain estimation, achieving $91.94\%$ accuracy using EDA signals.

\subsection{Audio-based}
\label{sec:audio_based}
A few studies have explored using audio information for pain detection and intensity estimation, as outlined in Table \ref{table:audio_studies}. These methods are especially relevant for neonates, where frequent facial and body occlusions make analyzing cries a more effective approach for pain detection. Chang and Li \cite{chang_li_2016} concentrated on infant cries to differentiate between hunger, pain, and sleepiness. They transformed the audio signals into 2D spectrograms using a fast Fourier transform (FFT) and trained a 2D CNN for feature extraction.
Similarly, \cite{salekin_zamzmi_2019} utilized spectrograms generated from recorded sounds, employing a model identical to that used in \cite{zamzmi_paul_salekin_2019}. Thiam and Schwenker \cite{thiam_schwenker_2019} focused on detecting adult pain by analyzing breathing sounds. They leveraged deep-learned features from spectrograms with Mel-scaled short-time Fourier transform, combined with various handcrafted cues. A CNN followed by a biLSTM was used to capture spatial and temporal dependencies, integrating both low- and high-level features.
In a different approach, Tsai \textit{et al.} \cite{tsai_weng_2017} examined pain events during emergency triage. They developed an LSTM autoencoder framework to extract temporal features from verbal behavior, reporting encouraging results.

\begin{landscape}
\renewcommand{\arraystretch}{1.5}
\begin{table}
\myfontsize
\begin{threeparttable}
\caption{Touch sensor-based studies.}
\label{table:contact_sensor_studies}
\centering
\begin{tabular}{ p{0.9cm} p{1.2cm} p{1.0cm} p{0.6cm} p{1.1cm} p{1.8cm} p{1.0cm} p{0.7cm} p{0.9cm} p{1.0cm} p{0.4cm} p{0.8cm} p{0.8cm} p{1.5cm} p{1.6cm}}
\hline
\toprule
\multirow{2}[3]{*}{Paper} &\multicolumn{4}{c}{Input}  &\multicolumn{4}{c}{Processing} & \multicolumn{6}{c}{Evaluation}\\
\cmidrule(lr){2-5} \cmidrule(lr){6-9} \cmidrule(lr){10-15}

\ & Modality  &Non deep \newline features &Fusion M/E &Temporal \newline Exploitation &Deep model  &Non deep \newline model &Learning Method &Classific. $/$Regres.  &Objective &GT &Number subjects &Validation Method &Dataset &Metrics\\
\hline\hline

\rowcolor{mygray}\ \textquotesingle 20 \cite{yu_sun_zhu_2020} &EEG &- &- FF &I &1D TCN &- &S 
&C &ID &S &32 &k-fold &other &97.30 ACC$^\ddagger$\\

\ \textquotesingle 20 \cite{wang_wei_2020} &EEG &- &- \space - &I &AE (TCN)  &LR  &UL, S &C &P &S &29 &LOSO &other &74.60 ACC\\

\rowcolor{mygray}\ \textquotesingle 21 \cite{rojas_romero_2021} &fNIRS &- &- \space - &E &biLSTM  &-  &SL 
&C &ID &S &18 &k-fold &other &90.60 ACC$^\ddagger$ \\

\ \textquotesingle 19 \cite{lim_kim_2019} &PPG  &- &- \space - &NL   &DBN &SBM &U, SL  &C &P$^1$, ID$^2$ &S &100 &k-fold &other &86.79$^1$ ACC, \newline 65.57$^2$ ACC\\

\rowcolor{mygray}\ \textquotesingle 18 \cite{hu_kim_2018} &kinematatics &- &- FF &E  &LSTM  &- &SL 
&C &P &LBP &44 &LOSO &other &97.20 ACC$^\ddagger$\\

\ \textquotesingle 19 \cite{mamontov_polonskaia_2019} &EDA &- &- FF &E &[RNN, LSTM, \newline GRU, NN]$^\cup$ &SelfCGA, selfCGP, PSOPB &SL &C &P &ST &40 &LOSO &Sense- \newline Emotion &81.94 ACC\\

\rowcolor{mygray}\ \textquotesingle 21 \cite{gouverneur_li_2021} &EDA &- &- \space - &I &NN &- &SL 
&C &P$^1$, I2 &ST &87, \newline 55 &LOSO &BioVid (A)$^1$, \newline PainMonit$^2$ &84.22$^{11}$ ACC$^\ddagger$, \newline 86.50$^{12}$ ACC$^\ddagger$\\

\bottomrule
\end{tabular}
\begin{tablenotes} 
 
\myfontsize
\item \textbf{Modality:} PPG: photoplethysmogram fNIRS: functional near-infrared spectroscopy EEG: electroencephalography EDA: electrodermal activity \textbf{Deep models:} DBN: Deep belief network RNN: recurrent neural network \textbf{Non deep models:} SBM: selective bagging model LR: Logistic Regression SelfCGA: Self-Configuring Genetic Algorithm SelfCGP: Self-Configuring Genetic Programming PSOPB: Particle Swarm Optimisation with parasitic behaviour \textbf{GT:} LBP: low back pain vs healthy population          
\end{tablenotes}
\end{threeparttable}
\end{table}


\renewcommand{\arraystretch}{1.5}
\begin{table}
\myfontsize
\begin{threeparttable}
\caption{Audio-based studies.}
\label{table:audio_studies}
\centering
\begin{tabular}{ p{0.9cm} p{1.3cm} p{1.0cm} p{0.6cm} p{1.1cm} p{1.8cm} p{1.0cm} p{0.7cm} p{0.9cm} p{1.0cm} p{0.4cm} p{0.8cm} p{0.8cm} p{1.5cm} p{1.5cm}}
\hline
\toprule
\multirow{2}[3]{*}{Paper} &\multicolumn{4}{c}{Input}  &\multicolumn{4}{c}{Processing} & \multicolumn{6}{c}{Evaluation}\\
\cmidrule(lr){2-5} \cmidrule(lr){6-9} \cmidrule(lr){10-15}

\ & Modality  &Non deep \newline features &Fusion M/E &Temporal \newline Exploitation &Deep model  &Non deep \newline model &Learning Method &Classific. $/$Regres.  &Objective &GT &Number subjects &Validation Method &Dataset &Metrics\\
\hline\hline

\rowcolor{mygray}\ \textquotesingle 16 \cite{chang_li_2016} &audio (cry) &- &- \space - &-  &2D CNN  &- &SL 
&C &P &O &? &k-fold &other &78.50 ACC\\

\ \textquotesingle 19 \cite{salekin_zamzmi_2019} &audio (cry)  &- &- \space - &-  &2D CNN &- &SL  &C &P &O &31 &LOSO$^\dagger$ &NPAD &96.77 ACC$^\ddagger$\\

\rowcolor{mygray}\ \textquotesingle 19 \cite{thiam_schwenker_2019} &audio \newline (breathing) &MFCCs, \newline RASTA-PLP, \newline DTD &- \,FF &E 
&[2D CNN, \newline LSTM]$^\cup$  &RFc  &SL &C &P &ST &40 &LOSO &Sense- \newline Emotion &64.39 ACC\\

\ \textquotesingle 17 \cite{tsai_weng_2017} &audio \newline(voice) &prosodic- spectral \newline features, SF &- \,FF &E &LSTM$^{+}$  &SVM  &UL, SL &C &P$^{1}$, ID$^{2}$ &S &63 &LOSO &other &72.30$^{1}$ UAR, \newline 54.20$^{2}$ UAR\\

\bottomrule
\end{tabular}
\begin{tablenotes}
\myfontsize
\item
\textbf{Non deep features:} MFCCs: Mel Frequency Cepstral Coefficients RASTA-PLT: Relative Spectral Perceptual Linear Predictive DTD: descriptors from temporal domain SF: statistical features      
\end{tablenotes}
\end{threeparttable}
\end{table}
\end{landscape}

\section{Multimodal studies}
\label{sec:multimodal}
Since pain is a multidimensional phenomenon, combining multiple modalities in a multimodal system offers a promising approach. Heterogeneous information sources can complement one another, enhancing specificity and sensitivity. As reported in \cite{werner_2019}, when individual modalities demonstrate good predictive performance, their fusion tends to yield improved outcomes. Moreover, integrating cues from various channels may be helpful and necessary, especially in clinical settings where specific modalities may become unavailable (for instance, if the patient turns and their face is occluded).
The information channels can originate from (1) the same hardware sensor but focus on different regions of interest, such as RGB facial images and RGB body images \cite{salekin_zamzmi_2019_b}, (2) different hardware sensors but the same region of interest, like RGB facial images and thermal facial images \cite{haque_bautista_2018}, or (3) different hardware sensors and information sources, such as RGB facial images and ECG signals \cite{naeini_shahhosseini_2019}. Table \ref{table:multimodal_studies} lists the studies utilizing multimodal approaches.

\subsection{Static Analysis}
\label{sec:non_temporal_exploitation}
A commonly used biosignal combination is those of EDA, EMG, and ECG, as these channels are found in all main pain reference databases. Thiam \textit{et al.} \cite{thiam_bellmann_kestler_2019} applied an early fusion method by merging these signals into a 2D representation and inputting it into a $9$-layer 2D CNN. Their results showed a strong correlation between EDA and pain intensity, and combining all three modalities did not outperform using EDA alone. Al-Qerem \textit{et al.} \cite{al-qerem_2020} used least generative adversarial networks (LSGANs) to enhance EMG, EDA, and ECG samples, reporting a notable improvement in classification when using an SVM on the augmented dataset. Haque \textit{et al.} \cite{haque_bautista_2018} introduced the \textit{MIntPAIN} dataset, which includes RGB, depth, and thermal videos for multi-class ($5$ levels) pain recognition. They combined these three visual modalities into a 5D matrix (RGB+D+T) and used it to train the pre-trained \textit{VGG-Face} model \cite{parkhi_vedaldi_2015}, leading to better classification performance in their experiments.

\subsection{Temporal Utilization}
\label{sec:temporal_exploitation}
Zhi \textit{et al.} \cite{zhi_zhou_yu_2021} proposed a multimodal stream-integrated neural network that leverages video and biosignal data. They combined raw facial video frames with optical flow images to capture spatio-temporal dependencies via 3D CNNs, integrating these with biosignal features extracted using LSTMs. The entire network was trained end-to-end, achieving superior results compared to their unimodal methods. Beyond facial analysis, Salekin \textit{et al.} \cite{salekin_zamzmi_2019_b} focused on assessing neonatal pain through body movements in videos. After identifying relevant body regions, video frames were fed into a pre-trained \textit{VGG-16} \cite{simonyan_karen_2015}, connected to an LSTM to capture temporal dynamics. In a follow-up study, Salekin \textit{et al.} \cite{salekin_zamzmi_goldgof_2021} fused three modalities---facial expressions, body movements, and crying sounds--demonstrating that this multimodal approach outperformed unimodal techniques. Similarly, Wang \textit{et al.} \cite{wang_xu_2020} explored combining EMG, EDA, and ECG biosignals with handcrafted and learned features from a biLSTM model. They applied the minimum relevance method (MRMR) to reduce the number of features, resulting in notable outcomes.

In addition to EDA, EMG, and ECG, other biosignal combinations have been explored. Zhao \textit{et al.} \cite{zhao_ly_hong_2020} integrated PPG, EDA, and temperature signals, using 2D convolutions for spatial feature extraction and time windows for capturing temporal information. Yuan \textit{et al.} \cite{yuan_mahmoud_2020} successfully estimated pain using whole-body MoCap sensors and EMG, utilizing LSTM layers with an attention mechanism in an autoencoder, which reduced training time by leveraging latent space representations of raw data. Similarly, Li \textit{et al.} \cite{li_ghosh_2020} employed MoCap and EMG as data sources and tested various LSTM configurations to predict pain intensity, achieving the best performance with a $3$-layer vanilla LSTM combined with a $3$-layer fully connected network.

\begin{landscape}
\renewcommand{\arraystretch}{1.5}
\begin{table}
\myfontsize
\begin{threeparttable}
\caption{Multimodal-based studies.}
\label{table:multimodal_studies}
\centering
\begin{tabular}{ p{0.9cm} p{1.2cm} p{1.0cm} p{0.6cm} p{1.1cm} p{1.8cm} p{1.0cm} p{0.7cm} p{0.9cm} p{1.0cm} p{0.4cm} p{0.8cm} p{0.8cm} p{1.4cm} p{1.6cm}}
\hline
\toprule
\multirow{2}[3]{*}{Paper} &\multicolumn{3}{c}{Input}  &\multicolumn{5}{c}{Processing} & \multicolumn{6}{c}{Evaluation} \\
\cmidrule(lr){2-4} \cmidrule(lr){5-9} \cmidrule(lr){10-15}

\ & Modality  &Non deep \newline features &Fusion M/E &Temporal \newline Exploitation &Deep model  &Non deep \newline model &Learning Method &Classific. $/$Regres.  &Objective &GT &Number subjects &Validation Method &Dataset &Metrics\\
\hline\hline

\rowcolor{mygray}\ \textquotesingle 18 \cite{haque_bautista_2018} &F (RGB, \newline thermal, depth)  &- &RF -  &-  &2D CNN$^{+}$ &- &SL   &C &ID &S &20 &k-fold &MIntPAIN &36.55 ACC\\

\ \textquotesingle 19 \cite{salekin_zamzmi_2019_b} & F, B (RGB) &- &FF - &E &[2D CNN$^{+}$, \newline LSTM]$^\cup$ &- &SL &C &P &O &31 &LOSO &other &92.48 ACC$^\ddagger$\\

\rowcolor{mygray}\ \textquotesingle 19 \cite{naeini_shahhosseini_2019} &F (RGB), \newline ECG, EDA &biosignals' features$^{\oslash}$ &FF FF &- &2D CNN$^{+}$  &RFc &SL &C &I2 &S &85 &k-fold &BioVid (A) &74.00 ACC\\

\ \textquotesingle 19 \cite{thiam_bellmann_kestler_2019} &EDA, EMG, \newline ECG &- &RF -  &- &2D CNN &- &SL &C &P$^{1}$ I2, \newline ID$^{2}$ &S &87, \newline 86 &LOSO &BioVid (A)$^{1}$ \newline BioVid (B) &84.40$^{11}$ ACC$^\ddagger$, \newline 36.54$^{12}$ ACC$^\ddagger$\\

\rowcolor{mygray}\ \textquotesingle 20 \cite{al-qerem_2020} &EDA, EMG, \newline ECG &Boruta \newline features &FF - &- &LSGAN &SVM &UL, SL &C &I2, ID$^{1}$ &S &85 &hold-out  &BioVid (A) &82.80$^{1}$ ACC\\

\ \textquotesingle 21 \cite{zhi_zhou_yu_2021} &F (RGB), EDA, \newline EMG, ECG &optical flow &FF FF &NL, \newline E, I &[3D CNN, \newline LSTM]$^\cup$ &- &SL &C, R &P$^{1}$, I2, \newline ID$^{2}$ &S &87, \newline 40 &k-fold$^\dagger$  &BioVid (A)$^{1}$, \newline MIntPain &68.20$^{11}$ ACC$^\ddagger$, \newline 28.10$^{21}$ ACC\\

\rowcolor{mygray}\ \textquotesingle 21 \cite{salekin_zamzmi_goldgof_2021} &F, B (RGB), \newline sound &- &DF - &E &[2D CNN$^{+}$, \newline LSTM]$^\cup$ &- &SL &C &P &O &45 &LOSO &NPAD &78.95 ACC$^\ddagger$\\

\ \textquotesingle 20 \cite{wang_xu_2020} &EDA, EMG, \newline ECG &MRMR, biosignals' features &RF FF &E &biLSTM &NN &SL &C &P$^{1}$, I2 &S &87 &LOSO &BioVid (A) &83.30$^{1}$ ACC\\

\rowcolor{mygray}\ \textquotesingle 20 \cite{thiam_kestler_schenker_2020_b} &EDA, EMG, \newline ECG &- &FF -  &I  &[DDCAE, \newline NN]$^\cup$ &- &UL, SL &C &P$^{1}$, I2 &S &87 &LOSO &BioVid (A) &83.99$^{1}$ ACC$^\ddagger$\\

\ \textquotesingle 21 \cite{thiam_hihn_2021} &EDA, EMG, \newline ECG, RSP &- &FF -  &I  &[DDCAE, \newline NN]$^\cup$ &- &UL, SL, SSL &C, R 
&P$^{1}$, ID$^{2}$, \newline IC &S &87, \newline 40 & LOSO &BioVid (A)$^{1}$, \newline Sense-Emotion &84.25$^{11}$ ACC$^\ddagger$, \newline 35.44$^{21}$ ACC$^\ddagger$\\

\rowcolor{mygray}\ \textquotesingle 21 \cite{subramaniam_dass_2021} &EDA, ECG &- &FF - &E  &1D CNN, \newline LSTM &- &UL &C &P$^{1}$, I2 &S &67 &hold-out &BioVid (A) &81.71$^{1}$ ACC\\

\ \textquotesingle 20 \cite{zhao_ly_hong_2020} &PPG, EDA, \newline temperature &- &RF - &I &2D CNN &- &SL 
&R$^{\circ}$ &P$^{1}$, ID$^{2}$ &S &21 &k-fold &other &96.30$^{1}$ ACC, \newline 95.23$^{2}$ ACC\\

\rowcolor{mygray}\ \textquotesingle 20 \cite{yuan_mahmoud_2020} &MoCap, EMG &- &RF - &E &AE, LSTM &- &UL, SL &C &ID &O & 23 &LOSO$^\dagger$ &EmoPain &52.60 ACC$^\ddagger$\\

\ \textquotesingle 20 \cite{li_ghosh_2020} &MoCap, EMG &- &RF -  &E &LSTM, NN &- &UL &C &ID &O &30 &hold-out &EmoPain &80.00 ACC$^\ddagger$\\

\rowcolor{mygray}\ \textquotesingle 21 \cite{li_ghosh_joshi_2021} &MoCap, EMG &- &RF - &E &LSTM, NN &- &SL &C &ID &O &30 &LOSO$^\dagger$ &EmoPain &54.60 ACC$^\ddagger$\\

\bottomrule
\end{tabular}
\begin{tablenotes}
\myfontsize
\item ${\oslash}$: Not specifically described 
${\circ}$: Ordinal
\textbf{Modality} F: face region B: body region \ EMG: electromyography \textbf{Non deep features:} MRMR: Minimum Redundancy Maximum Relevance method \textbf{Deep models:} LSGAN: Least Square Generative Adversarial Networks   
\end{tablenotes}
\end{threeparttable}
\end{table}
\end{landscape}

\section{Summary of Automatic Pain Assessment Methods}
This section presents an analysis of the reviewed studies, summarizing the main conclusions on current methods for automatic pain assessment, their advantages, and corresponding limitations. Additionally, it offers recommendations for future research directions that could advance the field of pain research from a computational perspective.

\subsection{Input}
First, we observe a clear imbalance between unimodal and multimodal approaches in pain assessment studies. More than $86\%$ of the reported research focuses on unimodal methods, even though the databases often contain multiple information channels. Notably, contact sensor-based and audio-based approaches are underrepresented, with only seven and four studies, respectively, compared to $84$ studies that utilize a vision-based approach.

Multimodal approaches are even less explored, with only $15$ studies falling into this category, making it difficult to draw strong conclusions about the effectiveness of specific modality combinations. However, there are indications that EDA sensor data is particularly valuable compared to other biopotentials. Researchers have primarily focused on visual data, likely due to the complexity of implementing multimodal frameworks or the impracticality of contact sensors in non-laboratory settings. Further exploration of diverse modality combinations is necessary to evaluate their potential for pain assessment fully---additionally, $28$ studies employed non-deep features to enhance deep-learned representations.

Finally, we identified three primary strategies in examining the approaches that utilize temporal information: non-machine learning-based, machine learning-based (implicit), and machine learning-based (explicit). Non-machine learning-based methods, such as motion history images \cite{thiam_kestler_schenker_2020} or temporal distillation \cite{tavakolian_bordallo_liu_2020}, rely on traditional computer vision techniques. These methods tend to be more straightforward but are generally less sophisticated. In contrast, machine learning-based approaches \cite{praveen_granger_cardinal_2020} \cite{zhi_wan_2019}  offer richer temporal information and the flexibility to adapt to specific requirements, such as emphasizing certain video frames. Among the studies reviewed, $55\%$ employed temporal features, with explicit methods---most commonly LSTM models---being the predominant choice. Given that many studies report superior performance when temporal information is incorporated, compared to non-temporal methods, it is evident that further emphasis on temporal approaches is warranted.

\subsection{Processing}
Regarding machine learning approaches, various models and techniques have been employed for pain estimation. CNN models remain the most widely used, with more than $75\%$ of studies utilizing 1D, 2D, or 3D filters, highlighting the central role of convolution operations in deep learning. Sequential models, such as RNNs, GRUs, LSTMs, and biLSTMs, follow closely behind in popularity. Almost half of the studies used pre-trained models to achieve their desired performance. This suggests that existing pain databases may not be adequate for training deep-learning models from scratch. Non-deep learning models have also been employed in $26$ studies as auxiliary decision components, with SVMs and shallow neural networks being the most common choices. There seems to be significant potential for adopting newer deep learning architectures, especially transformer-based models, which have demonstrated state-of-the-art results in various AI research fields and are particularly suited for exploiting temporal modality information \cite{lin_wang_2021}.

The predominant learning method used across studies is supervised learning. However, $16$ papers explored or adopted alternative methods such as unsupervised learning \cite{pedersen_2015,kharghanian_peiravi_2016,tavakolian_cruces_2019,tavakolian_hadid_2019,kalischek_thiam_2019, wang_wei_2020,lim_kim_2019,tsai_weng_2017,al-qerem_2020,thiam_kestler_schenker_2020_b,yuan_mahmoud_2020}, 
self-supervised \cite{tavakolian_bordallo_liu_2020,thiam_hihn_2021}, self-supervised learning \cite{tavakolian_bordallo_liu_2020,thiam_hihn_2021}, semi-supervised learning \cite{pedersen_2015}, weakly supervised learning \cite{gnana_praveen_granger_2020,praveen_granger_cardinal_2020}, and federated learning \cite{rudovic_tobis_2021}. Given the limited availability of pain data resources, self-supervised learning appears to be the most appropriate method for future research and should be further embraced by the community.

Lastly, it is notable that most studies---approximately $70\%$---treat pain assessment as a classification problem rather than a regression problem. However, we believe that regression more closely reflects the continuous nature of pain and is better suited to capturing the complexity of pain sensation.

\subsection{Evaluation}
\label{slr_evaluation}
The primary objectives of the reviewed studies were \textit{(i)} to estimate pain intensity on a discrete scale (multi-class classification), \textit{(ii)} to measure pain intensity on a continuous scale, and \textit{(iii)} to determine the presence or absence of pain (binary classification). Notably, $25$ studies focused on pain detection rather than pain intensity estimation, which, from a clinical standpoint, is less informative as it does not provide sufficient data for effective pain management. From an engineering perspective, detecting the presence or absence of pain is also a more straightforward and less demanding task.

A small subset of studies took a different approach to pain estimation. For instance, one study \cite{tavakolian_cruces_2019} sought to differentiate genuine pain from acted pain. Another \cite{tsai_weng_2017} explored pain events in emergency triage settings rather than controlled laboratory environments, while \cite{naeini_shahhosseini_2019} examined the feasibility of real-time pain detection on IoT devices. Additionally, \cite{semwal_londhe_b_2021} and \cite{semwal_londhe_c_2021} aimed to address the issue of occluded faces in pain estimation. Sociodemographic and psychological factors were also considered, as seen in studies like \cite{subramaniam_dass_2021}, which explored gender differences, and \cite{rezaei_moturu_2020}, which focused on pain assessment in elderly patients with dementia. The limited exploration of pain estimation in real-world settings or unconventional contexts suggests that current approaches may not be fully applicable in practical environments like clinics and hospitals.

Various annotation types are used regarding ground truth, such as self-reported ratings, FACS, and observer scales. Temporal features are critical for accurately estimating pain intensity, making the temporal granularity of the ground truth equally important. Several studies have questioned the objectivity of PSPI scores, as noted in \cite{werner_hamadi_ecklundt_2017}, which highlights that PSPI scores can be zero even when pain is present or that there may be no visible facial expressions in low-intensity pain. Pain expressions not captured by the FACS system, such as raising eyebrows or opening the mouth, further challenge the use of PSPI \cite{kunz_lautenbacher_2014}. Additionally, PSPI does not account for pain-related head and body movements, which are particularly valuable in newborn assessments \cite{ragnar_manon_2007}. For these reasons, we recommend moving away from PSPI as ground truth in favor of self-reports and observer scales at the video-segment level.

Around $54\%$ of the studies employed the leave-one-subject-out (LOSO) validation method, which is widely regarded as more objective and better for assessing the generalizability of models. However, LOSO can be less practical due to the increased model size and longer training times. When researchers use other validation methods, such as k-fold or hold-out, it is essential to ensure that consecutive, highly correlated frames from the same subject do not skew the training and validation results, leading to flawed estimations. Moreover, when researchers define their own validation or testing sets, comparing results across studies---especially between classification and regression models---becomes nearly impossible. We believe standardized evaluation protocols should be developed for each publicly available database for these reasons.

\subsection{Pain Databases for Evaluation}
The availability of suitable public databases is arguably the most crucial factor in addressing the challenge of automatic pain assessment. Several aspects must be considered in evaluating these datasets, including the number of subjects and their characteristics, such as age, sex, health status, and race. Moreover, the ground truth must be objective and offer meaningful insights into the subject's pain experience \cite{casti_mencattini_2019}.

Fig. \ref{fig:pain_datasets} illustrates the number of papers corresponding to the pain database utilized in each study. It is clear from this figure that the \textit{UNBC} and \textit{BioVid} databases were the most commonly used public datasets. However, the UNBC dataset does not record the subjects' ages, despite age being a known factor in pain expression \cite{Boerner_birnie_2014, gkikas_chatzaki_2022}. While the \textit{BioVid} dataset does document age, the oldest participants are only $65$ years old, which is notable since pain and its management are critical issues among individuals aged $65$ and older \cite{Jones_Ehrhardt_2016}. Similar limitations are found in other pain datasets, such as \textit{X-ITE} \cite{x-ite_databaase_2019}, \textit{EmoPain} \cite{emopain_dataset_2016}, and \textit{SenseEmotion} \cite{senseemotion_database_2017}. 

It is well known that aging causes skin changes, including texture, rigidity, and elasticity alterations, which can impact facial emotion recognition tasks \cite{ochi_midorikawa_2021}. Additionally, race-related factors can lead to inaccurate pain assessments due to variations in how pain is expressed \cite{forsythe_thorn_2011}. Notably, one study by Nerella \textit{et al.} \cite{nerella_cupka_2021} reported lower performance when their model was tested on African American patients. Furthermore, only one study \cite{rezaei_moturu_2020} was found that specifically addressed pain estimation in elderly individuals with dementia.

In summary, developing objective, automated, and generalizable deep learning-based pain assessment systems will only be possible if balanced and representative datasets are available for training and external validation.

\subsection{Interpretation of Results}
Recent advancements in AI have shown state-of-the-art performance across nearly every scientific discipline, often surpassing human accuracy in specific diagnostic tasks \cite{tschandl_codella_2019}. However, a significant drawback of AI solutions, particularly deep neural networks, is their lack of transparency, commonly called \textquotedblleft black box AI\textquotedblright. This term highlights how these models learn intricate functions that are opaque and frequently incomprehensible to humans \cite{yang_ye_2022}. This opacity is a primary reason for the criticism directed toward deep learning techniques \cite{lekadir_osuala_2021}. Various techniques, such as visualizations and gradients-backpropagation focusing on specific units, have been developed to offer insights into how these models function. For further reading, refer to the comprehensive review on explanatory techniques in deep learning \cite{linardos_2021}.

Table \ref{table:interpretation_approaches} outlines the different approaches used to interpret model decisions. Only a small fraction of the reviewed studies---$20$ out of $110$---implemented methods to explain how their models work and which features or elements they focus on. It is important to note that interpretable machine learning can be broadly defined as the \textit{\textquotedblleft extraction of relevant knowledge from a machine-learning model concerning relationships either contained in data or learned by the model\textquotedblright\space}\cite{murdoch_singh_2019}. To summarize: \textit{(i)} $18\%$ of the reviewed studies provided an approach to enhance the interpretability of the model's decision, \textit{(ii)} all of these methods were applied to studies using facial images as the input modality, and \textit{(iii)} around half of these studies were conducted by just three specific research groups. These findings suggest that the issue of interpretability and explainability within deep learning remains underexplored, particularly in the context of automatically classifying pain severity levels.

\renewcommand{\arraystretch}{1.5}
\begin{table}
\scriptsize
\begin{threeparttable}
\caption{Interpretation approaches.}
\label{table:interpretation_approaches}
\begin{tabular}{ p{0.9cm}p{0.5cm} p{1.4cm} p{5.0cm}}
\hline
\toprule
\textbf{Paper}  &\textbf{Year} &\textbf{Modality}  & \textbf{Method}\\
\hline
\hline
\ \cite{semwal_londhe_august_2021} &2021 &F (RGB) &visualization (saliency maps)\\\hline
\ \cite{tavakolian_hadid_2018_b} &2018 &F (RGB) &visualization (heat maps)\\\hline
\ \cite{semwal_londhe_2021}  &2021&F (RGB) &visualization (saliency map)\\\hline
\ \cite{kharghanian_peiravi_2016} &2016&F (RGB) &visualization (learned filters)\\\hline
\ \cite{kharghanian_peiravi_moradi_2021} &2021 &F (RGB) &visualization (learned filters)\\\hline
\ \cite{huang_xia_li_2019} &2019&F (RGB) &visualization (heat maps), \newline values of learned weights\\\hline
\ \cite{li_zhu_2018} &2018 &F (RGB) &visualization (saliency maps)\\\hline
\ \cite{xin_li_yang_2021} &2021 &F (RGB) &visualization (attention maps)\\\hline
\ \cite{semwal_londhe_b_2021} &2021 &F (RGB) &visualization (saliency map)\\\hline
\ \cite{semwal_londhe_c_2021} &2021 &F (RGB) &visualization (activation maps)\\\hline
\ \cite{xu_huang_sa_2020} &2020&F (RGB) &visualization  (pixels contributions)\\\hline
\ \cite{egede_valstar_2017} &2017 &F (RGB)&visualization (average saliency map)\\\hline
\ \cite{tavakolian_cruces_2019} &2019 &F (RGB) &visualization \newline (generated intermediate representation)\\\hline
\ \cite{rezaei_moturu_2020} &2020&F (RGB) & visualization (saliency maps)\\\hline
\ \cite{pandit_schmitt_2020} &2020&F (RGB) &weights per AU (contribution of AUs)\\\hline
\ \cite{zamzmi_paul_2019} &2019 &F (RGB) &visualization (feature maps)\\\hline
\ \cite{carlini_ferreira_2021} &2021 &F (RGB) &visualization (integrated gradients)\\\hline
\ \cite{vu_aimer_2021} &2021 &F (RGB) &visualization (heatmaps)\\\hline
\ \cite{huang_xia_2020} &2020&F (RGB) &visualization (attention maps), \newline values of learned weights\\\hline
\ \cite{yu_kurihara_2019} &2019 &F (RGB)  &visualization (attention maps)\\
\bottomrule
\end{tabular}
\end{threeparttable}
\end{table}

\section{Challenges and Future Directions}
This section discusses the existing challenges in automatic pain assessment and proposes future research directions to further progress in the field.

\subsection{Current Challenges in Automatic Pain Assessment \& Future Research Directions}
\label{chapter_3_challenges}

Several limitations exist in the current pain databases. Important demographic factors such as sex, gender, and age are often missing, and there is an apparent lack of racial diversity among subjects. For example, facial structures and emotional expressions vary across Caucasian, Asian, and African populations \cite{jack_2013}. Moreover, social interactions, such as the presence of a partner during assessments, could influence pain manifestation and should be included in future datasets \cite{McClelland_mccubibin_2008}. Estimating the location of pain, particularly for infants or individuals with communication impairments, is another vital aspect of pain assessment systems, which current databases largely overlook. Future datasets should incorporate stimuli targeting various body locations.
Furthermore, the videos in existing visual databases often have low to medium resolution and frame rates, which are inadequate for capturing facial micro-expressions. Audio data is also sparsely represented, though it holds potential as a valuable modality. From an audio perspective, integrating natural language processing (NLP) methods to extract linguistic features and create multimodal systems is a promising direction, as shown in affective computing research \cite{gambria_2013}. Finally, specific validation protocols should be provided with present and future datasets to ensure objective and consistent comparisons across studies.

From an engineering perspective, several issues must be addressed to advance automatic pain assessment. Developing multimodal approaches is essential for creating robust systems with enhanced capabilities. Not only do multimodal methods demonstrate better performance than unimodal ones, but they are also crucial in real-world scenarios where a specific modality may become unavailable. Additionally, it is essential to exploit each modality's temporal aspects fully. We encourage using machine learning models or other techniques that can accommodate the dynamic nature of pain. More work is needed to improve the accuracy of multi-level and low-intensity pain estimation. Another area of research involves the relationship between pain and other affective states, such as negative emotions, which often coexist during painful events. Detecting these emotions could improve pain assessment. Addressing challenges like occlusions or poor lighting conditions in vision-based systems also requires attention. Researchers should explore these scenarios, even if current databases do not account for them. Real-time application of pain assessment systems is another critical factor, so future studies should measure throughput, such as the number of images processed per second during inference. Generalization is another crucial concern for AI systems, and evaluating trained models across different pain databases could be valuable. Finally, to facilitate the clinical adoption of AI-based pain assessment systems, the models' decisions need greater explainability. Developing or adopting methods that improve interpretability will enhance their clinical viability.

    \chapter{Demographic Variables: Their Role and Impact}
\minitoc  
\label{chapter_4}

\section{Chapter Overview: Introduction \& Related Work}
This chapter includes the findings published in \cite{gkikas_chatzaki_2022, gkikas_chatzaki_2023}. As discussed in Section \ref{clinical_pain_assessment}, research has demonstrated that biological and psychological differences can lead to variations in how pain is perceived between men and women.
Regarding age, it is known that infants who cannot express themselves directly and older adults with health issues require specific care due to their unique needs. However, a significant question remains unanswered in pain research, both from clinical and biological perspectives: Does the sensation of pain change as individuals age? Specifically, does a person in pain perceive their situation differently as they grow older than others in different age groups? Is the sensation of pain evolving throughout life? To the best of our knowledge, this question has not yet been definitively explored thoroughly.
This chapter investigates the differences among males and females, as well as age groups, using ECG signals. In addition, it proposes a computational framework that utilizes these two demographic elements to improve pain assessment performance. 
This chapter analyzes ECG signals to explore the differences in pain perception between males and females and various age groups. Additionally, it introduces a computational framework incorporating these two demographic factors to enhance the accuracy of pain assessments.

From a computational standpoint, the literature concerning the use of demographic factors in pain assessment is scarce. The study in \cite{hinduja_canavan_2020} utilized a range of biosignals, including EDA, respiration rate, diastolic blood pressure, and facial action units, to demonstrate differences in pain perception between males and females. Similarly, in research \cite{subramaniam_dass_2021}, the authors utilized a hybrid CNN-LSTM model that processed ECG and EDA data, highlighting gender-based variations in pain response.
Following the publications of our research, another study by Ricken \textit{et al.} \cite{ricken_bellmann_2023} was released, which explored the differences in adaptation and habituation between men and women. This study extracted handcrafted features from various biosignals (including ECG and EDA) and employed random forest classifiers to analyze the data.

\section{ECG Analysis with Classical Machine Learning}
\label{chapter_4_paper_1}
We explore a pain estimation process using ECG signals and examine variations across different demographic groups, focusing on gender and age. Specifically, we analyze how pain manifestation differs between males and females, investigate variations in pain perception across different age groups, and consider the combined effects of age and gender on pain perception.

\subsection{Methodology}
This section will describe the electrocardiography processing algorithm and the methods used for feature extraction and classification algorithms.

\subsubsection{ECG signal Processing and Analysis}
\label{pan_tompkins}
An ECG signal captures the heart's electrical activity over time. Typically, a normal ECG displays a sequence of waves, identified as P, Q, R, S, T, and occasionally U. These waves and their intervals provide crucial insights into heart function. The P wave indicates atrial depolarization, the QRS complex signifies ventricular depolarization and contraction, and the T wave corresponds to the repolarization of the ventricles. Each heartbeat is depicted through the PQRST complex (refer to Figure 1). Accurately detecting the R wave within the QRS complex is especially critical as it is the most pronounced peak in the complex. Precise detection of the R wave allows for the calculation of heart rate (HR) and heart rate variability (HRV), the latter of which measures the time intervals between successive R waves, known as the R-R or Interbeat interval.
The \textit{Pan-Tompkins} algorithm, developed in 1985, is one of the most extensively used real-time QRS detection algorithms \cite{pan_tompkins_1985}. Over the years, both the original algorithm and its variations have been rigorously tested, consistently proving their effectiveness even with noisy and low-quality data \cite{fariha_ikeura_2020, liu_wei_2017}. The Pan-Tompkins Algorithm is frequently cited as a benchmark in the field due to its robust performance, making it a standard against which new QRS detection methods are compared \cite{zhao_li_2021}. Our research employed the original Pan-Tompkins Algorithm to identify the QRS complex. We integrated the algorithm in two primary phases: preprocessing and decision-making. The preprocessing stage is crucial for conditioning the ECG by eliminating noise and artifacts, smoothing the signal, and enhancing the QRS slope. The preprocessing steps of the \textit{Pan-Tompkins} algorithm are depicted in the flow diagram shown in Figure \ref{pan_tom}.

\begin{figure}
\centering
\includegraphics[scale=0.45]{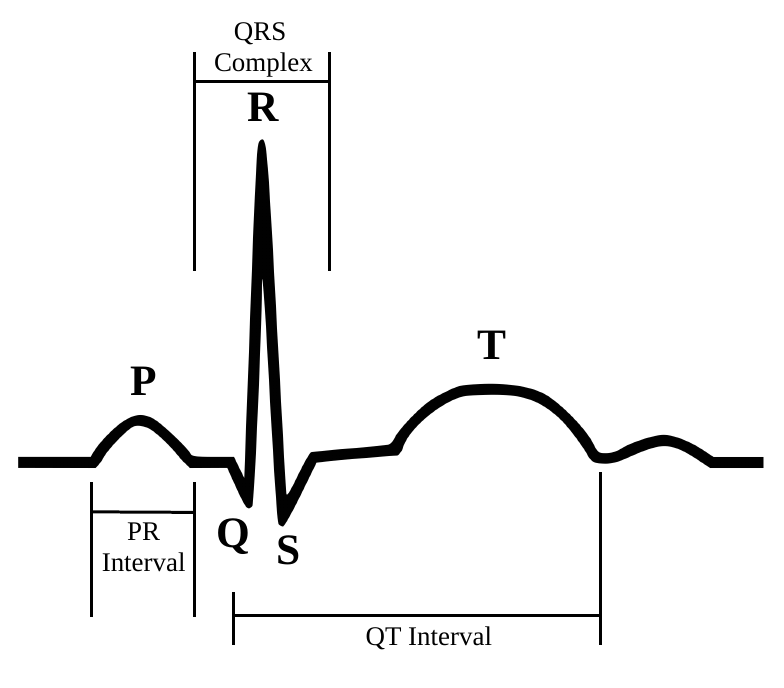}
\caption{The PQRST waveform.}
\label{qrs}
\end{figure}

\begin{figure}
\centering
\includegraphics[scale=1.2]{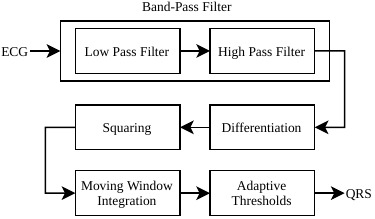}
\caption{The flowchart of the Pan-Tompkins algorithm's pre-processing procedure.}
\label{pan_tom}
\end{figure}

\subsubsection{Feature Extraction}
\begin{figure}
\centering
\includegraphics[scale=0.18]{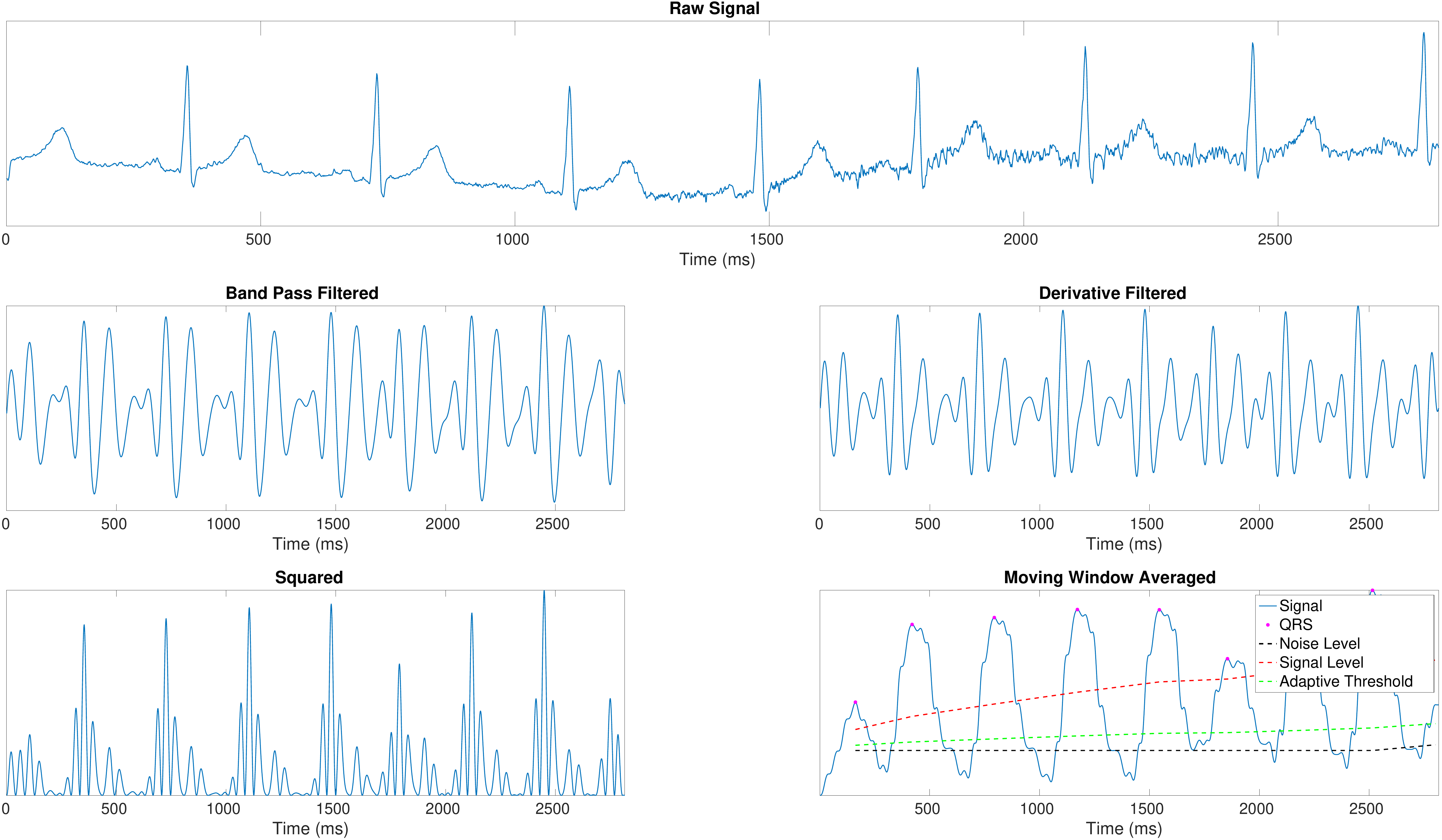}
\caption{The signal preprocessing using the \textit{Pan-Tompkins} algorithm.}
\label{pan_tompkins_signal}
\end{figure}

The subsequent phase involves extracting specific features based on the inter-beat intervals (IBIs). In our study, we calculated several metrics, including the mean of IBIs, the root mean square of successive differences (RMSSD), the standard deviation of IBIs (SDNN), the slope of the linear regression of IBIs, the ratio of SDNN to RMSSD, and the heart rate, as outlined below:
\begin{enumerate}
\item Mean of IBIs 
\begin{equation} 
\mu=\frac{1}{n}\sum_{i=1}^{n}(RR_{i+1}-RR_{i}),
\end{equation}
where $RR$ represents consecutive $R$ peaks.

\item Root mean square of successive differences
\begin{equation} 
RMSSD=\sqrt{\frac{1}{n-1}\sum_{i=1}^{n-1}(RR_{i+1}-RR_{i})^2}
\end{equation}

\item Standard deviation of IBIs
\begin{equation} 
SDNN=\sqrt{\frac{1}{n-1}\sum_{i=1}^{n}(RR_{i}-\mu)^2}
\end{equation}

\item Slope of the linear regression of IBIs
\begin{equation}
A^{T}Ax=A^{T}b,
\end{equation}
where is calculated using the least-square approximation, where $b$ is the vector of $RR$ peak intervals and $A$ is the corresponding time series.

\item Ratio of SDNN to RMSSD
\begin{equation}
SR=\frac{SDNN}{RMSSD}
\end{equation}

\item Heartbeat rate
\begin{equation}
HR=\frac{60 \cdot FS}{\mu},
\end{equation}
where $FS$ is the sampling frequency of the ECG recording, typically $512$ Hz. Figure \ref{pan_tompkins_signal} illustrates the raw ECG signal and the algorithm's stages.  
\end{enumerate}

\subsubsection{Classification Methods}
For the classification phase, three widely recognized classifiers were utilized: Linear Discriminant Analysis (LDA), Support Vector Machine (SVM) with a linear kernel, and SVM with a Radial Basis Function (RBF) kernel.

\begin{enumerate}
\item Linear Discriminant Analysis
\begin{equation} 
P(X\vert y=k)= \dfrac{exp\Big(-\dfrac{1}{2}(X-\mu_{k})^t\Sigma_{k}^{-1}(X-\mu_{k})^t\Big)}{(2\pi)^{d/2}\vert \Sigma_{k} \vert^{1/2}},
\end{equation}
where $P$ denotes the probability density function for the feature set $X$, conditional on the target class $y=k$.
\item SVM with linear kernel
\begin{equation} 
K(x_{1},x_{2})=x_{1}^Tx_{2},
\end{equation}
where $x_1$ and $x_2$ represent feature vectors from two separate classes.  
\item SVM with Radial Basis Function (RBF) kernel 
\begin{equation} 
K(x_{1},x_{2})=exp\Bigg(-\frac{\vert\vert x_{1}-x_{2} \vert\vert^2}{2\sigma^2}\Bigg),
\end{equation}
where $\sigma$ is the parameter defining the width of the RBF kernel.
\end{enumerate}

\subsubsection{Dataset Details}
In this study, we utilized the publicly available \textit{\textquotedblleft BioVid Heat Pain Database\textquotedblright} \cite{biovid_2013}, which contains facial videos and biosignals (ECG, EMG, EDA) from $87$ participants ($44$ males and $43$ females, aged $20-65$). This dataset is unique because it is the only publicly accessible resource that includes the subjects' age and gender. The data collection involved applying a heat stimulus to the right arm of each participant using a thermode. Prior to recording, the pain threshold (the temperature at which the participant first perceives heat as pain) and pain tolerance (the temperature at which the pain becomes intolerable) were established for each participant. The study defined specific thresholds as the temperatures for the lowest and highest pain levels. Also, it included two intermediate levels, resulting in five pain conditions: No pain (NP), mild pain (P\textsubscript{1}), moderate pain (P\textsubscript{2}), severe pain (P\textsubscript{3}), and very severe pain (P\textsubscript{4}). Each participant was exposed to $20$ stimulations for each intensity level, generating $100$ samples across the four modalities.

\subsection{Experiments}
\label{demographic_experiments_paper_1}
In the following experiments, we specifically used \textit{Part A} of the \textit{BioVid}, which includes pre-processed ECG samples filtered through a Butterworth band-pass filter, totaling $8700$ samples ($87\times 100=8700$).
All experiments were conducted in triplicate under identical conditions, using a distinct classifier for each iteration to compare their effectiveness. This was based on the leave-one-subject-out (LOSO) cross-validation method, utilizing all available subjects and ECG samples. The performance of each classifier was evaluated based on accuracy.

Using the previously mentioned classification algorithms, we conducted experiments to recognize pain and its relationship with demographic factors. The classification tasks were structured around the pain conditions in multi-class and binary classification formats. Specifically, five distinct experiments were executed:
\textit{(i)} multi-class pain classification, \textit{(ii)} NP vs. P\textsubscript{1}, \textit{(iii)} NP vs. P\textsubscript{2}, \textit{(iv)} NP vs. P\textsubscript{3}, \textit{(v)} NP vs. P\textsubscript{4}. 
In experiment \textit{(i)}, the goal was to categorize an ECG signal into one of the five pain conditions, while experiments \textit{(ii)-(v)} aimed to classify signals into one of two pain conditions, either no pain or the specified pain level. Furthermore, considering the gender and age of the subjects, we developed four different experimental schemes: 
\textit{(i)} the \textit{Basic Scheme}, utilizing the entire dataset, \textit{(ii)} the \textit{Gender Scheme}, where data were segmented by the gender of the subjects into males and females, (iii) the \textit{Age Scheme}, which grouped subjects into three age categories: \textit{\textquoteleft 20-35\textquoteright}, \textit{\textquoteleft 36-50\textquoteright}, \textit{\textquoteleft 51-65\textquoteright}, and (iv) the \textit{Gender-Age Scheme}, which combined both demographic factors, resulting in six distinct groups: \textit{\textquoteleft Males 20-35\textquoteright}, \textit{\textquoteleft Females 20-35\textquoteright}, \textit{\textquoteleft Males 36-50\textquoteright}, \textit{\textquoteleft Females 36-50\textquoteright}, \textit{\textquoteleft Males 51-65\textquoteright}, \textit{\textquoteleft Females 51-65\textquoteright}. The most successful classification results are displayed in Figures \ref{gender}-\ref{age} for each task and classification method, while Tables \ref{table:basic}-\ref{table:gender_age_females} detail the outcomes of each individual experiment.

\subsection{Results}
\label{results_chapter_4_paper_1}
\begin{figure}
\center
\includegraphics[scale=0.50]{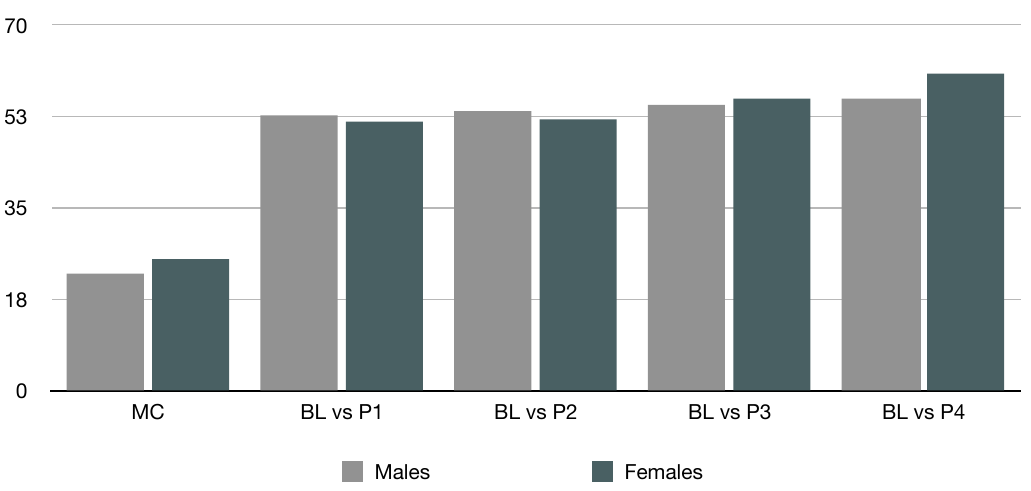}
\caption{Results for the \textit{Gender Scheme}.}
\label{gender}
\end{figure}

Table \ref{table:basic} shows the results from the entire dataset, where the multi-class pain classification achieved a $23.79\%$ accuracy, and performance scores generally increased with pain intensity, peaking at $58.62\%$ for NP vs. P\textsubscript{4}. This progression highlights the difficulty in detecting lower levels of pain severity. Regarding the classification algorithms, SVM (linear) was more effective, except for the highest pain level task, where SVM (RBF) was less successful.
In the \textit{Gender Scheme} (see Table \ref{table:gender}), notable differences were observed between males and females. Overall, females showed a $1.12\%$ higher accuracy variation than males, with females achieving $60.69\%$ in NP vs. P\textsubscript{4} over males' $56.07\%$. This $4.62\%$ increase suggests that females are more sensitive to higher levels of pain than males. Interestingly, in NP vs. P\textsubscript{1} and NP vs. P\textsubscript{2}, males outperformed females by $1.16\%$ and $1.78\%$, respectively. Consistent with the first scheme, SVM (linear) yielded better results in most tasks.
Figure \ref{gender} illustrates the gender differences in classification accuracy.

\renewcommand{\arraystretch}{1.2}
\begin{table}
\center
\small
\begin{threeparttable}
\caption{Results for the \textit{Basic Scheme} (1).}
\label{table:basic} 
\begin{tabular}{ P{0.5cm}  P{1.5cm}  P{2.0cm}  P{2.0cm}  P{2.0cm}}
\toprule
\multirow{2}[3]{*}{\rotatebox{90}{Group}}  &\multirow{2}[3]{*}{Task} &\multicolumn{3}{c}{Algorithm}\\
\cmidrule(lr){3-5}
\ & &LDA  &SVM (LN) &SVM (RBF)\\
\midrule
\midrule
\multirow{5}[1]{*}{\rotatebox{90}{All}}  
&MC                        &23.72 &\cellcolor{mygray}23.79 &22.77\\
&NP vs P\textsubscript{1}  &50.97 &\cellcolor{mygray}52.38 &49.97\\
&NP vs P\textsubscript{2}  &52.55 &\cellcolor{mygray}52.78 &52.70\\
&NP vs P\textsubscript{3}  &55.20 &\cellcolor{mygray}55.37 &53.87\\
&NP vs P\textsubscript{4}  &\cellcolor{mygray}58.62 &58.39 &57.41\\
\bottomrule
\end{tabular}
\begin{tablenotes}
\scriptsize
\item MC: multi-classification \space NP: no pain \space P\textsubscript{1}: mild pain \space P\textsubscript{2}: moderate pain \space P\textsubscript{3}: severe pain \space P\textsubscript{4}: very severe pain \space LDA: Linear Discriminant Analysis \space LN:Linear  \space RBF: Radial Basis Function
\end{tablenotes}
\end{threeparttable}
\end{table}

\renewcommand{\arraystretch}{1.2}
\begin{table}
\center
\small
\begin{threeparttable}
\caption{Results for the \textit{Gender Scheme} (1).}
\label{table:gender}
\begin{tabular}{ P{0.5cm}  P{1.5cm}  P{2.0cm}  P{2.0cm}  P{2.0cm}}
\toprule
\multirow{2}[3]{*}{\rotatebox{90}{Group}} &\multirow{2}[3]{*}{Task} &\multicolumn{3}{c}{Algorithm}\\
\cmidrule(lr){3-5}
\ & & LDA  &SVM (LN) &SVM (RBF)\\
\midrule
\midrule
\multirow{5}[1]{*}{\rotatebox{90}{Males}}  
&MC                       &22.13 &\cellcolor{mygray}22.25 &20.70\\
&NP vs P\textsubscript{1} &51.53 &\cellcolor{mygray}52.61 &47.72\\
&NP vs P\textsubscript{2} &53.12 &\cellcolor{mygray}53.69 &52.15\\
&NP vs P\textsubscript{3} &\cellcolor{mygray}54.94 &54.71 &51.36\\
&NP vs P\textsubscript{4} &55.28 &\cellcolor{mygray}56.07 &51.36\\
\hline
\multirow{5}[1]{*}{\rotatebox{90}{Females}}  
&MC                       &\cellcolor{mygray}25.11 &24.41 &23.41\\
&NP vs P\textsubscript{1} &50.23 &\cellcolor{mygray}51.45 &49.06\\
&NP vs P\textsubscript{2} &51.62 &51.86 &\cellcolor{mygray}51.91\\
&NP vs P\textsubscript{3} &\cellcolor{mygray}55.98 &55.87 &55.29\\
&NP vs P\textsubscript{4} &60.17 &\cellcolor{mygray}60.69 &59.82\\
\bottomrule
\end{tabular}
\end{threeparttable}
\end{table}

\begin{figure}
\center
\includegraphics[scale=0.50]{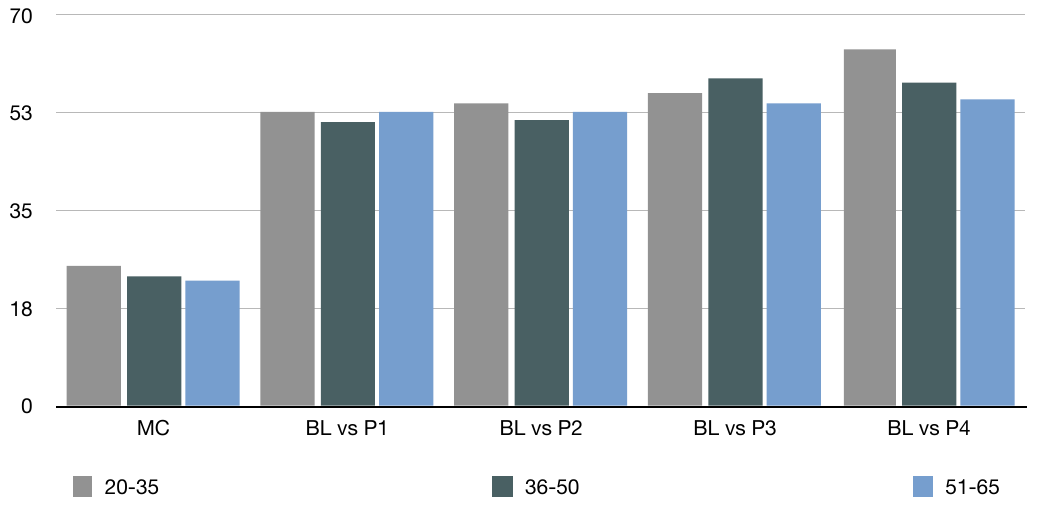}
\caption{Results for the \textit{Age Scheme}.}
\label{age}
\end{figure}

In the \textit{Age-Scheme} (refer to Table \ref{table:age}), the \textit{\textquoteleft 20-35\textquoteright}\space age group achieved $25.06\%$ accuracy in multi-level classification, compared to $23.27\%$ and $22.35\%$ for the \textit{\textquoteleft 36-50\textquoteright}\space and \textit{\textquoteleft 51-65\textquoteright}\space groups, respectively, indicating that age significantly affects pain perception. The nearly $9\%$ difference in the NP vs P\textsubscript{4} task between the youngest and oldest groups was particularly notable. Similar to the gender-based results, minor differences in low pain intensities among the age groups became more pronounced as pain intensity increased. Specifically, the variance ($\sigma^2$) between the groups in NP vs. P\textsubscript{1} was $1.38\%$. At the same time, in the other tasks, it increased to $2.44\%$, $6.35\%$, and $20.42\%$, respectively, showing that high pain intensities are necessary to discern perceptual differences among age groups.
Regarding classification accuracy, the \textit{\textquoteleft 20-35\textquoteright}\space group showed the highest sensitivity, followed by \textit{\textquoteleft 36-50\textquoteright}\space and \textit{\textquoteleft 51-65\textquoteright}. Regarding classification methods, the SVM (RBF) performed best in the \textit{\textquoteleft 51-65\textquoteright} group across almost all tasks, while it underperformed in the \textit{\textquoteleft 20-35\textquoteright}\space group, suggesting it is better suited for more challenging, separable classes.
Figure \ref{age} displays the results from the age scheme.

In the final analysis, we examined the subjects more closely to gain deeper insights into the relationship between pain and the demographic factors of gender and age. As shown in Tables \ref{table:gender_age_males}-\ref{table:gender_age_females}, the \textit{\textquoteleft Females 20-35\textquoteright}\space group achieved the highest accuracy in multi-class pain classification at $24.80\%$, while \textit{\textquoteleft Females 51-65\textquoteright}\space led in NP vs. P\textsubscript{1} with $55.38\%$, again indicating higher pain sensitivity in females. Moreover, \textit{\textquoteleft Females 51-65\textquoteright}\space followed by \textit{\textquoteleft Males 51-65\textquoteright}\space topped the performance in NP vs. P\textsubscript{2}, and in NP vs. P\textsubscript{3}, \textit{\textquoteleft Females 36-50\textquoteright}\space surpassed the next best group, \textit{\textquoteleft Males 20-35\textquoteright}, by $3.5\%$. In the final NP vs. P\textsubscript{4} task, \textit{\textquoteleft Females 20-35\textquoteright}\space excelled with $67\%$ accuracy, whereas \textit{\textquoteleft Males 51-65\textquoteright}\space had the lowest at $54.50\%$, marking them as the most and least pain-sensitive groups, respectively. It is noted that sometimes classification accuracy decreases despite increased pain levels (\textit{e.g.}, \textit{\textquoteleft Females 36-50\textquoteright}). This might be attributed to the subjects becoming accustomed to the stimulus over time during the biosignal recording.

\begin{figure}
\center
\includegraphics[scale=0.70]{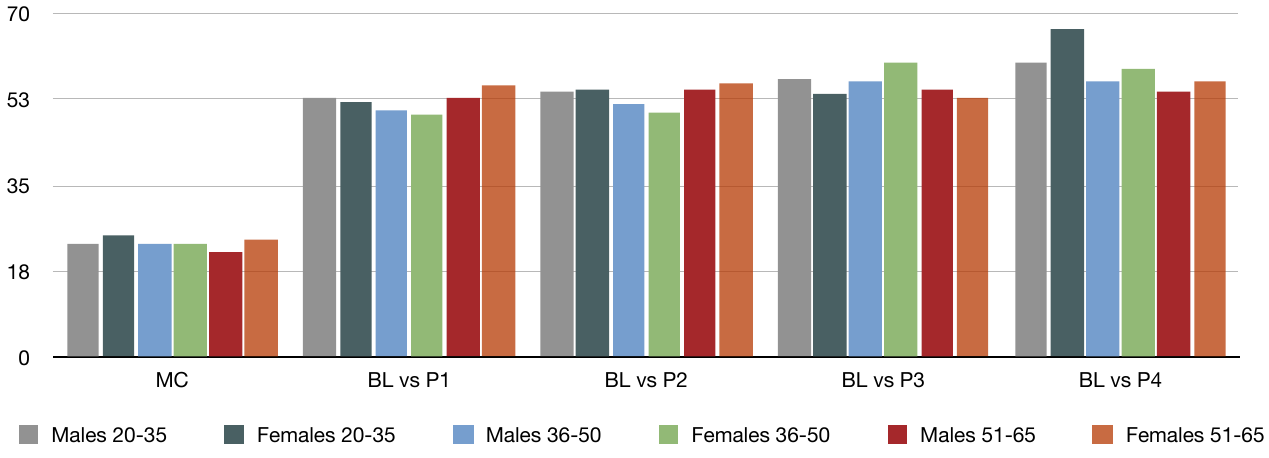}
\caption{Results for the \textit{Gender-Age Scheme}.}
\label{gender_age}
\end{figure}

Figure \ref{gender_age} illustrates the classification performance of the six groups in the \textit{Gender-Age Scheme}. Additionally, Table \ref{table:comparison} compares our results with other studies that used ECG signals from the \textit{BioVid} database and followed the same evaluation protocol, ensuring an objective comparison. Our study achieved the best classification performance in both the multi-class setting and the NP vs. P\textsubscript{1} and NP vs. P\textsubscript{2} tasks, with acceptable results in the remaining binary classification tasks.

\renewcommand{\arraystretch}{1.2}
\begin{table}
\center
\small
\begin{threeparttable}
\caption{Results for the \textit{Age Scheme} (1).}
\label{table:age}
\begin{tabular}{ P{0.5cm}  P{1.5cm}  P{2.0cm}  P{2.0cm}  P{2.0cm}}
\toprule
\multirow{2}[3]{*}{\rotatebox{90}{Group}} &\multirow{2}[3]{*}{Task} &\multicolumn{3}{c}{Algorithm}\\
\cmidrule(lr){3-5}
\ & & LDA  &SVM (LN) &SVM (RBF)\\
\midrule
\midrule
\multirow{5}[1]{*}{\rotatebox{90}{20-35}}  
&MC                        &\cellcolor{mygray}25.06 &24.73 &21.96\\
&NP vs P\textsubscript{1}  &\cellcolor{mygray}52.83 &\cellcolor{mygray}52.83 &49.90\\
&NP vs P\textsubscript{2}  &\cellcolor{mygray}54.33 &53.75 &52.75\\
&NP vs P\textsubscript{3}  &55.58 &\cellcolor{mygray}56.16 &54.66\\
&NP vs P\textsubscript{4}  &\cellcolor{mygray}63.83 &63.41 &60.75\\
\hline
\multirow{5}[1]{*}{\rotatebox{90}{36-50}}  
&MC        &\cellcolor{mygray}23.27 &22.06 &23.03\\
&NP vs P\textsubscript{1}  &50.34 &48.36 &\cellcolor{mygray}50.68\\
&NP vs P\textsubscript{2}  &49.13 &\cellcolor{mygray}51.20 &50.17\\
&NP vs P\textsubscript{3}  &58.10 &\cellcolor{mygray}58.70 &58.27\\
&NP vs P\textsubscript{4}  &\cellcolor{mygray}58.10 &57.75 &55.94\\
\hline
\multirow{5}[1]{*}{\rotatebox{90}{51-65}}  
&MC        &21.89 &22.07 &\cellcolor{mygray}22.35\\
&NP vs P\textsubscript{1}  &52.23 &51.87 &\cellcolor{mygray}52.58\\
&NP vs P\textsubscript{2}  &52.14 &51.69 &\cellcolor{mygray}52.76\\
&NP vs P\textsubscript{3}  &\cellcolor{mygray}53.66 &53.39 &54.10\\
&NP vs P\textsubscript{4}  &54.46 &54.19 &\cellcolor{mygray}54.91\\
\bottomrule
\end{tabular}
\end{threeparttable}
\end{table}

\renewcommand{\arraystretch}{1.2}
\begin{table}
\center
\small
\begin{threeparttable}
\caption{Results for the \textit{Gender-Age Scheme} (Males) (1).}
\label{table:gender_age_males}
\begin{tabular}{ P{0.5cm}  P{1.5cm}  P{2.0cm}  P{2.0cm}  P{2.0cm}}
\toprule
\multirow{2}[3]{*}{\rotatebox{90}{Group}} &\multirow{2}[3]{*}{Task} &\multicolumn{3}{c}{Algorithm}\\
\cmidrule(lr){3-5}
\ & & LDA  &SVM (LN) &SVM (RBF)\\
\midrule
\midrule
\multirow{5}[1]{*}{\rotatebox{90}{Males 20-35}}  
&MC                        &23.13 &\cellcolor{mygray}23.20 &18.73\\
&NP vs P\textsubscript{1}  &52.50 &\cellcolor{mygray}52.83 &45.83\\
&NP vs P\textsubscript{2}  &\cellcolor{mygray}54.00 &53.50 &53.16\\
&NP vs P\textsubscript{3}  &56.33 &\cellcolor{mygray}56.50 &54.83\\
&NP vs P\textsubscript{4}  &\cellcolor{mygray}60.00 &59.00 &53.66\\
\hline

\multirow{5}[1]{*}{\rotatebox{90}{Males 36-50}}  
&MC        &\cellcolor{mygray}23.21 &22.21 &20.92\\
&NP vs P\textsubscript{1}  &\cellcolor{mygray}50.53 &\cellcolor{mygray}50.53 &46.42\\
&NP vs P\textsubscript{2}  &50.00 &\cellcolor{mygray}51.78 &47.50\\
&NP vs P\textsubscript{3}  &54.64 &\cellcolor{mygray}56.25 &47.32\\
&NP vs P\textsubscript{4}  &55.53 &\cellcolor{mygray}56.25 &51.96\\
\hline

\multirow{5}[1]{*}{\rotatebox{90}{Males 51-65}}  
&MC        &20.06 &\cellcolor{mygray}21.60 &19.60\\
&NP vs P\textsubscript{1}  &\cellcolor{mygray}52.66 &51.66 &50.66\\
&NP vs P\textsubscript{2}  &54.00 &\cellcolor{mygray}54.66 &51.50\\
&NP vs P\textsubscript{3}  &53.00 &\cellcolor{mygray}54.66 &51.50\\
&NP vs P\textsubscript{4}  &53.33 &\cellcolor{mygray}54.50 &49.83\\
\bottomrule
\end{tabular}
\end{threeparttable}
\end{table}

\renewcommand{\arraystretch}{1.2}
\begin{table}
\center
\small
\begin{threeparttable}
\caption{Results for the \textit{Gender-Age Scheme} (Females) (1).}
\label{table:gender_age_females}
\begin{tabular}{ P{0.5cm}  P{1.5cm}  P{2.0cm}  P{2.0cm}  P{2.0cm}}
\toprule
\multirow{2}[3]{*}{\rotatebox{90}{Group}} &\multirow{2}[3]{*}{Task} &\multicolumn{3}{c}{Algorithm}\\
\cmidrule(lr){3-5}
\ & & LDA  &SVM (LN) &SVM (RBF)\\
\midrule
\midrule
\multirow{5}[1]{*}{\rotatebox{90}{Females 20-35}}  
&MC                        &24.73 &\cellcolor{mygray}24.80 &23.26\\ 
&NP vs P\textsubscript{1}  &49.83 &51.50 &\cellcolor{mygray}52.00\\
&NP vs P\textsubscript{2}  &\cellcolor{mygray}54.50 &53.66 &46.50\\
&NP vs P\textsubscript{3}  &\cellcolor{mygray}53.50 &52.83 &49.00\\
&NP vs P\textsubscript{4}  &65.83 &\cellcolor{mygray}67.00 &62.16\\
\hline
\multirow{5}[1]{*}{\rotatebox{90}{Females 36-50}}
&MC                        &\cellcolor{mygray}23.06 &22.73 &21.93\\
&NP vs P\textsubscript{1}  &48.16 &\cellcolor{mygray}49.33 &48.33\\
&NP vs P\textsubscript{2}  &48.66 &\cellcolor{mygray}49.83 &47.83\\
&NP vs P\textsubscript{3}  &57.50 &\cellcolor{mygray}60.00 &55.00\\
&NP vs P\textsubscript{4}  &\cellcolor{mygray}59.00 &58.83 &56.16\\
\hline
\multirow{5}[1]{*}{\rotatebox{90}{Females 51-65}}
&MC                        &21.23 &21.84 &\cellcolor{mygray}23.92\\
&NP vs P\textsubscript{1}  &48.84 &49.80 &\cellcolor{mygray}55.38\\
&NP vs P\textsubscript{2}  &51.15 &48.65 &\cellcolor{mygray}55.96\\
&NP vs P\textsubscript{3}  &\cellcolor{mygray}53.07 &\cellcolor{mygray}53.07 &50.96\\
&NP vs P\textsubscript{4}  &52.69 &55.00 &\cellcolor{mygray}56.34\\
\bottomrule
\end{tabular}
\end{threeparttable}
\end{table}

\renewcommand{\arraystretch}{1.2}
\begin{table}
\center
\small
\begin{threeparttable}
\caption{Comparison of studies utilizing \textit{BioVid}, ECG signals \\ and LOSO validation (1).}
\label{table:comparison}
\begin{tabular}{P{4.0cm} P{1.6cm} P{1.1cm}}
\toprule
Method  &Task  &Results\\
\midrule
\midrule
Martinez and Picard \cite{martinez_picard_2018} &NP vs P\textsubscript{4}  &57.69\\
\hline

\multirow{4}[1]{*}{Werner \textit{et al.} \cite{werner_hamadi_2014}} 
&NP vs P\textsubscript{1} &48.70\\
&NP vs P\textsubscript{2} &51.60\\
&NP vs P\textsubscript{3} &\cellcolor{mygray}56.50\\
&NP vs P\textsubscript{4} &\cellcolor{mygray}62.00\\
\hline

\multirow{5}[1]{*}{Thiam \textit{et al.} \cite{thiam_bellmann_kestler_2019}}
&MC                       &23.23\\
&NP vs P\textsubscript{1} &49.71\\
&NP vs P\textsubscript{2} &50.72\\
&NP vs P\textsubscript{3} &52.87\\
&NP vs P\textsubscript{4} &57.04\\
\hline

\multirow{5}[1]{*}{Ours} 
&MC                       &\cellcolor{mygray}23.79\\
&NP vs P\textsubscript{1} &\cellcolor{mygray}52.38\\
&NP vs P\textsubscript{2} &\cellcolor{mygray}52.78\\
&NP vs P\textsubscript{3} &55.37\\
&NP vs P\textsubscript{4} &58.62\\       
\bottomrule
\end{tabular}
\end{threeparttable}
\end{table}

\subsection{Discussion}
We analyzed ECG biosignals using the \textit{Pan-Tompkins} algorithm to detect QRS complexes and extracted features about inter-beat intervals. We also evaluated three machine learning techniques, assessing their performance in multi-class and binary pain classification across various pain intensities. We also examined the influence of gender and age on pain perception, discovering significant differences: males generally showed lower sensitivity to high pain levels. 
Regarding the age factor, significant variations suggest that pain sensitivity tends to diminish with age, potentially increasing the risk of further injury. In certain demographic groups, the difference in pain perception exceeded $12\%$, underscoring the variability of pain sensation among individuals.


\section{ECG Analysis with Multitask Neural Networks}
In this section, we build on previous analysis \ref{chapter_4_paper_1} that explored variations in pain manifestation across different demographic groups using ECG signals. It expands this investigation by implementing neural networks as the primary machine learning model and introduces a novel multi-task learning (MTL) neural network. This network leverages demographic information to estimate age and gender in addition to pain levels, aiming to enhance the automatic pain estimation system.

\subsection{Methodology}
\label{paper_2_feature_extraction}
The following method we developed is a neural network-based approach. The feature extraction process remains the same as previously described in \ref{pan_tompkins}, utilizing the \textit{Pan-Tompkins} algorithm for ECG signal processing.

\subsubsection{Neural Network}
The proposed neural network was designed and trained using two distinct approaches: single-task learning (STL) and multi-task learning (MTL). In the multi-task learning framework, the network is simultaneous training for pain estimation and predicting age and/or gender.

\paragraph{Single-Task Neural Network:}
The proposed neural network comprises two components: the encoder, which maps the original feature vectors into a higher dimensional space, and the task-specific classifier. In our design, both the encoder and the classifier utilize fully-connected (FC) layers, which are defined as follows:
\begin{equation}
z_{i}(s)=b_{i}+\sum_{j=1}^{n_{in}}W_{ij}s_{j} \qquad \textrm{for} \quad i=1,..,n_{out},
\end{equation}
where $z_{i}$ represents the result of linearly combining the incoming inputs $s_{j}$, with each input being weighted by $W_{ij}$ and adjusted by a bias $b_{i}$. Each layer in the encoder is followed by a nonlinear activation function, specifically the rectified linear unit (ReLU), which is defined as:
\begin{equation}
\sigma(z)=
	\begin{cases}
	1, \quad z\geq0 \\
	0, \quad z<0
	\end{cases}
\end{equation}
The classifier's layers are connected without nonlinearity, the encoder comprises four fully connected (FC) layers with $256$, $512$, $1024$, and $1024$ neurons respectively. The classifier includes $2$ layers with $1024$ and $n$ neurons, where $n$ represents the number of distinct pain classes being classified. The hyperparameters of the network are detailed in Table \ref{table:parameters}.

\begin{table}
\center
\small
\renewcommand{\arraystretch}{1.2}
\begin{threeparttable}
\caption{Hyper-parameters used in our approach.} 
\label{table:parameters}
\begin{tabular}{P{1.2cm} P{1.6cm} P{1.6cm} P{1.1cm} P{1.4cm} P{1.5cm} P{1.4cm} P{1.1cm}}
\toprule
Epochs  &Optimizer  &Learning rate &LR decay &Weight decay &Warmup epochs &Label smooth &EMA\\
\midrule
\midrule
300 &\textit{AdamW} &1e-3 &\textit{cosine} &0.1 &50 &0.1 &$\checkmark$\\
\bottomrule
\end{tabular}
\end{threeparttable}
\end{table}

\begin{figure}
\center
\includegraphics[scale=0.76]{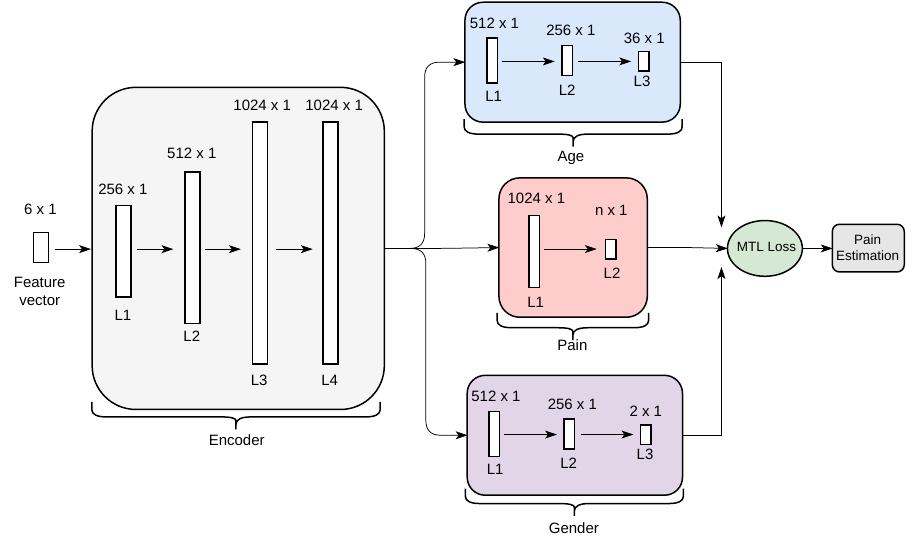}
\caption{The proposed MTL network: The sizes of the extracted vectors for the network are as follows: for the Pain classifier, \(n \times 1\), where \(n\) is the number of pain estimation tasks (\textit{e.g.}, $2$ for binary classification, $5$ for multi-class classification); for the Age classifier, \(36 \times 1\), where $36$ represents the possible age values of the subjects; for the Gender classifier, \(2 \times 1\), corresponding to the two possible gender categories (\textit{i.e.}, males and females).}
\label{mtl}
\end{figure}

\paragraph{Multi-task neural network:}
\label{mt_nn}
This proposed machine learning method is based on the principle of sharing representations across related tasks, which helps the model better generalize to the primary task of pain estimation in this case. 
We kept the same encoder and pain classifier in this configuration but introduced two additional auxiliary networks for age and gender estimation. The architecture of the proposed multi-task learning (MTL) neural network is illustrated in Fig. \ref{mtl}. 
The objective of this network is to simultaneously minimize three different losses. We adopt and expand upon the framework suggested by \cite{kendall_2018} for the multi-task learning loss, where learned weights are applied to each loss function based on the homoscedastic uncertainty of each task:
\begin{equation}
L_{total}= [e^{w1}L_{Pain}+w_{1}]c_{1} + [e^{w2}L_{Age}+w_{2}]c_{2} + [e^{w3}L_{Gender}+w_{3}]c_{3}.
\end{equation}
Here, \(L\) represents the corresponding loss, \(w\) denotes the weights, and \(c\) are the coefficients that modulate the losses \(L_{Age}\) and \(L_{Gender}\) to prioritize learning in the pain estimation task. It should be noted that all tasks are treated as classification problems, utilizing \textit{cross-entropy} loss with label smoothing:
\begin{equation}
L_{D} = -\sum_{\delta \in D}\sum^{n_{out}}_{i=1}p(i|x_\delta)\log[q(i|x_\delta)].
\end{equation}
Here, \(D\) denotes the pain database, \(p(i|x_\delta) = 1 - \epsilon\) represents the probability of the true class \(i\) given the input \(x_\delta\), and \(p(i \neq i_\delta|x_\delta) = \epsilon / (n_{out} - 1)\) is the probability distribution across the other classes. This formulation spreads a small portion \(\epsilon\) of the probability across classes other than the true class to implement label smoothing. Furthermore, \(q(i|x_\delta)\) is the probability distribution over the classes \(i\) as predicted by the network's output.

\subsection{Experiments}
\label{experiments_results}
Similar to \ref{demographic_experiments_paper_1}, we utilized the \textit{BioVid} database, specifically focusing on its ECG signals. Employing a single-task neural network (ST-NN), we conducted an initial series of experiments to assess the impact of demographic factors. Building on the concept we proposed in the previous section, we devised five experimental schemes: \textit{(i)} the \textit{Basic Scheme}, which included all subjects from the database; \textit{(ii)} the \textit{Gender Scheme}, which segregated subjects into male and female groups; \textit{(iii)} the \textit{Age Scheme}, which categorized subjects into three age groups---\textit{\textquoteleft 20-35\textquoteright}, \textit{\textquoteleft 36-50\textquoteright}, and \textit{\textquoteleft 51-65\textquoteright}; and \textit{(iv)} the \textit{\textquoteleft Gender-Age Scheme\textquoteright}, which combined both demographic factors, resulting in six distinct groups: \textit{\textquoteleft Males 20-35\textquoteright}, \textit{\textquoteleft Females 20-35\textquoteright}, \textit{\textquoteleft Males 36-50\textquoteright}, \textit{\textquoteleft Females 36-50\textquoteright}, \textit{\textquoteleft Males 51-65\textquoteright}, and \textit{\textquoteleft Females 51-65\textquoteright}. All experiments were conducted in both binary and multi-class classification formats. Specifically, the binary classification tasks were \textit{(i)} NP vs. P\textsubscript{1}, \textit{(ii)} NP vs. P\textsubscript{2}, \textit{(iii)} NP vs. P\textsubscript{3}, and (4) NP vs. P\textsubscript{4}, and the multi-class task utilized all available pain classifications from the database.

\subsection{Results}
\subsubsection{Demographic Groups}
\begin{figure}
\centering
\begin{subfigure}{\textwidth}
\centering
\includegraphics[scale=0.38]{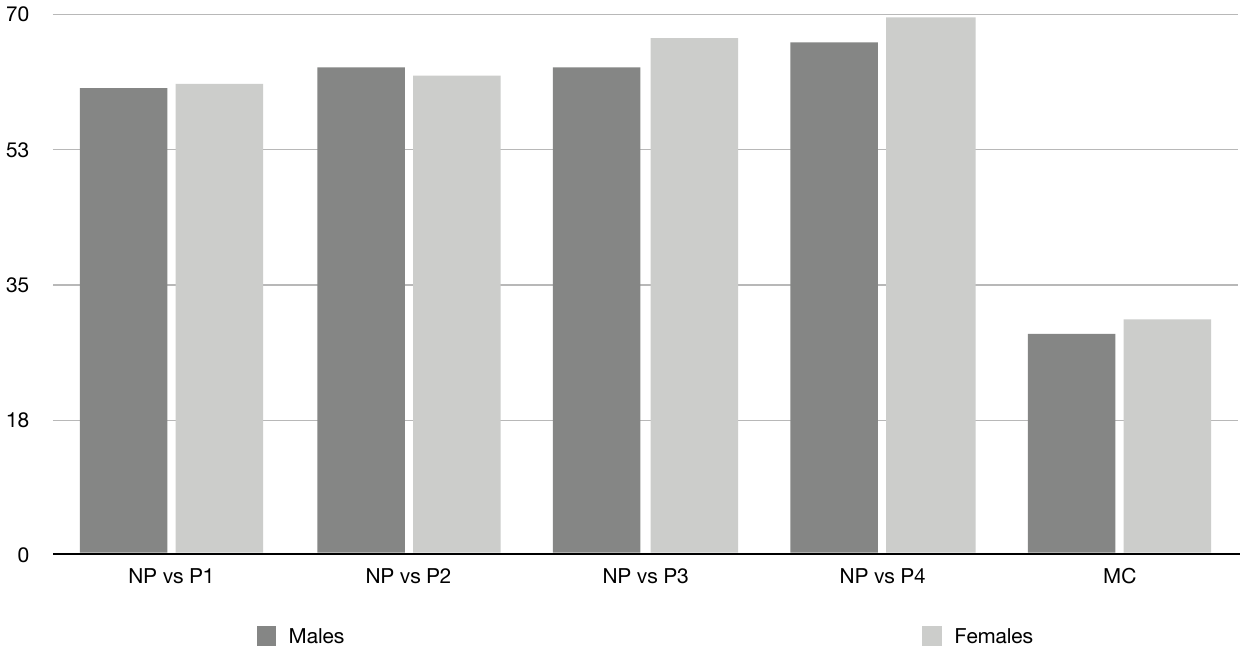}
\caption{Gender}
\label{2_gender}
\end{subfigure}
\par\bigskip

\begin{subfigure}{\textwidth}
\centering
\includegraphics[scale=0.38]{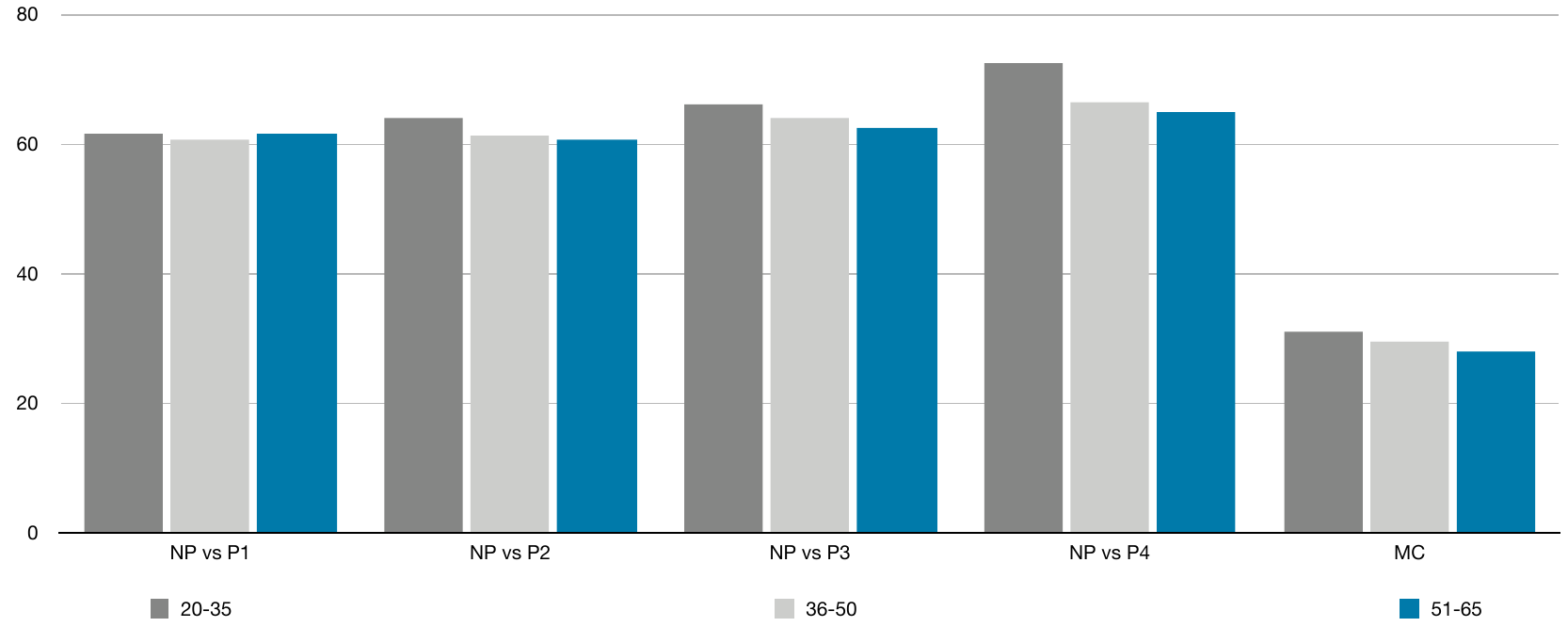}
\caption{Age}
\label{2_age}
\end{subfigure}
\par\bigskip

\begin{subfigure}{\textwidth}
\centering
\includegraphics[scale=0.38]{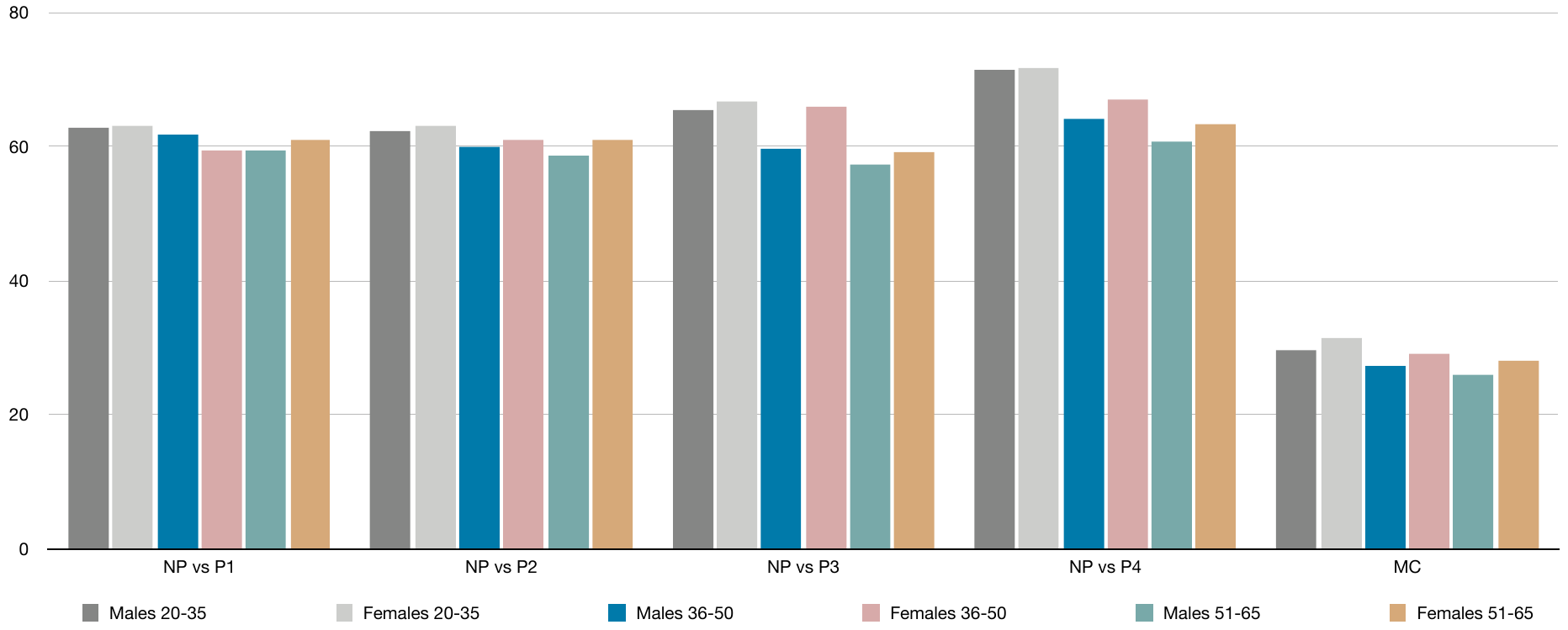}
\caption{Gender-Age}
\label{2_gender_age}
\end{subfigure}

\caption{Results for the proposed Schemes.}
\label{schemes}
\end{figure}

Table \ref{table:2_basic} presents the classification results of the \textit{Basic Scheme}, which utilized all subjects in the database. For the multi-class pain classification, we achieved an accuracy of $29.43\%$, with NP vs. P\textsubscript{1} scoring $61.15\%$ and NP vs. P\textsubscript{4} reaching $68.82\%$. These results indicate that as pain intensity increases, so performs, highlighting the difficulty in recognizing less severe pain.
According to the \textit{Gender Scheme} (refer to Table \ref{table:2_gender}), notable differences emerge between males and females, particularly at higher pain intensities. Specifically, in NP vs. P\textsubscript{4}, females achieved an accuracy of $69.48\%$ compared to $66.48\%$ for males, with an overall variance of $1.63\%$ between genders, suggesting that females exhibit higher pain sensitivity. Figure \ref{2_gender} illustrates these gender-based classification disparities.
In the \textit{Age Scheme} (see Table \ref{table:2_age}), the \textit{\textquoteleft 20-35\textquoteright}\space age group outperformed the \textit{\textquoteleft 36-50\textquoteright}\space and \textit{\textquoteleft 51-65\textquoteright}\space groups in NP vs. P\textsubscript{4}, with accuracies of $72.58\%$, $66.29\%$, and $64.91\%$, respectively. While the differences are less pronounced at lower pain intensities, this scheme still shows that age significantly impacts pain perception, particularly among the older population. Figure \ref{2_age} shows the results from the age scheme.

In the final scheme, by dividing subjects into more specific groups, we can analyze them more precisely and gain better insights into the relationship between gender, age, and pain perception. Table \ref{table:2_gender_age} reveals that in the NP vs. P\textsubscript{4} task, the group \textit{\textquoteleft Females 20-35\textquoteright}\space reached the highest accuracy of $71.67\%$, significantly outperforming the \textit{\textquoteleft Males 51-65\textquoteright}\space group, which scored the lowest at $60.67\%$, marking them as the least sensitive group. This pattern is consistent across the multi-class classification and other pain tasks, with \textit{\textquoteleft Females 20-35\textquoteright}\space and \textit{\textquoteleft Males 51-65\textquoteright}\space exhibiting the highest and lowest accuracies, respectively. This supports that females generally experience more pronounced pain responses, while older males have a reduced pain sensation. It is noted that in some instances, such as with \textit{\textquoteleft Males 20-35\textquoteright}\space and \textit{\textquoteleft Males 36-50\textquoteright}, higher pain levels do not necessarily correlate with higher classification accuracy, a phenomenon also noted in our previous experiments, in Section \ref{results_chapter_4_paper_1}. A possible explanation could be the subjects' habituation to pain stimuli, especially at lower intensities. Figure \ref{2_gender_age} visualizes the performance outcomes of the \textit{Gender-Age Scheme}.

\renewcommand{\arraystretch}{1.2}
\begin{table}
\center
\small
\begin{threeparttable}
\caption{Results for the \textit{Basic Scheme} (2).}
\label{table:2_basic}
\begin{tabular}{ P{1.0cm} P{1.7cm} P{1.5cm} P{1.5cm} P{1.5cm} P{1.5cm}  P{1.5cm}}
\toprule
\multirow{2}[3]{*}{Group}  &\multirow{2}[3]{*}{Algorithm} &\multicolumn{5}{c}{Task}\\
\cmidrule(lr){3-7}
\ & &NP vs P\textsubscript{1}  &NP vs P\textsubscript{2} &NP vs P\textsubscript{3} &NP vs P\textsubscript{4} &MC\\
\midrule
\midrule
All &ST-NN &61.15 &62.87 &65.14 &68.82 &29.43 \\
\bottomrule
\end{tabular}
\begin{tablenotes}
\scriptsize
\item ST-NN: single-task neural network \space NP: no pain \space P1: mild pain \space P2: moderate pain \space P3: severe pain \space P4: very severe pain \space MC: multi-classification 
\end{tablenotes}
\end{threeparttable}
\end{table}

\renewcommand{\arraystretch}{1.2}
\begin{table}
\center
\small
\begin{threeparttable}
\caption{Results for the \textit{Gender Scheme} (2).}
\label{table:2_gender}   
\begin{tabular}{ P{1.3cm} P{1.7cm}  P{1.5cm}  P{1.5cm}  P{1.5cm}  P{1.5cm}  P{1.5cm}}
\toprule
\multirow{2}[3]{*}{Group}  &\multirow{2}[3]{*}{Algorithm} &\multicolumn{5}{c}{Task}\\
\cmidrule(lr){3-7}
\ & &NP vs P\textsubscript{1}  &NP vs P\textsubscript{2} &NP vs P\textsubscript{3} &NP vs P\textsubscript{4} &MC\\
\midrule
\midrule
Males &ST-NN &60.40 &63.24 &63.18 &66.48 &28.61\\
Females &ST-NN &60.87 &62.15 &66.98 &69.48 &30.59\\
\bottomrule
\end{tabular}
\end{threeparttable}
\end{table}

\renewcommand{\arraystretch}{1.2}
\begin{table}
\center
\small
\begin{threeparttable}
\caption{Results for the \textit{Age Scheme} (2).}
\label{table:2_age}    
\begin{tabular}{ P{1.3cm} P{1.7cm}  P{1.5cm} P{1.5cm} P{1.5cm}  P{1.5cm} P{1.5cm}}
\toprule
\multirow{2}[3]{*}{Group}  &\multirow{2}[3]{*}{Algorithm} &\multicolumn{5}{c}{Task}\\
\cmidrule(lr){3-7}
\ & &NP vs P\textsubscript{1}  &NP vs P\textsubscript{2} &NP vs P\textsubscript{3} &NP vs P\textsubscript{4} &MC\\
\midrule
\midrule
20-35 &ST-NN &61.58 &64.08 &66.08 &72.58 &31.07\\
36-50 &ST-NN &60.52 &61.38 &64.05 &66.29 &29.59\\
51-65 &ST-NN &61.70 &60.80 &62.50 &64.91 &27.82\\
\bottomrule
\end{tabular}
\end{threeparttable}
\end{table}

\renewcommand{\arraystretch}{1.2}
\begin{table}
\center
\small
\begin{threeparttable}
\caption{Results for the \textit{Gender-Age} Scheme (2).}
\label{table:2_gender_age}
    
\begin{tabular}{ P{2.5cm}  P{1.7cm}  P{1.5cm}  P{1.5cm}  P{1.5cm}  P{1.5cm}  P{1.5cm}}
\toprule
\multirow{2}[3]{*}{Group}  &\multirow{2}[3]{*}{Algorithm} &\multicolumn{5}{c}{Task}\\
\cmidrule(lr){3-7}
\ & &NP vs P\textsubscript{1}  &NP vs P\textsubscript{2} &NP vs P\textsubscript{3} &NP vs P\textsubscript{4} &MC\\
\midrule
\midrule
Males   20-35 &ST-NN &62.83 &62.33 &65.50 &71.33 &29.73\\
Males   36-50 &ST-NN &61.79 &60.00 &59.64 &64.11 &27.14\\
Males   51-65 &ST-NN &59.50 &58.67 &57.33 &60.67 &26.07\\
Females 20-35 &ST-NN &63.17 &63.17 &66.83 &71.67 &31.53\\
Females 36-50 &ST-NN &59.50 &61.00 &65.83 &67.00 &29.13\\
Females 51-65 &ST-NN &60.96 &60.96 &59.23 &63.27 &27.69\\
\bottomrule
\end{tabular}
\end{threeparttable}
\end{table}

\subsubsection{Augmentation of Feature Vectors}
Building on the findings from the previous experiments about the impact of demographic factors on pain perception, we explored the practical use of subjects' demographic data. Experiments were conducted using the Single-Task Neural Network (ST-NN) and feature vectors enhanced with demographic attributes. Initially, the feature vectors, which originally consisted of six features (see \ref{paper_2_feature_extraction}), were augmented by adding either one additional feature (\textit{i.e.}, the subject's gender or age) or two additional features (both the subject's gender and age).
We conducted the same pain estimation tasks using this enhanced set of features. As shown in Table \ref{table:2_augm}, the results demonstrate improved performance with the augmented feature vectors. Specifically, the most effective augmentation involved combining gender and age features, which increased the average pain estimation performance by $0.55\%$. Using these demographic features individually also improved classification accuracy, albeit partially.

\renewcommand{\arraystretch}{1.2}
\begin{table}
\center
\small
\begin{threeparttable}
\caption{Comparison of results adopting the feature augmentation approach.}
\label{table:2_augm}
\begin{tabular}{ P{1.0cm}  P{1.7cm}  P{1.1cm}  P{1.5cm}  P{1.5cm}   P{1.5cm}  P{1.5cm}  P{1.5cm}}
\toprule
\multirow{2}[3]{*}{Group}  &\multirow{2}[3]{*}{Algorithm} &\multirow{2}[3]{*}{Aux.} &\multicolumn{5}{c}{Task}\\
\cmidrule(lr){4-8}
\ & & &NP vs P\textsubscript{1}  &NP vs P\textsubscript{2} &NP vs P\textsubscript{3} &NP vs P\textsubscript{4} &MC\\
\midrule
\midrule
All &ST-NN &- &61.15 &62.87 &65.14 &68.82 &29.43 \\
\hline
All &ST-NN &F(G)  &\cellcolor{mygray}61.44 &63.19 &65.00 &68.79 &29.68\\
All &ST-NN &F(A)  &61.21 &62.67 &65.66 &\cellcolor{mygray}69.57 &29.71\\
All &ST-NN &F(GA) &61.09 &\cellcolor{mygray}63.48 &\cellcolor{mygray}66.21 &69.54 &\cellcolor{mygray}29.86\\
\bottomrule
\end{tabular}
\begin{tablenotes}
\scriptsize
\item Aux: Auxiliary information \space -: original feature vectors \space F(G): feature vectors with the additional feature of gender \space F(A): feature vectors with the additional feature of age \space F(GA): feature vectors with the additional features of gender and age
\end{tablenotes}
\end{threeparttable}
\end{table}

\subsubsection{Multi-Task Neural Network} 
The final experiments utilized a multi-task learning framework with the proposed Multi-Task Neural Network (MT-NN) outlined in \ref{mt_nn}. The classification results for MT-NN, incorporating additional tasks of (1) gender estimation, (2) age estimation, and (3) simultaneous gender and age estimation, are detailed in Table \ref{table:2_mtl}. For comparison, the results from earlier experiments using the Single-Task Neural Network (ST-NN) method are also included in Table \ref{table:2_mtl}.
We noted that the task of gender estimation alone performed less effectively than the other tasks. In contrast, the combined gender and age estimation delivered the highest performance across four tasks. Specifically, in the multi-class classification, it achieved $30.24\%$, and in NP vs. P\textsubscript{1}, it reached $62.8\%$, marking the best results of any method presented in this study. In NP vs. P\textsubscript{3} and NP vs. P\textsubscript{4}, the combined tasks outperformed the individual gender and age tasks but were slightly less effective than the ST-NN approaches using augmented features.
Interestingly, in NP vs P\textsubscript{2}, the age estimation task alone excelled, achieving $63.97\%$, the highest result recorded in this study.

\begin{figure}
\centering
\begin{subfigure}{\textwidth}
\centering
\includegraphics[scale=0.60]{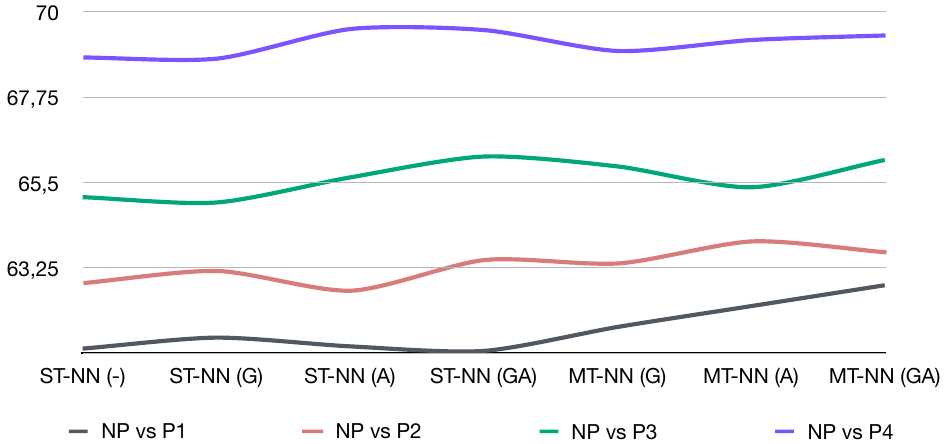}
\caption{Binary classification}
\label{2_np}
\end{subfigure}
\par\bigskip
     
\begin{subfigure}{\textwidth}
\centering
\includegraphics[scale=0.60]{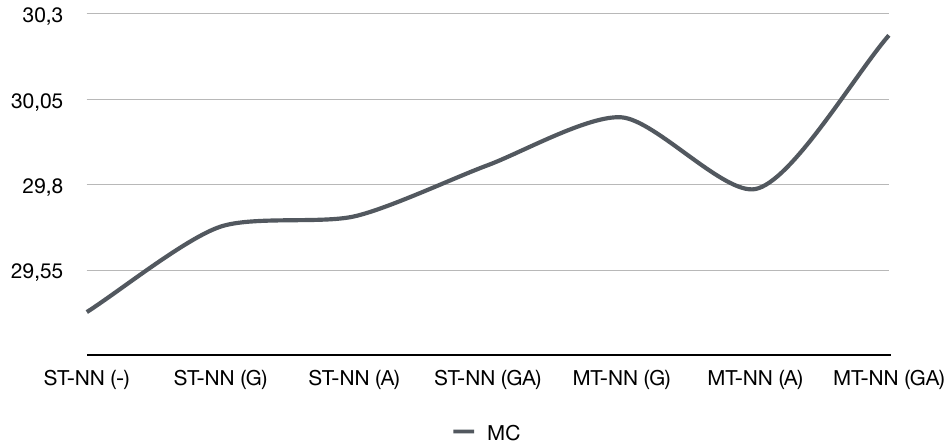}
\caption{Multi-class classification}
\label{2_mc}
\end{subfigure}

\caption{Comparison of performances utilizing various neural networks approaches.}
\label{2_all_nns}
\end{figure}

Comparing the overall performances of MT-NN with the ST-NN approaches (using both original and augmented feature vectors), there is a noticeable improvement of $0.71\%$ and $0.39\%$, respectively, in average pain estimation accuracy across all tasks. Figure \ref{2_all_nns} visually compare each neural network approach used in this study, encompassing multi-class and binary classification tasks.

\renewcommand{\arraystretch}{1.2}
\begin{table}
\center
\small
\begin{threeparttable}
\caption{Comparison of results adopting the MT-NN approach.}
\label{table:2_mtl}  
\begin{tabular}{ P{1.0cm}  P{1.7cm}  P{1.1cm}   P{1.5cm}  P{1.5cm}  P{1.5cm}  P{1.5cm} P{1.5cm}}
\toprule
\multirow{2}[3]{*}{Group}  &\multirow{2}[3]{*}{Algorithm} &\multirow{2}[3]{*}{Aux.} &\multicolumn{5}{c}{Task}\\
\cmidrule(lr){4-8}
\ & & &NP vs P\textsubscript{1}  &NP vs P\textsubscript{2} &NP vs P\textsubscript{3} &NP vs P\textsubscript{4} &MC\\
\midrule
\midrule
All &ST-NN &- &61.15 &62.87 &65.14 &68.82 &29.43 \\
\hline
All &ST-NN &F(G)  &61.44 &63.19 &65.00 &68.79 &29.68\\
All &ST-NN &F(A)  &61.21 &62.67 &65.66 &\cellcolor{mygray}69.57 &29.71\\
All &ST-NN &F(GA) &61.09 &63.48 &\cellcolor{mygray}66.21 &69.54 &29.86\\
\hline
All &MT-NN &T(G)  &61.72 &63.39 &65.95 &68.99 &30.00\\
All &MT-NN &T(A)  &62.72 &\cellcolor{mygray}63.97 &65.40 &69.28 &29.79\\
All &MT-NN &T(GA) &\cellcolor{mygray}62.82 &63.68 &66.12 &69.40 &\cellcolor{mygray}30.24\\
\bottomrule
\end{tabular}
\begin{tablenotes}
\scriptsize
\item T(G): MT-NN with the additional task of gender estimation \space T(A): MT-NN with the additional task of age estimation \space T(GA): MT-NN with the additional task of gender and age estimation  
\end{tablenotes}
\end{threeparttable}
\end{table}

\subsubsection{Comparison with Existing Approaches}
In this section, we benchmark the results achieved using the Multi-Task Neural Network (MT-NN), which incorporated additional tasks of gender and age estimation against relevant studies. These comparative studies also utilized electrocardiography signals from \textit{Part A} of the \textit{BioVid} database with all $87$ participants. To ensure a fair comparison, they followed the same evaluation protocol, specifically the leave-one-subject-out (LOSO) cross-validation. The comparative results are detailed in Table \ref{table:2_comparison} and include research that employed hand-crafted features with traditional machine learning algorithms \cite{gkikas_chatzaki_2022}\cite{werner_hamadi_2014}, \textit{end-to-end} deep learning models \cite{huang_dong_2022}\cite{thiam_bellmann_kestler_2019}, and finally, hybrid approaches combine hand-crafted features with deep learning classifiers \cite{martinez_picard_2018}.
Our approach, which leverages hand-crafted engineered ECG features and a high-dimensional mapping from the encoder in combination with multi-task learning neural networks, demonstrated superior performance across all pain estimation tasks, whether in binary or multi-class classification settings.

\renewcommand{\arraystretch}{1.2}
\begin{table}
\center
\small
\begin{threeparttable}
\caption{Comparison of studies utilizing \textit{BioVid}, ECG signals and LOSO validation (2).}
\label{table:2_comparison}
\begin{tabular}{ P{4.0cm}  P{1.5cm}  P{1.5cm}  P{1.5cm}  P{1.5cm} P{1.5cm}}
\toprule
\multirow{2}[3]{*}{Study}  &\multicolumn{5}{c}{Task}\\
\cmidrule(lr){2-6}
\ &NP vs P\textsubscript{1}  &NP vs P\textsubscript{2} &NP vs P\textsubscript{3} &NP vs P\textsubscript{4} &MC\\
\midrule
\midrule
Gkikas \textit{et al}. \cite{gkikas_chatzaki_2022}$^\dagger$ &52.38 &52.78 &55.37 &58.62 &23.79\\
Huang \textit{et al}. \cite{huang_dong_2022}$^{\star\odot}$ &- &- &- &65.00 &28.50\\
Martinez and Picard \cite{martinez_picard_2018_b}$^\divideontimes$ &- &- &- &57.69 &-\\
Thiam \textit{et al}. \cite{thiam_bellmann_kestler_2019}$^\star$ &49.71 &50.72 &52.87 &57.04 &23.23\\
Wernel \textit{et al}. \cite{werner_hamadi_2014}$^\dagger$ &48.70 &51.60 &56.50 &62.00 &-\\
This study$^\divideontimes$ &\cellcolor{mygray}62.82 &\cellcolor{mygray}63.68 &\cellcolor{mygray}66.12 &\cellcolor{mygray}69.40 &\cellcolor{mygray}30.24\\
\bottomrule
\end{tabular}
\begin{tablenotes}
\scriptsize
\item $\dagger$:hand-crafted features and classic machine learning \space $\star$: end-to-end deep learning \space 				$\divideontimes$: hand-crafted features with deep learning classification algorithms \space $\odot$: pseudo heart rate gain extracted from visual modality
\end{tablenotes}
\end{threeparttable}
\end{table}

\subsection{Discussion}
We explored multi-task learning neural networks for automatic pain estimation from electrocardiography signals. By implementing the \textit{Pan-Tompkins} algorithm to identify QRS complexes, we extracted features associated with inter-beat intervals (IBIs). Numerous experiments were conducted to explore how gender and age influence pain perception, highlighting their significant impact. Additionally, we introduced two approaches to enhance pain estimation results by leveraging demographic information. Firstly, we augmented the original feature vectors by incorporating the subjects' demographic data, improving classification accuracy. Secondly, we employed a multi-task learning neural network that combined the tasks of pain, gender, and age estimation. This approach yielded superior results compared to methods previously discussed in this chapter and other related research. These findings indicate that domain-specific features can achieve excellent outcomes when combined with well-designed deep-learning architectures and demographic factors.

\section{Summary}
In this chapter, we examine the impact of age and gender on pain perception using ECG signals to extract relevant features. Our study involved a series of experiments where subjects were categorized into different groups based on gender (males and females) and age (20-35, 36-50, and 51-65 years). Additionally, we created combined groups that segregated age groups within each gender. Our findings from both approaches provided strong evidence of significant differences in pain perception among these groups. Notably, we observed a $12.5\%$ disparity in pain sensitivity between young females and older males. Generally, our results confirm that females exhibit higher pain sensitivity than males, aligning with findings from other studies in pain research. A critical discovery from our study is that pain sensitivity appears to decrease with age, which may increase the risk of unnoticed injuries.
We presented two methods of incorporating demographic information into our models from a computational perspective. First, we augmented the feature vectors derived from ECGs with demographic data. Second, we utilized a multi-task neural network approach to estimate pain, gender, and age simultaneously. Both methods demonstrated improved performance compared to the standard approach, indicating that integrating demographic information can enhance the accuracy of automatic pain assessment systems. 

We recommend that clinical pain assessment tools be designed for specific demographic groups to account for the distinct ways pain manifests across different populations. Additionally, we emphasize to researchers developing new pain databases the importance of including demographic factors and information on the social context and psychological conditions of subjects to enhance the quality and applicability of the data collected.
Our study focused on analyzing pain sensation through biosignals, specifically ECGs. We propose that further research should explore pain expressivity through visual mediums such as video. As previously discussed in Section \ref{clinical_pain_assessment}, the expression of pain is a crucial and complex issue. People vary in expressiveness; for various reasons, they might exaggerate or even feign pain, making accurate assessment challenging.

\chapter{Optimization: Balancing Efficiency and Performance}
\minitoc  
\label{chapter_5}

\section{Chapter Overview: Introduction \& Related Work}
This chapter presents the findings from the studies published in \cite{gkikas_tsiknakis_embc,gkikas_tachos_2024}. As outlined in \ref{slr_evaluation}, research in automatic pain assessment has rarely considered real-world situations. For example, \cite{tsai_weng_2017} implemented their study in an emergency triage setting, while \cite{naeini_shahhosseini_2019} tested their approach within IoT devices.
Additionally, we highlighted that the scarcity of studies exploring pain estimation in real-world settings or unconventional contexts suggests that current methodologies might not be entirely suitable for practical environments like clinics or hospitals due to issues with generalization or operational factors such as efficiency and inference time.
For these reasons, this chapter's objective is to explore approaches that \textit{(i)} utilize modalities readily available and applicable in the market and \textit{(ii)} examine the impact of model size and computational cost on performance.
In this context, our methodologies and experiments exclusively utilize RGB videos, a universally available modality, particularly on mobile devices. Additionally, we incorporate heart rate data, which is commonly accessible from various types of wearable technology. It is important to note that although we employ established pain datasets in our experiments, the videos and heart rate data extracted from ECGs are akin to those that could be obtained from smartphones and wearables, serving as a proof of concept for application in real-world environments. Furthermore, with respect to efficiency and the speed of inference, our goal is to develop the most compact pain assessment frameworks possible, ensuring they maintain adequate performance levels.

Numerous studies highlight the capabilities of automated systems that utilize behavioral or physiological modalities for pain assessment \cite{werner_survey_2019}. Sario \textit{et al.} \cite{sario_haider_2023} demonstrate the feasibility of accurately detecting and quantifying pain through facial expressions, establishing their value in clinical settings. The use of multimodal sensing appears especially promising, offering increased accuracy in pain monitoring systems \cite{rojas_brown_2023}. An important aspect of pain monitoring involves wearable devices that record biopotentials to estimate pain levels. Few studies have investigated the use of mainstream, wearable technology for this purpose, possibly due to a research preference for more costly, highly precise medical equipment. Leroux \textit{et al.} \cite{leroux_lynn_2021} state, \textit{\textquotedblleft The challenge is not whether wearable devices will provide useful clinical information but rather when we will start to use them in practice to enhance the field of pain.\textquotedblright}\space Additionally, Claret \textit{et al.} \cite{claret_casali_2023} explore the potential of using cardiac signals from wearable sensors for automatic emotion recognition, confirming the effectiveness of such methods.

In this chapter, our deep learning approaches are founded on transformer-based architectures. Convolutional Neural Networks (CNNs) have been the cornerstone of mainstream neural architectures in computer vision (CV), especially in the field of automatic pain assessment using images and videos, as we discussed in Section \ref{slr}. Inspired by the success of transformer architecture in natural language processing (NLP), where the self-attention mechanism is a fundamental element \cite{vanswani_ashich_2017}, researchers have developed similar models for visual tasks. The introduction of Vision Transformers (ViT) \cite{dosovitskiy_beyer_2021} has established a new paradigm in the computer vision domain. This has led to a plethora of new approaches based on ViT, such as the \textit{Transformer in Transformer} (TNT) \cite{han_xiao_2021}, which enhances local feature representation by subdividing image patches into smaller sub-patches. 
While transformer-based models have shown impressive results and offer great flexibility, they tend to scale poorly with input size and incur higher computational costs due to the self-attention layers that compute interactions between all input pairs. Efforts to mitigate these challenges include replacing self-attention with cross-attention \cite{lee_lee_2019} or combining both techniques \cite{jaegle_perceiver_2021} to improve the efficiency and reduce the complexity of these architectures.

\section{Video Analysis with Vision Transformers}
\label{chapter_5_paper_1}
We introduce a framework incorporating a vision transformer as a module extracting spatial features for individual video frames, combined with a transformer-based model equipped with cross and self-attention blocks, extracting temporal features from the video feature sequences. This configuration enables the effective utilization of the temporal dimensions of video data to deliver more accurate and reliable estimation of the continuous nature of pain.

\subsection{Methodology}
This section outlines the preprocessing methods employed, the design of our framework, the implementation details concerning the training procedure, and the database used.

\subsubsection{Pre-processing}
\begin{figure}
\centering
\includegraphics[scale=0.2]{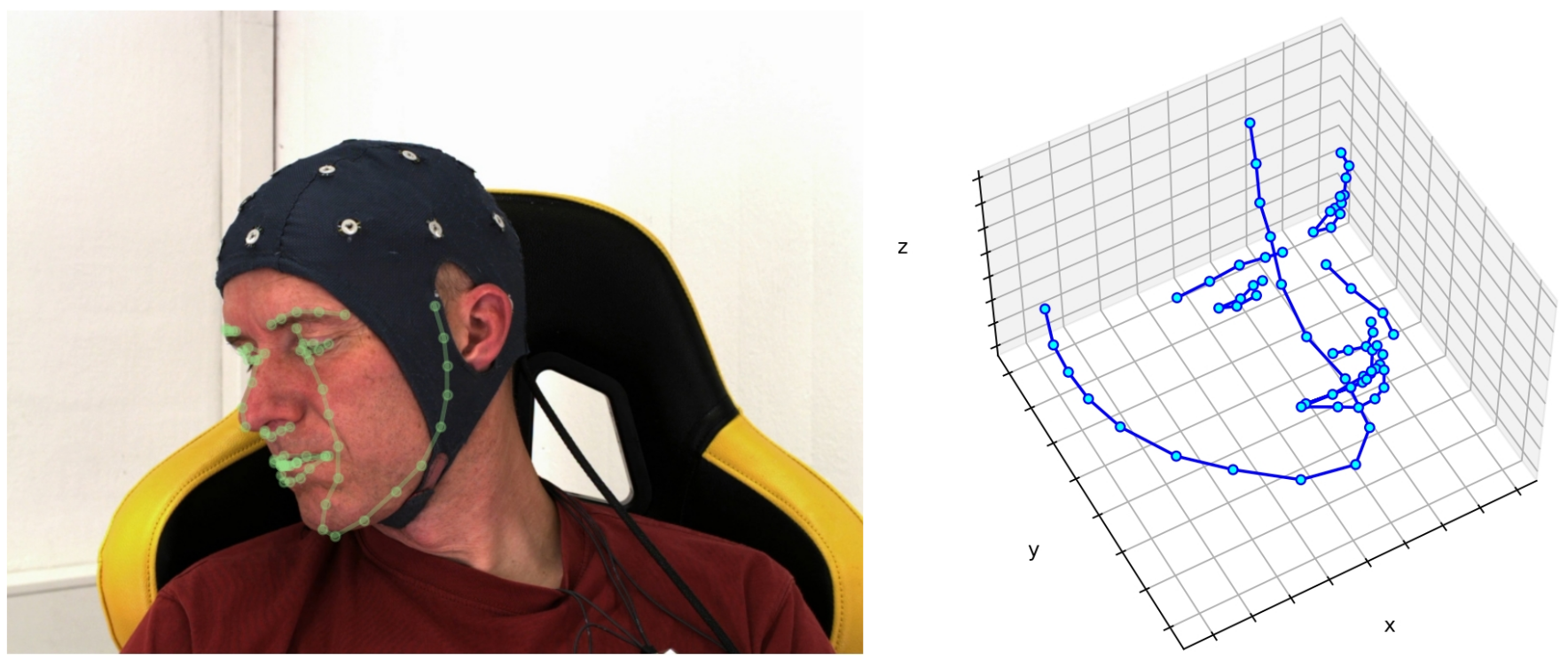}
\caption{The application of face alignment illustrates landmarks in 2D (left) and 3D (right) space.}
\label{1_alignment}
\end{figure}

Before processing videos for pain estimation, applying face detection and alignment was crucial to enhance performance and computational efficiency. We utilized the well-known face detector MTCNN \cite{zhang_2016} in combination with the Face Alignment Network (FAN) \cite{bulat_tzimiropoulos_2017}, which leverages 3D landmarks. This 3D approach is critical for addressing our specific challenges, as head movements tend to increase, particularly during instances of high-intensity pain, which can lead to inaccurate alignments with 2D methods. Additionally, it should be noted that all experiments were carried out using video frames with a resolution of $224\times224$ pixels. Figure \ref{1_alignment} illustrates the facial alignment process applied to a video frame.

\subsubsection{Transformer-based Framework}

\begin{figure}[b]
\centering
\includegraphics[scale=0.20]{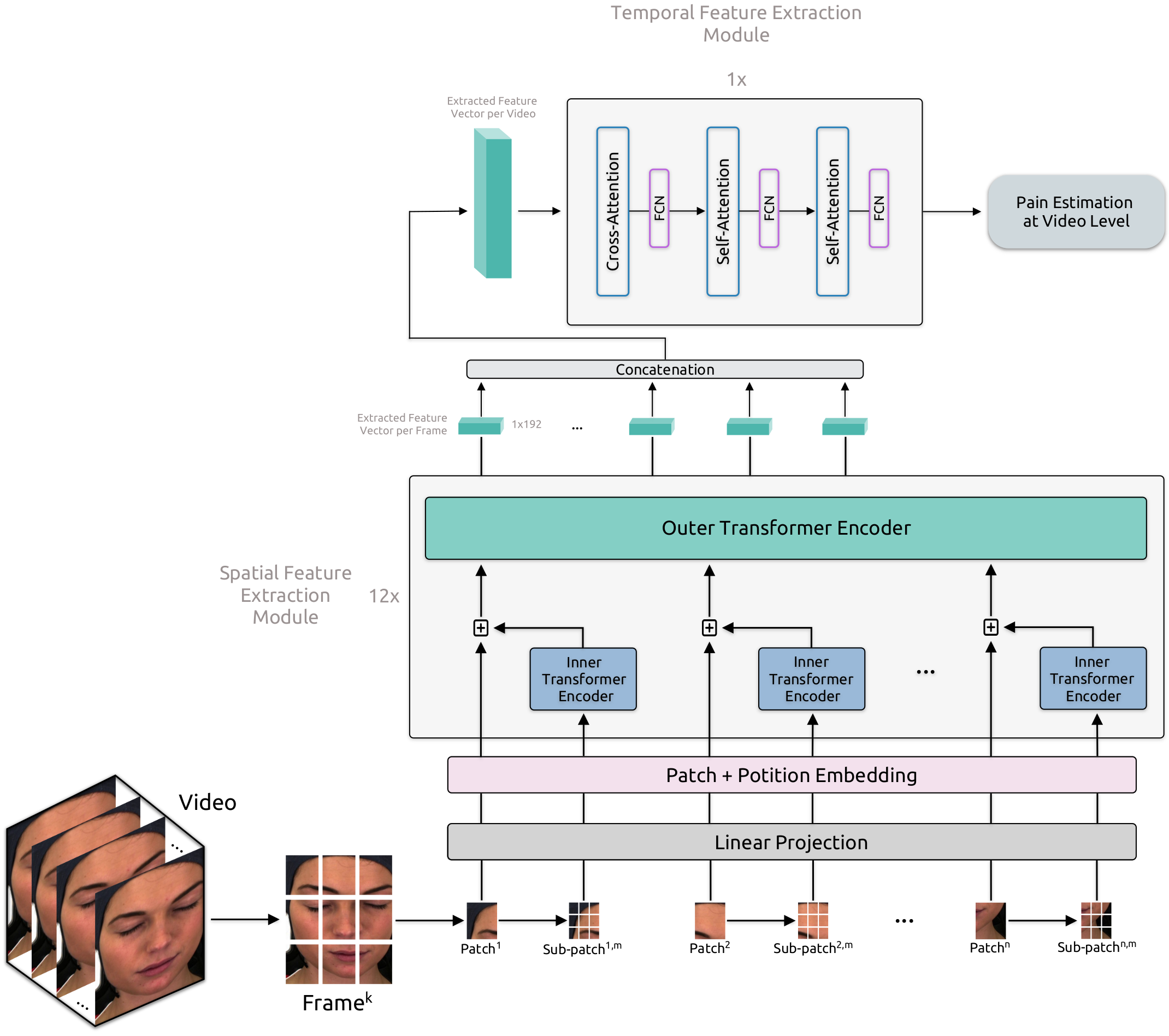}
\caption{An overview of our proposed transformer-based framework for automatic pain assessment.}
\label{1_tnt}
\end{figure}

Our framework is composed of two primary components: the \textit{\textquotedblleft spatial feature extraction module\textquotedblright}, specifically a TNT (Transformer in Transformer) model, and the \textit{\textquotedblleft temporal feature extraction module\textquotedblright}, which is a transformer with both cross and self-attention blocks. This framework, illustrated in Figure \ref{1_tnt}, includes approximately $24$ million parameters and performs operations at $4.2$ GFLOPS.

\paragraph{\textit{Spatial feature extraction module}:} Each frame is initially divided into \( n \) patches represented as 
\( \mathscr{F}_k = [{F}_k^1, {F}_k^2, \ldots, {F}_k^n] \in {R}^{n \times p \times p \times 3} \), where \( p \times p \) denotes the resolution of each patch (\textit{i.e.}, $16\times16$) and $3$ represents the number of color channels. 
These patches are then subdivided into $m$ sub-patches to facilitate the model's learning of global and local features. Each frame is thus transformed into a sequence of patches and sub-patches:
\begin{equation}
\mathscr{F}_k \rightarrow \left[{F}_{k,n,1}, {F}_{k,n,2}, \ldots, {F}_{k,n,m}\right],
\end{equation}
where \( {F}_{k,n,m} \in \mathbb{R}^{s \times s \times 3} \) is the \( m \)-th sub-patch of the \( n \)-th patch of the \( k \)-th frame, with each sub-patch having a resolution of \( s \times s \) (\textit{i.e.}, $4\times4$).
Following this, the patches and sub-patches undergo a linear projection and are transformed into embeddings \( {Z} \) and \( {Y} \). 
Position embeddings are then added to retain spatial information:
\begin{equation}
{Z}_0 \leftarrow {Z}_0 + {E}_{\text{patch}},
\end{equation}
where \( {E}_{\text{patch}} \) are the position encodings for the patches. Correspondingly, for each sub-patch within a patch, a position encoding is also added:
\begin{equation}
{Y}_i^0 \leftarrow {Y}_i^0 + {E}_{\text{sub-patch}},
\end{equation}
where \( {E}_{\text{sub-patch}} \) are the sub-patch position encodings and $i = 1, 2, \ldots, m$ denotes the index of a sub-patch. These sub-patches are then processed through an \textit{\textquotedblleft Inner Transformer Encoder\textquotedblright}, which consists of two multi-head self-attention blocks, crucial for dot product attention. The attention mechanism is defined as:
\begin{equation}
\text{Attention}({Q}, {K}, {V}) = \text{softmax}\left(\frac{{Q}{K}^T}{\sqrt{d_k}} {V}\right),
\label{1_attention}
\end{equation}
where \( {Q} \in \mathbb{R}^{M \times D}, {K} \in \mathbb{R}^{M \times C}, \) and \( {V} \in \mathbb{R}^{M \times C} \) (\( M \) is the input dimension, \( C \) and \( D \) are channel dimensions) are projections of the corresponding input and represent the Query, Key, and Value matrices. They defined as \( Q = XW_Q \), \( K = XW_K \), and \( V = XW_V \), where \( W \) are the learnable weight matrices and \( X \) is the input. The output embedding from the \textit{\textquotedblleft Inner Transformer Encoder\textquotedblright}\space is then added to the patch embedding and forwarded to the \textit{\textquotedblleft Outer Transformer Encoder\textquotedblright}. This encoder comprises three multi-head self-attention blocks, and its output is a feature vector $d=192$. The \textit{\textquotedblleft spatial feature extraction module\textquotedblright}\space as a whole encompasses a depth of $12$ blocks.

\paragraph{\textit{Temporal feature extraction module}:} The extracted embeddings of each input video frame are concatenated into a unified vector $\mathscr{D}$, representing the entire video as $\mathscr{V} \Rightarrow \mathscr{D} = (d^{1^{\frown}}d^{2^{\frown}}, \ldots, d^k)$. This vector is then processed through the temporal module, a transformer architecture consisting of $1$ cross-attention and $2$ self-attention mechanisms, each followed by a fully connected neural network (FCN). The introduction of cross-attention, which employs asymmetry in the attention mechanism, helps reduce computational complexity and increase the model's efficiency. Specifically, rather than projecting the input with dimensions \( M \times D \), the \( Q \) in cross-attention is a learned matrix with dimensions \( N \times D \), where \( N < M \). This module's self-attention components function as detailed in Equation \ref{1_attention}, with the cross and self-attention units comprising $1$ and $8$ heads, respectively. In addition, we incorporate Fourier feature position encoding \cite{jaegle_perceiver_2021}. 

\paragraph{Training Details:}
Before starting the automatic pain estimation training process, we pre-trained the \textit{\textquotedblleft spatial feature extraction module\textquotedblright}\space using the \textit{VGGFace2} dataset \cite{cao_shen_2018}, incorporating over three million facial images from more than nine thousand individuals. Table \ref{table:1_training_details} details the hyperparameters of our method and the applied augmentation techniques.

\renewcommand{\arraystretch}{1.2}
\begin{table}
\center
\small
\caption{Training details for the automatic pain assessment.}
\label{table:1_training_details}
\begin{center}
\begin{threeparttable}
\begin{tabular}{ P{2.0cm} P{1.5cm}  P{1.5cm} P{1.5cm} P{1.5cm} P{1.5cm} }
\toprule
Epochs &Optimizer & Learning Rate &LR decay &Weight decay &Warmup epochs\\ \hdashline\hdashline
200 &\textit{AdamW}  &\textit{1e-4} &\textit{cosine}  &0.1 &5\\
\midrule
\end{tabular}
\begin{tabular}{ P{2.0cm} P{1.5cm}  P{1.5cm} P{1.5cm} P{3.5cm}}
Label Smoothing &DropPath & Attention DropOut &Loss function &Augmentation methods\\ \hdashline\hdashline
0.1 &0.1 &0.1 &\textit{Cross Entropy} &\textit{AugMix}\cite{augmix} \& \textit{TrivialAugment}\cite{trivialAugment}\\
\bottomrule
\end{tabular}
\begin{tablenotes}
\scriptsize
\item DropPath applied to the \textit{\textquotedblleft spatial feature extraction module\textquotedblright}, Attention DropOut
applied to the \textit{\textquotedblleft temporal feature extraction module\textquotedblright}
\end{tablenotes}
\end{threeparttable}
\end{center}
\end{table}

\paragraph{Database Details:}
For this approach, we used the publicly available \textit{BioVid} dataset \cite{biovid_2013}, as described in the previous chapters.

\subsection{Experiments}
In this section, we detail the experiments conducted for pain estimation. 
Our experiments were carried out in both binary and multi-level classification formats. Specifically, we conducted binary classification tasks: \textit{(i)} NP vs. P\textsubscript{1}, \textit{(ii)} NP vs. P\textsubscript{2}, \textit{(iii)} NP vs. P\textsubscript{3}, \textit{(iv)} NP vs. P\textsubscript{4}, and \textit{(v)} a multi-level pain classification utilizing all available pain classes from the database. We employed the leave-one-subject-out (LOSO) cross-validation method as our evaluation protocol. Additionally, the classification metrics used in this study include micro-average accuracy, macro-average precision, macro-average recall (sensitivity), and macro-average F1 score.

\subsection{Results}
\subsubsection{Pain Estimation}
\label{1_pain_estimation}
In evaluating pain estimation tasks, we noted the following results: For the NP vs. P\textsubscript{1} task, accuracy reached $65.95\%$, with precision almost identical at $65.90\%$. The F1 score was slightly lower at $65.04\%$, and the recall stood out at $67.85\%$. In the NP vs. P\textsubscript{2} task, accuracy increased to $66.87\%$, and all related metrics improved, with the F1 score climbing by over $1.15\%$, highlighting enhanced detection of true positives.
The results were notably better for the NP vs. P\textsubscript{3} task, with an accuracy of $69.22\%$ and a sensitivity of $70.84\%$. This is expected as the pain at this level is considered severe, eliciting more pronounced responses from subjects. In the highest pain task, NP vs. P\textsubscript{4}, the recall was particularly high at $74.75\%$, with an accuracy of $73.28\%$, demonstrating that the detection of very severe pain is relatively more straightforward due to the pain reaching tolerance thresholds, making it more visibly evident through subjects' facial expressions.
However, in the multi-level classification task, performance metrics were lower, illustrating the complexity of estimating all pain levels concurrently; accuracy was only $31.52\%$, with a recall of $29.94\%$, pointing to significant challenges in accurately identifying true positives across multiple pain levels.

It should be noted that our framework, encompassing both the architectural and procedural aspects of training, was consistent across all binary and multi-level classification tasks. This was done to evaluate the generalization potential of our approach across all possible scenarios provided by the database, akin to real-world clinical settings. The detailed classification outcomes are presented in Table \ref{table:paper_1_results_pain_estimation}.

\renewcommand{\arraystretch}{1.2}
\begin{table}
\centering
\small
\caption{Results on the pain estimation tasks.}
\label{table:paper_1_results_pain_estimation}
\begin{center}
\begin{threeparttable}
\begin{tabular}{ P{2cm} P{1.5cm} P{1.5cm} P{1.5cm} P{1.5cm} P{1cm}}
\toprule
\multirow{2}{*}{Metric} & \multicolumn{5}{c}{Task} \\ 
\cmidrule(lr){2-6}
& NP vs P\textsubscript{1} & NP vs P\textsubscript{2} & NP vs P\textsubscript{3} & NP vs P\textsubscript{4} & MC\\
\midrule
\midrule
Accuracy &65.95 &66.87 &69.22 &73.28 &31.52\\
Precision &65.90 &66.89 &69.18 &73.31 &31.48 \\
Recall &67.85 &68.34 &70.84 &74.75 &29.94  \\
F1 &65.04 &66.19 &68.54 &72.75 &27.82  \\
\bottomrule 
\end{tabular}
\begin{tablenotes}
\scriptsize
\item NP: no pain P\textsubscript{1} : mild pain P\textsubscript{2} : moderate pain P\textsubscript{3} : severe pain P\textsubscript{4} : very severe pain MC: multi-level classification
\end{tablenotes}
\end{threeparttable}
\end{center}
\end{table}

\subsubsection{Video Sampling}
\begin{figure}
\centering
\includegraphics[scale=0.15]{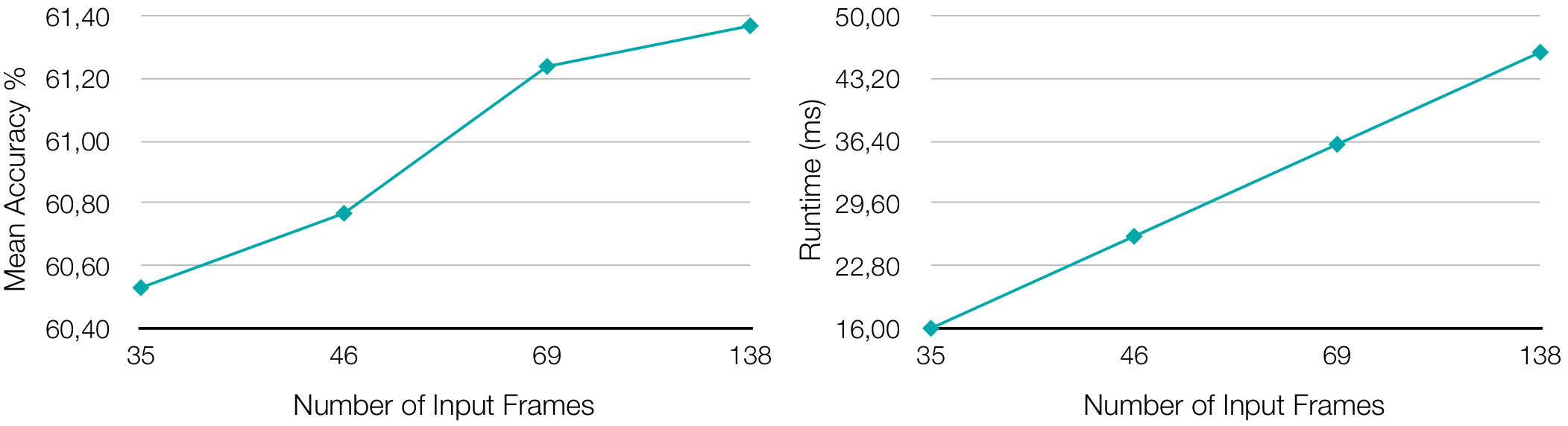}
\caption{The impact of the number of input frames on accuracy (left) and on runtime in milliseconds (right). Runtime calculated during inference on a \textit{NVIDIA RTX-3090}.}
\label{1_video_sampling}
\end{figure}

In this section, we explore the impact of video frame sampling on automatic pain estimation. Experiments detailed in Section \ref{1_pain_estimation} utilized all available frames ($138$) from each video. Subsequent experiments employed frame sampling with strides of $2$, $3$, and $4$. Starting with all $138$ frames, the video feature representation $\mathscr{D}$ has dimensions $138\times192$, totaling $26,496$. A stride of $2$ reduces this to $69$ frames, with $\mathscr{D}$ having dimensions $69\times192$ and totaling $13,248$. With strides $3$ and $4$, the frame counts reduce to $46$ and $35$, resulting in 
$\mathscr{D}$ sizes of $8,832$ and $6,720$, respectively. Table \ref{table:paper_1_results_video_sampling} displays the classification accuracies achieved with these varying frame counts for each pain estimation task. Concurrently, Figure \ref{1_video_sampling} demonstrates how the number of frames affects mean accuracy across the five tasks and mean runtime during inference. We noted a performance increase of approximately $1.38\%$ when using $138$ frames compared to $35$ frames. Additionally, the runtime increased by a factor of $3$. Despite the longer runtime, each sampling rate allows for real-time automatic pain estimation when necessary.

\renewcommand{\arraystretch}{1.2}
\begin{table}
\centering
\small
\caption{Results for the pain estimation tasks using various numbers of input frames.}
\label{table:paper_1_results_video_sampling}
\begin{center}
\begin{threeparttable}
\begin{tabular}{ P{2cm} P{1.5cm} P{1.5cm} P{1.5cm} P{1.5cm} P{1cm}}
\toprule
\multirow{2}[3]{*}{\shortstack{Number of\\ Frames}} & \multicolumn{5}{c}{Task} \\ 
\cmidrule(lr){2-6}
& NP vs P\textsubscript{1} & NP vs P\textsubscript{2} & NP vs P\textsubscript{3} & NP vs P\textsubscript{4} & MC\\
\midrule
\midrule
138 &65.95 &66.87 &69.22 &73.28 &31.52\\
69  &65.76 &66.74 &69.15 &73.25 &31.29\\
46  &65.66 &66.70 &68.50 &71.78 &31.20 \\
35  &65.40 &66.12 &68.32 &72.01 &30.80  \\
\bottomrule 
\end{tabular}
\begin{tablenotes}
\scriptsize
\item 
\end{tablenotes}
\end{threeparttable}
\end{center}
\end{table}

\subsubsection{Interpretation}
\begin{figure}
\centering
\includegraphics[scale=0.16]{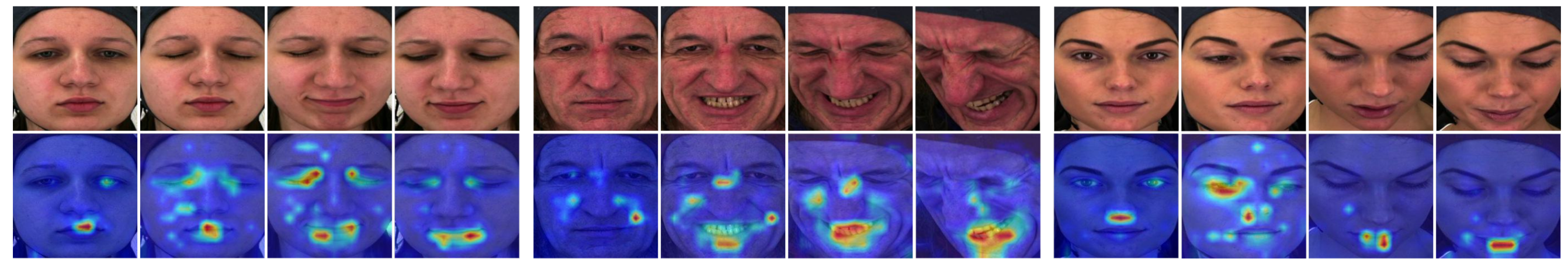}
\caption{Relevance Maps.}
\label{1_relevance_maps}
\end{figure}

Research in deep learning, particularly relevant to healthcare, increasingly focuses on model interpretability to explain decision-making processes. This is crucial for enhancing the transparency of models, a key factor for their acceptance and integration into clinical settings. In our study, we implemented the technique described in \cite{chefer_gur_2021} to generate relevant maps illustrating which facial areas our model---the \textit{\textquotedblleft spatial feature extraction module\textquotedblright} ---focuses on. As shown in Figure \ref{1_relevance_maps}, the model's attention is distributed over \textquotedblleft arbitrary\textquotedblright\space areas at the onset of a facial expression sequence. However, as the expression of pain intensifies, the focus sharpens on specific regions indicative of pain. It is important to note from our relevance maps that no universal facial expressions are unique to pain. However, there is a noticeable concentration on areas like the mouth and eyes.

\subsubsection{Comparison with existing methods}
In this section, we present a comparison of our results achieved using a transformer-based framework that utilizes all available video frames against other studies that also employed Part A of the \textit{BioVid} database with all $87$ subjects, following the same leave-one-subject-out (LOSO) cross-validation protocol. This ensures objective and accurate comparisons, with results detailed in Table \ref{table:paper_1_comparison}. The studies compared fall into three main categories: \textit{i}) those focusing exclusively on pain detection (NP vs. P\textsubscript{4}), \textit{ii}) those examining both pain detection and multi-level pain estimation, and \textit{iii}) those that cover all major pain-related tasks.

Our method, tested across all tasks, recorded the highest performance metrics in binary and multi-level pain estimations. Studies limited to specific aspects of pain detection or multi-level pain estimation often yielded comparable or superior results, as indicated in \cite{thiam_kestler_schenker_2020}, \cite{tavakolian_bordallo_liu_2020}, and \cite{patania_2022}. This highlights that while focused studies often show high performance, the broader impact lies in developing systems that perform well across all potential scenarios.

\renewcommand{\arraystretch}{1.2}
\begin{table}
\centering
\small
\caption{Comparison of studies utilizing \textit{BioVid}, RGB videos, and LOSO validation.}
\label{table:paper_1_comparison}
\begin{center}
\begin{threeparttable}
\begin{tabular}{ P{3.5cm} P{1.5cm} P{1.5cm} P{1.5cm} P{1.5cm} P{1cm}}
\toprule
\multirow{2}[3]{*}{Study} & \multicolumn{5}{c}{Task} \\ 
\cmidrule(lr){2-6}
& NP vs P\textsubscript{1} & NP vs P\textsubscript{2} & NP vs P\textsubscript{3} & NP vs P\textsubscript{4} & MC\\
\midrule
\midrule
Thiam \textit{et al.} \cite{thiam_kestler_schenker_2020}        &- & & &69.25 &- \\
Tavakolian \textit{et al.} \cite{tavakolian_bordallo_liu_2020}  &- &- &- &71.02 &-\\
Patania \textit{et al.} \cite{patania_2022}                     &- &- &- &73.20 &-  \\
\hline
Huang \textit{et al.} \cite{huang_dong_2022} &- &- &- &77.50 &34.30  \\
Xin \textit{et al.} \cite{xin_li_yang_2021}  &- &- &- &86.65 &40.40  \\
\hline
Zhi and Wan \cite{zhi_wan_2019}                            &56.50 &57.10 &59.60 &61.70 &29.70  \\
Werner \textit{et al.} \cite{werner_hamadi_ecklundt_2017}  &53.30 &56.00 &64.00 &72.40 &30.80  \\
Our approach                                               &65.95 &66.87 &69.22 &73.28 &31.52  \\
\bottomrule 
\end{tabular}
\begin{tablenotes}
\scriptsize
\item 
\end{tablenotes}
\end{threeparttable}
\end{center}
\end{table}

\subsection{Discussion}
This research examined the application of transformer-based architectures for automatic pain estimation through video analysis. Our framework employed exclusively transformer models, leveraging the spatial and temporal aspects of the video frames. The experiments demonstrated the effectiveness of our approach in assessing pain, showing strong generalization across various pain estimation tasks with notable results, particularly with low-intensity pain where facial expressions are less apparent. Additionally, the framework demonstrated high efficiency, suitable for real-time applications. A significant contribution of our work includes developing relevance maps highlighting facial areas the model focuses on. We advocate for continued efforts within the affective computing field to enhance the interpretability of these deep-learning methods.

\vspace*{10pt}
\section{Video \& Heart Rate Analysis with Transformer Architectures}
\label{chapter_5_paper_2}
We introduce a proof of concept for an automatic pain assessment framework that integrates facial video data captured by an RGB camera with heart rate signals. We build and extend our previous analysis in \ref{chapter_5_paper_1}.
Our main objectives include (1) evaluating the effectiveness and limitations of video and heart rate data as standalone modalities in an unimodal setting, (2) exploring the efficacy of combining behavioral (video) and physiological (heart rate) markers to overcome challenges associated with their reliance on different sensing technologies and information representations, and (3) analyzing the performance and efficiency of recently introduced transformer-based architectures.

\subsection{Methodology}
This section details the preprocessing methods for video and ECG, the design of the proposed framework, the augmentation techniques developed, and the specifics of implementing the pretraining process.

\subsubsection{Pre-processing}
Preparatory processing was essential before feeding data into the pain assessment framework, particularly for the raw ECG data used to compute heart rate. We focused on exploring heart rate as the primary feature due to its benefits: It's readily obtainable from wearable devices, cost-effective, and easily accessible. These advantages position heart rate as a potentially valuable feature for automated pain assessment.

\paragraph{Video Preprocessing:}
Video preprocessing included face detection to isolate the facial region using the MTCNN face detector \cite{zhang_2016}, which employs multitask cascaded convolutional networks to predict facial and landmark locations. Predicting landmarks facilitates face alignment, which is crucial for accurate facial analysis. However, we observed that face alignment reduced the expressiveness linked to head movements, a common behavioral indicator of pain. Consequently, face alignment was omitted from our proposed pipeline. Additionally, the resolution of frames post-face detection was standardized at $448\times448$ pixels.

\paragraph{ECG preprocessing \& analysis:}
\label{ecg}
Similar to Section \ref{chapter_4}, we utilize the \textit{Pan-Tompkins} \cite{pan_tompkins_1985} algorithm to detect the QRS complex, the most prominent wave complex in an ECG signal. This algorithm operates in two phases: preprocessing and decision-making. The preprocessing phase focuses on noise removal, artifact elimination, signal smoothing, and enhancing the QRS slope. The decision-making phase involves initial QRS detection using adaptive thresholds, a retrospective search to identify any missed QRS complexes, and a method for distinguishing T waves. Following the accurate detection of R waves, the estimation of inter-beat intervals (IBIs) was conducted, leading to the extraction of key features. Specifically, we calculated the mean of the IBIs as follows:
\begin{equation} 
\mu=\dfrac{1}{n}\sum_{i=1}^{n}(RR_{i+1}-RR_{i}),
\end{equation}
where $n$ is the total number of IBIs, and $RR_{i}$ denotes the consecutive $R-R$ intervals. Subsequently, the heart rate was calculated using the following formula:
\begin{equation}
HR=\dfrac{60\cdot FS}{\mu},
\end{equation}
where $FS$ denotes the sampling frequency of the ECG recording.

\subsubsection{Framework architecture}
\begin{figure}
\centering
\begin{subfigure}{0.70\textwidth}
\includegraphics[width=\linewidth]{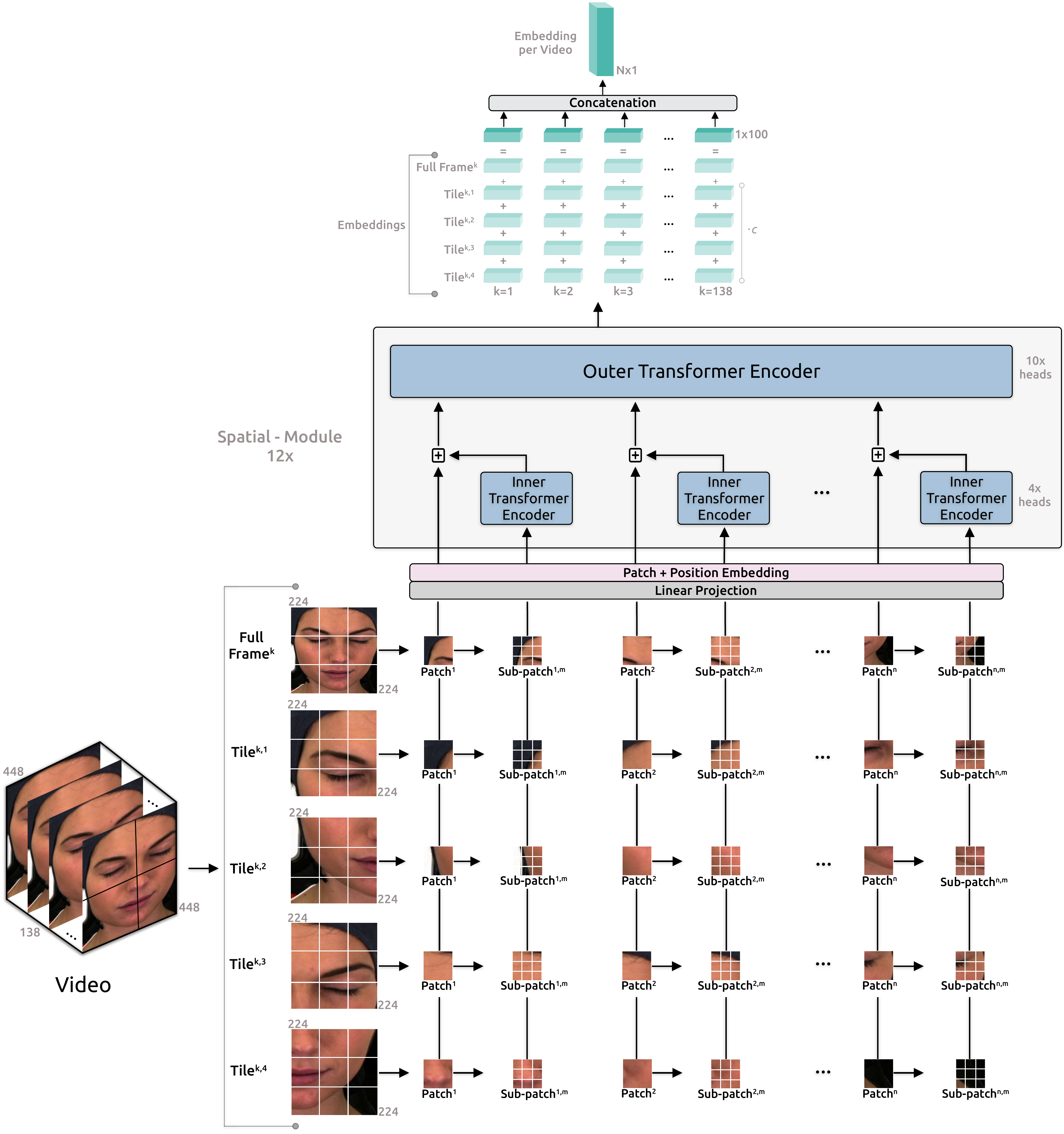}
\caption{Video analysis pipeline.}
\label{2_video_pipeline}
\end{subfigure}
\vspace{10pt} 

\begin{subfigure}{0.50\textwidth}
\centering
\includegraphics[width=\linewidth]{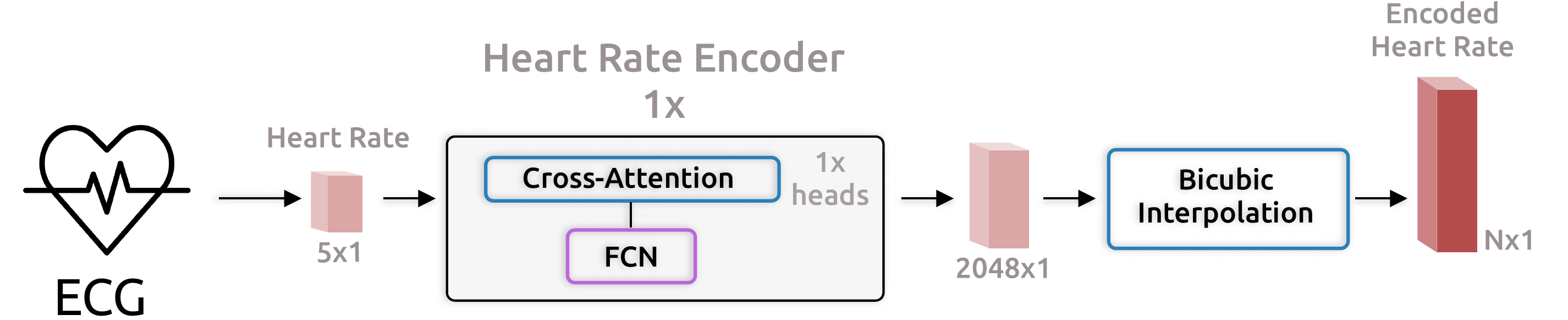}
\caption{ECG analysis pipeline.}
\label{2_ecg_pipeline}
\end{subfigure}
\vspace{10pt} 

\begin{subfigure}{0.60\textwidth}
\centering
\includegraphics[width=\linewidth]{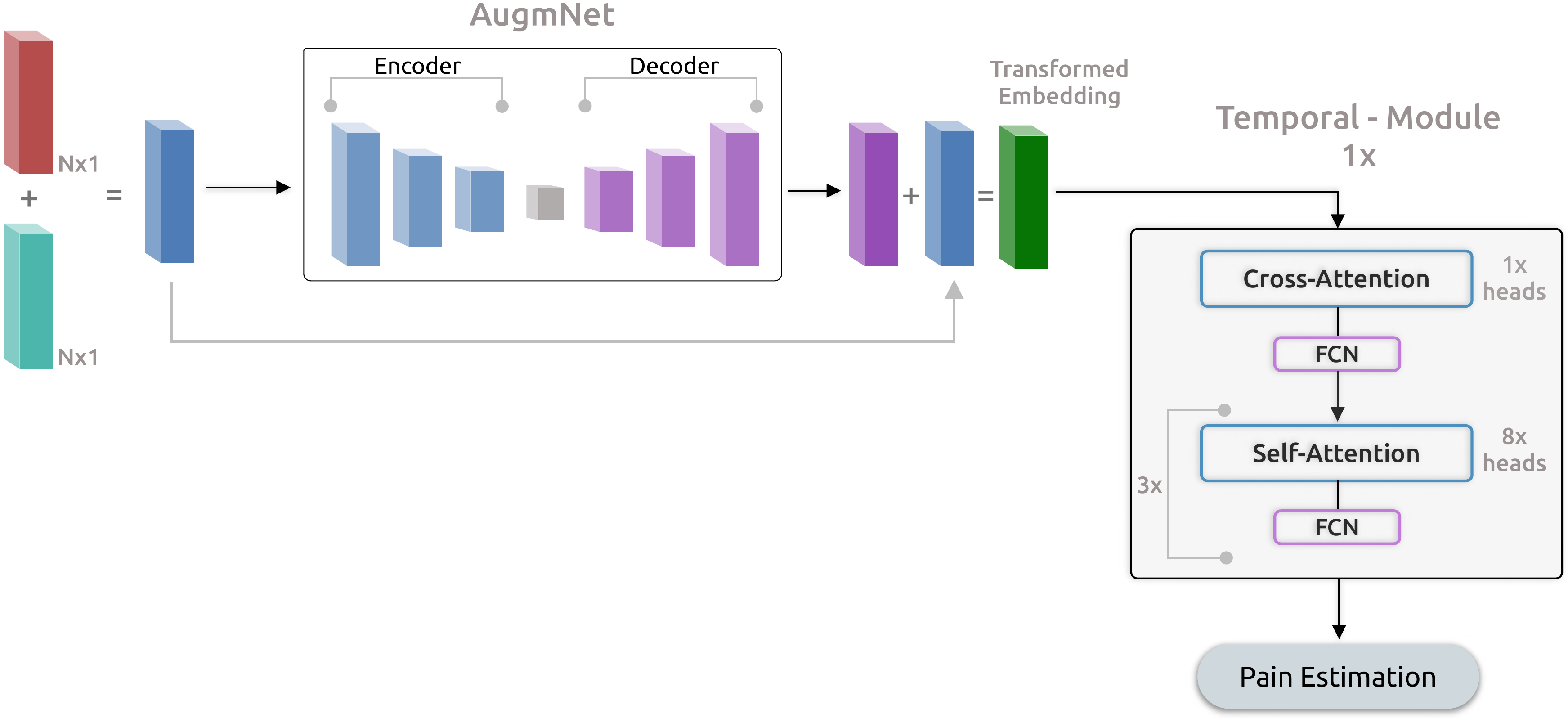}
\caption{Fusion analysis pipeline.}
\label{2_fusion_pipeline}
\end{subfigure}

\caption{Outline of the proposed framework.}
\label{2_framework}
\end{figure}

The proposed framework, as illustrated in Figure \ref{2_framework}, consists of four key components: the \textit{Spatial-Module} that extracts embeddings from video data, the \textit{Heart Rate Encoder} which maps heart rate signals into a higher dimensional space, the \textit{AugmNet} that generates augmentations in the latent space, and the \textit{Temporal-Module} performs with the final assessment of pain.

\paragraph{Spatial-Module:}
\label{spatial_module}
The architecture for this module draws inspiration from the \textit{Transformer in Transformer} approach as detailed by \cite{han_xiao_2021}. The process begins with the initial video frame at a resolution of $448 \times 448$ pixels, segmented into $4$ quadrants, each at $224 \times 224$ pixels. This tiling method, which maximizes the utilization of the frame's resolution, is influenced by approaches seen in satellite imaging analysis. Our framework incorporates the $4$ tiles and the original frame---resized to $224 \times 224$ pixels---into our analysis pipeline. Thus, each video frame transforms into $5$ distinct images, denoted as:
\begin{equation} 
\mathcal{F}^k=[F^{k,1},F^{k,2},\ldots,F^{k,t}],
\end{equation}
where $k$ stands for the frame number, and $t$ encompasses the tile count, including the resized full frame.
Each tile is initially split into $n$ patches, expressed as:
\begin{equation}
\mathcal{F}^{k,t}=[F^{k,t,1},F^{k,t,2},\ldots,F^{k,t,n}]\in\mathbb{R}^{n \times p \times p \times 3 },
\end{equation}
where $p \times p$ specifies the resolution of each patch ($16 \times 16$), and $3$ denotes the RGB channels. These patches are further segmented into $m$ sub-patches, allowing the model to capture the image's global and localized features. Each tile from a frame thus transitions into a sequence of patches and sub-patches, represented as 
$$\mathcal{F}^{k,t}= [F^{k,t,n,1},F^{k,t,n,2},\ldots,F^{k,t,n,m}].$$ 
Consequently, each video frame is characterized by:
\begin{equation}
\mathcal{F}^{k} \rightarrow \left[ F^{k,t,n,m} \mid t \in [1, 5], n \in [1, 196], m \in [1, 16] \right], 
\end{equation}
where $F^{k,t,n,m}\in\mathbb{R}^{s \times s \times 3 }$ defines the $m$-th sub-patch within the $n$-th patch of the $t$-th tile for the $k$-th frame, with each sub-patch having a resolution of $s \times s$ ($4 \times 4$). Each frame consists of $5$ image representations, encompassing $196$ patches, and each patch contains $16$ sub-patches.
The patches and sub-patches are then linearly projected into embeddings $Z$ and $Y$. Positional embedding is applied to maintain spatial information, employing 1D learnable position encodings:
\begin{equation}
Z_0\leftarrow Z_0+E_{patch},
\end{equation}
where $E_{patch}$ indicates the position encoding. Each sub-patch also receives its specific positional encoding:
\begin{equation}
Y_0^i\leftarrow Y_0^i+ E_{sub-patch},
\end{equation}
where $E_{sub-patch}$ denotes the positional encodings for sub-patches, and $i$ represents the index of a sub-patch within a patch. The sub-patches are processed in the \textit{Inner Encoder}, which consists of four self-attention heads \cite{vaswani_shazeer_2017}, utilizing dot product attention:
\begin{equation}
Attention(Q,K,V)= \text{softmax}\left( \frac{QK^T}{\sqrt{d_k}}V \right).
\end{equation}
The output from the \textit{Inner Encoder} integrates into the patch embedding, advancing to the \textit{Outer Encoder}, which mimics the \textit{Inner Encoder} with ten self-attention heads. The \textit{Spatial-Module} consists of twelve parallel blocks, generating embeddings of dimensionality $d=100$.
For each input video frame, $5$ distinct output embeddings of dimensionality $100$ are produced and combined to form a comprehensive frame representation:
\begin{equation}
\mathcal{D} = d_{Full Frame}+(d_{Tile^1}+d_{Tile^2}+d_{Tile^3}+d_{Tile^4})\cdot{c},  \qquad \mathcal{D}\in\mathbb{R}^{100},
\end{equation}
where $c$ adjusts the contribution from the tile embeddings, subsequently, the embedding for each frame, $\mathcal{D}$, is concatenated with those of other frames to construct a complete video representation:
\begin{equation}
\label{vd}
\mathcal{V}_D = [\mathcal{D}_1 \Vert \mathcal{D}_2 \Vert \ldots \Vert \mathcal{D}_{f}],  \qquad \mathcal{V}_D\in\mathbb{R}^{N},
\end{equation}
where $f$ represents the total number of frames in the video, and $N$ denotes the dimensionality of the final video embedding.

\paragraph{Heart Rate Encoder:}
\label{hr_encoder_architecture}
As mentioned in Section \ref{ecg}, heart rate is computed from the original ECG every second, producing a heart rate vector of length $h=\theta$ for recordings lasting $\theta$ seconds. It should be noted that when beats per minute (BPM) fall below $60$ within any $1$-second segment of the ECG, making direct heart rate calculation impractical, the method averages heart rates from the data points immediately before and after to maintain a uniform set of $\theta$ data points.
The \textit{Heart Rate Encoder}, which is part of a transformer-based architecture similar to the \textit{Inner} and \textit{Outer Encoders}, utilizes one cross-attention head instead of self-attention followed by a fully connected neural network (FCN). This use of cross-attention introduces an asymmetry that reduces computational load and increases efficiency. Unlike traditional input projections that are $M \times D$ in dimension, as detailed in Section \ref{spatial_module}, the $Q$ matrix in cross-attention is a learnable matrix sized $N \times D$ where $N < M$.
The encoder's internal embeddings are set to a dimensionality of $512$ and contain only a single block depth. Fourier feature position encodings \cite{jaegle_perceiver_2021} are also implemented to handle positional information. The main goal of this encoder is to transform the initial heart rate vector $h$ into a more complex and richer feature space,
$$h \in \mathbb{R}^\theta \rightarrow E_h \in \mathbb{R}^{2048},$$
where $E_h$ represents the enhanced output embedding of this encoder.
In the next step, the output from the heart rate encoder is expanded dimensionally via a bicubic interpolation module. This process enhances the original heart rate's feature representation, allowing it to integrate smoothly with the video's embedding representation through addition. The need for identical dimensions in both embedding vectors is critical and is adeptly addressed by this interpolation module. This non-learning-based approach proves to be efficient and effective for encoding purposes. Additionally, interpolation provides the flexibility to dynamically set the dimensionality of the final output embedding, unlike the fixed dimensions typically seen in neural network-based methods.
Specifically: 
\begin{equation}
\label{bh}
\mathcal{B}_h = \sum_{i=0}^{3} \sum_{j=0}^{3} a_{ij}(E_h)   \cdot (x-x_0(E_h))^i \cdot 
(y-y_0(E_h))^j,
\end{equation}
where $a_{ij}$ are the coefficients used for interpolation, and $\mathcal{B}_{h}$ is the resulting vector from the bicubic interpolation process. The dimension of $\mathcal{B}_{h}$ is $N$, which is the same as that of $\mathcal{V}_D$.

\paragraph{AugmNet:}
Inspired by recent developments in the augmentation literature \cite{cheung_yeung_2021}, employs a learning-based technique to identify augmentation patterns within the latent space. Unlike conventional methods that perform image augmentations (\textit{e.g.,} rotation, cropping) in the pixel space, \textit{AugmNet} universally applies transformations to the embeddings. This method eliminates the necessity for specific transformations tailored individually to each modality, \textit{e.g.,} image, signal, and text.
Incorporating this module within the automatic pain assessment framework helps to regularize the learning process and address overfitting issues. Moreover, corrupting the input embeddings compels the following model, especially the \textit{Temporal-Model}, to derive more precise and representative features, thereby improving performance in the pain assessment task.
The modality-agnostic method effectively applies to embedding representations from any original modality, including video and heart rate signals.
\textit{AugmNet} adopts an encoder-decoder architecture, where both the encoder and decoder consist of only $2$ fully connected layers with the \textit{ELU} nonlinear activation function applied after each layer.

For a session lasting $\theta$ seconds, it produces $\theta\times frames\ per\ second$ frames and 
$\theta\times FS$ data points for video and ECG, respectively. In the video analysis pipeline, the \textit{Spatial-Module} constructs an embedding representation, $\mathcal{V}_D$ \eqref{vd}, from the original video, dimensioned at $d\times FPS=N$. In the ECG analysis pipeline, a feature vector with dimension $\theta$ is created after extracting the heart rate, one data point per second. The \textit{Heart Rate Encoder} and bicubic interpolation then produce an embedding, $\mathcal{B}_h$ \eqref{bh}, with dimension $N$. 
The fusion of video and heart rate embeddings at the session level is performed, where $\mathcal{V_D}$ and $\mathcal{B}_h$ are merged by addition, integrating the data from the initial input modalities. The resulting combined embedding is then processed by \textit{AugmNet}:
\begin{linenomath}
\begin{DispWithArrows}
\mathcal{A}_D & \leftarrow \textit{AugmNet}(\mathcal{V}_D + \mathcal{B}_h)\Arrow{} \\
\mathcal{P} & \leftarrow\mathcal{A}_D+(\mathcal{V}_D + \mathcal{B}_h),
\end{DispWithArrows}
where $\mathcal{P}$ represents the transformed embedding vector, serving as input for the final module, the \textit{Temporal-Module}. \textit{AugmNet} is active only during training as a standard augmentation method and remains inactive during inference.
\end{linenomath}

\paragraph{Temporal-Module:}
Like the \textit{Heart Rate Encoder}, this component operates on a transformer-based architecture. It employs a combination of multi-head cross-attention and multi-head self-attention mechanisms. The architecture consists of one multi-head cross-attention block with a single attention head and three subsequent multi-head self-attention blocks, each featuring eight attention heads. An FCN follows each attention block. The internal embeddings in this module have a dimensionality of $128$ and encompass a single block in depth. The position encoding method used here mirrors that of the \textit{Heart Rate Encoder}, utilizing Fourier feature position encoding. This module processes the input embedding $\mathcal{P}$ or the sum $(\mathcal{V}_D + \mathcal{B}_h)$ if \textit{AugmNet} is inactive, to derive the final classification outcome. The learning error is computed during this phase, and the framework undergoes further training.

\subsubsection{Supplementary augmentation methods}
We also introduced additional augmentation strategies alongside the \textit{AugmNet} module, which applies learned transformations to embeddings. The initial technique, dubbed \textit{Basic}, combines polarity inversion and noise addition to manipulate the original inputs by reversing their polarity and injecting noise. Another technique, named \textit{Masking}, involves nullifying elements within the embeddings using randomly sized and placed masks that zero out $10-20\%$ of the embedding elements. These methods function within the latent space, similar to \textit{AugmNet}.

\subsubsection{Pre-training}
Before starting the training for automatic pain assessment, we individually pretrained all modules except \textit{AugmNet}. The \textit{Spatial-Module} underwent a dual-phase pretraining process. Initially, it was pre-trained on \textit{VGGFace2} \cite{cao_shen_2018}, a dataset designed for facial recognition to learn basic facial features. This was followed by an advanced training phase involving emotion recognition datasets in a multi-task learning framework. These datasets include the widely used \textit{AffectNet} \cite{mollahosseini_hasani_2019}, \textit{Compound Facial Expressions of Emotions Database} \cite{du_tao_2014}, \textit{RAF Face Database basic} \cite{li_deng_2017}, and \textit{RAF Face Database compound} \cite{li_deng_2017}, enabling the module to adapt to specific emotional expressions related to pain manifestations.
During this phase, the model is trained across these datasets simultaneously, employing a multi-task learning loss approach as suggested by \cite{cipolla_gal_2018}, where learned weights scale the individual losses to account for the homoscedastic uncertainty of each task:
$$
L_{total}= [e^{w1}L_{S_1}+w_{1}]+ [e^{w2}L_{S_2}+w_{2}] + [e^{w3}L_{S_3}+w_{3}]+[e^{w4}L_{S_4}+w_{4}],
$$
where $L_S$ denotes the loss for each dataset and $w$ are the weights adjusting the learning focus to optimize the overall loss $L_{total}$. The \textit{Temporal-Module} is trained solely on the \textit{VGGFace2} dataset. For this module, images are converted into 1D vectors prior to processing. The \textit{Heart Rate Encoder} is pre-trained using the \textit{ECG Heartbeat Categorization Dataset} \cite{kachuee_fazeli_2018}, which includes heartbeat signals from the \textit{MIT-BIH Arrhythmia Dataset} \cite{moody_mark_2001} and the \textit{PTB Diagnostic ECG Database} \cite{bousseljot_kreiseler_1995}\cite{goldberger_amaral_2000}. Table \ref{table:datasets} provides a detailed list of the datasets used in our training process.

\renewcommand{\arraystretch}{1.2}
\begin{table}
\center
\small
\caption{Datasets utilized for the pre-training process of the framework.}
\label{table:datasets}
\begin{center}
\begin{threeparttable}
\begin{tabular}{p{5.0cm} p{1.5cm} p{1.5cm} p{1.5cm}}
\toprule
Dataset &\#  samples &\# classes &Task\\
\midrule
\midrule
\textit{VGGFace2} \cite{cao_shen_2018}   &3.31M  &9,131  &Face\\
\textit{AffectNet} \cite{mollahosseini_hasani_2019} &0.40M &8 &Emotion\\
\textit{Compound FEE-DB}  \cite{du_tao_2014}&6,000 &26 &Emotion\\
\textit{RAF-DB basic} \cite{li_deng_2017}&15,000 &7 &Emotion\\
\textit{RAF-DB compound} \cite{li_deng_2017}&4,000 &11 &Emotion \\
\textit{ECG HBC Dataset} \cite{kachuee_fazeli_2018}&0.45M &5 &Arrhythmia\\
\bottomrule 
\end{tabular}
\begin{tablenotes}
\scriptsize
\item Task: all tasks involve classification
\end{tablenotes}
\end{threeparttable}
\end{center}
\end{table}

\subsubsection{Dataset Details}
\label{biovid}
In this study, to assess the developed framework, we utilized the \textit{BioVid Heat Pain Database} \cite{biovid_2013}, which includes facial videos, electrocardiograms, electromyograms, and skin conductance levels from $87$ healthy individuals ($44$ males and $43$ females, aged $20-65$). 
The dataset employs a thermode to induce varying pain levels in the participants' right arm. Initially, each participant's pain threshold (the transition from heat sensation to pain) and pain tolerance (the point at which pain becomes unbearable) were determined. These thresholds delineated the minimum and maximum pain levels, along with two intermediary levels, forming five distinct pain intensities: No Pain (NP), Mild Pain (P\textsubscript{1}), Moderate Pain (P\textsubscript{2}), Severe Pain (P\textsubscript{3}), and Very Severe Pain (P\textsubscript{4}).
Pain stimuli temperatures ranged from P\textsubscript{1} to P\textsubscript{4}, capped at $50.5$\textdegree$C$. Each subject experienced $20$ stimulations at each of the four intensities (P\textsubscript{1} to P\textsubscript{4}). Each stimulation lasted $4$ seconds, interspersed with recovery intervals of $8$ to $12$ seconds. This protocol, along with $20$ baseline measurements at NP ($32$\textdegree$C$), culminated in randomly ordered $100$ stimulations per participant.
The data was preprocessed to capture $5.5$-second windows starting $1$-second after reaching the target temperature for each stimulation. This process produced $8,700$ samples, each $5.5$ seconds in duration, distributed evenly across the five classes and modalities for all $87$ subjects.

\subsection{Experiments}
The study leveraged videos and electrocardiograms from Part A of the \textit{BioVid} dataset, using all available samples from the $87$ participants.
The videos were recorded at a rate of $25$ frames per second (FPS), and the electrocardiogram (ECG) signals were sampled at $512$ Hertz (Hz).
Each recording session lasted $5.5$ seconds, generating $138$ video frames and ECG vectors containing $2,816$ elements each, then converted into heart rate vectors of $5$ data points.
The complete set of frames and data points from both videos and cardiac signals was utilized in the experiments. 
The experimental approach included iterative refinement of techniques, with the most effective combinations selected for extended training periods ranging from $500$ to $800$ epochs to improve feature extraction and performance outcomes.
Table \ref{table:training_details} details the training configurations for the automatic pain assessment tasks.

Pain assessment experiments were structured around binary and multi-level classification setups, testing each modality individually and in combination.
The binary classification task differentiated between No Pain (NP) and Very Severe Pain (P\textsubscript{4}), whereas the multi-level classification (MC) involved categorizing all pain intensities available in the dataset. The evaluation strategy adopted was the leave-one-subject-out (LOSO) cross-validation, and the assessment metrics included accuracy, precision, recall (sensitivity), and F1 score.
Notably, a consistent training regimen was applied across both the binary (NP vs. P\textsubscript{4}) and multi-level (MC) classification tasks without varying the training schedule or optimization strategies.

\renewcommand{\arraystretch}{1.2}
\begin{table}
\center
\small
\caption{Training details for the automatic pain assessment.}
\label{table:training_details}
\begin{center}
\begin{threeparttable}
\begin{tabular}{ P{1.7cm}  P{1.5cm}  P{1.3cm} P{1.3cm} P{1.3cm} P{1.3cm}}
\toprule
Optimizer & Learning rate &LR decay &Weight decay &Warmup epochs &Batch size\\
\midrule
\midrule
\textit{AdamW}   &\textit{1e-4} &\textit{cosine}  &0.1 &50 &32\\
\bottomrule
\end{tabular}
\end{threeparttable}
\end{center}
\end{table}

\subsection{Results}
\subsubsection{Video modality}
\label{2_videos}
Experiments related to the video modality explored the effects of pretraining on the \textit{Spatial-Module}, the video analysis pipeline's performance, particularly the impact of tiling, and the implementation of new augmentation techniques. These experiments are detailed in Table \ref{table:video_modality}.
Performance enhancements are evident when comparing the first and second pretraining stages of the \textit{Spatial-Module}. For instance, in the NP vs. P\textsubscript{4} task, initial pretraining alone achieved $72.56\%$ accuracy, while including the second emotion-focused pretraining stage increased accuracy to $74.25\%$. This trend is also notable in the multi-level classification, where the second stage added $1.12\%$ to the performance, totaling $33.34\%$.
Further experiments assessed the effect of using tiles in the video representation. Initially, employing four tiles led to a performance decrease of over $6\%$ in the binary classification task and $1.85\%$ in the multi-level task. This reduction likely results from the localized information in each tile, which may capture irrelevant details like non-expressive facial areas or background elements. Including the resized full-frame ($224\times 224$ pixels) alongside tiles further decreased accuracy to $65.11\%$ and $27.84\%$ for binary and multi-level tasks, respectively. However, introducing a coefficient ($c=0.1$) to adjust the tile embeddings restored some performance, achieving $74.86\%$ and $33.86\%$ in respective tasks.

The integration of two augmentation techniques, \textit{Masking} and \textit{AugmNet}, along with the \textit{Basic} method, was then tested. \textit{Masking} reduced performance by $1.81\%$ and $1.72\%$, and \textit{AugmNet} showed smaller declines of $0.03\%$ and $0.13\%$. Using both techniques together resulted in better outcomes than \textit{Masking} alone but did not independently surpass the performance of \textit{AugmNet}.
Despite these initial results, combining all augmentation methods proved advantageous for extended training periods. This approach addresses the risk of overfitting through a robust regularization strategy. Ultimately, this comprehensive strategy led to final accuracy rates of $77.10\%$ and $35.39\%$ for binary and multi-level classifications, demonstrating its effectiveness in an unimodal, vision-based pain assessment framework.

\renewcommand{\arraystretch}{1.2}
\begin{table}
\center
\footnotesize
\caption{Results utilizing the video modality.}
\label{table:video_modality}
\begin{center}
\begin{threeparttable}
\begin{tabular}{ P{0.8cm} P{1.0cm} P{1.0cm} P{1.9cm} P{0.9cm} P{0.9cm}  P{0.9cm} P{1.5cm}  P{1.4cm}  P{0.9cm}}
\toprule
\multirow{2}[2]{*}{\shortstack{Epochs}}
&\multicolumn{2}{c}{Pretraining stage}
&\multicolumn{2}{c}{Pipeline}
&\multicolumn{3}{c}{Augmentations} 
&\multicolumn{2}{c}{Task}\\ 
\cmidrule(lr){2-3}\cmidrule(lr){4-5}\cmidrule(lr){6-8}\cmidrule(lr){9-10}
&\nth{1}  &\nth{2} &Full frame &Tiles  &Basic &Mask &AugmNet &NP vs P\textsubscript{4} &MC\\
\midrule
\midrule
500 &\checkmark &- &\checkmark &- &\checkmark &- &- &72.56  &31.22 \\
500 &- &\checkmark &\checkmark &- &\checkmark &- &- &74.25  &33.34 \\
\hline
500 &-  &\checkmark &- &\checkmark &\checkmark &- &- &68.07 &31.49\\
500 &- &\checkmark &\checkmark &\checkmark &\checkmark &- &- &65.11 &27.84\\
500 &-  &\checkmark &\checkmark &$\checkmark^c$ &\checkmark &- &- &74.86 &33.86\\
\hline
500 &-  &\checkmark &\checkmark &$\checkmark^c$ &\checkmark &\checkmark &- &73.05 &32.14\\
500 &-  &\checkmark &\checkmark &$\checkmark^c$ &\checkmark &- &\checkmark &74.83 &33.73\\
500 &-  &\checkmark &\checkmark &$\checkmark^c$ &\checkmark &\checkmark &\checkmark &73.16 &32.87\\
\hline
800 &-  &\checkmark &\checkmark &$\checkmark^c$ &\checkmark &\checkmark &\checkmark &\textbf{77.10} &\textbf{35.39}\\
\bottomrule 
\end{tabular}
\begin{tablenotes}
\scriptsize
\item Stage: referring to pretraining process for \textit{Spatial-Module} \space\space Mask: Masking \space\space c: constant-coefficient applied exclusively to the tiles \space\space NP: No Pain \space\space P\textsubscript{4}: Very Severe Pain \space\space MC: multiclass pain level
\end{tablenotes}
\end{threeparttable}
\end{center}
\end{table}

\subsubsection{Heart rate modality}
\label{hr_encoder}
The experiments concerning the heart rate modality explore the use of the encoder and various augmentation methods. Table \ref{table:hr_modality} details all experiments involving the heart rate modality.
Initially, using the original heart rate vectors with a dimensionality of $h=5$, we achieved classification scores of $61.70\%$ for NP vs P\textsubscript{4} and $27.60\%$ for the multi-level task. 
After applying the \textit{Heart Rate Encoder} to map these vectors to a higher-dimensional space of $h=2048$, we noted a slight improvement: a $0.23\%$ increase for the binary task and $0.08\%$ for the multi-level task.
Despite the considerable increase in embedding size, this modest enhancement suggests that the intrinsic information within the limited heart rate data points does not significantly enhance the feature representation.
Nonetheless, the encoder's use is vital for producing larger embeddings, especially for our multimodal approach integrating video and heart rate data, which will be discussed in subsequent sections.

We also tested augmentation methods on the heart rate data. Applying \textit{Masking} yielded a slight improvement of $0.02\%$ for the binary task and $0.05\%$ for the multi-level task. Implementing \textit{AugmNet} further enhanced performance to $62.09\%$ and $28.11\%$ for binary and multi-level tasks, respectively. However, combining all augmentations decreased performance to $61.87\%$ and $27.96\%$.
During an extended $800$-epoch training period, we achieved $64.87\%$ accuracy for the binary task and $29.81\%$ accuracy for the multi-level task using all augmentations.
Despite these gains, we found that augmentations pose more challenges for accurate heart rate classification than video. Therefore, we repeated the extended training without \textit{Basic} and \textit{Masking}, keeping only \textit{AugmNet}, which improved binary task performance to $67.04\%$ and multi-level to $31.22\%$. This reduction in heart rate embedding corruption significantly enhanced performance.
The differing effects of augmentations between heart rate and video modalities highlight the challenges of using a single, isolated feature in a machine learning system.
We infer that heart rate embeddings with limited informational content are more vulnerable to significant performance degradation from augmentations than richer video embeddings.

\renewcommand{\arraystretch}{1.2}
\begin{table}
\center
\small
\caption{Results utilizing the heart rate modality.}
\label{table:hr_modality}
\begin{center}
\begin{threeparttable}
\begin{tabular}{ P{1.3cm} P{1.6cm} P{1.1cm} P{1.1cm} P{1.7cm}  P{1.5cm} P{1.1cm}}
\toprule
\multirow{2}[2]{*}{\shortstack{Epochs}}
&\multirow{2}[2]{*}{\shortstack{HR \\Encoder}}
&\multicolumn{3}{c}{Augmentations} 
&\multicolumn{2}{c}{Task}\\ 
\cmidrule(lr){3-5}\cmidrule(lr){6-7}
&&Basic &Mask &AugmNet &NP vs P\textsubscript{4} &MC\\
\midrule 
\midrule
500 &- &\checkmark &- &- &61.70  &27.60 \\
500 &\checkmark &\checkmark &- &- &61.93  &27.68 \\
\hline
500 &\checkmark &\checkmark &\checkmark &- &61.95  &27.73 \\
500 &\checkmark &\checkmark &- &\checkmark &62.09  &28.11 \\
500 &\checkmark &\checkmark &\checkmark &\checkmark &61.87  &27.96 \\
\hline
800 &\checkmark &\checkmark &\checkmark &\checkmark &64.84  &29.81 \\
800 &\checkmark &- &- &\checkmark &\textbf{67.04}  &\textbf{31.22} \\
\bottomrule 
\end{tabular}
\begin{tablenotes}
\scriptsize
\item  \space 
\end{tablenotes}
\end{threeparttable}
\end{center}
\end{table}

\subsubsection{Multimodality}
The results of integrating video and heart rate modalities are detailed in Table \ref{table:multimodal}. Based on the insights gained from separate experiments with each modality, we extended the training duration to $800$ epochs. For this integrated approach, we utilized the tiles with a coefficient $c=0.1$ and applied \textit{AugmNet} as the sole augmentation method.
This fusion strategy resulted in a classification accuracy of $82.74\%$ for NP vs. P\textsubscript{4} and $39.77\%$ for the multi-level classification task.
These results mark a substantial enhancement, with improvements of $5.64\%$ and $15.70\%$ over the individual performances of the video and heart rate modalities, respectively, for the binary classification. Similarly, for the multi-level classification, the integrated approach shows a $4.38\%$ and $8.55\%$ increase compared to the standalone modalities. The combination of these two pivotal modalities significantly boosts the efficacy of the pain assessment process, outperforming the results obtained by each modality on its own.

\renewcommand{\arraystretch}{1.2}
\begin{table}
\center
\small
\caption{Results utilizing the video \& the heart rate modality.}
\label{table:multimodal}
\begin{center}
\begin{threeparttable}
\begin{tabular}{ P{1.2cm} P{1.5cm} P{2.0cm} P{1.0cm} P{1.0cm} P{1.0cm} P{1.6cm}  P{1.5cm} P{1.0cm}}
\toprule
\multirow{2}[2]{*}{\shortstack{Epochs}}
&\multirow{2}[2]{*}{\shortstack{HR \\Encoder}}
&\multicolumn{2}{c}{Pipeline}
&\multicolumn{3}{c}{Augmentations} 
&\multicolumn{2}{c}{Task}\\ 
\cmidrule(lr){3-4}\cmidrule(lr){5-7}\cmidrule(lr){8-9}
& &Full frame &Tiles &Basic &Mask &AugmNet &NP vs P\textsubscript{4} &MC\\
\midrule
\midrule
800 &\checkmark &\checkmark &$\checkmark^c$ &- &- &\checkmark &82.74  &39.77 \\
\bottomrule 
\end{tabular}
\begin{tablenotes}
\scriptsize
\item  \space 
\end{tablenotes}
\end{threeparttable}
\end{center}
\end{table}

\subsubsection{Comparison with existing methods}
In this section, a comparative analysis is performed to evaluate the performance of our method against other existing approaches documented in the literature. This evaluation is based on Part A of the \textit{BioVid} dataset, including all $87$ participants. The same evaluation protocol---leave-one-subject-out (LOSO) cross-validation---is adhered to for all comparisons to ensure fairness and accuracy. Our method is contrasted with both unimodal and multimodal approaches, divided into (1) video-based, (2) ECG-based, and (3) multimodal studies regardless of the modalities involved. The outcomes are summarized in Table \ref{table:literature}.

For video-based studies, our approach, achieving $77.10\%$ in binary and $35.39\%$ in multi-level classification tasks, is recognized as one of the highest-performing methods. It exceeds the average results of comparative studies by about $4.7\%$ for binary and $3.4\%$ for multi-level tasks. 
Regarding ECG-based studies, our method shows superior performance, exceeding the average by $8.5\%$ and $18.1\%$ for binary and multi-level classifications, respectively. Remarkably, it records the highest classification accuracy in the multi-level task at $31.22\%$. These findings underscore the effectiveness of using heart rate data extracted from ECG as a standalone feature, establishing our method's capability to assess pain accurately and achieve state-of-the-art results.
In multimodal studies, our approach records an impressive $82.74\%$ accuracy in the NP vs. P\textsubscript{4} task, making it one of the top-performing methods. It is slightly outperformed by studies \cite{huang_dong_2022} and \cite{thiam_kestler_schenker_2020_b}, which achieved $88.10\%$ and $83.99\%$, respectively.
For multi-level tasks, comparisons are scarce; however, study \cite{huang_dong_2022} reached $42.20\%$, and \cite{thiam_bellmann_kestler_2019} reported $36.54\%$, positioning our method favorably within this context.

\renewcommand{\arraystretch}{1.2}
\begin{table}
\center
\scriptsize
\caption{Comparison of studies utilizing \textit{BioVid} \& LOSO validation, reported on accuracy \%.}
\label{table:literature}
\begin{center}
\begin{threeparttable}
\begin{tabular}{ P{0.7cm} P{2.3cm}  P{2.2cm} P{2.2cm} P{1.4cm} P{1.3cm} P{1.1cm} P{0.7cm}}
\toprule
\multirow{2}[2]{*}{\shortstack{Study}}
&\multirow{2}[2]{*}{\shortstack{Modality}}
&\multicolumn{4}{c}{Method} 
&\multicolumn{2}{c}{Task}\\ 
\cmidrule(lr){3-6}\cmidrule(lr){7-8}
&&Features &Machine Learning &Params (M) &FLOPS (G) &NP vs P\textsubscript{4} &MC\\

\midrule
\midrule
\rowcolor{mygray}
\cite{werner_hamadi_walter_2017}  &Video &optical flow &RF &- &- &70.20  &- \\
\cite{zhi_wan_2019}  &Video &raw &SLSTM &- &- &61.70  &29.70 \\
\rowcolor{mygray}
\cite{huang_dong_2022}  &Video &raw &3D CNN, &- &- &\textbf{77.50}  &34.30 \\
\cite{thiam_kestler_schenker_2020}  &Video &raw &2D CNN, biLSTM &- &- &69.25  &- \\
\rowcolor{mygray}
\cite{tavakolian_bordallo_liu_2020}  &Video &raw &2D CNN &25.00$^\varocircle$ &4.00 &71.00  &- \\
\cite{werner_2016}  &Video &facial action\newline descriptors &Deep RF &- &- &72.40  &30.80 \\
\rowcolor{mygray}
\cite{werner_2016}  &Video &facial 3D distances &Deep RF &- &- &72.10  &30.30 \\
\cite{patania_2022}  &Video &fiducial points &GNN &- &- &73.20  &- \\
\rowcolor{mygray}
\cite{huang_xia_li_2019}$^\dagger$  &Video &raw &2D CNN &- &- &71.30  &37.60 \\
\cite{huang_xia_2020}$^\dagger$  &Video &raw &2D CNN, GRU &150.00$^\varocircle$ &- &73.90  &\textbf{39.10} \\
\rowcolor{mygray}
\cite{gkikas_tsiknakis_embc} &Video &raw &Transformer &24.00 &4.20 &73.28 &31.52\\
\cite{werner_hamadi_2014}  &Video &facial landmarks,\newline 3D distances &RF & & &71.60  &- \\
\rowcolor{mygray}
Our &Video &raw &Transformer &4.20$^\varoast$ &1.62 &77.10 &35.39\\
\Xhline{2\arrayrulewidth}
\cite{thiam_bellmann_kestler_2019}  &ECG &raw &1D CNN &1.80$^\varocircle$ &- &57.04  &23.23 \\
\rowcolor{mygray}
\cite{martinez_picard_2018_b}  &ECG &domain-specific$^\divideontimes$ &LR &- &- &57.69  &- \\
\cite{gkikas_chatzaki_2022}  &ECG & domain-specific$^\divideontimes$ &SVM &- &- &58.39 &23.79 \\
\rowcolor{mygray}
\cite{huang_dong_2022}  &ECG &heart rate$^{\star}$ &3D CNN &- &- &65.00 &28.50 \\
\cite{werner_hamadi_2014}  &ECG &domain-specific &RF &- &- &62.00  &- \\
\rowcolor{mygray}
\cite{gkikas_chatzaki_2023}  &ECG &domain-specific &FCN &4.09$^\varodot$ &0.40 &\textbf{69.40}  &30.24 \\
\cite{kachele_werner_2015} &ECG & domain-specific$^{\divideontimes}$ &SVM &- &- &63.50  &- \\
\rowcolor{mygray}
Our &ECG &heart rate &Transformer &6.03$^\varoast$ &1.25 &67.04  &\textbf{31.22}\\
\Xhline{2\arrayrulewidth}
\cite{thiam_bellmann_kestler_2019}  &ECG, EMG, GSR &raw &2D CNN &10.00$^\varocircle$ &- &76.72 &36.54 \\
\rowcolor{mygray}
\cite{martinez_picard_2018_b}  &ECG, GSR &domain-specific$^\divideontimes$ &SVM &- &- &72.20 &- \\
\cite{huang_dong_2022}  &Video$^1$,  ECG$^2$ &raw$^1$,  heart rate$^{2\star}$ &3D CNN &- &- &\textbf{88.10} &\textbf{42.20} \\
\rowcolor{mygray}
\cite{werner_hamadi_2014}  &ECG$^1$,  EMG$^1$, GSR$^1$ &domain-specific$^{1\divideontimes}$ &RF &- &- &74.10  &- \\
\cite{werner_hamadi_2014}  &Video$^1$,ECG$^2$, EMG$^2$, GSR$^2$ &facial landmarks$^1$, \newline 3D distances$^1$, domain-specific$^{2\divideontimes}$ &RF &- &- &77.80  &- \\
\rowcolor{mygray}
\cite{kachele_werner_2015} &Video$^1$, ECG$^2$,  GSR$^2$ &facial landmarks$^1$, \newline 3D distances$^1$, domain-specific$^{2\divideontimes}$ &RF &- &- &78.90  &- \\
\cite{kachele_werner_2015} &Video$^1$,  ECG$^2$,  EMG$^2$, GSR$^2$ &facial landmarks$^1$, \newline 3D distances$^1$, domain-specific$^{2\divideontimes}$ &SVM &- &- &76.60  &- \\
\rowcolor{mygray}
\cite{thiam_kestler_schenker_2020_b} &ECG, EMG, GSR&raw &DDCAE &4.00$^\varocircle$ &- &83.99 &- \\
Our &Video$^1$,ECG$^2$ &raw$^1$, heart rate$^2$ &Transformer &8.60$^\varoast$ &2.44 &82.74 &39.77 \\
\bottomrule 
\end{tabular}
\begin{tablenotes}
\scriptsize
\item M: Millions \space G: Giga \space $\dagger$: reimplemented for pain intensity estimation on \textit{BioVid} by \cite{huang_dong_2022} \space 
${\star}$: pseudo heart rate gain \space $\divideontimes$: numerous features\space $\varocircle$: parameter count estimated from provided paper details \space $\varoast$: \textit{AugmNet} excluded from parameter count, not used in inference \space $\varodot$: parameter count not mentioned in study, provided directly by authors \space -: missing value \space RF: Random Forest\space AE-ATT: Autoencoder Attention \space SVM: Support Vector Machines \space LR: Logistic Regression
\end{tablenotes}
\end{threeparttable}
\end{center}
\end{table}

\subsubsection{Inference time}
\begin{figure}[h]
\begin{center}
\includegraphics[scale=0.4]{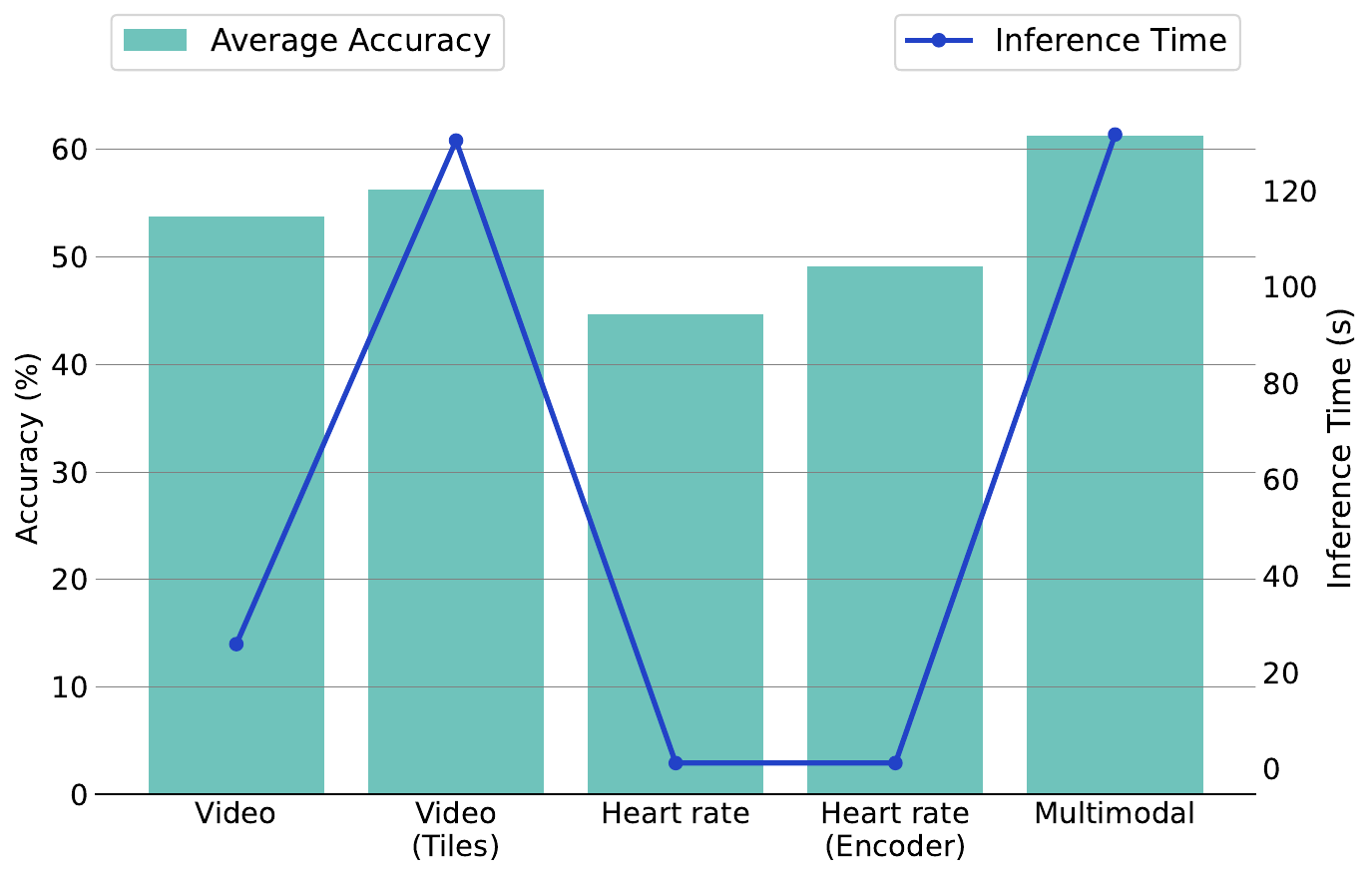}
\end{center}
\caption{Comparison of mean accuracy and inference period for unimodal and multimodal strategies across NP versus P\textsubscript{4} and MC tasks. The diagram adopts a dual-y-axis configuration---accuracy measurements on the left and time metrics on the right---to outline the balance between performance efficacy and computational load, categorizing the methodologies along the x-axis.}
\label{inference_diagram}
\end{figure}

We explored several methodologies in this study, including a video-based approach, video incorporating tiles, a heart rate-based approach, heart rate analysis with an encoder, and a combined multimodal strategy. Figure \ref{inference_diagram} illustrates each method's inference time in seconds and the corresponding average accuracy performances across binary and multi-level tasks. Table \ref{table:module_parameters} details the number of parameters and the computational cost of floating-point operations (FLOPS) for each component.
Inference tests were conducted on an \textit{Intel Core i7-8750H} CPU, including the time for face detection in each frame but excluding the extraction of heart rate from electrocardiography, focusing on the potential use of automatically provided cardiac features from wearables.

The inference time for the video modality employing the standard pipeline is approximately $26$ seconds. Utilizing the tile pipeline  increases inference time significantly, soaring to about $130$ seconds due to processing five image representations per frame---one full frame and four tiles---compared to a single image representation in the non-tiled approach.
In the context of heart rate signals, completing a pain assessment requires only $1.2$ seconds. With the integration of the \textit{Heart Rate Encoder}, the processing time remains virtually unchanged, showing a negligible increase of less than half a second, highlighting this specific encoder's efficiency.
Lastly, the comprehensive multimodal framework incorporating the tiles and the \textit{Heart Rate Encoder} demands about $131$ seconds, illustrating the increased complexity and computational requirements when combining multiple modalities.

\renewcommand{\arraystretch}{1.2}
\begin{table}
\center
\small
\caption{Module parameters and computational cost in FLOPS for the proposed framework.}
\label{table:module_parameters}
\begin{center}
\begin{threeparttable}
\begin{tabular}{ P{3.3cm}  P{2.0cm}  P{2.0cm}}
\toprule
Module & Params (M) &FLOPS  (G) \\
\midrule
\midrule
Spatial-Module   &2.57 &1.19  \\
Heart Rate Encoder  &4.40 &0.82 \\
AugmNet  &1.02 &0.02 \\
Temporal-Module  &1.63 &0.43\\
\hline
Total &9.62 &2.46\\
\bottomrule
\end{tabular}
\begin{tablenotes}
\scriptsize
\item 
\end{tablenotes}
\end{threeparttable}
\end{center}
\end{table}

\subsubsection{Interpretation}
\label{interpretation}
\begin{figure}
\centering
\begin{subfigure}{0.55\textwidth}
\includegraphics[width=\linewidth]{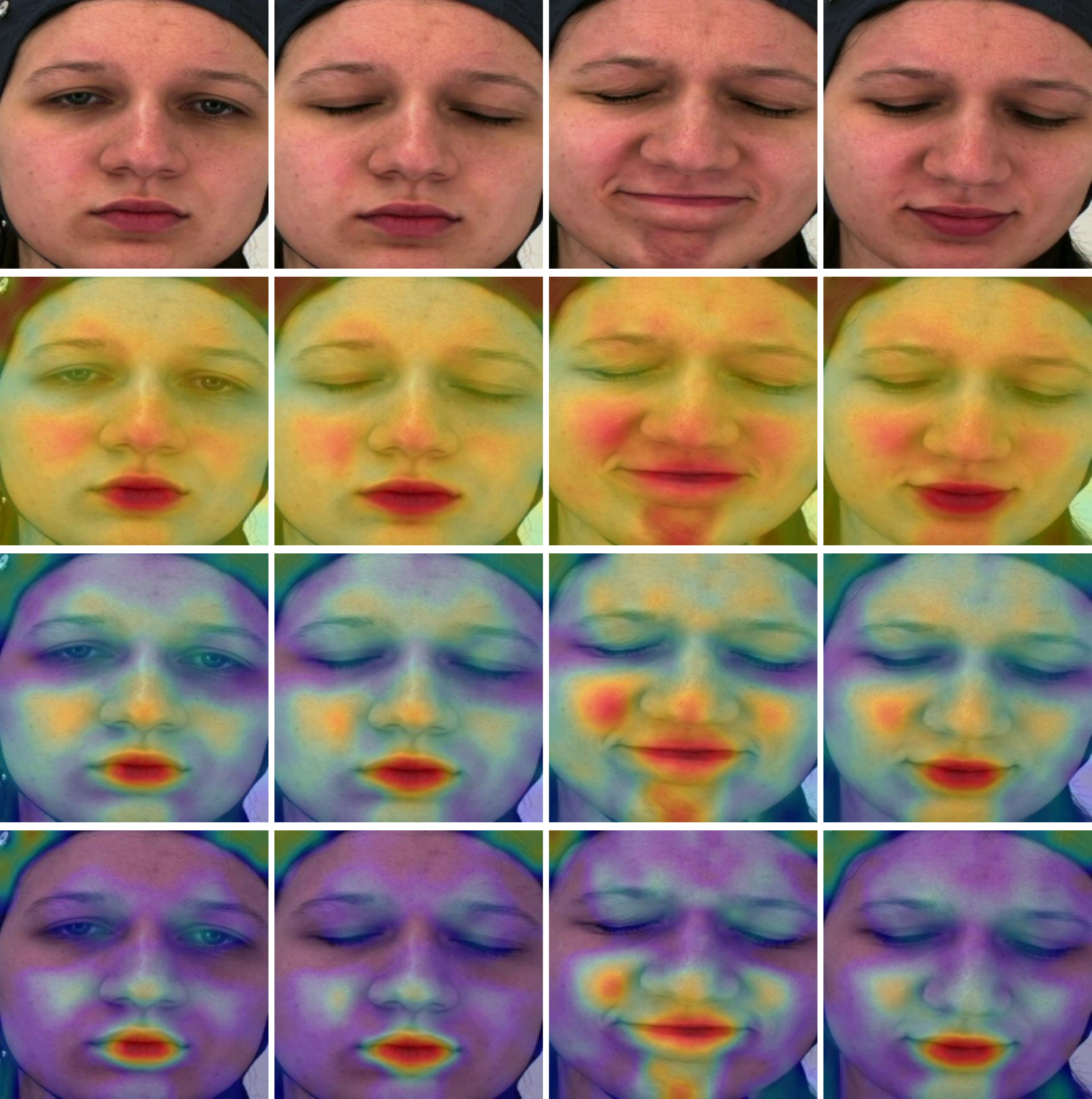}
\caption{Attention maps from the \textit{Spatial-Module}.}
\label{attention_map_spatial}
\end{subfigure}  
\vspace{10pt}
   
\begin{subfigure}{0.55\textwidth}
\includegraphics[width=\linewidth]{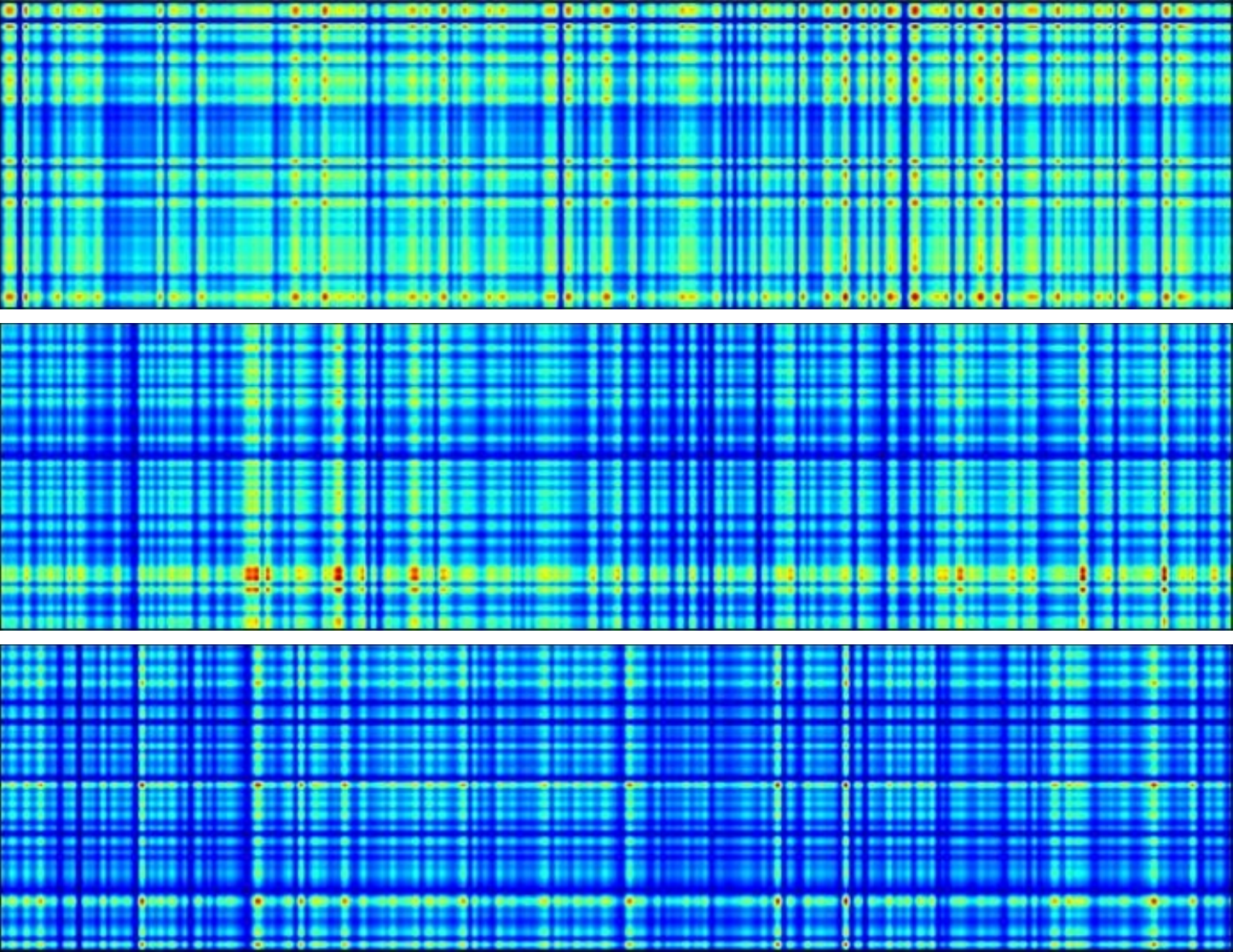}
\caption{Attention maps from the \textit{Temporal-Module}.}
\label{attention_map_temporal}
\end{subfigure}


\caption{\textit{Regions highlighted in yellow and red denote areas of significant attention.} \textbf{(a)} \textbf{(1\textsuperscript{st} row)} Sequence of original frames. \textbf{(2\textsuperscript{nd} row)} Derived from the \textit{Spatial-Module} after initial stage pretraining. \textbf{(3\textsuperscript{rd} row)} Derived from the \textit{Spatial-Module} post second stage pretraining. \textbf{(4\textsuperscript{th} row)} Derived from the \textit{Spatial-Module} trained on the \textit{BioVid} dataset. 
\textbf{(b)} \textbf{(1\textsuperscript{st} row)} Derived from the \textit{Temporal-Module} incorporating video embeddings. \textbf{(2\textsuperscript{nd} row)} Derived from the \textit{Temporal-Module} with heart rate embeddings. \textbf{(3\textsuperscript{rd} row)} Derived from the \textit{Temporal-Module} using a combined embedding of video and heart rate.}
\label{attention_maps}
\end{figure}

Improving model interpretability is essential for their acceptance and integration into clinical settings. This study generates attention maps from both the \textit{Spatial-Module} and the \textit{Temporal-Module}, as illustrated in Figure \ref{attention_maps}.
For the \textit{Spatial-Module}, attention maps are derived from the last fully connected layer's weight contributions, which are then interpolated onto the images to highlight the model's focal areas.
Figure \ref{attention_map_spatial} displays an original frame sequence along with three attention map variations: (1) post-initial pretraining, (2) after the second pretraining phase, and (3) post-training on the \textit{BioVid}. Based on face recognition tasks, the \textit{Spatial-Module} produces maps focusing broadly on the facial area, particularly the zygomatic, buccal, oral, mental, and nasal regions. The second stage, oriented towards multi-task emotion recognition, refines the focus, sharpening attention on specific facial areas, highlighted in the first stage but with greater clarity and emphasis. After training on the \textit{BioVid} for pain assessment, the attention maps show further refined focus on specific facial areas, with reduced attention to less relevant regions, ensuring concentrated focus on key areas. These maps consistently demonstrate the model's capability to adjust its focus based on pain-related facial expressions.

Attention maps from the \textit{Temporal-Module}, based on input embeddings, illustrate the weight contributions in the module's final layer, forming easy-to-visualize rectangular patterns.
Figure \ref{attention_map_temporal} shows examples for three scenarios: (1) video embedding, (2) heart rate embedding, and (3) a combined embedding of video and heart rate.
The attention maps exhibit a grid-like pattern, possibly due to the Fourier position encoding used, akin to those seen in perceiver-like architectures. The video embedding map shows intense attention across the input. In contrast, the heart rate embedding map focuses attention more sparsely, with a notable concentration in specific areas indicated by intense red coloration. The combined embedding map displays moderate intensity, consistent with the blended nature of the input. These maps tend to emphasize the latter part of the session, aligning with the timing of pain manifestation towards the session's end, indicating the model's responsiveness to real-time pain expressions.

\subsection{Discussion}
\label{discussion}
This research introduced a multimodal framework that integrates video and heart rate signals to assess pain automatically. Our innovative approach includes four main modules, each characterized by effectiveness and efficiency. The \textit{Spatial Module}, particularly notable for its compact size of only $2.57$ million parameters, ranks as one of the most efficient vision-based models in automatic pain assessment literature.
Despite limited comparative studies, our model has shown it can match or exceed the performance of larger models. Its high efficiency and robust performance are primarily due to a thorough pretraining regime on datasets related to affective responses, crucial for enhancing model capabilities in pain estimation tasks.
The \textit{Heart Rate Encoder}, with $4.40$ million parameters, excels at transforming heart rate data into complex, high-dimensional embeddings, which integrate seamlessly with video data during inference, all within under half a second. This quick processing underscores the encoder's efficiency, supported by bicubic interpolation that modifies input dynamically to achieve variable output dimensions without predefined constraints.
\textit{AugmNet}, a novel augmentation module, learns to modify latent space representations directly, preventing the need for specific augmentation techniques designed for each data type. However, this module requires careful application to avoid overfitting and other training challenges.
The \textit{Temporal-Module}, consisting of $1.63$ million parameters, is crucial for final pain level assessments. It leverages a mix of cross- and self-attention mechanisms to enhance efficiency and accuracy, demonstrating the potential of transformers in streamlined settings contrary to their typical use in large-scale applications.

Our experiments demonstrate that videos are invaluable for discerning individual pain experiences by capturing diverse behavioral indicators like facial expressions, eye movements, and even slight color changes under stress. Utilizing video data, our methodology reached an accuracy of $77.10\%$ in binary classification, effectively distinguishing between no pain and very severe pain scenarios. Moreover, it achieved $35.39\%$ accuracy in a multi-level classification spanning five distinct pain intensities.
The heart rate signal, tested as a standalone feature from electrocardiography, showed that remarkable outcomes are possible with this single data feature, which is pivotal for validating the feasibility of heart rate as a viable pain indicator. This is crucial as heart rate data is readily accessible from most wearable technologies, reducing the need for specialized algorithms to handle cardiac signals or raw biosignals, thereby conserving both time and computational resources. Solely using heart rate, our model excelled, registering accuracies of $67.04\%$ and $31.22\%$ for binary and multi-level classifications, respectively, among the highest reported.
Incorporating video and heart rate data, our multimodal method yielded superior results---$82.74\%$ and $39.77\%$ for binary and multi-level classifications, respectively. These figures significantly enhance video-only results by roughly $9\%$ and heart rate-only outcomes by about $24\%$. Furthermore, with a total parameter count of just $9.62$ million, our approach stands out for its efficiency and effectiveness.
Overall, this study showcases the synergy achievable by merging video and heart rate data, leading to a system outperforming its unimodal counterparts and emphasizing the potential of integrated multimodal pain assessment tools.

Using attention maps generated by the \textit{Spatial-Module}, our framework analysis identified crucial facial areas like the zygomatic and oral regions as significant for automatic pain assessment. These maps demonstrated that different pretraining stages refine the focus, showing more targeted attention with specialized training. Similarly, attention maps from the \textit{Temporal-Module} focused on the latter part of the input image, corresponding to where pain manifestations are typically observed in the particular dataset.

\section{Summary}
This chapter explores the interplay between model efficiency and performance in automatic pain assessment tasks. We also aimed to mirror real-world conditions by leveraging readily accessible and applicable modalities without relying on costly precision medical devices. Consequently, we utilized RGB videos with a resolution of approximately 1080x1080 and heart rate data.
The videos utilized are of medium quality, comparable to those captured with mobile phone cameras, and the heart rate data simulates readings from wearable devices. It is important to note that wearables across various price ranges automatically provide heart rate information. Consequently, exploring the potential of using this readily available modality for pain assessment is crucial. This proof of concept is significant as it could enable cost-effective and accessible pain assessment solutions without dependence on specialized medical equipment. Additionally, our study focuses on developing compact and efficient models that maintain robust performance.

The experiments detailed in this chapter reveal that reducing the number of frames used in a video-based pipeline by a factor of four minimizes the accuracy loss, under $1.5\%$, while cutting inference time threefold, facilitating near real-time pain assessment. Furthermore, our initially proposed framework, which incorporates $24$ million parameters and operates at $4.3$ GFLOPS, demonstrated exceptionally high performance.
In the subsequent experiments, we showcased that heart rate data alone can be effectively used for pain assessment, achieving impressive results. This finding underscores the practical utility of data available from wearables. Additionally, combining video data with heart rate information yielded the highest accuracy in our tests, illustrating that an integrated approach using both behavioral and physiological modalities can significantly enhance performance.
Additionally, we demonstrated that creating one of the smallest models documented, with only $9.62$ million parameters and $2.46$ GFLOPS, allowed us to achieve top-tier results. This highlights that large models, commonly favored in the current era of AI, are not always necessary for effective performance. However, it is important to note that extensive multi-stage pretraining across various datasets greatly aided this framework's success, which was critical in achieving such high efficiency and effectiveness.

In this chapter, we aimed to explore the effectiveness of compact models in achieving high performance with rapid inference times. We exclusively used heart rate data, mimicking the information typically available from wearable devices, though our experiments did not extend to real-world conditions. Instead, they relied on the sole publicly accessible dataset explicitly designed for pain assessment, including facial videos and cardiac signals collected in highly controlled laboratory settings. Participants were positioned facing forward under optimal lighting conditions, with physiological sensors precisely affixed.
Recognizing the potential challenges of applying these findings to clinical settings is essential. Issues like inconsistent lighting, unforeseen facial movements, occlusions, or difficulties with sensor placement must be meticulously addressed to tailor these systems for real-world use. Moreover, depending solely on heart rate as a cardiac feature could be restrictive in more demanding scenarios, highlighting the necessity to integrate multiple extracted features or uutilizeraw biosignals for comprehensive assessments.

    \chapter{Synthetic Data: The Role of Thermal Imaging}
\minitoc  
\label{chapter_6}

\section{Chapter Overview: Introduction \& Related Work}
This chapter presents the findings from the study published in \cite{gkikas_tsiknakis_thermal_2024}.
In recent years, synthetic data generation has gained traction as a viable approach to addressing data scarcity and privacy issues while also meeting the requirements for training AI algorithms on unbiased data with adequate sample size and statistical robustness. Additionally, synthetic data can increase the availability and diversity of real data, particularly in rare modalities, which can be essential for training AI-driven diagnostic and predictive models. This enhancement supports healthcare research and improves patient outcomes.\cite{pezoulas_zaridis_2024}.
In the literature on automatic pain assessment, no studies have been reported concerning creating synthetic modalities. This chapter introduces the generation process of synthetic thermal videos using \textit{Generative Adversarial Networks} (GANs).

Regarding thermal modality, in recent years, the field of affective computing research has increasingly adopted thermal imaging techniques \cite{qudah_2021}. This shift was motivated by studies showing that stress and cognitive load significantly affect skin temperature \cite{ioannou_2014}, attributable to the autonomic nervous system's (ANS) regulation of physiological signals such as heart rate, respiration rate, blood perfusion, and body temperature. These signals are vital indicators of human emotions and affect \cite{qudah_2021}. Moreover, muscle contractions can alter facial temperature by transferring heat to the facial skin \cite{jarlier_2011}. Consequently, thermal imaging has emerged as a viable method for recording transient facial temperatures \cite{merla_2004}. Research by the authors in \cite{youssef_2023} on thermal imaging and facial action units to evaluate emotions, such as frustration, boredom, and enjoyment, indicated that a multimodal approach yielded the most accurate results. Within pain research, thermal imaging has been explored in limited studies. For instance, \cite{erel_ozkan_2017} reported increased facial temperature in response to painful stimuli, suggesting thermal cameras as effective tools for monitoring pain. Another study \cite{haque_2018} introduced a pain dataset consisting of RGB, thermal, and depth videos, finding that while the RGB modality slightly outperformed the others, integrating all modalities provided the best results.
This prompted us to explore the specific modality of thermal imagery through the prism of synthesis using generative deep learning.

\section{Synthetic Thermal Videos using Generative Adversarial Networks}
\label{chapter_6_paper_1}
We present a process of creating synthetic thermal videos using GANs in unimodal and multimodal settings alongside RGB video modalities. Integrating a Vision Multilayer Perceptron (MLP) model with a transformer-based module is at the core of our automatic pain assessment framework. Key contributions of this research include (1) generating synthetic thermal videos to enhance pain assessment as an additional vision modality, (2) assessing the effectiveness of RGB and synthetic thermal videos as standalone modalities, (3) examining the utility of thermal-related information for pain assessment, and (4) evaluating the performance and implications of the newly developed Vision-MLP architectures.

\subsection{Methodology}
This section outlines the generation of synthetic thermal videos, which will be utilized subsequently and incorporated into an automatic assessment pipeline.

\subsubsection{Synthetic Thermal Videos}
\begin{figure}
\begin{center}
\includegraphics[scale=0.20]{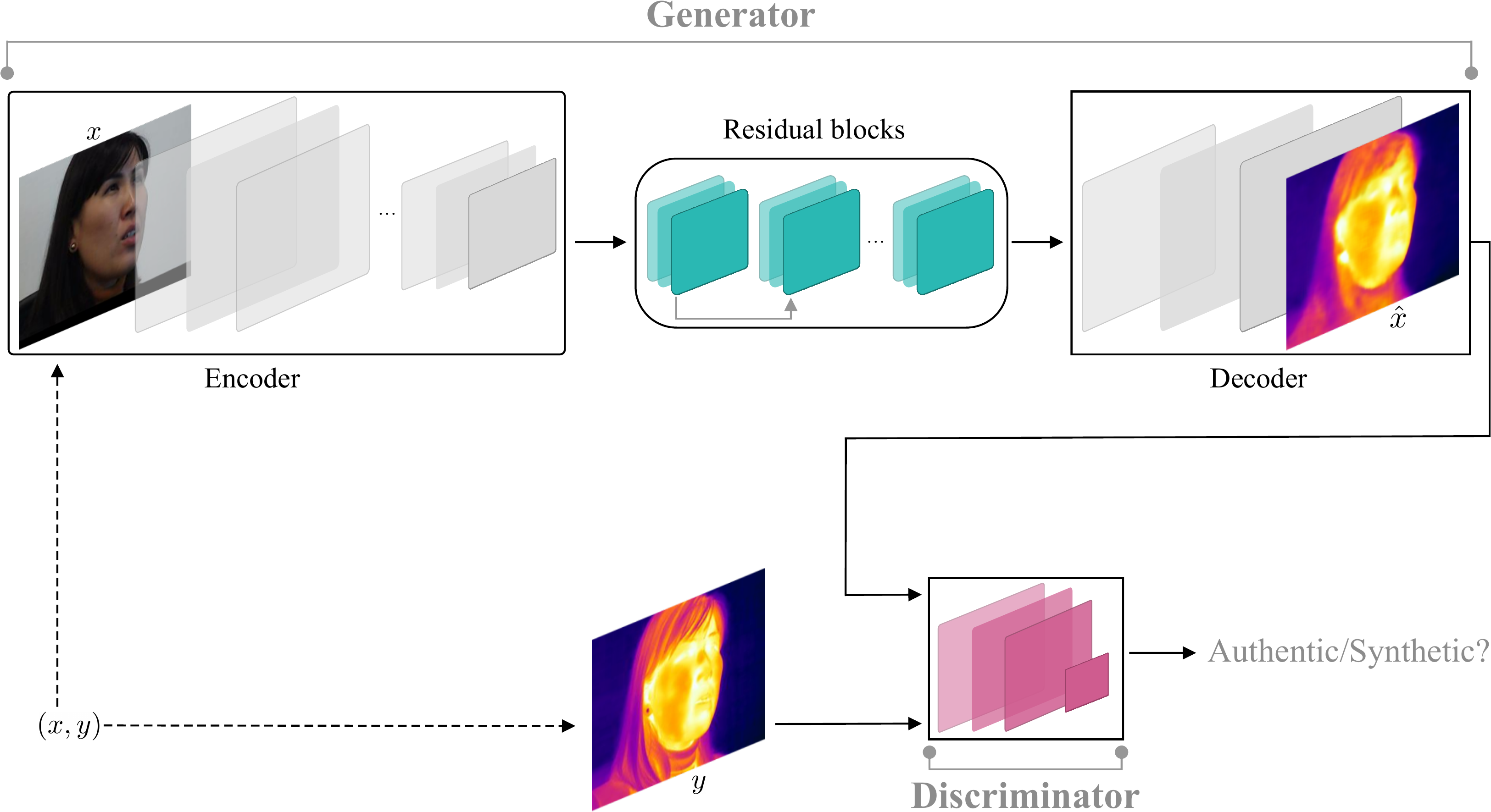}
\end{center}
\caption{Illustration of the procedure for creating thermal images, featuring the architecture of the Generator $G$ (Encoder, mid-stage ResNet, Decoder), and the Discriminator $D$.}
\label{gans}
\end{figure}

An image-to-image translation (I2I) approach has been utilized to create synthetic thermal videos. I2I generative models are designed to bridge different image domains by learning the data distributions inherent to each domain. Here, the source domain comprises RGB images; the target domain is thermal images.
In this research, conditional generative adversarial networks (cGANs) \cite{mirza_2014} were employed and trained in a supervised manner using paired images. Fig. \ref{gans} provides a high-level overview of the method. The generator \(G\) produces images that appear realistic, whereas the discriminator \(D\) works to differentiate between genuine and synthetic images through the following minimax game:
\begin{equation}
\min_{G} \max_{D} \mathcal{L}_{\text{cGAN}}(G, D),
\end{equation}
where the objective function $\mathcal{L}_{\text{cGAN}}(G, D)$ is defined as:
\begin{equation}
\mathbb{E}_{x, y}[\log D(x, y)] + \mathbb{E}_{x, z}[\log(1 - D(x, G(x, z)))],
\end{equation}
with $x$ denoting the actual data, $y$ indicating the target data, and $z$ representing the random noise vector. Here, $G$ seeks to minimize the objective function, whereas $D$ operates in opposition, striving to maximize it.
Additionally, we incorporated the \textit{Wasserstein} gradient penalty (WGAN-GP) \cite{wgans} to enhance the stability of training. The overall objective is articulated as:
\begin{equation}
\mathcal{L}_{\text{cGAN}}(G, D) + \lambda \mathbb{E}_{\hat{x}, y}[(\|\nabla_{\hat{x}} D(\hat{x}, y)\|_2 - 1)^2],
\end{equation}
where $\lambda$ is the penalty coefficient.
In the architectural design of our proposed method, which draws inspiration from \cite{wang_liu_2018}, the generator $G$ is divided into three distinct modules: an encoder, which includes two convolutional layers that downsample the input; a middle ResNet module, featuring nine residual blocks, each equipped with two convolutional layers; and a decoder that upsamples the feature maps back to the final resolution (\textit{i.e.,} $256 \times 256$) for the synthetic sample. The discriminator $D$, based on the approach outlined in \cite{isola_2017}, employs a pixel-level PatchGAN strategy using $1\times1$ kernels and consists of two convolutional layers.

\section{Combination of RGB and Synthetic Thermal Videos}

\subsection{Methodology}
This section presents the structure of the proposed automatic pain assessment framework, the augmentation techniques developed, the pre-processing methods employed, and the pre-training strategy for the modules.

\begin{figure}
\begin{center}
\includegraphics[scale=0.7]{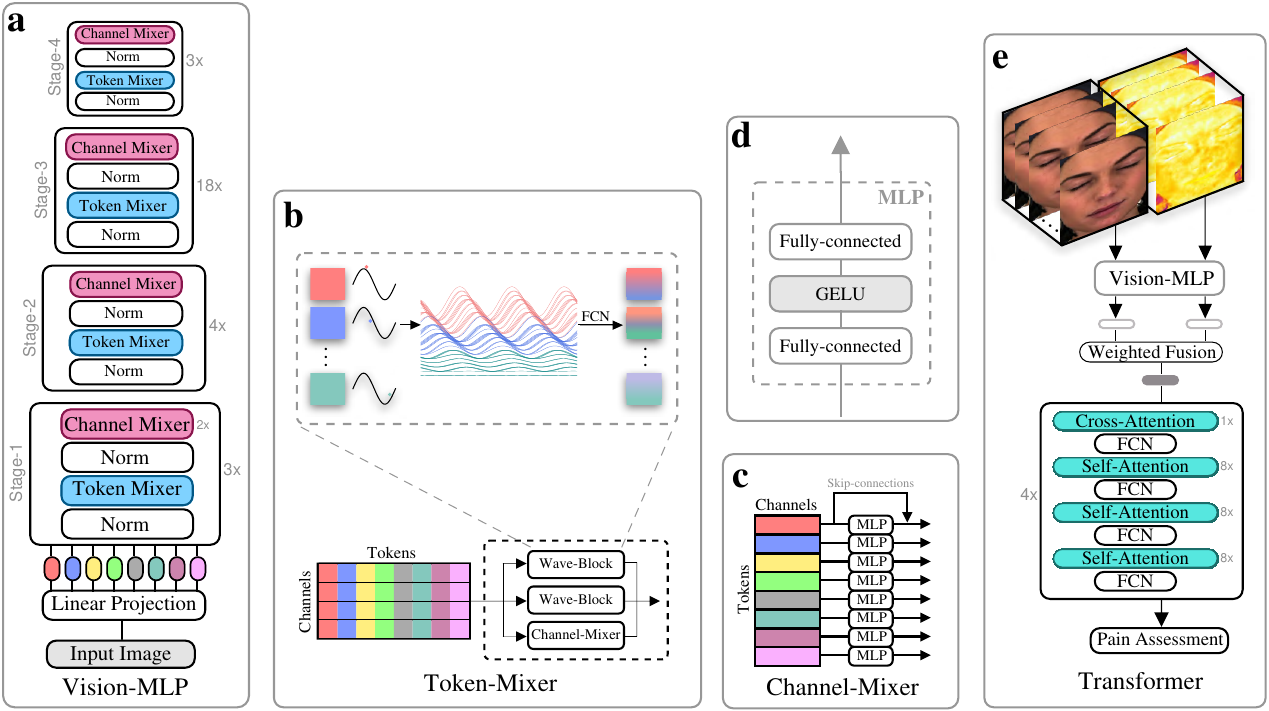}
\end{center}
\caption{Representation of the proposed framework, illustrating its components and their main functions: \textbf{(a)} The \textit{Vision-MLP} module, tasked with extracting feature embeddings from video frames. \textbf{(b)} The \textit{Token-Mixer}, an important sub-module of \textit{Vision-MLP}, generates the wave representation for the tokens. \textbf{(c)} The \textit{Channel-Mixer}, a crucial sub-module within \textit{Vision-MLP}. \textbf{(d)} The MLP, a core component of the \textit{Channel-Mixer}. \textbf{(e)} The fusion procedure that combines RGB and synthetic thermal embeddings, succeeded by the \textit{Transformer} module, which conducts the final pain assessment.}
\label{framework}
\end{figure}

\subsubsection{Framework Architecture}
The proposed framework consists of two primary modules: a Vision-MLP model that acts as a spatial embedding extractor for individual video frames and a transformer-based model that functions as a temporal module, using the embedded representations of the videos for temporal analysis and final pain assessment. Fig. \ref{framework} displays the modules and their main components.

\paragraph{Vision-MLP:}
MLP-like models have recently emerged as a novel class of vision models, providing an alternative to traditional Convolutional Neural Networks (CNNs) and Vision Transformers (ViT). These models are characterized by their straightforward architectures, which consist of fully connected layers combined with activation functions. They possess a lower level of inductive bias and rely on basic matrix multiplication operations. Our methodology is grounded in the principles outlined in \cite{mlp_mixer} that introduced the Vision-MLP, and \cite{wave_vision_mlp} that incorporates a wave representation for the patches (also known as tokens).
Each video frame is initially partitioned into \(n\) non-overlapping tokens 
\(\mathcal{F}_{m}  = [f_{m,1}, f_{m,2}, \ldots, f_{m,n}]\in\mathbb{R}^{n\times p\times p\times 3 }\), where \(p\) specifies the resolution of each token, \textit{i.e.,} \(16\times16\) pixels, and \(3\) represents the number of color channels.
Each token is subsequently linearly projected into a dimension \(d=768\) prior to entering the \textit{Vision-MLP} (refer to \hyperref[framework]{Fig. 2a}).
The first principal sub-module, the \textit{Channel-Mixer} (\hyperref[framework]{Fig. 2c}), operates independently on each token \(f_{j}\), enabling interactions among different channels, and is formulated as:
\begin{equation}
\text{\textit{Channel-Mixer}}(f_j, W^c) = W^c f_j
\label{channel_mixer}
\end{equation}
where \(W^c\) denotes the weight matrix with learnable parameters, and \(j = 1, 2, \ldots, n\).
Following this, the next significant sub-module, the \textit{Token-Mixer} (\hyperref[framework]{Fig. 2b}), facilitates communication among various tokens, aiding in the extraction of features from different spatial locations. Typically, in MLP-based models, the token mixers are defined as:
\begin{equation}
\text{\textit{Token-Mixer}}(\mathcal{F}, W^t)_j = \sum_{k} W^t_{jk} \odot f_{k},
\label{token_mixer}
\end{equation}
where \(W^t\) is the corresponding weight matrix for the tokens, and \(\odot\) represents element-wise multiplication.
Our proposed approach modifies tokens into wave-like representations to dynamically adjust the interactions between tokens and weights based on their semantic content. To depict a token \(f_j\) as a wave \(\tilde{f}_j\) via a wave function, both amplitude and phase information is necessary:
\begin{equation}
\tilde{f}_j = |f_j| \odot e^{i \theta_j}.
\label{wave_function}
\end{equation}
Here, \(i\) is the imaginary unit satisfying \(i^2 = -1\). The term \(|f_j|\) indicates the amplitude of the signal, while \(e^{i\theta_j}\) is a periodic function, with \(\theta_j\) representing the phase of the signal.
The absolute value operation is typically omitted and substituted with \ref{channel_mixer} for simplicity.
Each token's phase \(\theta_j\) reflects its position within the wave cycle and can, therefore, be characterized using fixed parameters, which are adjustable during the training phase. As such, \ref{channel_mixer} is also applied for phase estimation.
Given that \ref{wave_function} characterizes a wave within the complex domain, the Euler formula is employed to embed tokens within the neural network architecture:
\begin{equation}
\tilde{f}_j = |f_j| \odot \cos \theta_j + i |f_j| \odot \sin \theta_j.
\label{euler}
\end{equation}
Combining \ref{token_mixer} and \ref{euler}, a token is represented as:
\begin{equation}
f_j = \sum W^t_{jk} f_k \odot \cos \theta_k + W^i_{jk} f_k \odot \sin \theta_k
\label{eq:first_equation}
\end{equation}
\begin{equation}
\Longrightarrow \sum W^t_{jk} f_k \odot \cos(W^c f_k) + W^i_{jk} f_k \odot \sin(W^c f_k)
\label{eq:second_equation}
\end{equation}
where \(W^t\), \(W^c\), and \(W^i\) are the learnable weight matrices. The described process focuses on wave-like representations and is conducted within the \textit{Token-Mixer}, particularly in the \textit{Wave-Block}. The \textit{Token-Mixer} architecture consists of three blocks: two \textit{Wave-Blocks} and one \textit{Channel-Mixer} operating in parallel.
The \textit{Vision-MLP} module is organized into four stages, each featuring a sequence of a \textit{Token-Mixer} and a \textit{Channel-Mixer} block, preceded by a normalization layer. The depth of parallel blocks in each stage is \(3\), \(4\), \(18\), and \(3\), respectively, allowing for hierarchical embeddings extraction with corresponding dimensions across stages \(64\), \(128\), \(320\), and \(100\).

\paragraph{Fusion:}
For each input frame, the \textit{Vision-MLP} extracts an embedding of dimensionality \(d=100\). The embeddings from the individual frames of a specific video are then concatenated to form a comprehensive embedding representation of the video:
\begin{equation}
\mathcal{V}_D = [d_1 \| d_2 \| \cdots \|d_m], \quad \mathcal{V}_D \in \mathbb{R}^N,
\end{equation}
where \(m\) represents the number of frames in the video, and \(N\) is the dimensionality of the ultimate embedding.
Following this, the embeddings from the RGB and synthetic thermal videos are combined through a weighted fusion process:
\begin{equation}
\label{eq:fusion}
\mathcal{V}_{Fused} = w_1\cdot\mathcal{V}_{RGB} + w_2\cdot\mathcal{V}_{Thermal},   \quad \mathcal{V}_{Fused}\in\mathbb{R}^N.
\end{equation}
This fusion strategy integrates the respective embeddings using learned weights \(w_1\) and \(w_2\), which adjust the influence of the RGB and thermal embeddings, respectively. This weighted summation achieves an effective integration, capturing the relevance of each modality in the resultant fused representation \(\mathcal{V}_{Fused}\).

\paragraph{Transformer:}
The fused embeddings are then input into a transformer-based module that includes self-attention and cross-attention blocks (\hyperref[framework]{Fig. 2e}). The self-attention mechanism is expressed as:
\begin{equation}
Attention(Q,K,V) = \text{softmax}\left( \frac{QK^T}{\sqrt{d_k}} \right)V.
\end{equation}
In this formulation, \(Q\in\mathbb{R}^{M \times C}\), \(K\in\mathbb{R}^{M \times C}\), and \(V\in\mathbb{R}^{M \times C}\) are the Query, Key, and Value matrices respectively, where \(M\) indicates the input dimension, and \(C\) represents the channel dimension.
The cross-attention mechanism also utilizes a dot product operation; however, \(Q\) for cross-attention is dimensioned \(N \times C\) rather than \(M \times C\), with \(N<M\) providing a reduction in computational cost.
Each self-attention and cross-attention block features \(1\) and \(8\) attention heads, respectively, and the entire \textit{Transformer} module consists of \(4\) parallel blocks.
The output embeddings, with a dimensionality of \(340\), are used to perform the final pain assessment via a fully connected neural network.

\paragraph{Augmentation Methods:}
Two augmentation techniques have been implemented within the framework. Firstly, the method known as \textit{Basic} is utilized, which combines polarity inversion with the addition of noise. This approach modifies the original input embedding by inverting the polarity of data elements and adding random noise from a Gaussian distribution, thus introducing variability and perturbations. Secondly, the \textit{Masking} technique involves applying zero-valued masks to the embeddings, effectively eliminating sections of the vectors. The dimensions of these masks are randomly determined, ranging from 10\% to 50\% of the embedding's total dimensions, and are randomly positioned within the embeddings.

\paragraph{Pre-processing:}
The pre-processing involves face detection to isolate the facial region. The MTCNN face detector \cite{zhang_2016} is used, which employs multitask cascaded convolutional neural networks to identify faces and landmarks. It is essential to highlight that the face detector was applied exclusively to the individual RGB frames, and the coordinates of the detected face were then applied to the corresponding synthetic thermal frames. The resolution of all frames was standardized at $224\times 224$ pixels.

\paragraph{Pre-training:}
For the I2I approach, the \textit{SpeakingFaces} dataset \cite{speakingfaces} was employed to train the proposed GAN model for translating RGB to synthetic thermal videos. 
Additionally, before commencing the automatic pain assessment training, the \textit{Vision-MLP} and \textit{Transformer} modules underwent pre-training. The \textit{Vision-MLP} was subject to a three-stage pre-training strategy: initially, it was trained on \textit{DigiFace-1M} \cite{digiface1m} to acquire basic facial features. It was then trained on \textit{AffectNet} \cite{mollahosseini_hasani_2019} and \textit{RAF Face Database basic} \cite{li_deng_2017} to learn features related to basic emotions through multi-task learning. Lastly, the \textit{Compound Facial Expressions of Emotions Database} \cite{du_tao_2014} and the \textit{RAF Face Database compound} \cite{li_deng_2017} were used to learn features of compound emotions in a similar multi-task framework.
The multi-task learning process is represented as:
\begin{equation}
L_{total}= [e^{w1}L_{S_1}+w_{1}]+ [e^{w2}L_{S_2}+w_{2}],
\end{equation}
where $L_S$ is the loss for the specific task associated with different datasets, and $w$ are the learned weights that guide the learning process in minimizing the collective loss $L_{total}$, integrating all the individual losses.
The \textit{Transformer} was pre-trained solely on the \textit{DigiFace-1M} \cite{digiface1m}, adapting the input images into 1D vectors to suit its architectural needs. 
Table \ref{table:chapter_6_datasets} outlines the datasets utilized in the pre-training phase.

\begin{table}
\center
\small
\caption{Datasets utilized for the pretraining process of the framework.}
\label{table:chapter_6_datasets}
\begin{center}
\begin{threeparttable}
\begin{tabular}{ p{4.0cm} p{1.5cm} p{1.5cm} p{1.5cm} }
\toprule
Dataset &\#  samples &\# classes &Task\\
\midrule
\midrule
\textit{SpeakingFaces} \cite{speakingfaces}   &4.58M  &142  &Face$^\varocircle$\\
\textit{DigiFace-1M} \cite{digiface1m}   &1.00M  &10,000  &Face$^\varoast$\\
\textit{AffectNet} \cite{mollahosseini_hasani_2019} &0.40M &8 &Emotion$^\varoast$\\
\textit{Compound FEE-DB}  \cite{du_tao_2014}&6,000 &26 &Emotion$^\varoast$\\
\textit{RAF-DB basic} \cite{li_deng_2017}&15,000 &7 &Emotion$^\varoast$\\
 \textit{RAF-DB compound} \cite{li_deng_2017}&4,000 &11 &Emotion$^\varoast$\\
\bottomrule 
\end{tabular}
\begin{tablenotes}
\scriptsize
\item $\varocircle$: includes face image pairs for the I2I task $\varoast$: includes images for face or emotion recognition tasks
\end{tablenotes}
\end{threeparttable}
\end{center}
\end{table}

\subsection{Experiments}
The proposed framework was assessed using the \textit{BioVid Heat Pain Database} \cite{biovid_2013}, which comprises facial videos, electrocardiograms, electromyograms, and skin conductance levels from $87$ healthy subjects. Participants underwent heat-induced pain via a thermode on their right arm at five different intensities: no pain (NP), mild pain (P\textsubscript{1}), moderate pain (P\textsubscript{2}), severe pain (P\textsubscript{3}), and very severe pain (P\textsubscript{4}). Each level was applied $20$ times to each subject, resulting in $100$ samples per modality and a total of $1740$ samples per class.
Experiments were structured around binary and multi-level classification schemes to assess pain, analyzing each modality individually and collectively. Binary classification aimed to distinguish between no pain (NP) and very severe pain (P\textsubscript{4}), while multi-level classification (MC) was tasked with categorizing all levels of pain intensity present in the dataset.
The leave-one-subject-out (LOSO) method was utilized for validation, and accuracy served as the metric for performance evaluation.
Table \ref{table:details} outlines the training details for the automatic pain assessment, including parameter number and the computational cost measured in floating-point operations (FLOPS) for each module.

\begin{table}
\center
\small
\caption{Training specifications, and number of parameters and FLOPS for each module.}
\label{table:details}
\begin{tabular}{ p{3.0cm}  p{2.5cm}  p{2.5cm} }
\toprule
Training Details & Vision-MLP & Transformer \\
\midrule
\midrule
Optimizer: \textit{AdamW}          & Params: 7.35 M       & Params: 7.96 M\\
Learning rate: \textit{2e-5}       & FLOPS: 30.95 G       & FLOPS: 30.90 G\\
LR decay: \textit{cosine}          &                     &\\
Weight decay: 0.1         &                     &\\
Warmup epochs: 5          &                     &\\
Batch size: 32            &                     &\\ \hline
Total            & \multicolumn{2}{l|}{Params: 15.31 Millions \space FLOPS: 61.85 Giga} \\
\bottomrule 
\end{tabular}
\end{table}

\subsection{Results}
\subsubsection{RGB Videos}
\label{rgb_videos}
Within the context of the RGB video modality, we recorded an accuracy of \(69.37\%\) for the binary classification task (NP vs. P\textsubscript{4}) and \(30.23\%\) for the multi-class classification (MC). The use of the \textit{Masking} augmentation method, which obscured \(20-50\%\) of the input embeddings, yielded a slight increase in binary classification accuracy by \(0.89\%\). However, it led to a reduction in multi-class classification accuracy. 
By extending the training period to \(300\) epochs, modifying the \textit{Masking} method to cover \(30-50\%\) of the embeddings, and applying a \(90\%\) probability to both augmentation methods, the accuracies improved to \(70.05\%\) and \(30.02\%\) for the binary and multi-class tasks, respectively. This represents an average accuracy gain of just under \(0.5\%\). The classification outcomes are detailed in Table \ref{table:rgb}.

\begin{table}
\center
\small
\caption{Results utilizing the RGB video.}
\label{table:rgb}
\begin{center}
\begin{threeparttable}
\begin{tabular}{ P{1.5cm} P{1.5cm} P{1.5cm}  P{1.5cm}  P{1.5cm}  P{1.5cm}}
\toprule
\multirow{2}[2]{*}{\shortstack{Epochs}}
&\multicolumn{3}{c}{Augmentations} 
&\multicolumn{2}{c}{Task}\\ 
\cmidrule(lr){2-4}\cmidrule(lr){5-6}
&\textit{Basic} &\textit{Masking} &P(Aug) &NP vs P\textsubscript{4} &MC\\
\midrule
\midrule
200  &\checkmark &10-20 &0.7 &69.37 &\textbf{30.23} \\
200  &\checkmark &20-50 &0.7 &\textbf{70.26} &28.50 \\
300  &\checkmark &30-50 &0.9 &70.05 &30.02 \\
\bottomrule 
\end{tabular}
\begin{tablenotes}
\scriptsize
\item Masking: indicates the percentage of the input embedding to which zero-value masking is applied \space\space P(Aug): represents the probability of applying the augmentation methods of Basic \& Masking  \space\space NP: No Pain \space\space P\textsubscript{4}: Very Severe Pain \space\space MC: multiclass pain level
\end{tablenotes}
\end{threeparttable}
\end{center}
\end{table}

\subsubsection{Synthetic Thermal Videos}
\label{synthetic_thermal_videos}
In the synthetic thermal modality experiments conducted under identical conditions, the initial accuracies were \(69.97\%\) for the binary classification and \(30.04\%\) for the multi-class classification. Enhancing the intensity of the masking method yielded modest gains in accuracy of \(0.23\%\) for the binary classification and \(0.46\%\) for the multi-class classification. Subsequent final accuracies were \(70.69\%\) for the binary classification and \(29.60\%\) for the multi-class classification, reflecting an average increment of \(0.28\%\). This discrepancy likely arises from the challenge of detecting nuanced facial changes linked to low-level pain, exacerbated by more aggressive augmentation that potentially reduces performance.
The summarized results are presented in Table \ref{table:thermal}.

\renewcommand{\arraystretch}{1.2}
\begin{table}
\center
\small
\caption{Results utilizing the synthetic thermal video.}
\label{table:thermal}
\begin{center}
\begin{threeparttable}
\begin{tabular}{ P{1.5cm} P{1.5cm} P{1.5cm}  P{1.5cm}  P{1.5cm}  P{1.5cm}}
\toprule
\multirow{2}[2]{*}{\shortstack{Epochs}}
&\multicolumn{3}{c}{Augmentations} 
&\multicolumn{2}{c}{Task}\\ 
\cmidrule(lr){2-4}\cmidrule(lr){5-6}
&\textit{Basic} &\textit{Masking} &P(Aug) &NP vs P\textsubscript{4} &MC\\
\midrule
\midrule
200  &\checkmark &10-20 &0.7 &69.97 &30.04 \\
200  &\checkmark &20-50 &0.7 &70.20 &\textbf{30.50} \\
300  &\checkmark &30-50 &0.9 &\textbf{70.69} &29.60 \\
\bottomrule 
\end{tabular}
\begin{tablenotes}
\scriptsize
\item 
\end{tablenotes}
\end{threeparttable}
\end{center}
\end{table}

\subsubsection{Additional Analysis on RGB \& Synthetic Thermal Videos}
The results from the previous section showed a surprising similarity in performance between the RGB and synthetic thermal modalities. Specifically, the RGB modality achieved maximum accuracies of $70.26\%$ for the NP vs. P4 task and $30.23\%$ for the MC task. Similarly, the synthetic thermal modality reached top accuracies of $70.69\%$ and $30.50\%$ for the same tasks, respectively. On average, the thermal video performances were about $1\%$ higher than those for the RGB modality. This was unexpected, considering the synthetic modality was initially considered inferior to the original. This prompted further investigation into why synthetic modalities might perform comparably to or better than the original RGB modality. A key question involved the relevance and efficacy of the thermal-related data featured in the synthetic videos. The theory proposed that minimizing facial expressions in thermal videos could enhance the clarity of thermal data assessment.

\begin{figure}
\begin{center}
\includegraphics[scale=0.35]{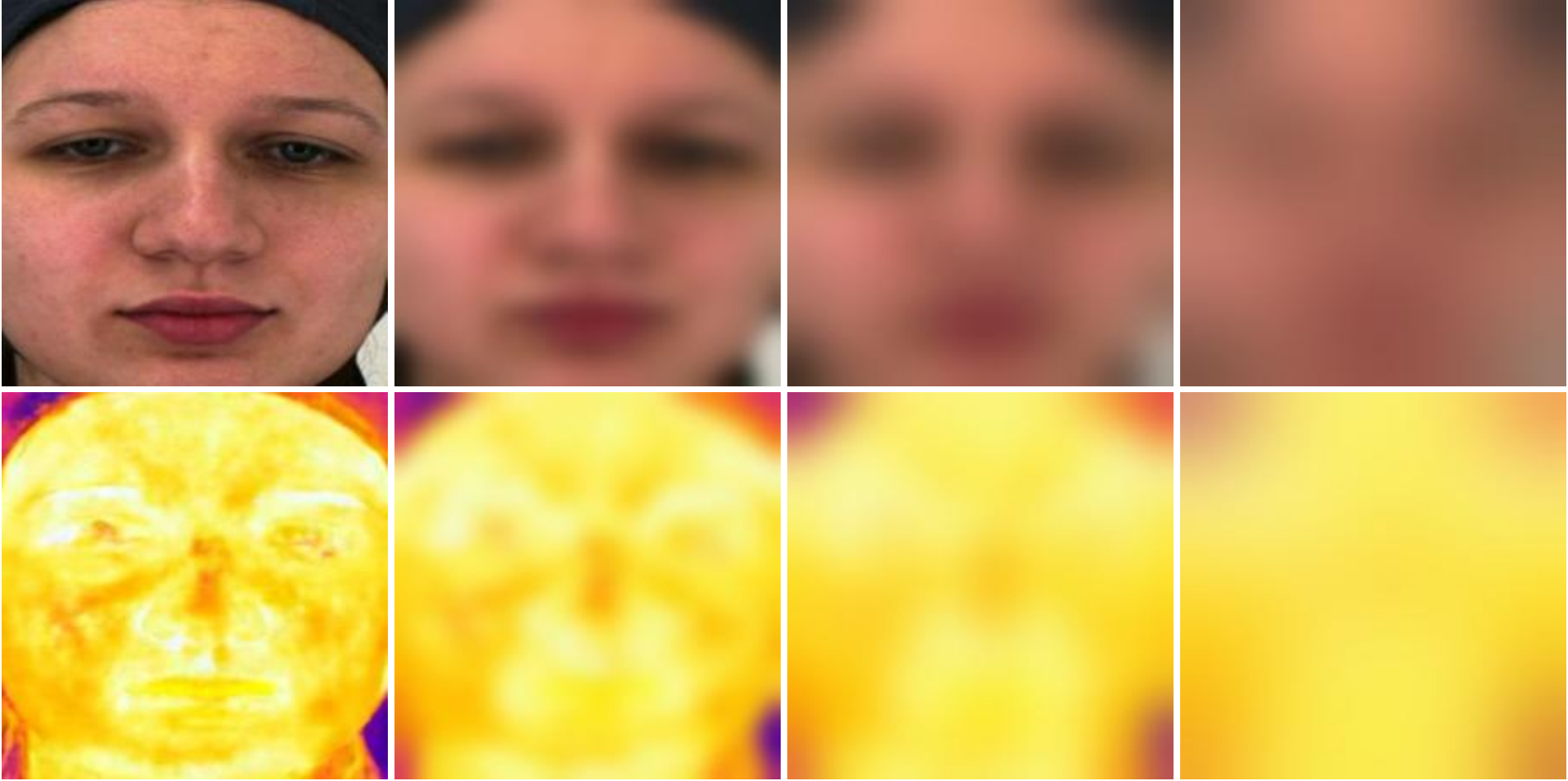} 
\end{center}
\caption{Gradual blurring of RGB and synthetic thermal facial images: a series displaying varying levels of Gaussian blur applied, with kernel sizes gradually increased from $k = 0$ (no blur) to $k = 191$ (extensively blurred).}
\label{faces}
\end{figure}

\begin{figure}
\begin{center}
\includegraphics[scale=0.12]{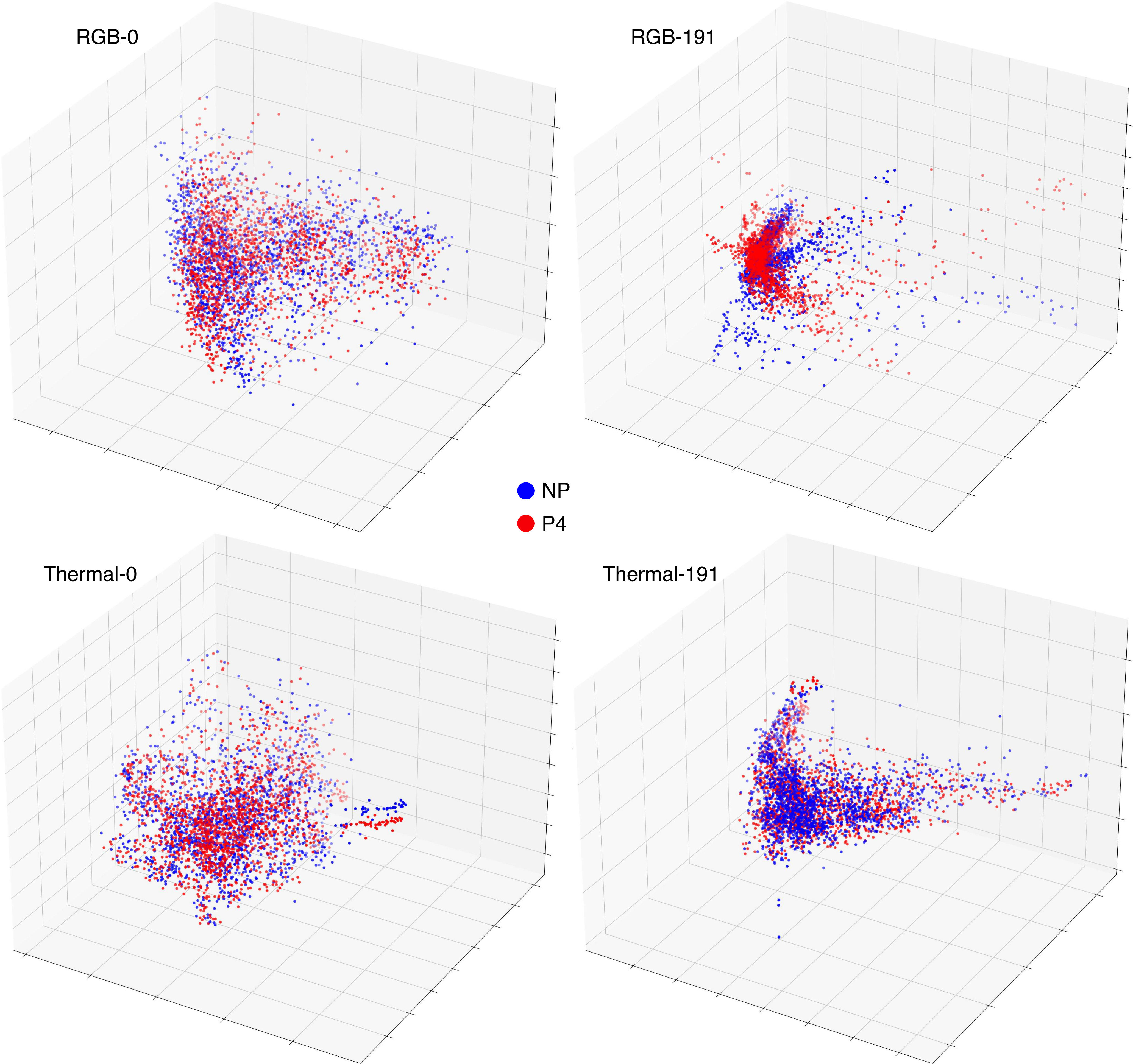} 
\end{center}
\caption{Distributions of 3D embeddings for NP (no pain) and P4 (very severe pain) classes in RGB and synthetic thermal videos, for \(k = 0\) (clear) and \(k = 191\) (heavily blurred).}
\label{emb}
\end{figure}

Gaussian blurring was incrementally applied to RGB and synthetic thermal videos, as shown in Fig. \ref{faces}, with kernel sizes increasing from $0$ to $191$. According to Table \ref{table:blur}, at a kernel size of $k=0$, a performance differential of $0.47\%$, favoring the thermal modality, confirms prior findings. This gap slightly expands to $0.49\%$ at $k=41$. Remarkably, at $k=91$, the disparity enlarges to $2.13\%$ and increases to $5.90\%$ at $k=191$, where blurring is most intense. The results reveal that as facial expressions become less visible through blurring, synthetic thermal videos consistently outperform RGB videos, with respective accuracies of $66.24\%$ versus $60.34\%$. As blurring intensifies from $k=0$ to $k=191$, accuracy rates for synthetic thermal and RGB modalities decrease by $1.81\%$ and $7.13\%$, respectively. This suggests that critical information, such as visually represented facial temperature in the synthetic modality, is unaffected or slightly impacted. Fig. \ref{emb} displays the embedding distribution for the RGB and synthetic thermal modalities at $k=0$ and $k=191$, highlighting a distinct variation in distribution patterns, albeit with ambiguous data point separation.
For $k=191$, while RGB embeddings tend to cluster and potentially overlap, many points conspicuously stray from the central mass without any distinct arrangement. Conversely, the data points for the synthetic modality spread more uniformly, possibly indicating better class differentiation.

\renewcommand{\arraystretch}{1.2}
\begin{table}
\center
\small
\caption{Results utilizing the RGB  \& the synthetic thermal video.}
\label{table:blur}
\begin{center}
\begin{threeparttable}
\begin{tabular}{ P{1.2cm} P{1.5cm} P{1.5cm} P{1.0cm} P{1.0cm}  P{1.5cm}  P{1.5cm} }
\toprule
\multirow{2}[2]{*}{\shortstack{Epochs}}
&\multirow{2}[2]{*}{\shortstack{Modality}}
&\multirow{2}[2]{*}{\shortstack{Blur}}
&\multicolumn{3}{c}{Augmentations} 
&\multicolumn{1}{c}{Task}\\ 
\cmidrule(lr){4-6}\cmidrule(lr){7-7}
& & &\textit{Basic} &\textit{Masking} &P(Aug) &NP vs P\textsubscript{4}\\
\midrule
\midrule
100  &RGB     &0 &\checkmark &10-20 &0.7 &67.47 \\
100  &Thermal &0 &\checkmark &10-20 &0.7 &68.05  \\\hline
100  &RGB     &41 &\checkmark &10-20 &0.7 &66.61 \\
100  &Thermal &41 &\checkmark &10-20 &0.7 &67.10  \\\hline
100  &RGB     &91 &\checkmark &10-20 &0.7 &64.80 \\
100  &Thermal &91 &\checkmark &10-20 &0.7 &66.93  \\\hline
100  &RGB &191 &\checkmark &10-20 &0.7 &60.34 \\
100  &Thermal &191 &\checkmark &10-20 &0.7 &66.24 \\
\bottomrule 
\end{tabular}
\begin{tablenotes}
\scriptsize
\item Blur: Gaussian blurring with kernel sizes $k$
\end{tablenotes}
\end{threeparttable}
\end{center}
\end{table}

\subsubsection{Fusion}
Three fusion strategies were assessed for multimodal analysis involving RGB and synthetic thermal video data. Initially, the fusion strategy referenced in \ref{eq:fusion} utilized learned weights \(w_1\) and \(w_2\) to scale the contributions of each modality. A second strategy incorporated an additional weight, \(w_3\), modifying the formula to \(w_3\cdot(w_1\cdot\mathcal{V}_{RGB} + w_2\cdot\mathcal{V}_{Thermal})\). The third method bypassed learned weights altogether, directly combining the embedding vectors from the modalities. The results, detailed in Table \ref{table:fusion}, indicate that omitting learned weights achieved accuracies of \(64.92\%\) and \(26.40\%\) for the binary and multi-class tasks, respectively. The introduction of \(w_3\) reduced \(0.5\%\) in accuracy for both tasks. The strategy using only weights \(w_1\) and \(w_2\) yielded the best outcomes, with accuracies of \(65.08\%\) and \(26.50\%\) for the binary and multi-class tasks, respectively.
By maintaining the use of weights \(w_1\) and \(w_2\) and increasing the training duration from \(100\) to \(300\) epochs, while consistent with the augmentation settings, accuracies improved to \(69.50\%\) and \(29.80\%\) for the binary and multi-class tasks, respectively.
Further prolonging the training to \(500\) epochs, with no evidence of overfitting, led to further improved performances, with final accuracies of \(71.03\%\) and \(30.70\%\) for the respective tasks.

\renewcommand{\arraystretch}{1.2}
\begin{table}
\center
\small
\caption{Results utilizing the fusion of RGB \& synthetic thermal video.}
\label{table:fusion}
\begin{center}
\begin{threeparttable}
\begin{tabular}{ P{1.5cm} P{1.5cm} P{1.5cm} P{1.5cm}  P{1.5cm}  P{1.5cm}  P{1.5cm}}
\toprule
\multirow{2}[2]{*}{\shortstack{Epochs}}
&\multirow{2}[2]{*}{\shortstack{Fusion\\weights}}
&\multicolumn{3}{c}{Augmentations} 
&\multicolumn{2}{c}{Task}\\ 
\cmidrule(lr){3-5}\cmidrule(lr){6-7}
& &\textit{Basic} &\textit{Masking} &P(Aug) &NP vs P\textsubscript{4} &MC\\
\midrule
\midrule
100  &-- &\checkmark &10-20 &0.7 &64.92 &26.40 \\
100  &W2 &\checkmark &10-20 &0.7 &65.08 &26.50 \\
100  &W3 &\checkmark &10-20 &0.7 &64.42 &25.90 \\
\hline
300  &W2 &\checkmark &10-20 &0.7 &69.50 &29.80 \\
500  &W2 &\checkmark &10-20 &0.7 &\textbf{71.03} &\textbf{30.70} \\
\bottomrule 
\end{tabular}
\begin{tablenotes}
\scriptsize
\item W2: utilization of [w\textsubscript{1},w\textsubscript{2}] \space\space 
W3: utilization of [w\textsubscript{1},w\textsubscript{2},w\textsubscript{3}]
\end{tablenotes}
\end{threeparttable}
\end{center}
\end{table}

\subsubsection{Comparison with Existing Methods}
This section compares the proposed method with other methodologies reported in the literature. We utilize Part A of the \textit{BioVid} dataset, which includes all $87$ subjects, employing the LOSO cross-validation protocol for validation. The results are displayed in Table \ref{table:sota}. Our vision-based approach, which leverages RGB and synthetic thermal modalities, shows performances comparable to or surpass those of prior studies.
Relative to results in the literature from \cite{werner_hamadi_walter_2017,zhi_wan_2019,thiam_kestler_schenker_2020,tavakolian_bordallo_liu_2020}, our method achieved higher accuracy in both binary and multi-level tasks. Notably, the research in \cite{werner_2016} recorded accuracies of $72.40\%$ and $30.80\%$, marking an improvement of $1.37\%$ and $0.10\%$ over our method, respectively. The highest reported results are from \cite{gkikas_tsiknakis_embc}, which utilized a transformer-based architecture.

Additionally, Table \ref{table:mintpain} compares our results with those from \cite{haque_2018}, where the authors introduced the \textit{MIntPAIN} dataset comprising both RGB and thermal videos for automatic pain assessment across five intensity levels. Our analysis revealed that the accuracies of the RGB and thermal modalities were closely matched at $18.55\%$ and $18.33\%$, respectively, which parallels our observations of similar performance between RGB and synthetic thermal modalities. By integrating these modalities, the authors in \cite{haque_2018} reported a significant performance increase of $30.77\%$, surpassing our modest gains. It should be emphasized that in \cite{haque_2018}, the performance levels of the individual modalities were below the random guess prediction threshold of $20\%$. It was only through their combination that performance was elevated above this threshold.

\renewcommand{\arraystretch}{1.2}
\begin{table}
\center
\small
\caption{Comparison of studies that utilized \textit{BioVid}, videos, and LOSO cross-validation.}
\label{table:sota}
\begin{center}
\begin{threeparttable}
\begin{tabular}{ P{3.5cm} P{3.5cm} P{1.5cm}  P{1.5cm}}
\toprule
\multirow{2}[2]{*}{\shortstack{Study}}
&\multirow{2}[2]{*}{\shortstack{Method}}
&\multicolumn{2}{c}{Task}\\ 
\cmidrule(lr){3-4}
& &NP vs P\textsubscript{4} &MC\\
\midrule
\midrule
Werner \textit{et al.}\cite{werner_2016} &Deep RF &72.40&30.80\\
Werner \textit{et al.}\cite{werner_hamadi_walter_2017} &RF &70.20 &--\\
Zhi \textit{et al.} \cite{zhi_wan_2019} &SLSTM &61.70 &29.70\\
Thiam \textit{et al.} \cite{thiam_kestler_schenker_2020} &2D CNN, biLSTM &69.25 &--\\
Tavakolian \textit{et al.} \cite{tavakolian_bordallo_liu_2020} &2D CNN &71.00 &--\\
Gkikas \textit{et al.}\cite{gkikas_tsiknakis_embc} &Vision-Transformer &73.28 &31.52\\
Our &Vision-MLP &71.03 &30.70\\
\bottomrule 
\end{tabular}
\end{threeparttable}
\end{center}
\end{table}

\renewcommand{\arraystretch}{1.1}
\begin{table}
\center
\small
\caption{Comparison with the \textit{MIntPAIN} dataset.}
\label{table:mintpain}
\begin{center}
\begin{threeparttable}
\begin{tabular}{ P{3.0cm} P{2.0cm} P{2.0cm} P{1.5cm}}
\toprule
\multirow{2}[2]{*}{\shortstack{Study}}
&\multirow{2}[2]{*}{\shortstack{Dataset}}
&\multirow{2}[2]{*}{\shortstack{Modality}}
&\multicolumn{1}{c}{Task}\\ 
\cmidrule(lr){4-4}
& & &MC\\
\midrule
\midrule
\multirow{3}{*}{Haque \textit{et al.} \cite{haque_2018}} &\multirow{3}{*}{MIntPAIN} &RGB       &18.55\\
                                                                   & &Thermal$^{\circ}$ &18.33\\
                                                                   & &Fusion  &30.77 \\
\hline
\multirow{3}{*}{Our} &\multirow{3}{*}{BioVid} &RGB       &30.02\\
                             & &Thermal$^{\star}$ &29.69\\
                             & &Fusion  &30.70\\              
\bottomrule 
\end{tabular}
\begin{tablenotes}
\scriptsize
\item $\circ$:real\space $\star$: synthetic
\end{tablenotes}
\end{threeparttable}
\end{center}
\end{table}

\section{Summary}
This chapter investigated the creation of synthetic thermal imagery via GAN models to assess its utility in automatic pain evaluation. Additionally, a novel framework incorporating a \textit{Vision-MLP} and supported by a \textit{Transformer} module as the core of the assessment system was introduced. 
The experiments demonstrated the effectiveness of the synthetic thermal modality, which achieved performances comparable to or better than the original RGB modality. This research also delved into the factors contributing to this effectiveness, particularly the role of temperature color representations in the analysis.
Furthermore, various fusion techniques were used to evaluate the combination of the two vision modalities, highlighting the potential for performance improvements over single-modality approaches. 
It is important to note that further enhancements and experimental work, especially with the multimodal approach, could improve outcomes. The generation and integration of synthetic modalities, such as thermal imagery, within an automatic pain assessment framework exhibit considerable promise, warranting additional exploration and research.

    \chapter{General-Purpose Models}
\minitoc  
\label{chapter_7}

\section{Chapter Overview: Introduction \& Related Work}
This chapter discusses the approaches and findings from \cite{gkikas_tsiknakis_painvit_2024} and \cite{gkikas_rojas_foundation_2024}. In recent years, the literature on deep learning has shown a trend towards adopting general-purpose models. These models are characterized by architectures not tied to specific modalities or pre-training on datasets derived solely from a single domain. We will explore two approaches: first, a modality-agnostic pipeline for automatic pain assessment, and second, the development of a foundation model explicitly applied for automatic pain assessment tasks.

Regarding modality-agnostic approaches, in \ref{chapter_5_paper_2}, we introduced the concept specifically for augmentations. These augmentations were designed not directly for the image or biosignal space but for the latent space. Thus, regardless of the original input modality, whether images, biosignals, or others, the augmentations were applied to their feature representations. This chapter advances our work by developing a modality-agnostic multimodal fusion pipeline evaluated in pain assessment tasks.
The literature on modality-agnostic approaches remains limited. In \cite{zheng_lyu_2024}, the researchers pursued a novel approach by exploring modality-agnostic representations through knowledge distillation for semantic segmentation. Their goal was to reduce the modality gap and diminish semantic ambiguity, enabling the combination of various modalities under any visual conditions.
In \cite{yang_li_2024}, the authors addressed the persistent challenges of temporal asynchrony and modality heterogeneity in multimodal sequence fusion, often leading to performance bottlenecks. To overcome these issues, they developed a strategy integrating modality-exclusive and modality-agnostic representations for multimodal ideo sequence fusion. This approach enabled them to capture reliable context dynamics within individual modalities and enhance distinctive features across modality-exclusive spaces. Additionally, they designed a hierarchical cross-modal attention module to uncover significant element correlations across different modalities within the modality-agnostic space.

The literature on foundation models is considerably more extensive. With the emerging paradigm of building AI systems around foundation models, there has been a shift toward creating more adaptable and scalable systems that generalize across various tasks and domains. A foundation model is defined as any model trained on vast datasets, often through extensive self-supervision, which can then be adapted---such as through fine-tuning---to a wide range of downstream tasks. Despite their reliance on traditional deep learning and transfer learning techniques, the extensive scale of foundation models fosters the development of new capabilities and enhances effectiveness across many tasks \cite{bommasani_husdon_2021}. Numerous examples have surfaced recently in academic literature. For instance, the \textit{SAM} model \cite{kirillov_mintun_2023}, a foundation model for image segmentation, was initially trained from scratch on $11$ million images. In later studies \cite{ma_he_2024,wu_ji_2023}, researchers have adapted \textit{SAM} for medical imaging by optimizing it for smaller, specialized datasets.
Additionally, a notable paradigm shift has occurred with the introduction of generalist models \cite{reed_zolna_2022}, a novel class of foundation models trained simultaneously on various tasks under a unified learning strategy, typically supervised. This approach is particularly advantageous in computer vision, enabling handling differing embedding representations across tasks and various visual modalities \cite{awais_naseer_2023}.

To the best of our knowledge, there are no modality-agnostic or foundation models in the field of automatic pain assessment. Currently, the majority of the approaches utilize pre-trained models; however, these models typically adhere to the traditional methodology of pre-training on a general, large-scale dataset and then fine-tuning for the specific task of pain assessment. Studies such as those detailed in \cite{haque_2018,rodriguez_cucurull_2022} rely on transfer learning techniques derived from facial recognition datasets. The majority of these studies are reviewed in Chapter \ref{slr}.

\section{A Modality-Agnostic Pipeline for Automatic Pain Assessment}
\label{chapter_7_paper_1}
We present a modality-agnostic multimodal framework that leverages both video and fNIRS data. The pipeline operates on a dual Vision Transformer (ViT) format, which obviates the necessity for modality-specific architectures or expansive feature engineering. It interprets the inputs as unified images through 2D waveform representations.

\subsection{Methodology}
This section outlines the pipeline for the proposed multi-modal automatic pain assessment framework, describing the model architectures, pre-processing methods, pre-training strategies, and augmentation techniques.

\subsubsection{Framework Architecture}
The proposed framework, \textit{Twins-PainViT}, includes two modules: \textit{PainViT--1} and \textit{PainViT--2}. Both models share uniform architectures and parameters and follow to identical pre-training protocol.
\textit{PainViT--1} is tasked with processing the individual input video frames and the visualized fNIRS waveforms, serving as an embedding extractor. Meanwhile, \textit{PainViT--2} manages the visual representation of these embeddings and performs the final pain classification task.

\paragraph{PainViT:} 
Vision Transformers (ViTs) \cite{vit} have established themselves as a leading framework in computer vision tasks, recognized for their impressive performance. However, despite their effectiveness, transformer-based models encounter scalability challenges with larger input sizes, significantly increasing computational demands. This inefficiency mainly stems from the element-wise operations in the multi-head self-attention mechanism.
Efforts to improve the efficiency and simplify the architecture of transformer-based models have included modifications to the self-attention module and overall structural adjustments \cite{mobile_former}\cite{pale_transformer}. Our methodology builds on the principles of \cite{swin_transformer}, which incorporates hierarchical architectures into vision transformers, and \cite{efficientvit}, which introduces mechanisms to enhance efficiency and processing speed.

\paragraph{PainViT--block:}
A block consists of two key elements: the \textit{Token-Mixer} and the \textit{Cascaded-Attention}. The architecture places the \textit{Cascaded-Attention} module centrally, combined with a \textit{Token-Mixer} module before and after it. For every input image $I$, overlapping patch embedding is utilized, generating $16\times 16$ patches. Each patch is then projected into a token with a dimensionality of $d$.

\subparagraph{Token-Mixer:}
To better integrate local structural information, the token $T$ is processed through a depthwise convolution layer:
\begin{equation}
Y_{c} = K_c * T_{c} + b_c.
\end{equation}
Here, $Y_{c}$ denotes the output of the depthwise convolution for channel $c$ of the token $T_c$. $K_c$ represents the convolutional kernel for channel $c$, $T_{c}$ indicates the $c$-th channel of the token $T$, and $b_c$ is the bias term added to the convolution output for channel $c$. The symbol $*$ indicates the convolution operation. After the depthwise convolution, batch normalization is then applied to the output:
\begin{equation}
Z_{c} = \gamma_c \left( \frac{Y_{c} - \mu_B}{\sqrt{\sigma_B^2 + \epsilon}} \right) + \beta_c.
\end{equation}
Here, $Z_{c}$ represents the batch-normalized output for channel $c$ of the token $T$. The learnable parameters $\gamma_c$ and $\beta_c$ are specific to channel $c$, adjusting the scale and shift of the normalized data. $\mu_B$ denotes the batch mean of $Y_{c}$, $\sigma_B^2$ indicates the batch variance of $Y_{c}$, and $\epsilon$ is a small constant included for numerical stability to prevent division by zero. Subsequently, a feed-forward network (FFN) is used to enhance communication between different feature channels:
\begin{equation}
\Phi^F(Z_{c}) = W_2 \cdot \text{ReLU}(W_1 \cdot Z_{c} + b_1) + b_2.
\end{equation}
Here, $\Phi^F(Z_{c})$ represents the output of the feed-forward network for the input $Z_{c}$. The weight matrices for the first and second linear layers are denoted by $W_1$ and $W_2$, respectively; $b_1$ and $b_2$ are the corresponding bias terms for these layers. The activation function employed here is $\text{ReLU}$.

\subparagraph{Cascaded-Attention}
Respecting the attention mechanism, it includes a single self-attention layer. For each input embedding:
\begin{equation}
X_{i+1}=\Phi^A(X_i).
\end{equation}
The $X_i$ refers to the entire input embedding for the $i$-th \textit{PainViT-block}. The \textit{Cascaded-Attention} module uses a cascaded mechanism that splits the full input embedding into smaller segments, with each segment directed to a specific attention head. This method distributes the computational load across the heads, improving efficiency by managing long input embeddings more effectively.
The attention mechanism operates as follows:
\begin{equation}
\widetilde{X}_{ij} = Attn({X}_{ij} W^Q_{ij}, {X}_{ij} W^K_{ij}, {X}_{ij} W^V_{ij}),
\end{equation}
\begin{equation}
\widetilde{X}_{i+1} = Concat[\widetilde{X}_{ij}]_{j=1:h}W^P_i,
\end{equation}
where each $j$-th head is responsible for computing the self-attention of $X_{i,j}$, the $j$-th segment of the full input embedding $X_i$. This embedding is organized as $[X_{i1}, X_{i2}, \dots, X_{ih}]$, with $j$ ranging from 1 to $h$, where $h$ is the total number of heads. The projection layers $W^Q_{ij}$, $W^K_{ij}$, and $W^V_{ij}$ transform each segment of the input embedding into distinct subspaces for queries, keys, and values, respectively. The process concludes with $W^P_i$, a linear layer that combines the outputs of all heads, restoring them to the original input dimensionality.
Additionally, the cascaded design aids in developing more complex representations for the $Q$, $K$, and $V$ layers. This enhancement occurs as the output from each head is fed into the subsequent head, allowing for a progressive accumulation of information throughout the layers. Specifically:
\begin{equation}
X^{'}_{ij} = X_{ij} + \widetilde{X}_{i(j-1)}.
\end{equation}
Here, $X'_{ij}$ denotes the sum of the $j$-th input segment $X_{ij}$ and the output $\tilde{X}_{i(j-1)}$ from the $(j-1)$-th head. This summation is the new input embedding for the $j$-th head in the self-attention computation. Additionally, depthwise convolution is applied to each $Q$ in every attention head, enhancing the self-attention process's ability to capture both global representations and local details.

\begin{figure}
\begin{center}
\includegraphics[scale=0.47]{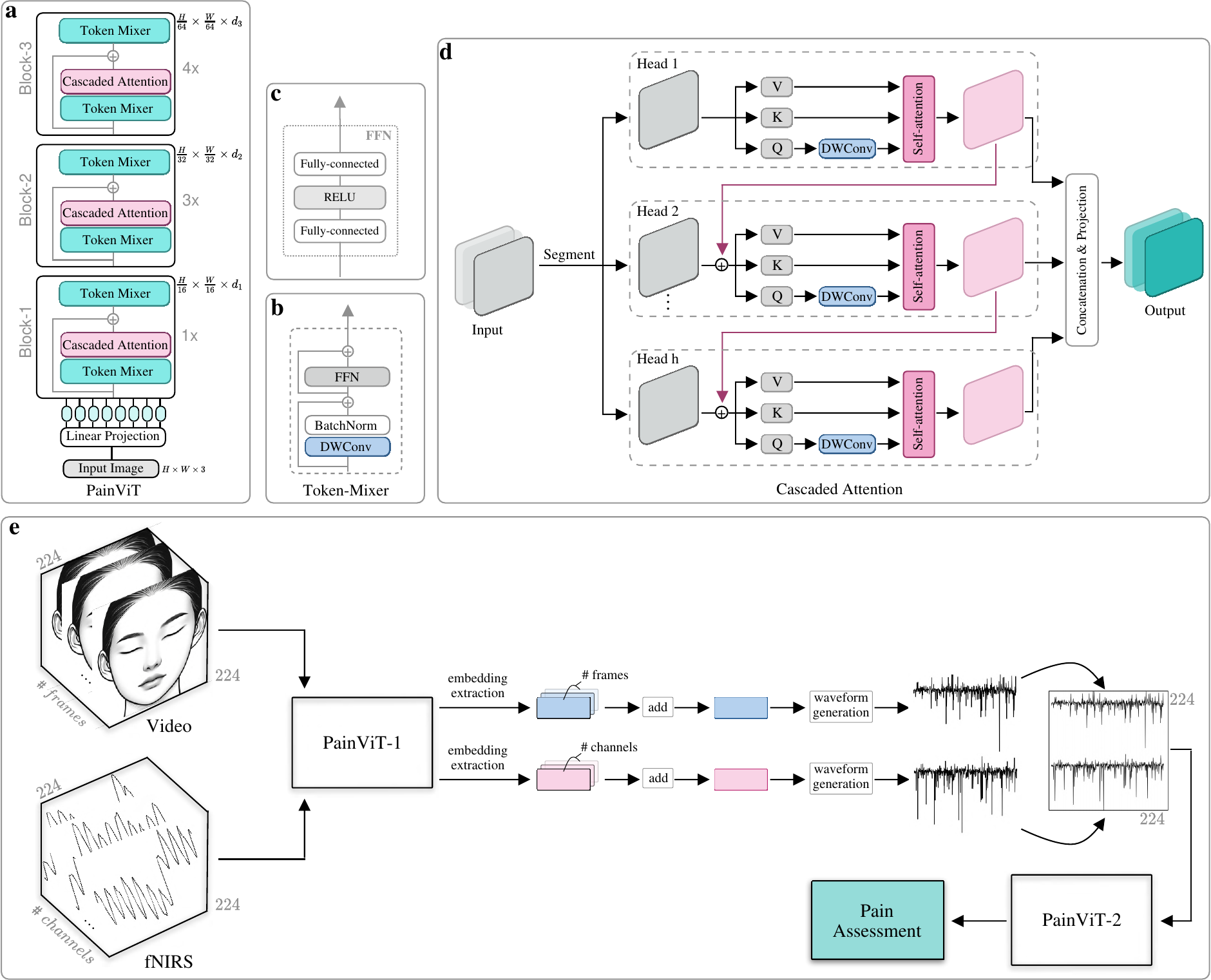} 
\end{center}
\caption{\textit{PainViT}: \textbf{(a)} Hierarchical arrangement of the \textit{PainViT} blocks, each layer having varying depths, showcasing how token resolution decreases at each stage; \textbf{(b)} Composition of the Token-Mixer module, featuring elements like depthwise convolution (DWConv) and batch normalization; \textbf{(c)} Architecture of the Feed-Forward Network (FFN) within the \textit{Token-Mixer}; \textbf{(d)} The \textit{Cascaded Attention} mechanism implemented across multiple heads, illustrating how outputs from preceding heads are incorporated to refine the self-attention process, culminating in the final output projection; \textbf{(e)} Configuration of the proposed multimodal pipeline, employing videos and fNIRS. The embeddings from \textit{PainViT--1} are represented as waveform diagrams, which are merged into a single diagram that illustrates both modalities before entering \textit{PainViT--2} for final pain evaluation.}
\label{1_full}
\end{figure}

The architecture consists of three \textit{PainViT--blocks} with depths of $1$, $3$, and $4$. This hierarchical design progressively reduces the number of tokens by subsampling the resolution by $2\times$ at each stage, enabling the extraction of embeddings with dimensions $d$ across the blocks, particularly $192$, $288$, and $500$. Each block also features a multihead self-attention mechanism, employing $3$, $3$, and $4$ heads, respectively.
\hyperref[1_full]{Fig. 1(a-d)} depicts the \textit{PainViT} architecture and its core components, while Table \ref{table:ch7_p1_module_parameters} details the number of parameters and the computational costs measured in floating-point operations (FLOPS).

\renewcommand{\arraystretch}{1.2}
\begin{table}
\small
\center
\caption{Number of parameters and FLOPS for the components of the proposed \textit{Twins-PainViT}.}
\label{table:ch7_p1_module_parameters}
\begin{center}
\begin{threeparttable}
\begin{tabular}{ P{3.3cm}  P{2.0cm}  P{2.0cm}}
\toprule
Module & Params (M) &FLOPS  (G) \\
\midrule
\midrule
PainViT--1   &16.46 &0.59\\
PainViT--2   &16.46 &0.59 \\
\hline
Total &32.92 &1.18\\
\bottomrule
\end{tabular}
\begin{tablenotes}
\scriptsize
\item
\end{tablenotes}
\end{threeparttable}
\end{center}
\end{table}

\subsubsection{Embedding extraction \& Fusion}
For every frame in a video, $V = [v_1, v_2, \ldots, v_n]$, \textit{PainViT--1} generates a corresponding embedding. These embeddings are combined to produce a composite feature representation of the video. Similarly, for each fNIRS signal channel, $C = [c_1, c_2, \ldots, c_m]$, \textit{PainViT--1} extracts embeddings which are then compiled to form a complete representation of the fNIRS signal. The following equations outline this procedure:
\begin{equation}
E_V \leftarrow \sum_{i=1}^n \text{\textit{PainViT--1}}(v_i),
\end{equation}

\begin{equation}
E_C \leftarrow \sum_{i=1}^m \text{\textit{PainViT--1}}(c_i),
\end{equation}
where $E_V$ and $E_C$ represent the embedding representations for the video and fNIRS, respectively. After these embeddings are extracted, $E_V$ and $E_C$ are visualized as waveform diagrams. The waveforms from both modalities---video and fNIRS---are then combined into a single image with a resolution of $224\times 224$. This integrated visual representation is input into \textit{PainViT--2} for final pain assessment. (\hyperref[1_full]{Fig. 1e}) provides a high-level overview of the multimodal proposed pipeline.

\subsubsection{Pre-processing}
\label{1_preprocessing}
\begin{figure}
\begin{center}
\includegraphics[scale=0.35]{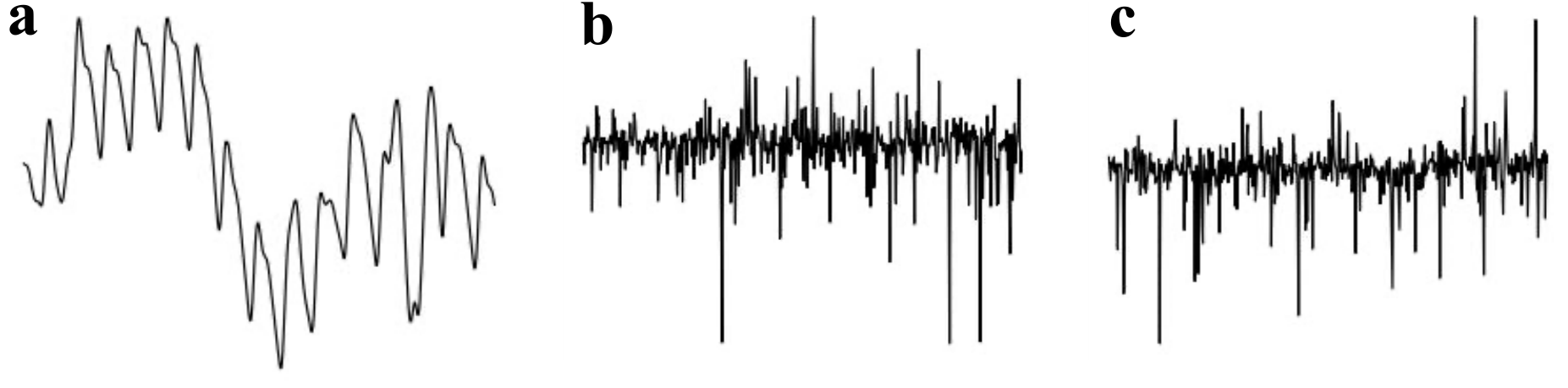}
\end{center}
\caption{Waveform illustrations for various data types: \textbf{(a)} original fNIRS signal, \textbf{(b)} video embedding derived from \textit{PainViT--1}, and \textbf{(c)} fNIRS embedding obtained from \textit{PainViT--1}.}
\label{1_waveforms}
\end{figure}

The preprocessing steps include face detection in video frames and generating waveform diagrams from the original fNIRS data. The MTCNN face detector \cite{zhang_2016}, which uses a series of cascaded convolutional neural networks, was employed to identify faces and facial landmarks with the faces set at a resolution of $224\times 224$ pixels.
All fNIRS channels are used to create waveform diagrams, which visually represent the signal's wave shape and form over time, displaying amplitude, frequency, and phase. This method offers a straightforward approach to visualizing a signal without requiring transformations or complex computations typical of spectrograms, scalograms, or recurrence plots.
Similarly, embeddings extracted by \textit{PainViT--1} are visualized using waveform diagrams. Although these embeddings are not signals per se, the 1D vectors can still be plotted in a 2D space for further analysis or use by deep-learning vision models. 
Waveform diagrams created from fNIRS data and embeddings are formatted as images with a $224\times 224$ pixels resolution. Fig. \ref{1_waveforms} shows waveform representations of channel-specific fNIRS signals, an embedding extracted from a video, and an embedding derived from a channel-specific fNIRS sample.

\subsubsection{Pre-training}
\label{1_pretraining}
Before the automatic pain assessment training, the \textit{Twins-PainViT} models underwent pre-training using a multi-task learning approach. This pre-training incorporated four datasets designed for emotion recognition tasks. The \textit{AffectNet} \cite{mollahosseini_hasani_2019} and \textit{RAF-DB basic} \cite{li_deng_2017} datasets supplied facial images to train on basic emotions, whereas the \textit{Compound FEE-DB} \cite{du_tao_2014} and \textit{RAF-DB compound} \cite{li_deng_2017} datasets focused on complex emotional states.
In addition, five datasets containing various biosignals were utilized. The \textit{EEG-BST-SZ} \cite{ford_2013} dataset includes electroencephalograms for schizophrenia analysis, and the \textit{Silent-EMG} \cite{gaddy_klein_2020} contains electromyograms that help pinpoint the origin of EMG activities, such as from the throat or mid-jaw areas. The \textit{BioVid} \cite{biovid_2013} dataset provided electrocardiogram, electromyogram, and galvanic skin response samples for pain assessment.
All biosignals were represented as waveforms, as detailed in \ref{1_preprocessing}.
The equation defines the multi-task learning framework:
\begin{equation}
L_{total} = \sum_{i=1}^9 \left[ e^{w_i} L_{S_i} + w_i \right],
\end{equation}
where $L_{S_i}$ represents the loss for each specific task from the various datasets, and $w_i$ are the learned weights that drive the learning process to minimize the combined loss $L_{total}$, considering all individual losses.
Table \ref{table:1_datasets} outlines the datasets involved in the pre-training phase.

\renewcommand{\arraystretch}{1.2}
\begin{table}
\small
\center
\caption{Datasets utilized for the pretraining process of the framework.}
\label{table:1_datasets}
\begin{center}
\begin{threeparttable}
\begin{tabular}{ p{3.8cm} p{1.5cm} p{1.5cm} p{2.0cm} }
\toprule
Dataset &\#  samples &\# classes &Modality\\
\midrule
\midrule
\textit{AffectNet} \cite{mollahosseini_hasani_2019} &0.40M &8 &Facial Images\\
\textit{RAF-DB basic} \cite{li_deng_2017}&15,000 &7 &Facial Images\\
\textit{RAF-DB compound} \cite{li_deng_2017}&4,000 &11 &Facial Images\\
\textit{Compound FEE-DB}  \cite{du_tao_2014}&6,000 &26 &Facial Images\\
\textit{EEG-BST-SZ} \cite{ford_2013}&1.5M &2 &EEG\\
\textit{Silent-EMG} \cite{gaddy_klein_2020}&190,816 &8 &EMG\\
\textit{BioVid} \cite{biovid_2013}&8,700 &5 &ECG\\
\textit{BioVid} \cite{biovid_2013}&8,700 &5 &EMG\\
\textit{BioVid} \cite{biovid_2013}&8,700 &5 &GSR\\
\bottomrule 
\end{tabular}
\begin{tablenotes}
\scriptsize
\item EEG: electroencephalogram\space EMG: electromyogram\space ECG: electrocardiogram\space GSR: galvanic skin response 
\end{tablenotes}
\end{threeparttable}
\end{center}
\end{table}

\subsubsection{Augmentation Methods \& Regularization}
Several augmentation methods have been employed to train the proposed framework. For the pre-training process, \textit{RandAugment} \cite{randAugment} and \textit{TrivialAugment} \cite{trivialAugment} were used, along with auxiliary noise from a uniform distribution and a custom \textit{MaskOut} technique that masks out random square sections of input images. In the automatic pain assessment task, \textit{AugMix} \cite{augmix} is used in addition to the previously mentioned methods. Moreover, \textit{Label Smoothing} \cite{label_smoothing} and \textit{DropOut} \cite{dropout} are implemented as regularization techniques to optimize the training outcome.

\subsection{Experimental Evaluation \& Results}
We employ the dataset provided by the challenge organizers \cite{ai4pain,rojas_hirachan_2023}, which includes facial videos and fNIRS data from $65$ participants. The dataset is partitioned into 41 training, 12 validation, and 12 testing subjects, all recorded at the Human-Machine Interface Laboratory, University of Canberra, Australia. Electrodes for transcutaneous electrical nerve stimulation, used as pain stimuli, were placed on the right hand's inner forearm and back. The study measures both pain threshold---the lowest stimulus intensity perceived as painful---and pain tolerance---the maximum intensity of pain a participant can tolerate.
For the fNIRS measurements, $24$ channels were utilized for both HbO and HbR, and each video in the dataset contains $30$ frames. This paper focuses on the results from the validation segment of the dataset, which are structured into a multi-level classification setting for \textit{No Pain}, \textit{Low Pain}, and \textit{High Pain}.
Table \ref{table:1_training_details} details the training framework for the automatic pain assessment. It should be noted that while numerous experiments were conducted across different modalities and their combinations, only the most successful outcomes are discussed in the subsequent sections and detailed in the corresponding tables.

\renewcommand{\arraystretch}{1.2}
\begin{table}
\small
\center
\caption{Training details for the automatic pain assessment.}
\label{table:1_training_details}
\begin{center}
\begin{threeparttable}
\begin{tabular}{ P{1.5cm} P{0.8cm} P{1.0cm} P{1.0cm} P{1.0cm} P{1.5cm} P{1.5cm} P{1.0cm}}
\toprule
Optimizer & LR &LR decay &Weight decay &Epochs &Warmup epochs &Cooldown epochs &Batch size\\
\midrule
\midrule
\textit{AdamW}   &\textit{2e-5} &\textit{cosine}  &0.1 &100 &10 &10 &32\\
\bottomrule
\end{tabular}
\begin{tablenotes}
\scriptsize
\item LR: learning rate
\end{tablenotes}
\end{threeparttable}
\end{center}
\end{table}

\subsubsection{Facial Videos}
For facial videos, two embedding fusion methods were implemented: the \textit{Addition} technique, which aggregates $30$ embeddings into a single vector of dimension $d=500$, and the \textit{Concatenation} approach, which merges the embeddings into a larger vector of $d=15,000$. With the \textit{Addition} method, the initial accuracy reached $41.90\%$ under basic augmentation and regularization settings. Enhancing the augmentation intensity and adjusting \textit{MaskOut} improved the accuracy incrementally, achieving a peak of $44.91\%$. Adjusting \textit{DropOut} and other augmentation parameters refined the performance to $43.52\%$. These findings are detailed in Table \ref{table:1_videos_addition}.
Employing the \textit{Concatenation} method with initial uniform augmentation probabilities resulted in a starting accuracy of $40.28\%$. Strategic increases in \textit{MaskOut} and maintaining other augmentations at moderate levels led to a gradual accuracy improvement, culminating in a high of $43.75\%$ when all augmentations were maximized except \textit{MaskOut}, paired with high regularization settings. The results of this approach are outlined in Table \ref{table:1_videos_concat}.

\renewcommand{\arraystretch}{1.0}
\begin{table}
\small
\center
\caption{Results utilizing the video modality \& \textit{Addition} method.}
\label{table:1_videos_addition}
\begin{center}
\begin{threeparttable}
\begin{tabular}{ P{1.0cm}  P{1.0cm} P{1.0cm} P{1.5cm}  P{0.5cm}  P{1.3cm}  P{1.0cm} }
\toprule
\multicolumn{4}{c}{Augmentation} 
&\multicolumn{2}{c}{Regularization} 
&\multicolumn{1}{c}{Task}\\ 
\cmidrule(lr){1-4}\cmidrule(lr){5-6}\cmidrule(lr){7-7}
\textit{AugMix} &\textit{Rand} &\textit{Trivial} &\textit{MaskOut} &\textit{LS} &\textit{DropOut} &MC\\
\midrule
\midrule
0.1 &0.1 &0.1 &0.1\textbar 3  &0.1 &0.5 &41.90\\
0.5 &0.5 &0.5 &0.7\textbar 3  &0.0 &0.5 &\textbf{44.91}\\
0.5 &0.5 &0.5 &0.7\textbar 10 &0.0 &0.5 &42.13\\
0.5 &0.5 &0.5 &0.7\textbar 3  &0.0 &0.6 &42.36\\
0.9 &0.9 &0.9 &0.7\textbar 3  &0.3 &0.7 &43.52\\
\bottomrule 
\end{tabular}
\begin{tablenotes}
\scriptsize
\item \textit{Rand}: \textit{RandAugment} \space  \textit{Trivial}: TrivialAugment \space \textit{LS}: \textit{Label Smoothing} \space MS: multiclass pain assessment. For Augmentation \& Regularization the first number represents the probability of application, while in \textit{MaskOut} the number followed \textbar \space indicates the number of square  sections applied.
\end{tablenotes}
\end{threeparttable}
\end{center}
\end{table}

\renewcommand{\arraystretch}{1.0}
\begin{table}
\small
\center
\caption{Results utilizing the video modality \& \textit{Concatenation} method.}
\label{table:1_videos_concat}
\begin{center}
\begin{threeparttable}
\begin{tabular}{ P{1.0cm}  P{1.0cm} P{1.0cm} P{1.5cm}  P{0.5cm}  P{1.3cm}  P{1.0cm} }
\toprule
\multicolumn{4}{c}{Augmentation} 
&\multicolumn{2}{c}{Regularization} 
&\multicolumn{1}{c}{Task}\\ 
\cmidrule(lr){1-4}\cmidrule(lr){5-6}\cmidrule(lr){7-7}
\textit{AugMix} &\textit{Rand} &\textit{Trivial} &\textit{MaskOut} &\textit{LS} &\textit{DropOut} &MC\\
\midrule
\midrule
0.3 &0.3 &0.3 &0.3\textbar 3 &0.1 &0.5 &40.28\\
0.5 &0.5 &0.5 &0.8\textbar 5 &0.0 &0.5 &41.44\\
0.9 &0.9 &0.9 &0.7\textbar 3 &0.2 &0.7 &42.13\\
0.9 &0.9 &0.9 &0.7\textbar 1 &0.4 &0.5 &41.90\\
0.9 &0.9 &0.9 &0.6\textbar 3 &0.4 &0.5 &\textbf{43.75}\\
\bottomrule 
\end{tabular}
\end{threeparttable}
\end{center}
\end{table}

\subsubsection{fNIRS}
Similar to facial videos, the \textit{Addition} and \textit{Concatenation} methods were also applied to the fNIRS channels, excluding two faulty ones from the original $24$. For the HbR with the \textit{Addition} method, the initial accuracy was $39.35\%$, set with uniform probabilities of $0.5$ for \textit{AugMix}, \textit{Rand}, and \textit{Trivial}, and \textit{MaskOut} adjusted to $0.6|5$. Modifying \textit{MaskOut} to $0.7|3$ and increasing \textit{LS} slightly reduced accuracy, while subsequent adjustments in \textit{LS} and \textit{DropOut} improved it to $41.20\%$ (refer to Table \ref{table:hhb_addition}).
In the HbR with the \textit{Concatenation} method, starting with \textit{MaskOut} at $0.7|3$ led to an accuracy of $40.97\%$. Escalating all augmentations to $0.9$, while keeping \textit{MaskOut} at $0.7|3$, achieved a peak accuracy of $42.13\%$ (refer to Table \ref{table:hhb_concat}).
For the HbO with the \textit{Addition} method, accuracies started at $43.06\%$ with uniform augmentation probabilities of $0.3$ and \textit{MaskOut} at $0.3|3$. Elevating \textit{MaskOut} to $0.7|3$ with minor adjustments in \textit{LS} and \textit{DropOut} maintained similar accuracies, while optimizing \textit{MaskOut} to $0.8|3$ enhanced performance to $44.68\%$ (refer to Table \ref{table:02hb_addition}).
The HbO with the \textit{Concatenation} method began with an accuracy of $42.13\%$ under an augmentation probability of $0.1$. A balanced augmentation setup at $0.9$ and \textit{MaskOut} at $0.7|3$ lifted the peak accuracy to $44.44\%$, demonstrating the effectiveness of increased overall augmentation coupled with high regularization. Further adjustments slightly reduced accuracy, highlighting the critical nature of optimal augmentation settings (refer to Table \ref{table:o2hb_concat}).
Generally, enhanced performance is observed with HbO compared to HbR, as noted in other studies \cite{rojas_huang_2017_c}, attributed to its superior signal-to-noise ratio. Combining HbR and HbO using the \textit{Addition} method initially showed an accuracy of $42.82\%$ with all augmentations at zero except for \textit{MaskOut} at $0.7|3$. Increasing \textit{AugMix}, \textit{Rand}, and \textit{Trivial} to $0.5$ while raising \textit{MaskOut} to $0.7|7$ slightly improved accuracy to $43.29\%$. Adjustments to \textit{MaskOut} back to $0.7|3$ with a slight increase in \textit{LS} led to a minor reduction in accuracy to $42.59\%$. However, further increasing all augmentations to $0.9$ and \textit{LS} to $0.3$ while maintaining \textit{MaskOut} at $0.7|3$ maximized accuracy to $43.75\%$. A reduction in \textit{DropOut} to $0.1$ in the final setup slightly reduced accuracy to $43.06\%$, emphasizing the importance of optimizing regularization alongside augmentation strategies for achieving optimal results (refer to Table \ref{table:hhb_o2hb_addition}).

\renewcommand{\arraystretch}{1.0}
\begin{table}
\small
\center
\caption{Results utilizing the HbR \& \textit{Addition} method.}
\label{table:hhb_addition}
\begin{center}
\begin{threeparttable}
\begin{tabular}{ P{1.0cm}  P{1.0cm} P{1.0cm} P{1.5cm}  P{0.5cm}  P{1.3cm}  P{1.0cm} }
\toprule
\multicolumn{4}{c}{Augmentation} 
&\multicolumn{2}{c}{Regularization} 
&\multicolumn{1}{c}{Task}\\ 
\cmidrule(lr){1-4}\cmidrule(lr){5-6}\cmidrule(lr){7-7}
\textit{AugMix} &\textit{Rand} &\textit{Trivial} &\textit{MaskOut} &\textit{LS} &\textit{DropOut} &MC\\
\midrule
\midrule
0.5 &0.5 &0.5 &0.6\textbar 5 &0.0 &0.5 &39.35\\
0.5 &0.5 &0.5 &0.7\textbar 3 &0.4 &0.5 &38.89\\
0.9 &0.9 &0.9 &0.7\textbar 3 &0.1 &0.9 &40.05\\
0.9 &0.9 &0.9 &0.7\textbar 5 &0.4 &0.5 &\textbf{41.20}\\
0.5 &0.5 &0.5 &0.7\textbar 3 &0.0 &0.4 &40.51\\
\bottomrule 
\end{tabular}
\end{threeparttable}
\end{center}
\end{table}

\renewcommand{\arraystretch}{1.0}
\begin{table}
\small
\center
\caption{Results utilizing the HbR \& \textit{Concatenation} method.}
\label{table:hhb_concat}
\begin{center}
\begin{threeparttable}
\begin{tabular}{ P{1.0cm}  P{1.0cm} P{1.0cm} P{1.5cm}  P{0.5cm}  P{1.3cm}  P{1.0cm} }
\toprule
\multicolumn{4}{c}{Augmentation} 
&\multicolumn{2}{c}{Regularization} 
&\multicolumn{1}{c}{Task}\\ 
\cmidrule(lr){1-4}\cmidrule(lr){5-6}\cmidrule(lr){7-7}
\textit{AugMix} &\textit{Rand} &\textit{Trivial} &\textit{MaskOut} &\textit{LS} &\textit{DropOut} &MC\\
\midrule
\midrule
0.0 &0.0 &0.7 &0.7\textbar 3 &0.0 &0.5 &40.97\\
0.5 &0.5 &0.5 &0.7\textbar 1 &0.0 &0.5 &41.44\\
0.9 &0.9 &0.9 &0.7\textbar 3 &0.1 &0.8 &\textbf{42.13}\\
0.9 &0.9 &0.9 &0.7\textbar 3 &0.4 &0.5 &41.20\\
0.5 &0.5 &0.5 &0.7\textbar 3 &0.0 &0.3 &39.81\\
\bottomrule 
\end{tabular}
\end{threeparttable}
\end{center}
\end{table}

\renewcommand{\arraystretch}{1.0}
\begin{table}
\small
\center
\caption{Results utilizing the HbO \& \textit{Addition} method.}
\label{table:02hb_addition}
\begin{center}
\begin{threeparttable}
\begin{tabular}{ P{1.0cm}  P{1.0cm} P{1.0cm} P{1.5cm}  P{0.5cm}  P{1.3cm}  P{1.0cm} }
\toprule
\multicolumn{4}{c}{Augmentation} 
&\multicolumn{2}{c}{Regularization} 
&\multicolumn{1}{c}{Task}\\ 
\cmidrule(lr){1-4}\cmidrule(lr){5-6}\cmidrule(lr){7-7}
\textit{AugMix} &\textit{Rand} &\textit{Trivial} &\textit{MaskOut} &\textit{LS} &\textit{DropOut} &MC\\
\midrule
\midrule
0.3 &0.3 &0.3 &0.3\textbar 3 &0.1 &0.5 &43.06\\
0.5 &0.5 &0.5 &0.7\textbar 3 &0.2 &0.5 &42.82\\
0.9 &0.9 &0.9 &0.7\textbar 3 &0.4 &0.8 &43.29\\
0.9 &0.9 &0.9 &0.7\textbar 9 &0.4 &0.5 &44.44\\
0.9 &0.9 &0.9 &0.8\textbar 3 &0.4 &0.5 &\textbf{44.68}\\
\bottomrule 
\end{tabular}
\end{threeparttable}
\end{center}
\end{table}

\renewcommand{\arraystretch}{1.0}
\begin{table}
\small
\center
\caption{Results utilizing the HbO \& \textit{Concatenation} method.}
\label{table:o2hb_concat}
\begin{center}
\begin{threeparttable}
\begin{tabular}{ P{1.0cm}  P{1.0cm} P{1.0cm} P{1.5cm}  P{0.5cm}  P{1.3cm}  P{1.0cm} }
\toprule
\multicolumn{4}{c}{Augmentation} 
&\multicolumn{2}{c}{Regularization} 
&\multicolumn{1}{c}{Task}\\ 
\cmidrule(lr){1-4}\cmidrule(lr){5-6}\cmidrule(lr){7-7}
\textit{AugMix} &\textit{Rand} &\textit{Trivial} &\textit{MaskOut} &\textit{LS} &\textit{DropOut} &MC\\
\midrule
\midrule
0.1 &0.1 &0.1 &0.1\textbar 3 &0.1 &0.5 &42.13\\
0.5 &0.5 &0.5 &0.0\textbar 0 &0.0 &0.5 &43.98\\ 
0.5 &0.5 &0.5 &0.7\textbar 1 &0.0 &0.5 &42.36\\ 
0.9 &0.9 &0.9 &0.7\textbar 3 &0.4 &0.9 &\textbf{44.44}\\ 
0.5 &0.5 &0.5 &0.7\textbar 3 &0.0 &0.8 &43.52\\
\bottomrule 
\end{tabular}
\end{threeparttable}
\end{center}
\end{table}

\renewcommand{\arraystretch}{1.0}
\begin{table}
\small
\center
\caption{Results utilizing the HbR, HbO \& \textit{Addition} method.}
\label{table:hhb_o2hb_addition}
\begin{center}
\begin{threeparttable}
\begin{tabular}{ P{1.0cm}  P{1.0cm} P{1.0cm} P{1.5cm}  P{0.5cm}  P{1.3cm}  P{1.0cm} }
\toprule
\multicolumn{4}{c}{Augmentation} 
&\multicolumn{2}{c}{Regularization} 
&\multicolumn{1}{c}{Task}\\ 
\cmidrule(lr){1-4}\cmidrule(lr){5-6}\cmidrule(lr){7-7}
\textit{AugMix} &\textit{Rand} &\textit{Trivial} &\textit{MaskOut} &\textit{LS} &\textit{DropOut} &MC\\
\midrule
\midrule
0.0 &0.0 &0.7 &0.7\textbar 3 &0.0 &0.5 &42.82\\
0.5 &0.5 &0.5 &0.7\textbar 7 &0.0 &0.5 &43.29\\
0.5 &0.5 &0.5 &0.7\textbar 3 &0.1 &0.5 &42.59\\
0.9 &0.9 &0.9 &0.7\textbar 3 &0.3 &0.9 &\textbf{43.75}\\
0.5 &0.5 &0.5 &0.7\textbar 3 &0.0 &0.1 &43.06\\
\bottomrule 
\end{tabular}
\end{threeparttable}
\end{center}
\end{table}

\renewcommand{\arraystretch}{1.0}
\begin{table}
\small
\center
\caption{Results utilizing the videos, HbO \& \textit{Addition} method.}
\label{table:video_o2hb_addition}
\begin{center}
\begin{threeparttable}
\begin{tabular}{ P{1.0cm}  P{1.0cm} P{1.0cm} P{1.5cm}  P{0.5cm}  P{1.3cm}  P{1.0cm} }
\toprule
\multicolumn{4}{c}{Augmentation} 
&\multicolumn{2}{c}{Regularization} 
&\multicolumn{1}{c}{Task}\\ 
\cmidrule(lr){1-4}\cmidrule(lr){5-6}\cmidrule(lr){7-7}
\textit{AugMix} &\textit{Rand} &\textit{Trivial} &\textit{MaskOut} &\textit{LS} &\textit{DropOut} &MC\\
\midrule
\midrule
0.5 &0.5 &0.5 &0.4\textbar 5 &0.0 &0.5 &42.36\\
0.5 &0.5 &0.5 &0.7\textbar 9 &0.0 &0.5 &41.67\\
0.9 &0.9 &09. &0.7\textbar 3 &0.1 &0.6 &42.59\\
0.9 &0.9 &0.9 &0.7\textbar 3 &0.3 &0.9 &43.06\\\
0.9 &0.9 &0.9 &0.7\textbar 5 &0.4 &0.5 &\textbf{43.75}\\
\bottomrule 
\end{tabular}
\end{threeparttable}
\end{center}
\end{table}

\renewcommand{\arraystretch}{1.0}
\begin{table}
\small
\center
\caption{Results utilizing the videos, HbO \& \textit{Single Diagram} method.}
\label{table:video_o2hb_1_image}
\begin{center}
\begin{threeparttable}
\begin{tabular}{ P{1.0cm}  P{1.0cm} P{1.0cm} P{1.5cm}  P{0.5cm}  P{1.3cm}  P{1.0cm} }
\toprule
\multicolumn{4}{c}{Augmentation} 
&\multicolumn{2}{c}{Regularization} 
&\multicolumn{1}{c}{Task}\\ 
\cmidrule(lr){1-4}\cmidrule(lr){5-6}\cmidrule(lr){7-7}
\textit{AugMix} &\textit{Rand} &\textit{Trivial} &\textit{MaskOut} &\textit{LS} &\textit{DropOut} &MC\\
\midrule
\midrule
0.5 &0.5 &0.5 &0.3\textbar 5 &0.0 &0.5 &45.83\\
0.9 &0.9 &0.9 &0.7\textbar 3 &0.1 &0.6 &\textbf{46.76}\\
0.9 &0.9 &0.9 &0.7\textbar 3 &0.3 &0.6 &46.53\\
0.9 &0.9 &0.9 &0.9\textbar 3 &0.4 &0.5 &45.83\\\
0.5 &0.5 &0.5 &0.7\textbar 3 &0.0 &0.7 &45.14\\
\bottomrule 
\end{tabular}
\end{threeparttable}
\end{center}
\end{table}

\subsubsection{Fusion}
This section explores the fusion of facial videos and fNIRS, explicitly utilizing HbO due to its demonstrated superior performance over HbR. Two fusion methods were employed: the \textit{Addition} method, which aggregates embeddings from video frames and fNIRS channels into a unified vector, and the \textit{Single-Diagram} method, where aggregated embeddings from both modalities are visualized simultaneously in a single image. For the \textit{Addition} method, initial configurations with moderate augmentation levels ($0.5$ for \textit{AugMix}, \textit{Rand}, \textit{Trivial}) and \textit{MaskOut} at $0.4|5$ achieved an accuracy of $42.36\%$. Increasing the augmentation levels to $0.9$ and adjusting the regularization parameters (\textit{LS} up to $0.4$ and \textit{DropOut} up to $0.9$) enhanced the accuracy, peaking at $43.75\%$ (refer to Table \ref{table:video_o2hb_addition}). For the \textit{Single Diagram} method, accuracy improvements were noted, as shown in Table \ref{table:video_o2hb_1_image}. Starting with lower \textit{MaskOut} levels at $0.3|5$ and standard augmentation probabilities ($0.5$), the accuracy was $45.83\%$. Utilizing augmentation probabilities to $0.9$ and \textit{MaskOut} adjustments to $0.7|3$ significantly improved performance, achieving a high of $46.76\%$.

\subsubsection{Interpretation \& Comparison}
In the framework's analysis, attention maps from the last layer of \textit{PainViT--2} were generated, illustrating the processed unified image that integrates both the video and HbO embedding waveforms. This layer consists of $500$ neurons, each specifically engaging with different input aspects. Figure \ref{1_attention} displays four examples demonstrating how specific neurons predominantly focus on the video embedding waveform. In contrast, others concentrate on the HbO waveform, and some attend to both, highlighting various details.
Table \ref{table:1_comparison} compares the proposed pipeline and the baseline results set by the challenge organizers. The video-based approach utilizing the \textit{Addition} method exceeded the baseline by $4.91\%$. Implementing the HbO with the \textit{Addition} method showed a minor improvement of $1.48\%$. However, the fusion of modalities through the \textit{Single Diagram} method achieved a more substantial gain of $6.56\%$.

\begin{figure}
\begin{center}
\includegraphics[scale=0.275]{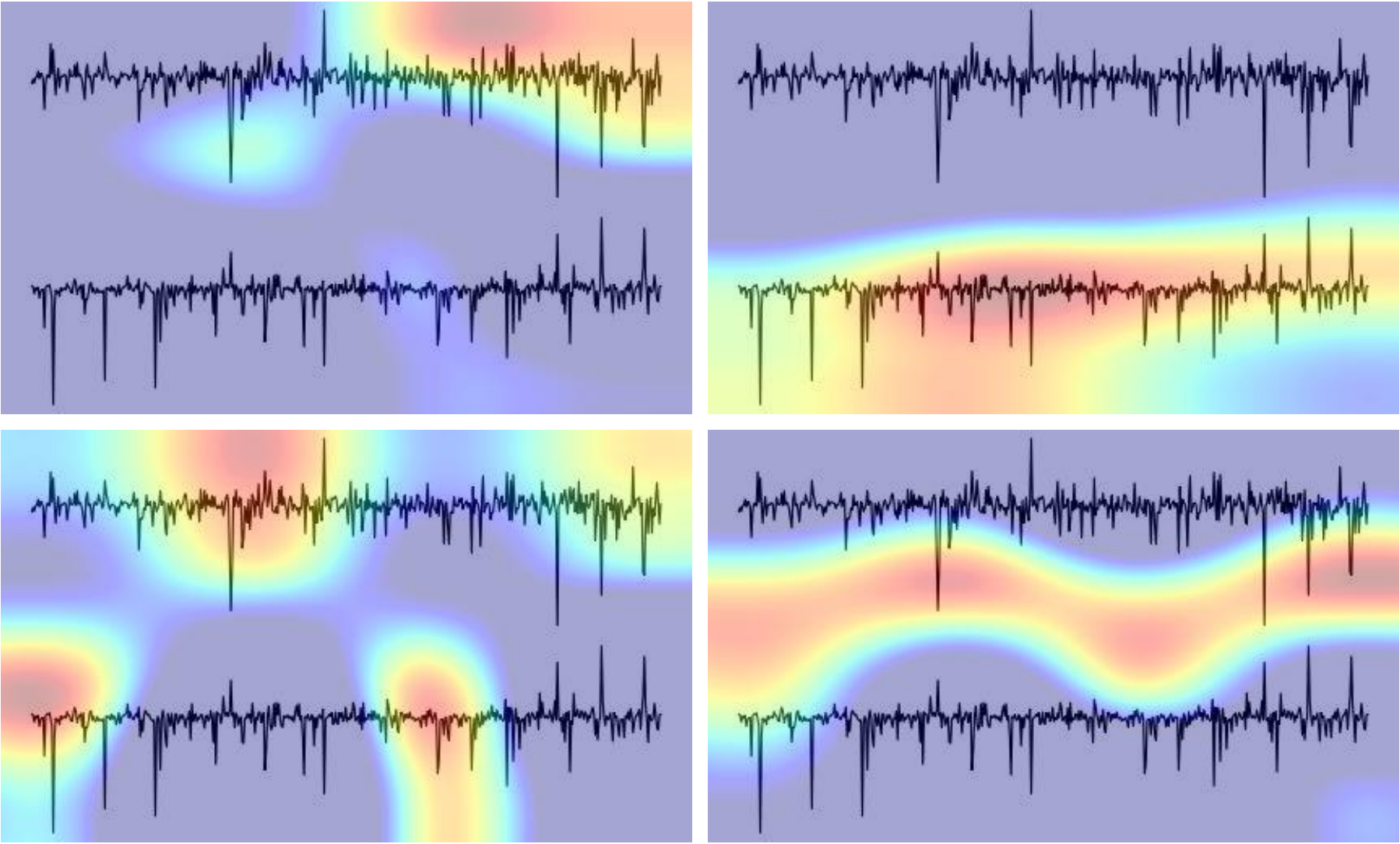}
\end{center}
\caption{Attention maps from the \textit{PainViT--2}.}
\label{1_attention}
\end{figure}

\renewcommand{\arraystretch}{1.0}
\begin{table}
\small
\center
\caption{Comparison with the validation baseline provided by the \textit{AI4PAIN} challenge organizers.}
\label{table:1_comparison}
\begin{center}
\begin{threeparttable}
\begin{tabular}{ P{1.8cm} P{1.8cm} P{1.8cm} P{1.8cm}}
\toprule
\multirow{2}[2]{*}{\shortstack{Approach}}
&\multicolumn{3}{c}{Modality}\\ 
\cmidrule(lr){2-4}
&Video &fNIRS &Fusion\\
\midrule
\midrule
Baseline &40.00 &43.20 &40.20\\
Our &44.91 &44.68 &46.76\\
\bottomrule 
\end{tabular}
\begin{tablenotes}
\scriptsize
\item 
\end{tablenotes}
\end{threeparttable}
\end{center}
\end{table}

\subsection{Discussion}
This chapter contributes to the \textit{First Multimodal Sensing Grand Challenge for Next-Gen Pain Assessment (AI4PAIN)}, utilizing facial videos and fNIRS in a modality-agnostic framework. The \textit{Twins-PainViT}, a dual Vision Transformer configuration, was pre-trained across multiple datasets using a multi-task learning approach. A key feature of our approach is the waveform representation applied to the original fNIRS data and the learned embeddings, allowing for their integration into a single image diagram. This method effectively eliminates the need for domain-specific models for each modality.
Our experiments demonstrated high performance in both unimodal and multimodal configurations, outperforming the established baselines. The analysis of \textit{PainViT--2} through attention maps further revealed that specific neurons specifically target different modalities or distinct aspects within them, suggesting a comprehensive analytical approach.
Future research should continue to explore multimodal strategies, as they have shown superior efficacy in real-world pain assessment settings. Developing interpretative methods is crucial for integrating these advanced frameworks into clinical practice.

\section{A Foundation Model for Automatic Pain Assessment}
We introduce \textit{PainFormer}, a multi-task learning vision foundation model tailored for automatic pain assessment. This initiative is the first to develop and deploy a foundation model for pain recognition, inspired by the frameworks discussed in \cite{reed_zolna_2022}. Our method involves training across various datasets and tasks, leveraging large-scale corpora to enhance representation learning for pain assessment applications.
This research makes three key contributions: (1) it introduces a foundation model capable of extracting robust embeddings from diverse modalities, (2) it incorporates synthetic thermal and estimated depth videos as innovative modalities, and (3) it evaluates the performance of these modalities in both unimodal and multimodal settings.

\subsection{Methodology}
This section outlines the structure and components of the suggested framework. It covers the foundation model's pretraining based on multi-task learning, the methods for augmentation, and the training configurations for pretraining and pain evaluation tasks. It also describes how synthetic thermal and depth videos are generated and details the visualization techniques for biosignal modalities employed in this research.

\subsubsection{Framework Architecture}
The framework integrates three models: the \textit{PainFormer}, a foundation model that extracts embeddings from input data; the \textit{Embedding-Mixer}, which applies these embeddings, either singly or in combination, to classify pain; and the \textit{Video-Encoder}, which reduces video data into a lower-dimensional latent space for use in the multimodal approaches that are explained later.
This framework operates in two separate stages: initially extracting embeddings and then deploying them according to the demands of specific modality pipelines.
Table \ref{table:ch7_p2_module_parameters} details the parameters and computational costs for each module's floating-point operations (FLOPS).

\renewcommand{\arraystretch}{1.0}
\begin{table}
\center
\small
\caption{Number of parameters and FLOPS for the modules of the proposed framework.}
\label{table:ch7_p2_module_parameters}
\begin{center}
\begin{threeparttable}
\begin{tabular}{ P{3.0cm}  P{2.5cm}  P{2.0cm}}
\toprule
Module & Params (Millions) &FLOPS  (Giga) \\
\midrule
\midrule
\textit{PainFormer}      &19.60 &5.82\\
\textit{Embedding-Mixer} &9.85  &2.94 \\
\textit{Video-Encoder}   &3.37  &0.86 \\
\hline
Total &32.82 &9.62\\
\bottomrule
\end{tabular}
\begin{tablenotes}
\scriptsize
\item
\end{tablenotes}
\end{threeparttable}
\end{center}
\end{table}

\begin{figure}
\begin{center}
\includegraphics[scale=0.445]{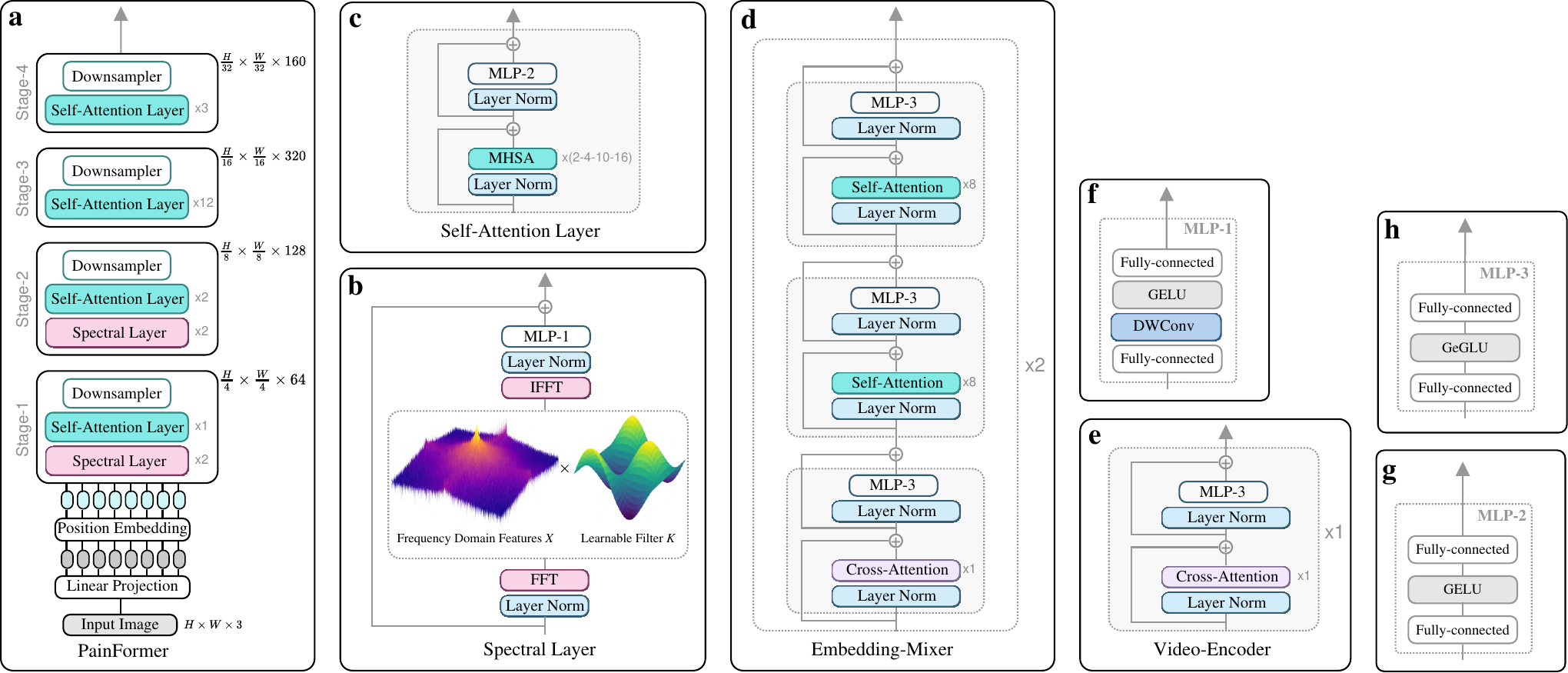} 
\end{center}
\caption{Overview of primary models and their components outlined in this research:
\textbf{(a)} \textit{PainFormer} is structured hierarchically into four stages, incorporating \textit{Spectral} and \textit{Self-Attention Layers} to extract embeddings from the inputs;
\textbf{(b)} The \textit{Spectral Layer}, a key element of \textit{PainFormer}, uses FFT to analyze frequency-specific data along with a learnable filter $K$ to highlight critical frequencies;
\textbf{(c)} The \textit{Self-Attention Layer}, crucial for \textit{PainFormer}, enables parallel processing of features and their interconnections;
\textbf{(d)} The \textit{Embedding-Mixer}, employing both cross and self-attention mechanisms, functions as the component for the final classification of embeddings in pain assessment;
\textbf{(e)} The \textit{Video-Encoder}, designed for compact and efficient encoding, compresses video data into a reduced dimensional form;
\textbf{(f)} The \textit{MLP-1} is part of the Spectral Layer;
\textbf{(g)} The \textit{MLP-2} is included in the \textit{Self-Attention Layer};
\textbf{(h)} The \textit{MLP-3} configuration is integrated into the \textit{Embedding-Mixer} and \textit{Video-Encoder}.}

\label{2_full}
\end{figure}

\paragraph{PainFormer:}
\label{painformer}
Vision Transformers (ViT) have become increasingly widespread for various image-processing tasks, demonstrating the effectiveness of their self-attention mechanisms. Moreover, developing Vision Multilayer Perceptron (Vision-MLP) models that use spectral mixing techniques---substituting self-attention layers with Fourier transformation layers---illustrates that simpler structures with fewer inductive biases can achieve similar outcomes.
Our strategy incorporates two key concepts: hierarchical Vision Transformers (ViT) \cite{swin_transformer}, which use multiple embedding extraction stages to boost performance and scalability, and the Fourier transform's efficient token information mixing as shown in \cite{guibas_mardani_2021}.
\textit{PainFormer} integrates spectral layers using the Fast Fourier Transform (FFT) with self-attention layers. Both spectral and self-attention layers are initially applied, whereas later stages rely solely on self-attention. The architecture of \textit{PainFormer} is illustrated in \hyperref[2_full]{Fig. 1(a)}.
Each 2D input image $I$ is segmented into $n$ non-overlapping patches, each patch $\in \mathbb{R}^{n \times h \times w \times 3}$, where $h$ and $w$ are the patch resolution set at $16 \times 16$, and $3$ represents the RGB channels. Each patch is linearly projected into a dimension $d = 768$, followed by positional encoding.
Applying Discrete Fourier Transform (DFT) to a 1D sequence of $N$ elements, $x[n]$, ranging from $0$ to $N-1$, yields:
\begin{equation}
X[k] = \sum_{n=0}^{N-1} x[n] \cdot e^{-i2\pi\tfrac{k}{N}n} := \sum_{n=0}^{N-1} x[n] \cdot W_N^{kn},
\label{dft}
\end{equation}
where $i$ is the imaginary unit, and $W_N$ is defined as $e^{-i\tfrac{2\pi}{N}}$.
The sequence can be transformed back to the time domain by applying the inverse Discrete Fourier Transform (IDFT):
\begin{equation}
x[n] = \frac{1}{N} \sum_{k=0}^{N-1} X[k] \cdot e^{i2\pi \tfrac{k}{N}n},
\label{idft}
\end{equation}
where $x[n]$ represents the original sequence.
Furthermore, for two-dimensional inputs, $x[m,n]$, with $0 \leq m \leq M-1$ and $0 \leq n \leq N-1$, the formula extends to:
\begin{equation}
X[u, v] = \sum_{m=0}^{M-1} \sum_{n=0}^{N-1} x[m, n] \cdot e^{-i 2\pi \big(\tfrac{um}{M} + \tfrac{vn}{N}\big)},
\label{2d_dft}
\end{equation}
where $X[u, v]$ is the frequency-domain representation of the input $x[m,n]$.

\subparagraph{Spectral Layer:}
For the tokens $x$ from image $I$, a 2D FFT is applied across the spatial dimensions to transform $x$ into the frequency domain:
\begin{equation}
X = \mathscr{F}[x] \in \mathbb{C}^{h\times w\times d}.
\label{tokens_fft}
\end{equation}
After applying the FFT to extract the various frequency components of the image, we employ a learnable filter, 
$K \in \mathbb{C}^{h\times w\times d}$ acts as a gate to regulate the significance of each frequency component. This spectrum modulation allows for the identification and learning of features such as lines and edges. Specifically: 
\begin{equation}
\tilde{X} = K \odot X,
\label{filter_k}
\end{equation}
where $\odot$ defines the element-wise multiplication.
Afterward, the inverse Fast Fourier Transform (IFFT) is applied, which converts the spectral space back into the physical space:
\begin{equation}
x \leftarrow \mathscr{F}^{-1}[\tilde{X}],
\label{ifft}
\end{equation}
where the physical space is referred to as the spatial domain in this case.
The final component of a spectrum layer is an MLP module, which enables efficient channel mixing communication:
\begin{equation}
\Phi(x) = W_2 \cdot \text{GELU}(\text{DWConv}(W_1 \cdot x + b_1)) + b_2,
\label{mlp_spectrum}
\end{equation}
where DWConv denotes a depthwise convolution layer.
In addition, layer normalization is employed before and after the FFT and IFFT processes, refer to \hyperref[2_full]{Fig. 1(b)}.

\subparagraph{Self-Attention Layer:}
In this layer, the standard self-attention mechanism characteristic of transformers is utilized. For a given token sequence $X$, the attention mechanism is defined as:
\begin{equation}
\text{Att}(X) := \text{softmax} \left( \frac{XW_q (XW_k)^T}{\sqrt{d}} \right) XW_v, 
\label{self_attention}
\end{equation}
where $\text{Att}$ maps $\mathbb{R}^{N\times d}$ to $\mathbb{R}^{N\times d}$, with $N$ representing $hw$. The matrices $W_q$, $W_k$, and $W_v$ in $\mathbb{R}^{d\times d}$ correspond to the query, key, and value weights, respectively. 
Layer normalization is applied both before and after the attention mechanism, mirroring the approach used in the spectral layer. Additionally, the MLP component within this layer is expressed as:
\begin{equation}
\Phi(x) = W_2 \cdot \text{GELU}(W_1 \cdot x + b_1) + b_2.
\label{mlp_attention}
\end{equation}
\hyperref[2_full]{Fig. 1(c)} illustrates the design of this layer.

\subparagraph{Stages:} 
A stage-based architecture was developed to create a hierarchical representation. \textit{PainFormer} is organized into four stages, each followed by a single-layer 2D CNN that downsamples the resolution by a factor of $2$ to reduce the token dimensions. Additionally, each stage incorporates a distinct combination of spectral and self-attention layers, with varying numbers of heads in the self-attention layers and different dimensions for the extracted tokens. Table \ref{table:ch7_stages} presents the relevant details.

\renewcommand{\arraystretch}{1.0}
\begin{table}
\caption{Details of the \textit{PainFormer's} architecture.}
\label{table:ch7_stages}
\begin{center}
\begin{threeparttable}
\begin{tabular}{ P{1.0cm}  P{2.0cm} P{2.5cm} P{2.5cm} P{1.5cm}}
\toprule
Stage & \# Spectral \newline Layers & \# Self-Attention Layers & \# Self-Attention Heads & Dimension \newline $d$\\
\midrule
\midrule
1  & 2  & 1  & 2  &64 \\
2  & 2  & 2  & 4  &128 \\
3  & -- & 12 & 10 &320 \\
4  & -- & 3  & 16 &160 \\

\bottomrule
\end{tabular}
\begin{tablenotes}
\scriptsize
\item $d$: token dimensions
\end{tablenotes}
\end{threeparttable}
\end{center}
\end{table}

\paragraph{Embedding-Mixer:}
The model is built on a transformer architecture, leveraging cross- and self-attention mechanisms. Echoing insights from prior research \cite{jaegle_perceiver_2021}, it employs an asymmetrical attention scheme using cross-attention with fewer latent variables to reduce computational demands and boost efficiency. Cross-attention operation parallels self-attention, as detailed in Eq. (\ref{self_attention}). Nonetheless, the dimensions for $W_q$, $W_k$, and $W_v$ adjust to $n\times d$ from $d\times d$, with $n<d$ and $n$ specifically set at $256$.
The \textit{Embedding-Mixer} includes $2$ layers, each equipped with $1$ cross-attention and $2$ self-attention modules. The head counts for cross- and self-attention are $1$ and $8$, respectively. The final classification task utilizes an output embedding of length $512$, as shown in \hyperref[2_full]{Fig. 1(d)}.

\paragraph{Video-Encoder:}
The design of this specific module mirrors the \textit{Embedding-Mixer}. However, it is simplified for efficiency by including only a $1$ layer that contains a single cross-attention module with a $1$ head. The number of latent variables, $n$, is maintained at $256$, and the output embedding length is set at $40$. This module functions exclusively within a particular framework as part of one of the multimodal strategies, integrating video and GSR embeddings. The module's architecture is depicted in \hyperref[2_full]{Fig. 1(e)}.

\subsubsection{Synthetic Thermal \& Depth Videos} 
\label{thermal_depth}
\begin{figure}
\begin{center}
\includegraphics[scale=0.18]{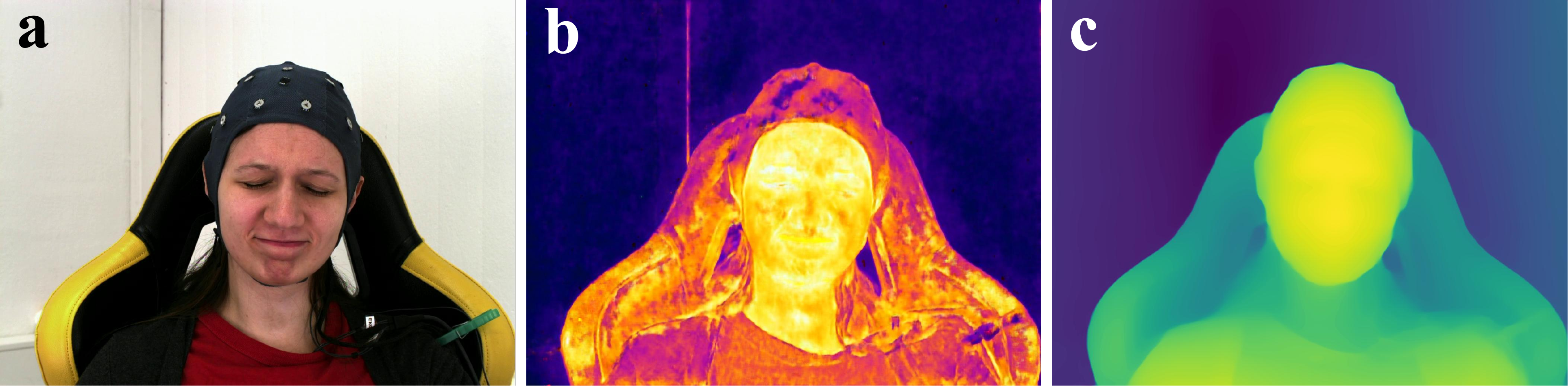} 
\end{center}
\caption{Examples of different vision modalities in frame samples: \textbf{(a)} RGB frame, \textbf{(b)} synthetic thermal frame, and \textbf{(c)} depth estimation frame.}
\label{2_frame_samples}
\end{figure}

This section incorporates thermal and depth vision modalities alandGB videos into our pain assessment frameworks. For the thermal modality, we utilize thermal videos from our earlier research \cite{gkikas_tsiknakis_thermal_2024}, described in Chapter \ref{chapter_6}, which introduced an image-to-image translation (I2I) method using a conditional generative adversarial network (cGAN). This network was designed and trained to map the data distribution from the RGB to the thermal domain, facilitating the creation of synthetic thermal images from new RGB videos. For the depth videos, we employ the \textit{\textquotedblleft Depth Anything\textquotedblright} \space technique \cite{yang_kang_2024}, which is a pioneering model for monocular depth estimation (MDE) that uses a vision transformer-based encoder-decoder architecture with semi-supervised learning. 
Fig. \ref{2_frame_samples} displays a frame sample from the RGB, synthetic thermal, and depth modalities.

\subsubsection{Biosignal Visualization} 
\label{biosignal_visualization}
\begin{figure}
\begin{center}
\includegraphics[scale=0.32]{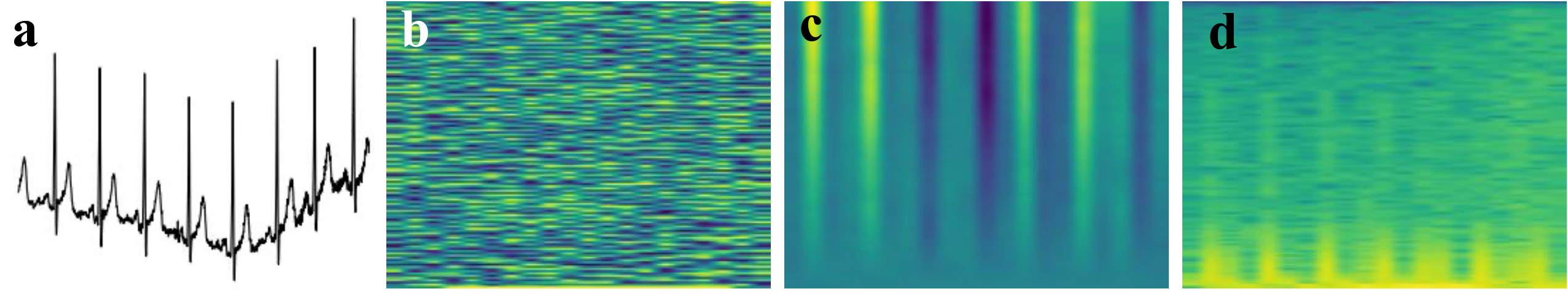} 
\end{center}
\caption{Examples of different visual representations for biosignals: \textbf{(a)} \textit{waveform}, \textbf{(b)} \textit{spectrogram-angle}, \textbf{(c)} \textit{spectrogram-phase}, and \textbf{(d)} \textit{spectrogram-PSD}.}
\label{2_biosignal_samples}
\end{figure}

Given that the core model in this research is vision-based, it necessitates using 2D representations for physiological modalities. We explore four distinct visualizations: (1) \textit{waveform} diagrams, which outline the signal's progression over time, showcasing its amplitude, frequency, and phase characteristics; (2) \textit{spectrogram-angle}, which displays the phase angles associated with different frequencies; (3) \textit{spectrogram-phase}, which reveals phase details and incorporates unwrapping to rectify discontinuities; and (4) \textit{spectrogram-PSD}, which delineates the power spectral density, indicating the distribution of power across frequencies over time. Fig. \ref{2_biosignal_samples} provides an example of each visualization type.

\subsubsection{Foundation Training}
\textit{PainFormer}, our proposed foundation model, serves as an embedding extractor as previously outlined. It has undergone extensive training on $14$ datasets, which include a total of $10.9$ million samples; for further details, see Table \ref{table:ch7_p2_datasets}. 
The training datasets cover a variety of human-centric data, ranging from facial recognition datasets such as \textit{VGGFace2} \cite{cao_shen_2018} and \textit{DigiFace-1M} \cite{digiface1m} to datasets aimed at recognizing basic and compound emotions, like \textit{AffectNet} \cite{mollahosseini_hasani_2019} and \textit{RAF-DB} \cite{li_deng_2017}. Furthermore, datasets based on biosignals like EEG, EMG, and ECG have been incorporated.
Regarding training methodology, \textit{PainFormer} employs a multi-task learning strategy, with each dataset representing a separate supervised learning task. Architecturally, the model adheres to its initial design as specified in \ref{painformer} but now includes additional task-specific auxiliary classifiers. These classifiers comprise a single-layer, fully connected network with an ELU (Exponential Linear Unit) activation function.
The objective during training is to simultaneously learn from all $14$ datasets/tasks. The following equation formalizes this approach:
\begin{equation}
L_{total} = \sum_{i=1}^{14} \left[ e^{w_i} L_{S_i} + w_i \right],
\end{equation}
where $L_{S_i}$ indicates the loss linked to each dataset/task, and $w_i$ are the adaptive weights that aim to minimize the overall loss $L_{total}$, encompassing all individual task losses. The model was trained using this methodology over $200$ epochs.

\renewcommand{\arraystretch}{1.1}
\begin{table}
\center
\small
\caption{Datasets utilized for the multitask learning-based pretraining process of the framework.}
\label{table:ch7_p2_datasets}
\begin{center}
\begin{threeparttable}
\begin{tabular}{ p{5.0cm} p{2.0cm} p{2.0cm} p{2.5cm} }
\toprule
Dataset &\#  samples &\# classes &Modality\\
\midrule
\midrule
\textit{VGGFace2}                              \cite{cao_shen_2018}    &3.31M  &9,131   &Facial Images \\
\textit{SpeakingFaces} RGB     \cite{speakingfaces}$^{\varocircle}$ &0.76M  &142     &Facial Images\\
\textit{SpeakingFaces} Thermal \cite{speakingfaces}$^{\varocircle}$ &0.76M  &142     &Facial Images\\
\textit{DigiFace-1M}           \cite{digiface1m}    &0.72M  &10,000  &Facial Images\\
\textit{DigiFace-1M}           \cite{digiface1m}    &0.50M  &100,000 &Facial Images\\

\textit{AffectNet}       \cite{mollahosseini_hasani_2019} &0.40M  &8      &Facial Images\\
\textit{SFace}           \cite{boutros_huber_2022}        &1.84M  &10,341 &Facial Images\\
\textit{CACIA-WebFace}   \cite{yi_lei_2014}               &0.50M  &10,575 &Facial Images\\
\textit{RAF-DB basic}    \cite{li_deng_2017}              &15,000 &7      &Facial Images\\
\textit{RAF-DB compound} \cite{li_deng_2017}              &4,000  &11     &Facial Images\\
\textit{Compound FEE-DB} \cite{du_tao_2014}               &6,000  &26     &Facial Images\\
\textit{EEG-BST-SZ}      \cite{ford_2013}$^{\varodot}$                 &1.50M   &2      &EEG signals\\
\textit{Silent-EMG}      \cite{gaddy_klein_2020}$^{\varodot}$          &0.19M  &8      &EMG signals\\
\textit{ECG HBC Dataset} \cite{kachuee_fazeli_2018}$^{\varodot}$       &0.45M  &5      &ECG signals\\

\Xhline{1.5\arrayrulewidth}
Total: 14 datasets--tasks &10.9M \\
 
\bottomrule 
\end{tabular}
\begin{tablenotes}
\scriptsize
\item EEG: electroencephalogram \space EMG: electromyography \space ECG \space $\varocircle$: The datasets were also used for the I2I process described in \ref{thermal_depth}, in addition to the training of the \textit{PainFormer} \space $\varodot$: The samples were transformed into spectrograms before being employed.
\end{tablenotes}
\end{threeparttable}
\end{center}
\end{table}

\subsubsection{Augmentation \& Regularization Methods}
Various augmentation and regularization strategies were applied during the pre-training of \textit{PainFormer} and in the downstream pain assessment tasks. For foundational training, \textit{TrivialAugment} \cite{trivialAugment} and \textit{AugMix} \cite{augmix} were used. A customized augmentation method that modifies brightness, contrast, and saturation and involves image cropping was also implemented.
The pre-training regime included adding random noise sourced from a Gaussian distribution. Moreover, a technique was devised to obscure random square portions of the input images. Regularization during pre-training was achieved using \textit{DropPath} \cite{droppath} and \textit{Label Smoothing} \cite{label_smoothing}.
Within the pain assessment framework, two specific augmentation methods were integrated. The first, termed \textit{Basic}, involves polarity inversion and the addition of noise, which alters the original input embeddings by reversing data elements' polarity and adding random noise from a Gaussian distribution, thereby inducing variability. The second method, \textit{Masking}, implements zero-valued masks on the embeddings, effectively nullifying sections of the vectors. These masks are randomly sized and placed, obscuring $10\%$ to $20\%$ of the embedding's total dimensions.
For further regularization, techniques such as \textit{DropOut} \cite{dropout} and \textit{Label Smoothing} \cite{label_smoothing} were employed. Additional specifics on the two training methodologies are detailed in Table \ref{table:ch7_p2_training_details}.

\renewcommand{\arraystretch}{1.0}
\begin{table}
\small
\center
\caption{Training details of the proposed framework.}
\label{table:ch7_p2_training_details}
\begin{center}
\begin{threeparttable}
\begin{tabular}{ P{0.8cm} P{1.0cm} P{1.0cm} P{1.5cm} P{1.0cm} P{1.5cm} P{1.5cm} P{1.7cm} }
\toprule
Task &Optimizer & LR &LR decay &Weight decay &Warmup epochs &Cooldown epochs &Batch size\\
\midrule
\midrule
MTL &\textit{AdamW} &\textit{2e-5} &\textit{cosine}   &0.1 &5  &10 &126$^{\varoast}$ \\
Pain &\textit{AdamW} &\textit{2e-5} &\textit{cosine}  &0.1 &10 &10 &32\\
\bottomrule
\end{tabular}
\begin{tablenotes}
\scriptsize
\item MTL: multi-task learning for pre-training the foundation model \space Pain: pain assessment task \space LR: learning rate \space $\varoast$: batch size is proportionally distributed across the 14 tasks
\end{tablenotes}
\end{threeparttable}
\end{center}
\end{table}

\begin{figure}
\begin{center}
\includegraphics[scale=0.45]{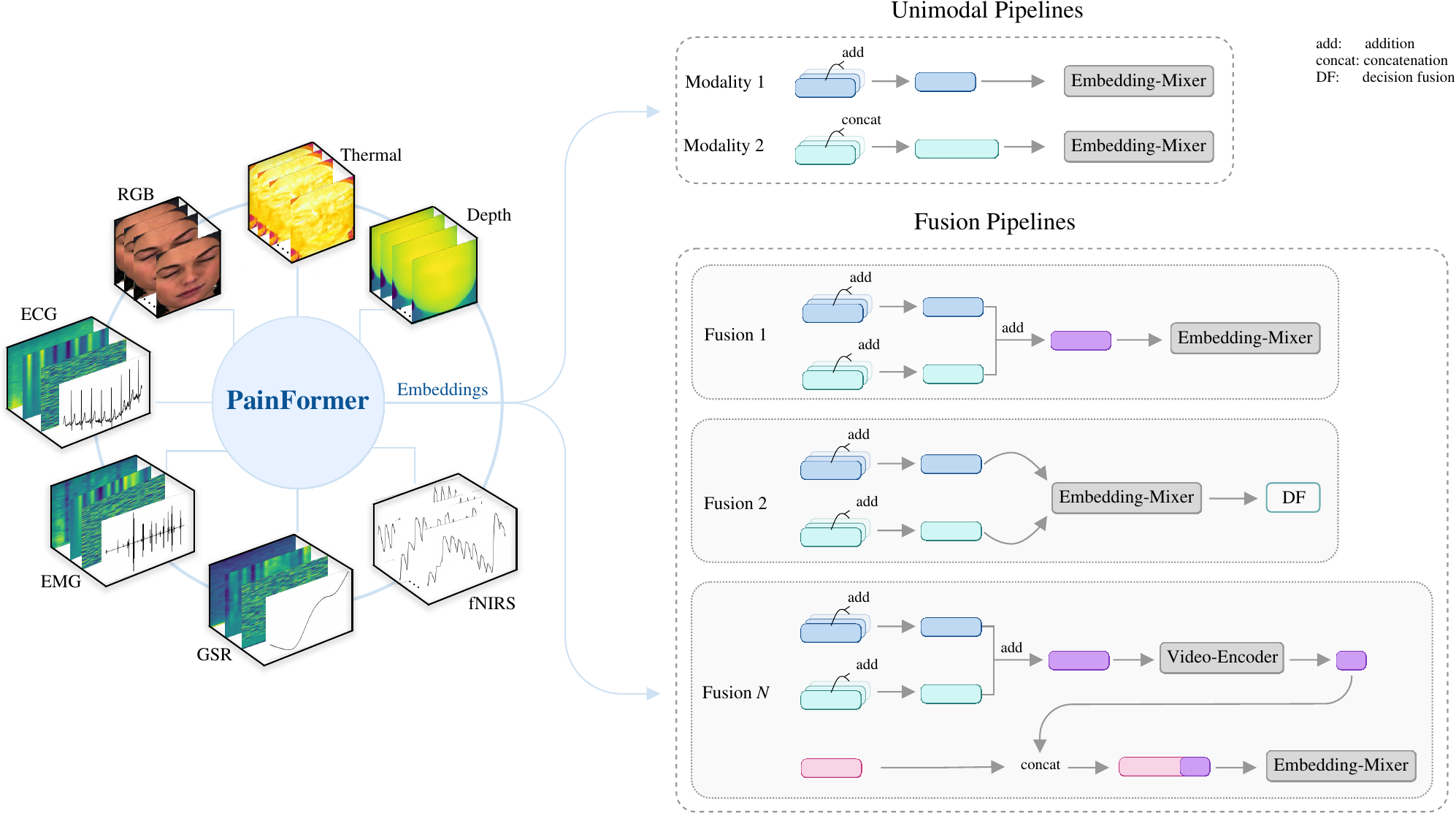} 
\end{center}
\caption{An overview of the presented framework. 
\textit{PainFormer}, the foundational model, excels in deriving high-quality embeddings from a diverse array of behavioral and physiological modalities.
The evaluation of RGB, thermal, and depth videos, alongside various representations of ECG, EMG, GSR, and fNIRS such as waveforms and spectrograms, underscores the rich information captured within these embeddings.
Leveraging the embeddings from \textit{PainFormer} facilitates the creation of various and diverse unimodal and multimodal pipelines designed for the pain assessment task.
Each pipeline can be customized to suit the specific modalities involved, dataset characteristics, and the demands of the intended application or clinical setting.
Our assessments included the development and implementation of several pipelines in both unimodal and multimodal contexts, achieving leading-edge results across various modalities and data representations.}
\label{2_pipeline}
\end{figure}

\subsubsection{Dataset Details}
To evaluate the effectiveness and resilience of our proposed framework, we performed tests on two specific pain datasets, \textit{BioVid} \cite{biovid_2013} and \textit{AI4Pain} \cite{ai4pain}. These datasets offer a varied and solid foundation for validating the performance of our model in pain assessment tasks.

\paragraph{BioVid Heat Pain Database:}
This dataset is recognized and well-established within the domain of pain research.
It encompasses facial videos, electrocardiograms, electromyograms, and galvanic skin response measurements from eighty-seven $(n\text{=}87)$ healthy participants ($44$ males and $43$ females, aged between $20$ and $65$). The experiment involved applying a thermode to the participants' right arm to induce pain. Before data collection, each participant's pain and tolerance thresholds were determined, defining the minimum and maximum levels of pain experienced. This setup included two additional intermediate levels, culminating in five distinct pain intensities: No Pain (NP), Mild Pain (P\textsubscript{1}), Moderate Pain (P\textsubscript{2}), Severe Pain (P\textsubscript{3}), and Very Severe Pain (P\textsubscript{4}). The temperature for inducing these pain levels ranged from P\textsubscript{1} to P\textsubscript{4} but did not exceed $50.5^\circ\text{C}$. Participants underwent $20$ inductions at each of the four specified intensity levels (P\textsubscript{1} to P\textsubscript{4}), with each stimulus lasting $4s$, followed by a recovery interval of $8$ to $12s$.
Additionally, $20$ baseline measurements at $32^\circ\text{C}$ (NP) were conducted, resulting in $100$ total stimulations per participant, administered randomly. After reaching the target temperature for each induction, the data was subsequently segmented into $5.5s$ intervals starting at $1s$. This segmentation generated $8,700$ samples, each $5.5s$ long, evenly distributed across the five pain intensity classes for each modality, encompassing all $87$ subjects.
The video recordings were captured at a frame rate of $25$ frames per second (FPS), while the biosignal (ECG, EMG, GSR) recordings were sampled at $512$ Hz.

\paragraph{AI4Pain Dataset:}
The AI4Pain Grand Challenge 2024 dataset is a recent addition tailored for advanced pain recognition tasks using fNIRS and facial video data. Sixty-five $(n\text{=}65)$ volunteers participated, including $23$ females, with ages ranging from $17$ to $52$ years.
The dataset additionally includes physiological signals like photoplethysmography (PPG), electrodermal activity (EDA), and respiration (RESP), though these are not currently publicly available. The dataset is segmented into three parts: training ($41$ volunteers), validation ($12$ volunteers), and testing ($12$ volunteers).
The data collection setup for this dataset involves comprehensive fNIRS and video recording to capture both brain activity and facial expressions. The fNIRS recordings were conducted using an \textit{Artinis} device (Artinis Medical Systems, Gelderland, the Netherlands), measuring fluctuations in oxygenated haemoglobin (HBO2) and deoxygenated haemoglobin (HHB) concentrations (in $\mu$mol/L). This fNIRS system uses $24$ channels to cover the prefrontal cortex with optodes ($10$ sources and $8$ detectors) placed $30$ mm apart. It emits near-infrared light at wavelengths of $760$ nm and $840$ nm and samples at a rate of $50$ Hz. The video data is recorded using a \textit{Logitech StreamCam} at a frame rate of $30$ FPS.
The \textit{AI4Pain} dataset categorizes pain into three levels: \emph{No Pain}, \emph{Low Pain}, and \emph{High Pain}. It includes $65$ instances of \emph{No Pain} (lasting $60$s each), $780$ instances of \emph{Low Pain} (lasting $10$s each), and $780$ instances of \emph{High Pain} (also lasting $10$s each). The \emph{No Pain} instances represent baseline data. In contrast, \emph{Low Pain} and \emph{High Pain} are derived from the pain tolerance tests, capturing subtle and significant changes in neurological and behavioral responses via fNIRS and video data.

\subsection{Experimental Evaluation \& Results}
This research devised various testing scenarios, including unimodal and multimodal settings, to assess the effectiveness of the proposed foundational model. The aim is to utilize a variety of behavioral and physiological modalities to ascertain the capability of \textit{PainFormer} to generate and provide high-quality embeddings for pain assessment. The experimental framework utilizes a comprehensive set of modalities, encompassing RGB, synthetic thermal imaging, depth videos, and physiological measurements such as ECG, EMG, GSR, and fNIRS, with waveform and spectrogram representations.
Additionally, specific pipelines were tailored to single modalities or their combinations based on each integration need.
This adaptability is a cornerstone of our approach, as different pipelines might be required depending on the specific demands, data availability, or intended application.
We aim to offer robust feature representations for any given input modality and excel in performance across all modalities and testing scenarios.
Figure \ref{2_pipeline} displays a high-level view of the proposed framework.
Note that all images, including video frames and biosignal visual representations, are standardized to a resolution of $224\times224$ pixels.

This research employed \textit{Part A} of the \textit{BioVid} dataset, focusing on pain assessment in binary terms, differentiating between \textit{No Pain} (NP) and \textit{Very Severe Pain} (P\textsubscript{4}). Validation was conducted using the leave-one-subject-out (LOSO) cross-validation technique.
In the case of \textit{AI4Pain}, a multilevel classification scheme was applied, categorizing pain into three levels: \textit{No Pain}, \textit{Low Pain}, and \textit{High Pain}. The challenge organizers' hold-out method (training, validation, testing) was utilized for validation.
For both datasets, the evaluation metrics included accuracy, recall (sensitivity), and F1 score.
It is also important to mention that all experiments used a deterministic approach, ensuring no influence from random initializations. This practice assures that any differences in performance observed are strictly attributable to specific optimization parameters, modalities, and other controlled variations rather than any random factors.

\subsubsection{BioVid}
Numerous experiments were performed using the \textit{BioVid} dataset.
Beyond the original RGB videos, synthetic thermal and depth videos were developed to provide additional visual contexts, as detailed in \ref{thermal_depth}. As specified in \ref{biosignal_visualization}, four distinct representations of ECGs, EMGs, and GSRs were assessed for biosignals. Combinations of these representations were also explored.

\paragraph{Video:} 
\label{biovid_video}
For behavioral modalities within the \textit{BioVid} dataset, \textit{PainFormer} generates an embedding of dimension $d=160$ for each video frame. These embeddings are concatenated to create a comprehensive representation of each video:
\begin{equation}
\mathcal{V}_D = [d_1 \| d_2 \| \cdots \|d_m], \quad D \in \mathbb{R}^{N_1},
\label{eq:videos_biovid}
\end{equation}
where $m$ represents the number of frames per video, and $N_1$ is the dimensionality of the total embedding, computed as $m \times d \rightarrow 138 \times 160 = 22,080$.
This unified embedding is fed into the \textit{Embedding-Mixer} for final pain assessment.
Starting with a training duration of $200$ epochs and using augmentation only on RGB videos, an accuracy of $71.83\%$ and a recall of $74.52\%$ were recorded. Thermal and depth videos achieved accuracies of $69.83\%$ and $69.00\%$, respectively.
When the training was extended to $300$ epochs with intensified augmentations and incorporating \textit{Label Smoothing} for regularization, RGB accuracy improved to $72.50\%$, although recall decreased slightly by $0.46\%$. Thermal modality performance decreased overall, highlighting its sensitivity to augmentations and regularization techniques. Conversely, depth modalities responded well to these changes, showing improved metrics with an accuracy of $70.08\%$, a recall of $71.27\%$, and an F1 score of $69.63\%$.
In the final experimental phase, training was extended to $600$ epochs, employing lighter augmentations at a $0.7$ probability, alongside $0.1$ \textit{Label Smoothing} and $0.5$ \textit{DropOut}. This regimen resulted in the highest performance for RGB videos, achieving an accuracy of $76.29\%$ and a recall of $77.56\%$. The F1 score also significantly increased, rising over $5\%$ to $75.56\%$.
Similar patterns were observed for the thermal and depth videos in this final experimental setup, albeit with minor improvements. Accuracy for thermal videos was $71.55\%$ and for depth videos $71.67\%$, with recall rates closely matching at $72.83\%$ and $72.84\%$, respectively. These findings demonstrate consistent enhancement across all visual modalities with refined training parameters and extended training durations.
Table \ref{table:ch7_p2_video} consolidates these experimental outcomes, indicating that the RGB modality consistently surpasses others, while the thermal and depth modalities show comparable performance levels. Moreover, although thermal and depth enhancements are modest, they suggest a plateau in potential performance increases.

\renewcommand{\arraystretch}{1.1}
\begin{table}
\center
\small
\caption{Results utilizing the video modality, NP vs. P\textsubscript{4} task, reported on accuracy, recall and F1 score.}
\label{table:ch7_p2_video}
\begin{center}
\begin{threeparttable}
\begin{tabular}{ P{0.8cm} P{1.0cm} P{1.0cm} P{1.5cm} P{1.0cm} P{1.5cm} P{0.7cm} P{0.7cm} P{0.7cm} }
\toprule
\multirow{2}[2]{*}{\shortstack{Input}}
&\multirow{2}[2]{*}{\shortstack{Epochs}}
&\multicolumn{2}{c}{Augmentation} 
&\multicolumn{2}{c}{Regularization} 
&\multicolumn{3}{c}{Metric}\\ 
\cmidrule(lr){3-4}\cmidrule(lr){5-6}\cmidrule(lr){7-9}
& &\textit{Basic} &\textit{Masking}  &\textit{LS} &\textit{DropOut} &Acc &Rec &F1 \\
\midrule
\midrule
\multirow{3}[1]{*}{\rotatebox{90}{RGB}}  
&200  &0.5 &0.5\textbar10-20\textbar &0.0 &0.0 &71.83 &74.52 &70.29 \\
&300  &0.7 &0.7\textbar15-20\textbar &0.1 &0.0 &72.50 &74.06 &70.93 \\
&600  &0.5 &0.5\textbar15-20\textbar &0.1 &0.5 &\textbf{76.29} &\textbf{77.56} &\textbf{75.56} \\
\Xhline{1.5\arrayrulewidth}
\multirow{3}[1]{*}{\rotatebox{90}{Thermal}}  
&200  &0.5 &0.5\textbar10-20\textbar &0.0 &0.0 &69.83 &71.51 &69.17 \\
&300  &0.7 &0.7\textbar15-20\textbar &0.1 &0.0 &68.83 &69.77 &68.41 \\
&600  &0.5 &0.5\textbar15-20\textbar &0.1 &0.5 &71.55 &72.83 &71.12 \\
\Xhline{1.5\arrayrulewidth}
\multirow{3}[1]{*}{\rotatebox{90}{Depth}}  
&200  &0.5 &0.5\textbar10-20\textbar &0.0 &0.0 &69.00 &69.44 &67.94 \\
&300  &0.7 &0.7\textbar15-20\textbar &0.1 &0.0 &70.08 &71.27 &69.63 \\
&600  &0.5 &0.5\textbar15-20\textbar &0.1 &0.5 &71.67 &72.84 &71.26 \\
\bottomrule 
\end{tabular}
\begin{tablenotes}
\scriptsize
\item \textit{LS: Label Smoothing} \space For Augmentation and Regularization, the number denotes the
probability of application, while in \textit{Masking}, the number in \textbar \space \textbar \space indicates the size of the mask applied.
\end{tablenotes}
\end{threeparttable}
\end{center}
\end{table}

\paragraph{ECG:} 
The training configuration used for the video data was similarly applied to the ECG signals. As previously indicated, four visual representations were utilized. Each representation corresponds to an image dimension of $224\times224$ pixels, from which embeddings of dimensionality $d=160$ are extracted and then inputted into the \textit{Embedding-Mixer}.
Starting with $200$ epochs and employing minimal augmentation without regularization, the \textit{waveform} representation reached an accuracy of $69.58\%$, with recall and F1 scores of $72.67\%$ and $68.10\%$, respectively. The \textit{spectrogram-angle} had lower performance in all metrics, achieving an accuracy of $65.58\%$. Meanwhile, the \textit{spectrogram-phase} showed better accuracy, surpassing the prior two by $0.5\%$ and $4.5\%$, respectively. The \textit{spectrogram-PSD} achieved the highest results, recording $71.08\%$ accuracy, $73.13\%$ recall, and $70.19\%$ F1 score.
Further improvements were seen in the $300$-epoch configuration across all visual representations and metrics. In the ultimate experimental setup extending to $600$ epochs, enhancements were noted universally, but the \textit{spectrogram-PSD} showed the most considerable gains, nearly $4\%$, achieving $75.49\%$ accuracy, $77.15\%$ recall, and $74.90\%$ F1 score. This indicates that integrating amplitude and frequency information, as the PSD representation provides, is particularly effective and valuable for analyzing ECG signals.
Table \ref{table:ch7_p2_ecg} documents the outcomes for the ECG modality.

\renewcommand{\arraystretch}{1.1}
\begin{table}
\center
\small
\caption{Results utilizing the ECG modality, NP vs. P\textsubscript{4} task, reported on accuracy, recall and F1 score.}
\label{table:ch7_p2_ecg}
\begin{center}
\begin{threeparttable}
\begin{tabular}{ P{0.8cm} P{1.0cm} P{1.0cm} P{1.5cm} P{1.0cm} P{1.5cm} P{0.7cm} P{0.7cm} P{0.7cm} }
\toprule
\multirow{2}[2]{*}{\shortstack{Input}}
&\multirow{2}[2]{*}{\shortstack{Epochs}}
&\multicolumn{2}{c}{Augmentation} 
&\multicolumn{2}{c}{Regularization} 
&\multicolumn{3}{c}{Metric}\\ 
\cmidrule(lr){3-4}\cmidrule(lr){5-6}\cmidrule(lr){7-9}
& &\textit{Basic} &\textit{Masking}  &\textit{LS} &\textit{DropOut} &Acc &Rec &F1 \\
\midrule
\midrule
\multirow{3}[1]{*}{\rotatebox{90}{Wave}}  
&200  &0.5 &0.5\textbar10-20\textbar &0.0 &0.0 &69.58 &72.67 &68.10 \\
&300  &0.7 &0.7\textbar15-20\textbar &0.1 &0.0 &71.08 &72.74 &70.22 \\
&600  &0.5 &0.5\textbar15-20\textbar &0.1 &0.5 &73.36 &74.75 &72.52 \\
\Xhline{1.5\arrayrulewidth}
\multirow{3}[1]{*}{\rotatebox{90}{Angle}}  
&200  &0.5 &0.5\textbar10-20\textbar &0.0 &0.0 &65.58 &66.68 &64.89 \\
&300  &0.7 &0.7\textbar15-20\textbar &0.1 &0.0 &66.33 &68.22 &65.22 \\
&600  &0.5 &0.5\textbar15-20\textbar &0.1 &0.5 &68.25 &71.24 &66.99 \\
\Xhline{1.5\arrayrulewidth}
\multirow{3}[1]{*}{\rotatebox{90}{Phase}}  
&200  &0.5 &0.5\textbar10-20\textbar &0.0 &0.0 &70.08 &71.54 &69.40 \\
&300  &0.7 &0.7\textbar15-20\textbar &0.1 &0.0 &72.33 &73.73 &71.69 \\
&600  &0.5 &0.5\textbar15-20\textbar &0.1 &0.5 &72.70 &74.19 &72.14 \\
\Xhline{1.5\arrayrulewidth}
\multirow{3}[1]{*}{\rotatebox{90}{PSD}}  
&200  &0.5 &0.5\textbar10-20\textbar &0.0 &0.0 &71.08 &73.13 &70.19 \\
&300  &0.7 &0.7\textbar15-20\textbar &0.1 &0.0 &71.50 &73.14 &70.18 \\
&600  &0.5 &0.5\textbar15-20\textbar &0.1 &0.5 &\textbf{75.49} &\textbf{77.15} &\textbf{74.90} \\
\bottomrule 
\end{tabular}
\end{threeparttable}
\end{center}
\end{table}

\paragraph{EMG:} 
For EMG signals, the initial training configuration of $200$ epochs demonstrated comparable accuracy across the \textit{waveform}, \textit{spectrogram-phase}, and \textit{spectrogram-PSD} representations, recording scores of $68.75\%$, $68.33\%$, and $69.25\%$ respectively. However, the \textit{spectrogram-angle} representation underperformed with an accuracy of $66.42\%$, mirroring its lower performance in the ECG modality.
In subsequent training sessions with increased epochs and enhanced augmentation and regularization, the \textit{spectrogram-angle} representation showed a notable decline in performance across all metrics. Despite some marginal improvements in the $300$-epoch configuration, it still trailed behind its initial results, posting an accuracy of $65.32\%$, with recall and F1 scores of $68.15\%$ and $63.17\%$, respectively. This pattern suggests that the angle representation, which lacks phase unwrapping, is less effective for pain assessment tasks in EMG signals.
Conversely, the other visual representations demonstrated consistent improvements in each training configuration. The \textit{spectrogram-PSD} achieved the highest accuracy at $72.10\%$ and an F1 score of $71.82\%$. The \textit{waveform} representation obtained the highest recall at $73.64\%$. 
These results are presented in Table \ref{table:ch7_p2_emg} for the EMG modality.

\renewcommand{\arraystretch}{1.0}
\begin{table}
\small
\center
\caption{Results utilizing the EMG modality, NP vs. P\textsubscript{4} task, reported on accuracy, recall and F1 score.}
\label{table:ch7_p2_emg}
\begin{center}
\begin{threeparttable}
\begin{tabular}{ P{0.8cm} P{1.0cm} P{1.0cm} P{1.5cm} P{1.0cm} P{1.5cm} P{0.7cm} P{0.7cm} P{0.7cm} }
\toprule
\multirow{2}[2]{*}{\shortstack{Input}}
&\multirow{2}[2]{*}{\shortstack{Epochs}}
&\multicolumn{2}{c}{Augmentation} 
&\multicolumn{2}{c}{Regularization} 
&\multicolumn{3}{c}{Metric}\\ 
\cmidrule(lr){3-4}\cmidrule(lr){5-6}\cmidrule(lr){7-9}
& &\textit{Basic} &\textit{Masking}  &\textit{LS} &\textit{DropOut} &Acc &Rec &F1 \\
\midrule
\midrule
\multirow{3}[1]{*}{\rotatebox{90}{Wave}}  
&200  &0.5 &0.5\textbar10-20\textbar &0.0 &0.0 &68.75 &70.55 &67.93 \\
&300  &0.7 &0.7\textbar15-20\textbar &0.1 &0.0 &69.83 &72.52 &68.68 \\
&600  &0.5 &0.5\textbar15-20\textbar &0.1 &0.5 &72.07 &\textbf{73.64} &71.48 \\
\Xhline{1.5\arrayrulewidth}
\multirow{3}[1]{*}{\rotatebox{90}{Angle}}  
&200  &0.5 &0.5\textbar10-20\textbar &0.0 &0.0 &66.42 &68.57 &65.26 \\
&300  &0.7 &0.7\textbar15-20\textbar &0.1 &0.0 &63.92 &66.33 &62.67 \\
&600  &0.5 &0.5\textbar15-20\textbar &0.1 &0.5 &65.32 &68.15 &63.77 \\
\Xhline{1.5\arrayrulewidth}
\multirow{3}[1]{*}{\rotatebox{90}{Phase}}  
&200  &0.5 &0.5\textbar10-20\textbar &0.0 &0.0 &68.33 &69.75 &67.68 \\
&300  &0.7 &0.7\textbar15-20\textbar &0.1 &0.0 &68.58 &70.00 &67.97 \\
&600  &0.5 &0.5\textbar15-20\textbar &0.1 &0.5 &69.37 &71.17 &68.66 \\
\Xhline{1.5\arrayrulewidth}
\multirow{3}[1]{*}{\rotatebox{90}{PSD}}  
&200  &0.5 &0.5\textbar10-20\textbar &0.0 &0.0 &69.25 &70.38 &68.84 \\
&300  &0.7 &0.7\textbar15-20\textbar &0.1 &0.0 &69.67 &71.06 &69.12 \\
&600  &0.5 &0.5\textbar15-20\textbar &0.1 &0.5 &\textbf{72.10} &72.82 &\textbf{71.82} \\
\bottomrule 
\end{tabular}
\end{threeparttable}
\end{center}
\end{table}

\paragraph{GSR:} 
\label{gsr}
For the GSR modality, distinct performance variations among the four representations are evident. The \textit{waveform}-based representations significantly outshine the others, starting with an initial accuracy of $87.75\%$ in the $200$-epoch configuration, which is over $14\%$ higher than other metrics.
With extended training sessions, a modest improvement is noted across all representations, indicating that the GSR modality might have reached its maximum potential performance. Among the spectrograms, the \textit{spectrogram-phase} proves to be the most informative, culminating in final accuracy, recall, and F1 scores of $76.41\%$, $77.23\%$, and $76.47\%$, respectively.
The \textit{waveform} representation emerges as the most effective, achieving the highest metrics with an accuracy of $88.99\%$, recall of $89.55\%$, and an F1 score of $88.88\%$. The distinct performance of these representations can be related to the inherent characteristics of the GSR signal. As depicted in Fig. \ref{2_pipeline}, GSR typically presented as a smooth curve with gradual slopes, indicative of slow and steady changes in skin conductivity due to variations in sweat gland activity triggered by stress or arousal.
In comparison, EMG signals are marked by sharp spikes and erratic fluctuations, reflecting rapid electrical activities from skeletal muscle contractions. On the other hand, ECG signals display distinct cyclical patterns, including the P and T waves and the QRS complex. These observations imply that the simpler patterns in GSR are not as well suited for spectral and frequency domain analyses, which are more effectively captured by spectrograms. However, \textit{waveform} representations excel in capturing critical physiological data from GSR signals, outperforming all other modalities and visual representations due to their ability to effectively represent the essential dynamics of GSR activity.
The results for the GSR modality are summarized in Table  \ref{table:ch7_p2_gsr}.

\renewcommand{\arraystretch}{1.0}
\begin{table}
\center
\small
\caption{Results utilizing the GSR modality, NP vs. P\textsubscript{4} task, reported on accuracy, recall and F1 score.}
\label{table:ch7_p2_gsr}
\begin{center}
\begin{threeparttable}
\begin{tabular}{ P{0.8cm} P{1.0cm} P{1.0cm} P{1.5cm} P{1.0cm} P{1.5cm} P{0.7cm} P{0.7cm} P{0.7cm} }
\toprule
\multirow{2}[2]{*}{\shortstack{Input}}
&\multirow{2}[2]{*}{\shortstack{Epochs}}
&\multicolumn{2}{c}{Augmentation} 
&\multicolumn{2}{c}{Regularization} 
&\multicolumn{3}{c}{Metric}\\ 
\cmidrule(lr){3-4}\cmidrule(lr){5-6}\cmidrule(lr){7-9}
& &\textit{Basic} &\textit{Masking}  &\textit{LS} &\textit{DropOut} &Acc &Rec &F1 \\
\midrule
\midrule
\multirow{3}[1]{*}{\rotatebox{90}{Wave}}  
&200  &0.5 &0.5\textbar10-20\textbar &0.0 &0.0 &87.75 &88.68 &87.56 \\
&300  &0.7 &0.7\textbar15-20\textbar &0.1 &0.0 &88.50 &89.16 &88.34 \\
&600  &0.5 &0.5\textbar15-20\textbar &0.1 &0.5 &\textbf{88.99} &\textbf{89.55} &\textbf{88.88} \\
\Xhline{1.5\arrayrulewidth}
\multirow{3}[1]{*}{\rotatebox{90}{Angle}}  
&200  &0.5 &0.5\textbar10-20\textbar &0.0 &0.0 &73.67 &75.00 &73.26 \\
&300  &0.7 &0.7\textbar15-20\textbar &0.1 &0.0 &73.08 &74.60 &72.66 \\
&600  &0.5 &0.5\textbar15-20\textbar &0.1 &0.5 &73.24 &75.02 &72.83 \\
\Xhline{1.5\arrayrulewidth}
\multirow{3}[1]{*}{\rotatebox{90}{Phase}}  
&200  &0.5 &0.5\textbar10-20\textbar &0.0 &0.0 &75.17 &76.13 &74.79 \\
&300  &0.7 &0.7\textbar15-20\textbar &0.1 &0.0 &75.92 &76.60 &75.57 \\
&600  &0.5 &0.5\textbar15-20\textbar &0.1 &0.5 &76.41 &77.23 &76.47 \\
\Xhline{1.5\arrayrulewidth}
\multirow{3}[1]{*}{\rotatebox{90}{PSD}}  
&200  &0.5 &0.5\textbar10-20\textbar &0.0 &0.0 &72.83 &73.91 &72.34 \\
&300  &0.7 &0.7\textbar15-20\textbar &0.1 &0.0 &73.08 &73.96 &72.68 \\
&600  &0.5 &0.5\textbar15-20\textbar &0.1 &0.5 &73.96 &74.81 &73.50 \\

\bottomrule 
\end{tabular}
\end{threeparttable}
\end{center}
\end{table}

\paragraph{Fusion:} 
Various fusion techniques were tested to evaluate whether combining different representations or modalities could enhance performance. In this research, using inputs from the same sensor type, such as RGB with depth-estimation videos or ECG \textit{waveforms} with ECG \textit{spectrogram-PSD}, was considered an unimodal fusion approach. On the other hand, combining inputs from different sensor types, like GSR with EMG, was treated as a multimodal fusion.
Three primary methods of fusion were explored: feature fusion and decision fusion. Feature fusion includes strategies such as addition, where embeddings from various inputs are summed before progressing to the following processing stage, and concatenation, which aligns them along the \textit{y-axis}.
Decision fusion, meanwhile, involves processing each embedding through the \textit{Embedding-Mixer}, which then aggregates the predictions from each input to generate a final decision.
All related experiments were conducted under the previously detailed $600$-epoch training configuration, with results compiled in Table \ref{table:ch7_p2_fusion}.

In video modality fusion, we assessed combinations of RGB with thermal, RGB with depth, and thermal with depth, plus a three-input amalgamation of RGB, thermal, and depth.
The RGB and thermal blend underperformed compared to RGB alone, with the best performance ($75.66\%$ accuracy) achieved through decision fusion. The RGB and depth combination similarly yielded optimal results through decision fusion, achieving $75.53\%$ accuracy but falling short of RGB-only performance. Notably, merging thermal and depth videos improved upon using depth alone, particularly via decision fusion, which attained a $73.02\%$ accuracy rate.
The combination of RGB, thermal, and depth inputs was the sole group that outperformed the standalone RGB setup, with decision fusion delivering the highest metrics: $76.55\%$ accuracy, $77.91\%$ recall, and $76.11\%$ F1 score, indicating marginal improvements across all measures. Decision fusion consistently outperformed the addition method in all video-based experiments.

For biosignals, experiments focused on ECG and EMG using the \textit{waveform} and the representations of \textit{spectrogram-PSD}. No fusion experiments were conducted for GSR due to the waveform's dominance in performance. For ECG, all fusion methods were less effective than the \textit{spectrogram-PSD} alone, except for the addition method, which slightly improved recall by $0.21\%$. EMG results were enhanced by all fusion techniques, with concatenation proving to be the most beneficial, leading to increases in accuracy, recall, and F1 score by $0.74\%$, $0.36\%$, and $0.64\%$, respectively.

Physiological and behavioral modalities were integrated into our multimodal setup, combining GSR signals with RGB, synthetic thermal, and estimated depth videos. The GSR's \textit{waveform} representation and video features, described in \ref{eq:videos_biovid}, were merged into a unified vector of dimension $22,080$. This vector was then processed through the \textit{Video-Encoder} into a smaller space of $40$. The resulting combined vector of $160+40=200$ dimensions was formed by concatenating the GSR and video embeddings, represented as:
\begin{equation}
\mathcal{M}_h = \mathcal{G}_d \| Enc\big[(\mathcal{V}^{\tiny{\text{RGB}}}_D + \mathcal{V}^{\tiny{\text{Thermal}}}_D + 
\mathcal{V}^{\tiny{\text{Depth}}}_D)\big],  \quad h \in \mathbb{R}^{N_2},
\end{equation}
where $\mathcal{G}$ denotes the GSR embedding and $\mathcal{M}$ the fused vector with $N_2$ equal to $200$. This approach, visualized in Fig. \ref{2_pipeline} (bottom right), achieved the highest performance in the study, with accuracy, recall, and F1 scores of $89.08\%$, $89.88\%$, and $88.87\%$, respectively. This method slightly surpassed the performance of GSR used independently, especially in accuracy and recall.

\renewcommand{\arraystretch}{1.1}
\begin{table}
\scriptsize
\center
\caption{Results on fusion settings$^*$, NP vs. P\textsubscript{4} task, reported on accuracy, recall and F1 score.}
\label{table:ch7_p2_fusion}
\begin{center}
\begin{threeparttable}
\begin{tabular}{ P{1.5cm} P{3.0cm} P{1.0cm} P{0.9cm} P{0.9cm} P{0.9cm} }
\toprule
\multirow{2}[2]{*}{\shortstack{Modality}}
& \multirow{2}[2]{*}{\shortstack{Input}} 
& \multirow{2}[2]{*}{\shortstack{Fusion}} 
&\multicolumn{3}{c}{Metric}\\ 
\cmidrule(lr){4-6}
& & &Acc &Rec &F1 \\
\midrule
\midrule
\multirow{15}[1]{*}{Video}
& \multirow{3}{*}{\raisebox{-1.5\height}{RGB, Thermal}} & Add & 75.09 \textcolor{myred_2}{\scriptsize \Minus 1.20} & 76.97 \textcolor{myred_2}{\scriptsize \Minus 0.59} & 73.98 \textcolor{myred_2}{\scriptsize \Minus 1.58} \\

& & DF & 75.66 \textcolor{myred_2}{\scriptsize \Minus 0.63} & 77.23 \textcolor{myred_2}{\scriptsize \Minus 0.33} & 75.08 \textcolor{myred_2}{\scriptsize \Minus 0.48} \\
\cdashline{2-6}[.5pt/1pt]

& \multirow{3}{*}{\raisebox{-1.5\height}{RGB, Depth}}          &Add &74.93 \textcolor{myred_2}{{\scriptsize \Minus 1.36}} &76.41 \textcolor{myred_2}{{\scriptsize \Minus 1.15}} &73.38 \textcolor{myred_2}{{\scriptsize \Minus 2.18}}\\

&  &DF &75.53 \textcolor{myred_2}{{\scriptsize \Minus 0.76}} &77.18 \textcolor{myred_2}{{\scriptsize \Minus 0.38}} &75.00 \textcolor{myred_2}{{\scriptsize \Minus 0.56}}\\
\cdashline{2-6}[.5pt/1pt]

& \multirow{3}{*}{\raisebox{-1.5\height}{Thermal, Depth}}      &Add &71.44 \textcolor{myred_2}{{\scriptsize \Minus 0.23}} &73.15 \textcolor{mygreen}{{\scriptsize \Plus 0.31}} &70.73 \textcolor{myred_2}{{\scriptsize \Minus 0.50}}\\

& &DF &73.02 \textcolor{mygreen}{{\scriptsize \Plus 1.35}} &74.46 \textcolor{mygreen}{{\scriptsize \Plus 1.62}} &72.59 \textcolor{mygreen}{{\scriptsize \Plus 1.33}}\\
\cdashline{2-6}[.5pt/1pt]

& \multirow{3}{*}{\raisebox{-1.5\height}{RGB, Thermal, Depth}} &Add &76.26 \textcolor{myred_2}{{\scriptsize \Minus 0.03}} &77.70 \textcolor{mygreen}{{\scriptsize \Plus 0.14}} &75.78 \textcolor{mygreen}{{\scriptsize \Plus 0.22}}\\

&  &DF &76.55 \textcolor{mygreen}{{\scriptsize \Plus 0.26}} &77.91 \textcolor{mygreen}{{\scriptsize \Plus 0.35}} &76.11 \textcolor{mygreen}{{\scriptsize \Plus 0.55}}\\
\Xhline{1.5\arrayrulewidth}

\multirow{3}{*}{\raisebox{-1.5\height}{ECG}}

& \multirow{3}{*}{\raisebox{-1.5\height}{Wave, PSD}} &Add &75.43 \textcolor{myred_2}{{\scriptsize \Minus 0.06}} &77.36 \textcolor{mygreen}{{\scriptsize \Plus 0.21}} &74.75 \textcolor{myred_2}{{\scriptsize \Minus 0.15}}\\

& &Concat &74.74 \textcolor{myred_2}{{\scriptsize \Minus 0.75}} &76.77 \textcolor{myred_2}{{\scriptsize \Minus 0.38}} &74.00 \textcolor{myred_2}{{\scriptsize \Minus 0.90}}\\

\Xhline{1.5\arrayrulewidth}

\multirow{3}{*}{\raisebox{-1.5\height}{EMG}}

& \multirow{3}{*}{\raisebox{-1.5\height}{Wave, PSD}} &Add &72.79 \textcolor{mygreen}{{\scriptsize \Plus 0.69}} &74.15 \textcolor{mygreen}{{\scriptsize \Plus 0.51}} &72.28 \textcolor{mygreen}{{\scriptsize \Plus 0.46}}\\

&  &Concat &72.84 \textcolor{mygreen}{{\scriptsize \Plus 0.74}} &74.00 \textcolor{mygreen}{{\scriptsize \Plus 0.36}} &72.46 \textcolor{mygreen}{{\scriptsize \Plus 0.64}}\\

\Xhline{1.5\arrayrulewidth}
\multirow{2}[1]{*}{Video, GSR}
&RGB, Thermal, Depth, Wave &Add \& Concat &89.08 \textcolor{mygreen}{{\scriptsize \Plus 0.09}} &89.88 \textcolor{mygreen}{{\scriptsize \Plus 0.33}} &88.87 \textcolor{myred_2}{{\scriptsize \Minus 0.01}}\\

\bottomrule 
\end{tabular}
\begin{tablenotes}
\scriptsize
\item $\ast$: All experiments follow the augmentation and regularization settings for the 600 epoch configuration outlined in the unimodal experiments. \textcolor{mygreen}{+} and \textcolor{myred_2}{-} indicate an increase or decrease in performance, respectively, compared to the best unimodal input approach. DF: Decision Fusion \space  Add: Addition \space Concat: Concatenation
\end{tablenotes}
\end{threeparttable}
\end{center}
\end{table}

\subsubsection{AI4Pain}
In the \textit{AI4Pain} dataset, experiments were conducted utilizing both unimodal and multimodal approaches. The original RGB videos were employed for the behavioral modality, while waveforms from the fNIRS's HBO2 channels were used for the physiological modality. It is important to note that out of the $24$ available HBO2 channels, $2$ were excluded due to malfunctions.
Table \ref{table:ch7_p2_ai4pain_validation} presents the corresponding results.

\paragraph{Video:}
Similar to \ref{biovid_video}, an embedding of dimension $d=160$ is extracted for every frame in the \textit{AI4Pain} dataset. However, in this instance, the extracted embeddings are aggregated into a unified vector:
\begin{equation}
\mathcal{V}_d = [d_1 + d_2 + \cdots +d_m], \quad d \in \mathbb{R}^{N_3},
\label{eq:videos_ai4pain}
\end{equation}
where $m$ represents the number of frames in a video, and $N_3$ is the dimensionality of the unified embedding, set at $160$.
After processing the embedding through the \textit{Embedding-Mixer} and employing the same $600$-epoch training configuration as used in prior experiments, this setup achieved an accuracy of $49.77\%$, with recall and F1 scores of $50.11\%$ and $49.77\%$, respectively. Increasing the \textit{DropOut} rate to $0.3$ improved the accuracy and F1 scores to $51.39\%$ and $51.31\%$. Further elevating the \textit{DropOut} rate to $0.8$ enhanced the recall to $52.74\%$.

\paragraph{fNIRS:}
For the fNIRS modality, embeddings were aggregated across the $22$ HBO2 channels to produce a feature representation of $\mathcal{O}_d=160$. The 600-epoch training setup initially yielded $43.06\%$ accuracy, $42.80\%$ recall, and $42.07\%$ F1 score. By increasing the \textit{DropOut} rate to $0.3$, peak performance metrics of $44.44\%$ accuracy, $45.55\%$ recall, and $43.74\%$ F1 score were achieved.

\paragraph{Fusion:}
For the fusion of video and fNIRS data, the following aggregation approach was utilized:
\begin{equation}
\mathcal{F}_d = \mathcal{V}_d + \mathcal{O}_d, \quad d \in \mathbb{R}^{N_3},
\label{eq:fusion_ai4pain}
\end{equation}
where $\mathcal{F}_d$ represents the combined feature representation. Starting with the same $600$-epoch training configuration, the initial results were $50.00\%$ accuracy, $51.01\%$ recall, and $48.54\%$ F1 score. Increasing the \textit{DropOut} rate to $0.8$ slightly improved the accuracy and F1 score by $0.23\%$ and $1.7\%$, respectively, though recall decreased by $0.75\%$. The optimal \textit{DropOut} setting of $0.6$ achieved peak performances of $51.85\%$ accuracy, $51.57\%$ recall, and $51.35\%$ F1 score.

\renewcommand{\arraystretch}{1.0}
\begin{table}
\small
\center
\caption{Results on the validation set of \textit{AI4Pain} dataset, multilevel classification task, reported on accuracy, recall and F1 score.}
\label{table:ch7_p2_ai4pain_validation}
\begin{center}
\begin{threeparttable}
\begin{tabular}{ P{0.8cm} P{1.0cm} P{1.0cm} P{1.5cm} P{1.0cm} P{1.5cm} P{0.7cm} P{0.7cm} P{0.7cm} }
\toprule
\multirow{2}[2]{*}{\shortstack{Input}}
&\multirow{2}[2]{*}{\shortstack{Epochs}}
&\multicolumn{2}{c}{Augmentation} 
&\multicolumn{2}{c}{Regularization} 
&\multicolumn{3}{c}{Metric}\\ 
\cmidrule(lr){3-4}\cmidrule(lr){5-6}\cmidrule(lr){7-9}
& &\textit{Basic} &\textit{Masking}  &\textit{LS} &\textit{DropOut} &Acc &Rec &F1 \\
\midrule
\midrule
\multirow{3}[1]{*}{\rotatebox{90}{Video}}  
&600  &0.5 &0.5\textbar15-20\textbar &0.1 &0.5 &49.77 &50.11 &49.77\\
&600  &0.5 &0.5\textbar15-20\textbar &0.1 &0.3 &\textbf{51.39} &51.50 &\textbf{51.31} \\
&600  &0.5 &0.5\textbar15-20\textbar &0.1 &0.8 &48.38 &\textbf{52.74} &46.69\\
\Xhline{1.5\arrayrulewidth}
\multirow{3}[1]{*}{\rotatebox{90}{fNIRS}}  
&600  &0.5 &0.5\textbar15-20\textbar &0.1 &0.5 &43.06 &42.80 &42.07 \\
&600  &0.5 &0.5\textbar15-20\textbar &0.1 &0.3 &\textbf{44.44} &\textbf{45.55} &\textbf{43.74} \\
&600  &0.4 &0.4\textbar15-20\textbar &0.1 &0.1 &43.06 &44.18 &42.44 \\
\Xhline{1.5\arrayrulewidth}
\multirow{3}[1]{*}{\rotatebox{90}{Fusion}}  
&600  &0.5 &0.5\textbar15-20\textbar &0.1 &0.5 &50.00 &51.01 &48.54 \\
&600  &0.1 &0.1\textbar15-20\textbar &0.1 &0.8 &50.23 &50.25 &50.24 \\
&600  &0.4 &0.4\textbar15-20\textbar &0.1 &0.6 &\textbf{51.85} &\textbf{51.87} &\textbf{51.35} \\
\bottomrule 
\end{tabular}
\begin{tablenotes}
\scriptsize
\item Fusion: the Addition method of the modalities applied
\end{tablenotes}
\end{threeparttable}
\end{center}
\end{table}

\subsection{Comparison with existing methods}
To evaluate \textit{PainFormer}, we compared it against studies from the literature that utilized the \textit{BioVid} dataset (\textit{Part A}), included all available subjects ($87$), conducted the same task, adhered to the leave-one-subject-out (LOSO) validation protocol, and reported accuracy metrics. For the \textit{AI4Pain} dataset, our comparisons were made with studies that strictly followed the evaluation guidelines outlined in the corresponding challenge.

In \textit{BioVid}, the proposed approach using RGB, thermal, and depth video inputs is among the top performers in video-based studies, achieving an accuracy of $76.55\%$. This surpasses all methods utilizing hand-crafted features, as referenced in \cite{werner_hamadi_niese_2014,werner_hamadi_walter_2017,patania_2022} and outperforms most deep learning-based methods cited in \cite{zhi_wan_2019,thiam_kestler_schenker_2020,tavakolian_bordallo_liu_2020,gkikas_tsiknakis_embc}. Exceptions include results from \cite{gkikas_tachos_2024} at $77.10\%$ and \cite{yang_guan_yu_2021} at $78.90\%$, with the study in \cite{huang_dong_2022} achieving $77.50\%$ using a 3D CNN approach, which, when combined with pseudo heart rate data extracted from videos, reached the highest reported result of $88.10\%$. These results are documented in Table \ref{table:video_based}.

Regarding biosignals, in ECG-based studies, \textit{PainFormer} achieved the highest accuracy in the literature at $75.49\%$ using the \textit{spectrogram-PSD} representation, significantly outperforming subsequent studies \cite{gkikas_chatzaki_2023,gkikas_tachos_2024} by over $6\%$ and $8\%$, respectively. In EMG-based studies utilizing \textit{waveform} and \textit{spectrogram-PSD} representations, we achieved a $72.84\%$ accuracy, significantly exceeding the nearest study \cite{werner_hamadi_niese_2014} at $63.10\%$. For GSR-based studies, utilizing solely \textit{waveform} representation led to the highest performance with an $88.99\%$ accuracy. Studies using raw biosignals instead of extracting domain-specific features generally exhibited better results, with the second \cite{lu_ozek_kamarthi_2023} and third \cite{phan_iyortsuun_2023} ranked studies achieving $85.56\%$ and $84.80\%$ accuracy, respectively. Table \ref{table:biosignal_based} presents these biosignal results.

In multimodal scenarios, our approach combining video inputs and GSR achieved the highest reported accuracy of $89.08\%$ (refer to Table \ref{table:multimodal_based}). Notably, with one exception \cite{gkikas_tachos_2024}, all studies incorporated the GSR signal. GSR is consistently recognized as the most effective modality for pain assessment, with the second-highest-performing study \cite{jiang_li_he_2024} using a GSR and ECG combination achieving $87.06\%$. A study including videos, ECG, EMG, and GSR \cite{zhi_yu_2019} reached $86.00\%$ accuracy.

For the \textit{AI4Pain} dataset, \textit{PainFormer} achieved a $53.67\%$ accuracy using the RGB video modality, outperforming \cite{prajod_schiller_2024} at $49.00\%$ but falling behind \cite{nguyen_yang_2024} at $55.00\%$ utilizing a transformer-based masked autoencoder. Using only fNIRS, an accuracy of $52.60\%$ was achieved. In a multimodal approach combining videos and waveform representations, an accuracy of $55.69\%$ was attained, surpassing \cite{gkikas_tsiknakis_painvit_2024} by over $9\%$ and establishing the highest performance on this dataset to date.

\renewcommand{\arraystretch}{1.1}
\begin{table}
\center
\scriptsize
\caption{Comparison of video-based studies utilizing \textit{BioVid (Part-A)}, NP vs. P\textsubscript{4} task and LOSO validation.}
\label{table:video_based}
\begin{center}
\begin{threeparttable}
\begin{tabular}{ P{1.0cm} P{3.5cm} P{2.3cm} P{0.6cm} }
\toprule
\multirow{2}[2]{*}{\shortstack{Study}}
&\multicolumn{2}{c}{Method} 
&\multirow{2}[2]{*}{\shortstack{Acc\%}}\\
\cmidrule(lr){2-3}
&Features &ML &\\

\midrule
\midrule

\cite{zhi_wan_2019}  &raw &SLSTM &61.70\\ \hdashline
\cite{thiam_kestler_schenker_2020} &raw &2D CNN, biLSTM  &69.25\\ \hdashline
\cite{werner_hamadi_walter_2017} &optical flow &RF &70.20\\ \hdashline
\cite{tavakolian_bordallo_liu_2020} &raw &2D CNN &71.00\\ \hdashline
\cite{gkikas_tsiknakis_thermal_2024} &raw$^{\varocircle}$ &Vision-MLP &71.03\\ \hdashline
\cite{huang_xia_li_2019}$^\dagger$  &raw &2D CNN &71.30\\ \hdashline
\cite{werner_hamadi_2014} &facial landmarks,  3D distances &RF &71.60\\ \hdashline
\cite{werner_2016} &facial 3D distances &Deep RF &72.10\\ \hdashline
\cite{werner_2016} &facial action descriptors &Deep RF &72.40\\ \hdashline 
\cite{kachele_werner_2015} &facial landmarks, 3D distances &RF &72.70\\ \hdashline
\cite{patania_2022}  &fiducial points &GNN &73.20\\ \hdashline
\cite{gkikas_tsiknakis_embc} &raw &Transformer &73.28\\ \hdashline
\cite{huang_xia_2020}$^\dagger$  &raw &2D CNN, GRU &73.90\\ \hdashline
\cite{werner_hamadi_niese_2014} &facial landmarks, head pose &RF &76.60\\ \hdashline
\cite{gkikas_tachos_2024} &raw &Transformer &77.10\\ \hdashline
\cite{huang_dong_2022}  &raw &3D CNN, &77.50\\ \hdashline
\cite{yang_guan_yu_2021} &raw, rPPG\textsuperscript{\ding{70}} &3D CNN &78.90\\ \hdashline
\cite{huang_dong_2022}  &raw,  heart rate\textsuperscript{\ding{72}} &3D CNN &\textbf{88.10}\\ \hdashline
\rowcolor{mygray}Our &raw\textsuperscript{\ding{95}} &Transformer &76.55\\

\bottomrule 
\end{tabular}
\begin{tablenotes}
\scriptsize
\item $\dagger$: reimplemented for pain intensity estimation on \textit{BioVid} by \cite{huang_dong_2022} \space 
$\varocircle$: RGB, synthetic thermal videos \space \ding{70}: remote  photo plethysmography (estimated from videos) \space \ding{72}: pseudo heart rate gain (estimated from videos) \space \ding{95}: RGB-thermal-depth (DF)  \space RF: Random Forest\space GNN: Graph Neural Networks \space MLP: Multi-Layer Perceptron
\end{tablenotes}
\end{threeparttable}
\end{center}
\end{table}

\renewcommand{\arraystretch}{1.1}
\begin{table}
\center
\scriptsize
\caption{Comparison of biosignal-based studies utilizing \textit{BioVid (Part-A)}, NP vs. P\textsubscript{4} task and LOSO validation.}
\label{table:biosignal_based}
\begin{center}
\begin{threeparttable}
\begin{tabular}{ P{1.0cm} P{1.0cm} P{2.2cm} P{2.2cm} P{0.7cm} }
\toprule
\multirow{2}[2]{*}{\shortstack{Study}}
&\multirow{2}[2]{*}{\shortstack{Modality}}
&\multicolumn{2}{c}{Method} 
&\multirow{2}[2]{*}{\shortstack{Acc\%}}\\
\cmidrule(lr){3-4}
&&Features &ML &\\

\midrule
\midrule

\cite{thiam_bellmann_kestler_2019}  &ECG &raw &1D CNN  &57.04 \\ \hdashline
\cite{patil_2024} &ECG &domain-specific$^\divideontimes$ &LR &57.40\\ \hdashline
\cite{martinez_picard_2018_b}  &ECG &domain-specific$^\divideontimes$ &LR &57.69\\ \hdashline
\cite{gkikas_chatzaki_2022}  &ECG & domain-specific$^\divideontimes$ &SVM &58.39\\ \hdashline
\cite{phan_iyortsuun_2023} &ECG &raw &1D CNN, biLSTM &61.20\\ \hdashline
\cite{werner_hamadi_2014}  &ECG &domain-specific$^\divideontimes$ &RF &62.00\\ \hdashline
\cite{kachele_werner_2015} &ECG & domain-specific$^{\divideontimes}$ &SVM &62.40\\ \hdashline
\cite{kachele_werner_2015} &ECG & domain-specific$^\divideontimes$ &SVM &62.40\\ \hdashline
\cite{patil_patil_2024} &ECG & raw &2D CNN, biLSTM  &63.20\\ \hdashline
\cite{werner_hamadi_niese_2014} &ECG &domain-specific$^\divideontimes$ &RF &64.00\\ \hdashline
\cite{huang_dong_2022}  &ECG &heart rate\textsuperscript{\ding{72}} &3D CNN &65.00\\ \hdashline
\cite{gkikas_tachos_2024} &ECG &heart rate &Transformer &67.04\\ \hdashline
\cite{gkikas_chatzaki_2023}  &ECG &domain-specific$^\divideontimes$ &FCN &69.40\\ \hdashline
\rowcolor{mygray}Our &ECG &raw\textsuperscript{\ding{70}} &Transformer &\textbf{75.49}\\
\Xhline{1.5\arrayrulewidth}

\cite{patil_2024} &EMG &domain-specific$^\divideontimes$ &LR &58.59\\
\cite{thiam_bellmann_kestler_2019}  &EMG &raw &2D CNN &58.65\\ \hdashline
\cite{werner_hamadi_niese_2014} &EMG &domain-specific$^\divideontimes$ &RF &63.10\\ \hdashline
\rowcolor{mygray}Our &EMG &raw\textsuperscript{\ding{96}} &Transformer &\textbf{72.84}\\
\Xhline{1.5\arrayrulewidth}

\cite{werner_hamadi_niese_2014} &GSR &domain-specific$^\divideontimes$ &RF &71.90\\ \hdashline
\cite{martinez_picard_2018_b} &GSR &domain-specific$^\divideontimes$ &LR &74.21\\ \hdashline
\cite{kachele_werner_2015} &GSR & domain-specific$^\divideontimes$ &RF &74.40\\ \hdashline
\cite{ji_zhao_li_2023} &GSR &domain-specific$^\divideontimes$ &RF &80.40\\ \hdashline
\cite{kachele_thiam_amirian_werner_2015} &GSR &domain-specific$^\divideontimes$ &RF &81.90\\ \hdashline
\cite{patil_2024} &GSR &domain-specific$^\divideontimes$ &LR &82.36\\ \hdashline
\cite{pouromran_radhakrishnan_2021} &GSR &domain-specific$^\divideontimes$ &SVM &83.30\\ \hdashline
\cite{patil_patil_2024} &GSR & raw &1D CNN, biLSTM  &83.60\\ \hdashline
\cite{gouverneur_li_2021} &GSR &domain-specific$^\divideontimes$ &MLP &84.22\\ \hdashline
\cite{thiam_bellmann_kestler_2019}  &GSR &raw &1D CNN  &84.57\\ \hdashline
\cite{phan_iyortsuun_2023} &GSR &raw &1D CNN, biLSTM &84.80\\ \hdashline
\cite{lu_ozek_kamarthi_2023} &GSR &raw &1D CNN, Transformer &85.56\\ \hdashline
\rowcolor{mygray}Our &GSR &raw\textsuperscript{\ding{95}} &Transformer &\textbf{88.99}\\

\bottomrule 
\end{tabular}
\begin{tablenotes}
\scriptsize
\item
$\divideontimes$: numerous features \space \ding{72}: pseudo heart rate gain (estimated from videos) \space \ding{70}: PSD \space \ding{96}: waveform-PSD (Concat) \space \ding{95}: waveform \space SVM: Support Vector Machines \space LR: Logistic Regression
\end{tablenotes}
\end{threeparttable}
\end{center}
\end{table}

\renewcommand{\arraystretch}{1.1}
\begin{table}
\center
\scriptsize
\caption{Comparison of multimodal-based studies utilizing \textit{BioVid (Part-A)}, NP vs. P\textsubscript{4} task and LOSO validation.}
\label{table:multimodal_based}
\begin{center}
\begin{threeparttable}
\begin{tabular}{ P{1.0cm} P{3.0cm}  P{4.5cm} P{2.0cm} P{0.6cm} }
\toprule
\multirow{2}[2]{*}{\shortstack{Study}}
&\multirow{2}[2]{*}{\shortstack{Modality}}
&\multicolumn{2}{c}{Method} 
&\multirow{2}[2]{*}{\shortstack{Acc\%}}\\
\cmidrule(lr){3-4}
&&Features &ML &\\
\midrule
\midrule
\cite{martinez_picard_2018_b}  &ECG, GSR &domain-specific$^\divideontimes$ &SVM &72.20\\ \hdashline
\cite{werner_hamadi_2014}  &ECG,  EMG, GSR &domain-specific$^{\divideontimes}$ &RF &74.10\\ \hdashline
\cite{werner_hamadi_2014}  &Video$^1$,ECG$^2$, EMG$^2$, GSR$^2$ &facial landmarks$^1$, 3D distances$^1$, domain-specific$^{2\divideontimes}$ &RF &77.80\\ \hdashline
\cite{kachele_werner_2015} &Video$^1$, ECG$^2$,  GSR$^2$ &facial landmarks$^1$, 3D distances$^1$, domain-specific$^{2\divideontimes}$ &RF &78.90\\ \hdashline
\cite{werner_hamadi_niese_2014} &Video$^1$, ECG$^2$, EMG$^2$, GSR$^2$ &facial landmarks$^1$, head pose$^1$, domain-specific$^2$ &RF &80.60\\ \hdashline
\cite{gkikas_tachos_2024} &Video$^1$, ECG$^2$ &raw$^1$, heart rate$^2$ &Transformer &82.74\\ \hdashline
\cite{kachele_thiam_amirian_werner_2015} &Video$^1$, ECG$^2$, EMG$^2$, GSR$^2$ &geometric$^1$, appearance$^1$, domain-specific$^2$ &RF &83.10\\ \hdashline
\cite{patil_2024} &ECG, EMG, GSR &domain-specific &LR &83.20\\ \hdashline
\cite{wang_xu_2020} &ECG, EMG, GSR &domain-specific &biLSTM &83.30\\ \hdashline
\cite{thiam_kestler_schenker_2020_b} &ECG, EMG, GSR&raw &DDCAE &83.99\\ \hdashline
\cite{thiam_hihn_braun_2021} &ECG, EMG, GSR &raw &DDCAE, NN &84.25\\ \hdashline
\cite{thiam_bellmann_kestler_2019}  &ECG, EMG, GSR &raw &2D CNN &84.40\\ \hdashline
\cite{jiang_rosio_2024} &GSR, ECG &domain-specific$^{\divideontimes}$ &NN &84.58\\ \hdashline
\cite{patil_patil_2024} &Video, GSR & raw &2D CNN, biLSTM &84.80\\ \hdashline
\cite{phan_iyortsuun_2023} &ECG, GSR &raw &1D CNN, biLSTM &84.80\\ \hdashline
\cite{kachele_thiam_amirian_2016} &ECG, EMG, GSR &domain-specific &RF &85.70\\ \hdashline
\cite{bellmann_Schwenker_2020} &ECG, EMG, GSR &domain-specific &RF &85.80\\ \hdashline
\cite{zhi_yu_2019} &Video$^1$, ECG$^2$, EMG$^2$, GSR$^2$ &facial descriptors$^1$, domain-specific$^2$ &RF &86.00\\\hdashline
\cite{jiang_li_he_2024} &GSR, ECG &domain-specific$^{\divideontimes}$ &NN &87.06\\
\hdashline
\rowcolor{mygray}Our &Video\textsuperscript{\ding{66}}, GSR\textsuperscript{\ding{93}} &raw &Transformer &\textbf{89.08}\\
\bottomrule 
\end{tabular}
\begin{tablenotes}
\scriptsize
\item
\ding{66}: RGB-thermal-depth \space \ding{93}: waveform \space $\divideontimes$: numerous features \space DDCAE: deep denoising convolutional autoencoders \space NN: neural network
\end{tablenotes}
\end{threeparttable}
\end{center}
\end{table}

\renewcommand{\arraystretch}{1.1}
\begin{table}
\small
\center
\caption{Comparison of studies on the testing set of \textit{AI4Pain} dataset.}
\label{table:ai4pain_test}
\begin{center}
\begin{threeparttable}
\begin{tabular}{ P{1.2cm} P{1.5cm} P{1.5cm} P{1.5cm} P{2.0cm} P{1.5cm}}
\toprule
\multirow{2}[2]{*}{\shortstack{Study}}
&\multicolumn{3}{c}{Modality}
&\multirow{2}[2]{*}{\shortstack{ML}}
&\multirow{2}[2]{*}{\shortstack{Acc\%}}\\
\cmidrule(lr){2-4}
&Video &fNIRS &Fusion\\
\midrule
\midrule
\cite{gkikas_tsiknakis_painvit_2024} &-- &-- &\checkmark &Transformer &46.67\\
\cite{prajod_schiller_2024}          &\checkmark &-- &-- &2D CNN &49.00\\
\cite{nguyen_yang_2024}              &\checkmark &-- &-- &Transformer &55.00\\
\hdashline
\multirow{3}{*}{Our}
&\checkmark &-- &--  &\multirow{3}{*}{Transformer} &53.67\\
&-- &\checkmark &--  & &52.60\\
&-- &-- &\checkmark  & &55.69\\
        
\bottomrule 
\end{tabular}
\begin{tablenotes}
\scriptsize
\item
\end{tablenotes}
\end{threeparttable}
\end{center}
\end{table}

\subsection{Interpretation} 
\label{interpretation}
\begin{figure}
\begin{center}
\includegraphics[scale=0.30]{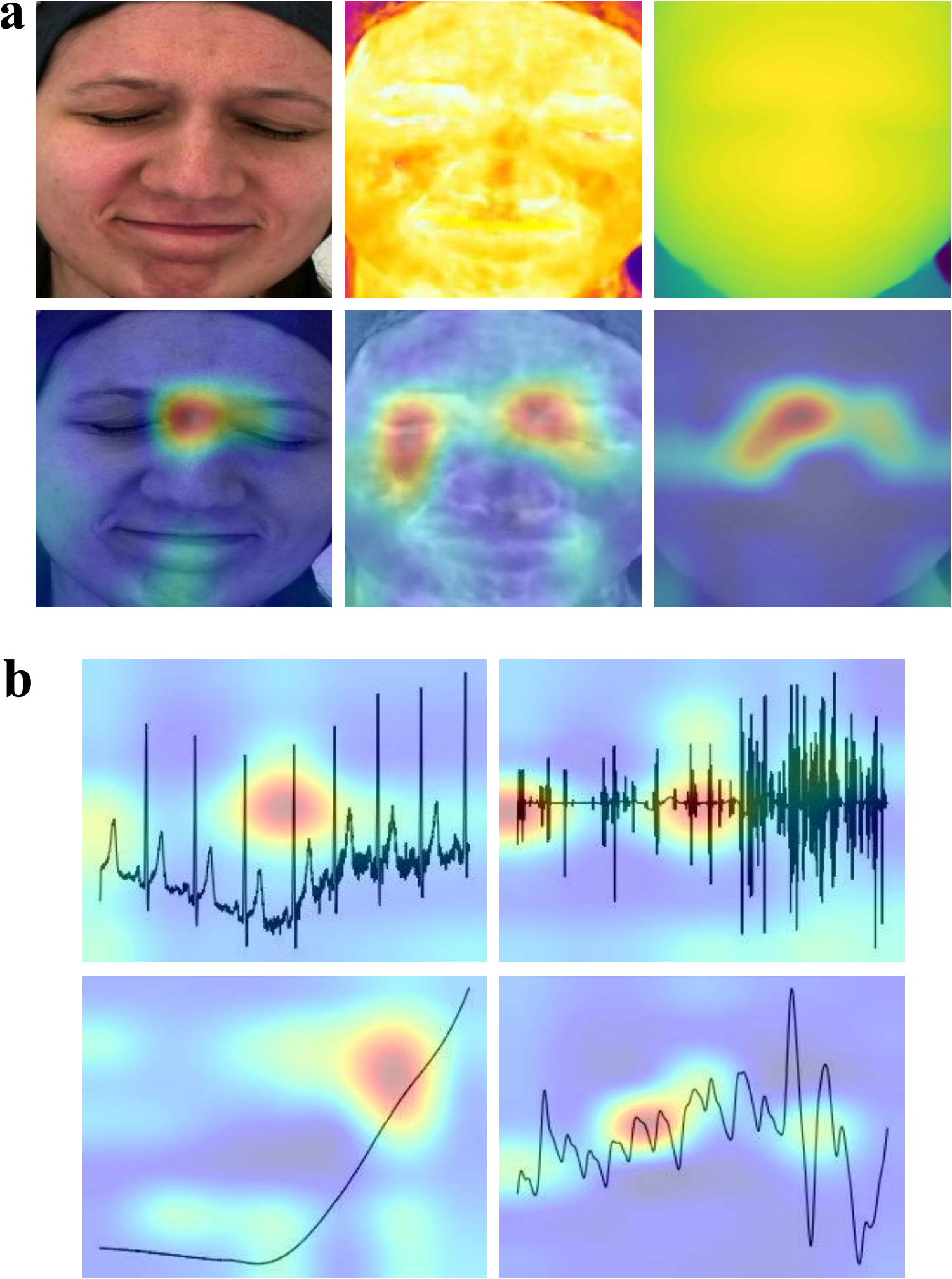} 
\end{center}
\caption{
Attention maps from the \textit{PainFormer}: \textbf{(a)(1\textsuperscript{st} row)} frames from RGB, thermal, and depth video modalities; \textbf{(a)(2\textsuperscript{nd} row)} corresponding attention maps; \textbf{(b)(1\textsuperscript{st} row)} attention maps for ECG and EMG; \textbf{(b)(2\textsuperscript{nd} row)} attention maps for EDA and fNIRS modalities.}
\label{2_attention_maps}
\end{figure}

Enhancing the interpretability of models is crucial for their acceptance and effective integration into clinical settings. In this study, \textit{PainFormer} generates attention maps, as illustrated in Fig. \ref{2_attention_maps}. The weights from the \textquotedblleft Stage 4\textquotedblright \space self-attention heads are applied by interpolating them onto the input images, enabling visualization of the model's focal areas.

In \hyperref[2_attention_maps]{Fig. 7.8(a)}, (1\textsuperscript{st} row), we display examples from the RGB, thermal, and depth modalities, and in \hyperref[2_attention_maps]{Fig. 7.8(a)}, (2\textsuperscript{nd} row), the corresponding attention maps are presented. Observations indicate that the model primarily focuses on the glabella region (the area between the eyebrows) in the RGB frame, a key area for facial expressions. Additional focus is observed on the mental protuberance area (the chin), which is also associated with expressions of pain. For the thermal frame, the model concentrates on areas around the eyes and the sides of the mouth, where brighter colors in the thermal imagery suggest higher temperatures, indicating that temperature variations rather than facial expressions drive the model's attention in the thermal modality. In the depth frame, the model targets areas showing variations in depth, particularly across the horizontal eye region, with slight attention to the frame's lower left and right edges, highlighting depth differences in body parts beyond the face, which illustrates a nuanced understanding of the model's representation of depth.

The ECG attention maps in \hyperref[2_attention_maps]{Fig. 7.8(b)}, (top left), primarily emphasizes a distinct R peak in the trace's center. Significant attention is also directed towards the T waves, especially those following the central R peak, highlighting the model's sensitivity to these elements in the signal. In the EMG attention maps of \hyperref[2_attention_maps]{Fig. 7.8(b)}, (top right), \textit{PainFormer} mainly focuses on the initial and middle sections of the signals. Despite a muscle contraction burst appearing later in the sequence, the model exhibits less attention to this portion. This observation may be related to the \textit{PainFormer}'s pre-training on the \textit{Silent-EMG} dataset \cite{gaddy_klein_2020}, which might influence its responsiveness to specific sections of the EMG signals.

For the GSR signal in \hyperref[2_attention_maps]{Fig. 7.8(b)}, (bottom left), mild attention is noted at the onset of the response, marking the start of the conductance increase, with the most intense attention near the peak amplitude, where conductance reaches its maximum level. In the fNIRS signal shown in \hyperref[2_attention_maps]{Fig. 7.8(b)} (bottom right), the attention map predominantly highlights regions aligning with peaks and rapid changes in HbO2 levels. Significant attention is concentrated in the map's left, middle, and right sections, where distinct peaks and dips in the signal are observed, indicating that \textit{PainFormer} consistently focuses on substantial fluctuations in the HbO2 signal, likely associated with pain conditions. Areas with lower or moderate attention correspond to segments of the time series with stable or minor variations in HbO2, reflecting lower levels of brain activation typically associated with mild or no pain responses.

\subsection{Discussion}
\label{discussion}
This research introduces \textit{PainFormer}, a foundation model utilizing a vision-transformer architecture tailored for pain assessment across diverse modalities. The model was pre-trained on $14$ datasets, totaling $10.9$ million samples, using a multi-task learning framework to enhance its capability in processing behavioral and physiological inputs. \textit{PainFormer} is complemented by the \textit{Embedding-Mixer} for analyzing embeddings and the \textit{Video-Encoder} for reducing the dimensionality of video embeddings. We tested our model using the \textit{BioVid} and \textit{AI4Pain} datasets, experimenting with modalities including RGB, synthetic thermal, depth videos, and biosignal representations such as ECG, EMG, GSR, and fNIRS.
The evaluation demonstrated that RGB videos provided the highest accuracy among behavioral modalities, recording $76.29\%$ accuracy in the \textit{BioVid} dataset. Thermal and depth modalities also performed well, with accuracies of $71.55\%$ and $71.67\%$, respectively. Interestingly, integrating thermal and depth modalities improved accuracy by $1.35\%$, suggesting their potential to match the efficacy of RGB while addressing privacy concerns associated with direct facial recordings.
ECG was particularly effective regarding physiological signals, with the \textit{spectrogram-PSD} achieving the highest accuracy at $75.49\%$. Although combining different ECG representations did not enhance performance, EMG signals showed exceptional accuracy above $72\%$ when combining \textit{waveform} and \textit{spectrogram-PSD}. GSR, known for its efficacy in pain assessment, achieved an impressive accuracy of $88.9\%$ using waveform representations alone. Our multimodal approach that integrated GSR with RGB, thermal, and depth video embeddings led to a notable accuracy of $89.08\%$ in the \textit{BioVid} dataset, underscoring the strength of combining multiple modalities.
Further, the creation of attention maps revealed \textit{PainFormer}'s consistent focus on key areas indicative of pain across all tested modalities. This ability highlights the model's utility in clinical settings, where understanding pain through various indicators is crucial. However, the influence of pre-training on specific areas requires further investigation to ensure the model's generalizability and accuracy in real-world applications.
Our findings place \textit{PainFormer} at the forefront of current methodologies, achieving state-of-the-art results across the board. While our model excels with video-based and biosignal inputs in unimodal and multimodal settings, continuous advancements and further explorations are needed, especially with the newer \textit{AI4Pain} dataset.

\section{Summary}
This chapter has introduced general-purpose models and pipelines for automatic pain assessment. Such approaches have attained popularity recently, driven by the development of advanced architectures and the availability of substantial data and computing resources necessary for training these models. While this combination has led to notable achievements in the broader fields of deep learning and AI, its effect on pain research has been virtually nonexistent. This distinction motivated our exploration of these methodologies in pain recognition tasks.
We presented a modality-agnostic method that homogenizes the pipeline irrespective of the input type. Our experiments with RGB videos and fNIRs showed promising results in both unimodal and multimodal settings. However, there is potential for further improvement. We believe increasing the pre-training data for models within a modality-agnostic framework could significantly improve performance.
It is generally accepted that foundation models deliver superior outcomes. Our study introduced the first foundation model specifically developed for and applied to pain assessment. The results demonstrated that this approach is highly effective, not only for well-established applications such as language understanding but also for automatic pain assessment. While further investigation is necessary, it is essential to acknowledge a fundamental challenge: the generally limited availability of pain-related data, which could restrict the effectiveness of these models.

    \chapter{Conclusions, Perspectives and Future Work}
\minitoc  

\section{Summary of Thesis Achievements}
The primary objective of this thesis was to explore and improve methods for automatic pain assessment. We developed innovative methods that either improved assessment accuracy or introduced new approaches, potentially paving the way for advanced methodologies in the future. Additionally, this thesis aimed to integrate ideas and insights from psychology, biology, and nursing, translating them into engineering concepts.

\subsection{Reporting on Deep Learning in Automatic Pain Assessment}
Chapter \ref{slr} presented a comprehensive review of deep learning-based approaches in the field. This systematic review, conducted at the beginning of this Ph.D. project, laid the groundwork for our understanding of the domain. We believe that this foundational work will continue to benefit other researchers interested in pain assessment from a machine learning and AI perspective. In addition to documenting existing approaches, this work identified emerging trends and suggested potential improvements, offering valuable insights for future research.

\subsection{Insights from Gender and Age Analysis}
In Chapter \ref{chapter_4}, we explored the impact of demographic factors on an automatic pain assessment pipeline, focusing specifically on gender and age. For decades, it has been recognized that these factors significantly affect pain expression, sensation, and perception in diverse and intriguing ways. Our study utilized ECG signals to explore variations in pain sensation among people. Our findings confirmed significant differences between males and females, with the latter group exhibiting higher sensitivity---a phenomenon well-documented in pain literature from psychological and biological perspectives.
Moreover, substantial variations were observed across different age groups. A critical discovery was that pain sensation tends to decrease with age. This observation is particularly concerning as it implies that older individuals might sustain injuries without adequate perception of pain due to neurological reasons, potentially leading to further harm. To our knowledge, such explorations and findings have not been previously addressed in automatic pain assessment. We expect our research will inspire more studies into these phenomena from computational and engineering perspectives.

\subsection{Pain Assessment with Compact, High-Performance Models}
In Chapter \ref{chapter_5}, we introduced methods aimed at developing effective and efficient approaches suitable for real-world applications, focusing on computational cost and efficiency. A primary concern addressed is the tendency in both automatic pain research and the broader field of deep learning to rely on large models that require high-end GPUs for basic inference. We investigated whether achieving comparable performance with more efficient and faster models is feasible. Additionally, we conducted one of the first studies to use heart rate as the sole feature for pain assessment. This choice was motivated by the widespread availability of heart rate data from consumer wearables, prompting us to examine its viability for pain assessment. Our findings indicate that well-designed and optimized models can deliver performance on par with, or even superior, systems that use more complex features or raw ECG signals. This is significant, as effective real-world applications must balance peak performance with manageable computational demands, particularly in clinical or home monitoring settings.

\subsection{Synthetic Data for Improved Pain Assessment and Privacy}
In Chapter \ref{chapter_6}, we explored the creation of synthetic data and its potential utility within an automatic pain assessment framework. Motivated by literature suggesting that thermal imagery can reflect skin temperature increases during arousal events like pain, we generated synthetic thermal videos. Our efforts led to the development of this new synthetic modality, and we assessed its effectiveness compared to authentic RGB videos. The results demonstrated that the synthetic videos we produced were not only of high quality but also reached the performance of RGB videos. This significant outcome may pave the way for new automatic systems that rely on synthetic data. Such systems could (\textit{i}) utilize modalities that are rare, challenging, or costly to record and (\textit{ii}) enhance privacy by concealing or partially concealing the identities of individuals involved. Furthermore, it opens the possibility for systems that integrate both authentic and synthetic modalities, leveraging the strengths of each.

\subsection{Universal Modeling for Automatic Pain Assessment}
In Chapter \ref{chapter_7}, we explored and presented general-purpose models and pipelines for automatic pain assessment tasks. Our initial focus was on conceiving a single pipeline applicable across all input modalities, simplifying the development process by eliminating the need for specialized models and modules for each modality. Our strategy employed vision-based models, which are highly effective since all inputs can be transformed into image formats---for instance, a video frame remains an image. Signals can be visualized as images, such as spectrograms or waveforms. This approach proved straightforward and effective in our experiments with videos and FNIRs, applying waveforms in unimodal and multimodal scenarios.
A major advancement in our research was the development of the \textit{PainFormer}, a foundation model specifically crafted for automatic pain assessment. This model is notable for being the first in its field and represents a significant step forward in pain assessment technology.
The most important finding from this study is the exceptional performance of the foundation model across a range of modalities, including RGB, thermal, depth videos, ECG, EMG, GSR, and fNIRs. It consistently delivered state-of-the-art results, demonstrating its versatility and effectiveness. We believe that this research will encourage further investigations into similar models, enhancing their performance in benchmarks and possibly leading to broader real-world applications by proving their effectiveness and utility.

\subsection{Explainable Deep Learning}
Although explainability was not the primary focus of the thesis, our research consistently aimed to provide insights and explanations on how the model performs and makes decisions. Using attention maps, we aimed to deliver these explanations, which were often clear and valuable. This approach is crucial, especially if such models are to be implemented in clinical settings in the future.

\section{Perspectives for Automatic Pain Assessment Methods}
Automatic pain assessment is a fascinating subject from engineering and research perspectives and a critical matter in healthcare and clinical environments. Our observations and critiques focus on the datasets, the cornerstone of any computational science research. In Section \ref{chapter_3_challenges}, we discussed the challenges in automatic pain assessment. The observations and statements remain largely unchanged a few years after this writing.

The comments provided here cannot be addressed with existing datasets; they are intended for future ones. The most superficial data to incorporate involves demographic factors such as age and gender. As explored in our research, both are crucial and including them should be considered a minimum standard for future datasets.
Another factor critical to the generalization of future models is the inclusion of individuals from various racial backgrounds. We have observed notable improvements in this area; for instance, the \textit{AI4Pain} dataset, a recent addition, features participants from diverse backgrounds. One important element not yet considered in any dataset is social interactions, which, as discussed, influence pain perception. Future datasets should incorporate this factor, designing experiments to assess responses to stimuli when individuals are alone, accompanied by a female and male person. Additionally, attractiveness also affects pain perception. Incorporating these aspects into experiments is feasible and should be relatively straightforward.
Recording sound and speech is essential in terms of modalities. This data will benefit signal processing and language analysis and be valuable for assessing pain. 

Lastly, we observe that even though significant funds are invested in equipment for recording biosignals, video capture quality in some datasets is only medium to low. Future research must utilize cameras capable of recording high-resolution videos, such as 4K. It is important to note that while mobile phones may meet these specifications on paper, the quality of their sensors and lenses is often inadequate. More professional video recording equipment is necessary. Additionally, a high frame rate is essential for capturing facial micro-expressions. Furthermore, as we have discussed, incorporating visual modalities beyond RGB, such as thermal and depth imaging, can be highly valuable. The cost of these technologies is comparable to that of biosignal recording equipment and should not be prohibitively expensive.

\section{Future Work}
Based on the work and findings presented in this thesis, we propose recommendations for future research. As emphasized earlier, any modality must incorporate and extract features that capture the temporal dimension. This is vital because pain is a dynamic condition, expressed and evolving over time, not static. This principle applies equally to many other affective-related tasks.

We have a primary recommendation regarding the computational methods used in automatic pain assessment. We have observed that researchers in this field often recycle the same methods repeatedly. This repetition does not appear to stem from efforts to enhance or refine existing approaches. We encourage future researchers to be more adventurous and explore new concepts and techniques. The literature on deep learning, which is readily accessible and constantly evolving, provides a solid foundation. Researchers should strive to adopt and adapt these ideas to their specific problems and tasks and seek to innovate and improve upon them.

A more specific technical recommendation regarding methods concerns the use of foundation models. In our research, we explored this concept and developed \textit{PainFormer}. It is important to note that the specific model is not large compared to others in the literature, especially the well-known language models; it can be considered reasonably compact. This demonstrates that small, efficient models can be developed for tasks like pain assessment and being effective. We believe greater efforts should be directed toward creating foundational, general-purpose models incorporating multimodality. This approach represents a promising and significant path for future research.
In addition, we believe that methods related to generation processes hold valuable potential. In particular, until new, high-quality pain datasets are developed, techniques like super-resolution (upscaling) to produce higher-resolution video frames and methods for generating auxiliary frames to increase FPS are promising areas for exploration.

    \bibliographystyle{unsrt}
    \cleardoublepage 
    \phantomsection
    \addcontentsline{toc}{chapter}{Bibliography}
    \bibliography{bib/library}
	\clearpage
	\newpage
	\phantomsection

\appendix

\addcontentsline{toc}{chapter}{Appendix}
\chapter*{Appendix}

\markboth{Appendix}{}

\section*{Supplementary Metrics}
\label{appendix_metrics}

\renewcommand{\arraystretch}{1.2}
\begin{table}[h]
\footnotesize
\center
\caption{Results utilizing the video modality reported on precision,  recall, and F1 score (refer to Section \ref{chapter_5_paper_2}).}
\label{table:appendix_1}
\begin{center}
\begin{threeparttable}
\begin{tabular}{ P{1.0cm} P{1.0cm} P{0.7cm} P{0.7cm} P{1.5cm} P{0.6cm} P{0.6cm}  P{0.6cm} P{1.3cm}  P{1.3cm}  P{0.7cm}}
\toprule
\multirow{2}[2]{*}{\shortstack{Epochs}}
&\multirow{2}[2]{*}{\shortstack{Metric}}
&\multicolumn{2}{c}{Pretraining stage}
&\multicolumn{2}{c}{Pipeline}
&\multicolumn{3}{c}{Augmentations} 
&\multicolumn{2}{c}{Task}\\ 
\cmidrule(lr){3-4}\cmidrule(lr){5-6}\cmidrule(lr){7-9}\cmidrule(lr){10-11}
& &\nth{1}  &\nth{2} &Full frame &Tiles  &Basic &Mask &AugmNet &NP vs P\textsubscript{4} &MC\\
\midrule
\midrule

\multirow{3}{*}{500}  
&Precision &\checkmark &- &\checkmark &- &\checkmark &- &- &72.53  &31.24 \\
&Recall &\checkmark &- &\checkmark &- &\checkmark &- &-       &74.31  &29.61 \\
&F1 &\checkmark &- &\checkmark &- &\checkmark &- &-             &71.95  &27.16\\
\hdashline

\multirow{3}{*}{500} 
&Precision &- &\checkmark &\checkmark &- &\checkmark &- &- &74.21  &33.36 \\
&Recall &- &\checkmark &\checkmark &- &\checkmark &- &-      &76.74  &33.41 \\
&F1 &- &\checkmark &\checkmark &- &\checkmark &- &-            &72.24  &28.77 \\
\Xhline{2\arrayrulewidth}

\multirow{3}{*}{500}  
&Precision &-  &\checkmark &- &\checkmark &\checkmark &- &- &68.11 &31.50\\
&Recall &-  &\checkmark &- &\checkmark &\checkmark &- &-      &72.15 &27.99\\
&F1 &-  &\checkmark &- &\checkmark &\checkmark &- &-            &65.92 &25.14\\
\hdashline

\multirow{3}{*}{500} 
&Precision &- &\checkmark &\checkmark &\checkmark &\checkmark &- &- &65.14 &27.78\\
&Recall &- &\checkmark &\checkmark &\checkmark &\checkmark &- &-      &70.36 &18.42\\
&F1 &- &\checkmark &\checkmark &\checkmark &\checkmark &- &-            &61.93 &18.86\\
\hdashline

\multirow{3}{*}{500} 
&Precision &-  &\checkmark &\checkmark &$\checkmark^c$ &\checkmark &- &- &74.88 &33.96\\
&Recall &-  &\checkmark &\checkmark &$\checkmark^c$ &\checkmark &- &-      &77.41 &34.31\\
&F1 &-  &\checkmark &\checkmark &$\checkmark^c$ &\checkmark &- &-            &73.90 &29.20\\
\Xhline{2\arrayrulewidth}

\multirow{3}{*}{500} 
&Precision &-  &\checkmark &\checkmark &$\checkmark^c$ &\checkmark &\checkmark &- &73.09 &32.17\\
&Recall &-  &\checkmark &\checkmark &$\checkmark^c$ &\checkmark &\checkmark &-      &75.72 &28.41\\
&F1 &-  &\checkmark &\checkmark &$\checkmark^c$ &\checkmark &\checkmark &-            &71.92 &26.02\\
\hdashline

\multirow{3}{*}{500}
&Precision &-  &\checkmark &\checkmark &$\checkmark^c$ &\checkmark &- &\checkmark &74.87 &33.88\\
&Recall &-  &\checkmark &\checkmark &$\checkmark^c$ &\checkmark &- &\checkmark      &77.80 &29.30\\
&F1 &-  &\checkmark &\checkmark &$\checkmark^c$ &\checkmark &- &\checkmark            &73.59 &27.74\\
\hdashline

\multirow{3}{*}{500}
&Precision &-  &\checkmark &\checkmark &$\checkmark^c$ &\checkmark &\checkmark &\checkmark &73.12 &32.79\\
&Recall &-  &\checkmark &\checkmark &$\checkmark^c$ &\checkmark &\checkmark &\checkmark      &76.18 &28.51\\
&F1 &-  &\checkmark &\checkmark &$\checkmark^c$ &\checkmark &\checkmark &\checkmark            &71.91 &26.57\\
\Xhline{2\arrayrulewidth}

\multirow{3}{*}{800} 
&Precision &-  &\checkmark &\checkmark &$\checkmark^c$ &\checkmark &\checkmark &\checkmark &\textbf{77.15} &\textbf{35.39}\\
&Recall &-  &\checkmark &\checkmark &$\checkmark^c$ &\checkmark &\checkmark &\checkmark      &7\textbf{9.35} &\textbf{35.11}\\
&F1 &-  &\checkmark &\checkmark &$\checkmark^c$ &\checkmark &\checkmark &\checkmark            &\textbf{76.33} &\textbf{31.70}\\
\bottomrule 
\end{tabular}
\end{threeparttable}
\end{center}
\end{table}

\renewcommand{\arraystretch}{1.2}
\begin{table}[h]
\footnotesize
\center
\caption{Results utilizing the heart rate modality reported on precision,  recall, and F1 score (refer to Section \ref{chapter_5_paper_2}).}
\label{table:appendix_2}
\begin{center}
\begin{threeparttable}
\begin{tabular}{ P{1.0cm} P{1.0cm} P{1.5cm} P{0.6cm} P{0.6cm} P{1.4cm}   P{1.3cm} P{0.7cm}}
\toprule
\multirow{2}[2]{*}{\shortstack{Epochs}}
&\multirow{2}[2]{*}{\shortstack{Metric}}
&\multirow{2}[2]{*}{\shortstack{HR \\Encoder}}
&\multicolumn{3}{c}{Augmentations} 
&\multicolumn{2}{c}{Task}\\ 
\cmidrule(lr){4-6}\cmidrule(lr){7-8}
&&&Basic &Mask &AugmNet &NP vs P\textsubscript{4} &MC\\
\midrule
\midrule

\multirow{3}{*}{500} 
&Precision &\checkmark &\checkmark &- &- &61.73  &27.66 \\
&Recall &\checkmark &\checkmark &- &-    &65.04  &20.91 \\
&F1 &\checkmark &\checkmark &- &-        &57.74  &19.73 \\
\hdashline

\multirow{3}{*}{500} 
&Precision &- &\checkmark &- &- &61.97  &27.71 \\
&Recall &- &\checkmark &- &-    &66.01  &22.13 \\
&F1 &- &\checkmark &- &-        &57.79  &20.61 \\

\Xhline{2\arrayrulewidth}

\multirow{3}{*}{500}  
&Precision &\checkmark &\checkmark &\checkmark &- &61.97  &27.80 \\
&Recall &\checkmark &\checkmark &\checkmark &-    &65.27  &20.98 \\
&F1 &\checkmark &\checkmark &\checkmark &-        &57.38  &20.97 \\
\hdashline

\multirow{3}{*}{500}  
&Precision &\checkmark &\checkmark &- &\checkmark &62.09  &28.00 \\
&Recall &\checkmark &\checkmark &- &\checkmark    &65.73  &21.27 \\
&F1 &\checkmark &\checkmark &- &\checkmark        &58.04  &21.61 \\
\hdashline

\multirow{3}{*}{500} 
&Precision &\checkmark &\checkmark &\checkmark &\checkmark &61.63  &27.86 \\
&Recall &\checkmark &\checkmark &\checkmark &\checkmark    &65.08  &21.24 \\
&F1 &\checkmark &\checkmark &\checkmark &\checkmark        &56.78  &21.17 \\
\Xhline{2\arrayrulewidth}

\multirow{3}{*}{800} 
&Precision &\checkmark &\checkmark &\checkmark &\checkmark &65.44  &29.73 \\
&Recall &\checkmark &\checkmark &\checkmark &\checkmark    &69.85  &27.40 \\
&F1 &\checkmark &\checkmark &\checkmark &\checkmark        &62.07  &23.71 \\
\hdashline

\multirow{3}{*}{800} 
&Precision &\checkmark &- &- &\checkmark &\textbf{67.07}  &\textbf{31.11} \\
&Recall &\checkmark &- &- &\checkmark    &\textbf{71.24}  &\textbf{29.33} \\
&F1 &\checkmark &- &- &\checkmark        &\textbf{63.97}  &\textbf{25.83} \\
\bottomrule 
\end{tabular}
\begin{tablenotes}
\scriptsize
\item  \space 
\end{tablenotes}
\end{threeparttable}
\end{center}
\end{table}

\renewcommand{\arraystretch}{1.2}
\begin{table}[h]
\footnotesize
\center
\caption{Results utilizing the video \& the heart rate modality reported on precision,  recall and F1 score (refer to Section \ref{chapter_5_paper_2}).}
\label{table:appendix_3}
\begin{center}
\begin{threeparttable}
\begin{tabular}{ P{1.0cm} P{1.0cm} P{1.3cm} P{1.5cm} P{1.0cm} P{0.6cm} P{0.6cm} P{1.4cm}  P{1.3cm} P{0.7cm}}
\toprule
\multirow{2}[2]{*}{\shortstack{Epochs}}
&\multirow{2}[2]{*}{\shortstack{Metric}}
&\multirow{2}[2]{*}{\shortstack{HR \\Encoder}}
&\multicolumn{2}{c}{Pipeline}
&\multicolumn{3}{c}{Augmentations} 
&\multicolumn{2}{c}{Task}\\ 
\cmidrule(lr){4-5}\cmidrule(lr){6-8}\cmidrule(lr){9-10}
& & &Full frame &Tiles &Basic &Mask &AugmNet &NP vs P\textsubscript{4} &MC\\
\midrule
\midrule
\multirow{3}{*}{800} 
&Precision &\checkmark &\checkmark &$\checkmark^c$ &- &- &\checkmark &82.69  &39.13 \\
&Recall &\checkmark &\checkmark &$\checkmark^c$ &- &- &\checkmark    &84.71  &37.67 \\
&F1 &\checkmark &\checkmark &$\checkmark^c$ &- &- &\checkmark        &81.44  &36.31 \\
\bottomrule 
\end{tabular}
\begin{tablenotes}
\scriptsize
\item  \space 
\end{tablenotes}
\end{threeparttable}
\end{center}
\end{table}

\renewcommand{\arraystretch}{1.2}
\begin{table}[h]
\footnotesize
\center
\caption{Results utilizing the RGB video modality, reported on recall and F1 score (refer to Section \ref{chapter_6_paper_1}).}
\label{table:appendix_4}
\begin{center}
\begin{threeparttable}
\begin{tabular}{ P{1.0cm} P{0.7cm} P{1.0cm} P{1.0cm} P{1.0cm} P{1.5cm} P{1.0cm}}
\toprule
\multirow{2}[2]{*}{\shortstack{Epochs}}
&\multicolumn{3}{c}{Augmentations} 
&\multirow{2}[2]{*}{\shortstack{Metric}}
&\multicolumn{2}{c}{Task}\\ 
\cmidrule(lr){2-4}\cmidrule(lr){6-7}
&Basic &Masking &P(Aug) & &NP vs P\textsubscript{4} &MC\\
\midrule
\midrule
\multirow{2}{*}{200}    &\multirow{2}{*}{\checkmark} 
&\multirow{2}{*}{30-50} &\multirow{2}{*}{0.9} &Recall    &71.29  &29.61\\
                                          && &  &F1      &68.53  &27.22\\                     
\hline
\multirow{2}{*}{200}    &\multirow{2}{*}{\checkmark} 
&\multirow{2}{*}{30-50} &\multirow{2}{*}{0.9} &Recall    &71.93  &24.43\\
                                              && & &F1   &69.61  &23.78\\
\hline
\multirow{2}{*}{300}    &\multirow{2}{*}{\checkmark} 
&\multirow{2}{*}{30-50} &\multirow{2}{*}{0.9} &Recall    &71.34  &30.64\\
                                          && &  &F1      &69.65  &26.12\\                      
\bottomrule 
\end{tabular}
\begin{tablenotes}
\scriptsize
\item  \space 
\end{tablenotes}
\end{threeparttable}
\end{center}
\end{table}

\renewcommand{\arraystretch}{1.2}
\begin{table}[h]
\footnotesize
\center
\caption{Results utilizing the synthetic thermal video modality, reported on recall and F1 score (refer to Section \ref{chapter_6_paper_1}).}
\label{table:appendix_5}
\begin{center}
\begin{threeparttable}
\begin{tabular}{ P{1.0cm} P{0.7cm} P{1.0cm} P{1.0cm} P{1.0cm} P{1.5cm} P{1.0cm}}
\toprule
\multirow{2}[2]{*}{\shortstack{Epochs}}
&\multicolumn{3}{c}{Augmentations} 
&\multirow{2}[2]{*}{\shortstack{Metric}}
&\multicolumn{2}{c}{Task}\\ 
\cmidrule(lr){2-4}\cmidrule(lr){6-7}
&Basic &Masking &P(Aug) & &NP vs P\textsubscript{4} &MC\\
\midrule
\midrule
\multirow{2}{*}{200}    &\multirow{2}{*}{\checkmark} 
&\multirow{2}{*}{30-50} &\multirow{2}{*}{0.9} &Recall    &72.04  &28.80\\
                                          && &  &F1      &69.16  &26.45\\                     
\hline
\multirow{2}{*}{200}    &\multirow{2}{*}{\checkmark} 
&\multirow{2}{*}{30-50} &\multirow{2}{*}{0.9} &Recall    &72.18  &30.89\\
                                              && & &F1   &69.44  &26.45\\
\hline
\multirow{2}{*}{300}    &\multirow{2}{*}{\checkmark} 
&\multirow{2}{*}{30-50} &\multirow{2}{*}{0.9} &Recall    &72.52  &24.96\\
                                              && & &F1   &70.01  &23.43\\                  
\bottomrule 
\end{tabular}
\begin{tablenotes}
\scriptsize
\item  \space 
\end{tablenotes}
\end{threeparttable}
\end{center}
\end{table}

\renewcommand{\arraystretch}{1.2}
\begin{table}[h]
\footnotesize
\center
\caption{Results utilizing the fusion of RGB \& synthetic thermal video modality, reported on recall and F1 score (refer to Section \ref{chapter_6_paper_1}).}
\label{table:appendix_6}
\begin{center}
\begin{threeparttable}
\begin{tabular}{ P{1.0cm} P{1.0cm} P{1.0cm} P{1.0cm}  P{1.0cm}  P{1.0cm} P{1.5cm} P{1.0cm}}
\toprule
\multirow{2}[2]{*}{\shortstack{Epochs}}
&\multirow{2}[2]{*}{\shortstack{Fusion\\weights}}
&\multicolumn{3}{c}{Augmentations} 
&\multirow{2}[2]{*}{\shortstack{Metric}}
&\multicolumn{2}{c}{Task}\\ 
\cmidrule(lr){3-5}\cmidrule(lr){7-8}
& &Basic &Masking &P(Aug) & &NP vs P\textsubscript{4} &MC\\
\midrule
\midrule
\multirow{2}{*}{100}    &\multirow{2}{*}{--} &\multirow{2}{*}{\checkmark} 
&\multirow{2}{*}{30-50} &\multirow{2}{*}{0.9} &Recall    &67.05  &21.68\\
                                          &&& &  &F1     &62.96  &18.29\\  
\hline                               
\multirow{2}{*}{100}    &\multirow{2}{*}{W2} &\multirow{2}{*}{\checkmark} 
&\multirow{2}{*}{30-50} &\multirow{2}{*}{0.9} &Recall    &68.72  &21.69\\
                                          &&& &  &F1     &62.98  &19.35\\ 
\hline         
\multirow{2}{*}{100}    &\multirow{2}{*}{W3} &\multirow{2}{*}{\checkmark} 
&\multirow{2}{*}{30-50} &\multirow{2}{*}{0.9} &Recall    &66.12  &23.12\\
                                          &&& &  &F1     &59.72  &19.67\\                     
\hline
\multirow{2}{*}{300}    &\multirow{2}{*}{W2} &\multirow{2}{*}{\checkmark} 
&\multirow{2}{*}{30-50} &\multirow{2}{*}{0.9} &Recall    &71.40  &26.39\\
                                              &&& & &F1  &68.82  &26.18\\
\hline
\multirow{2}{*}{500}    &\multirow{2}{*}{W2} &\multirow{2}{*}{\checkmark} 
&\multirow{2}{*}{10-20} &\multirow{2}{*}{0.7} &Recall    &73.20  &29.69\\
                                            & & & & &F1  &70.30  &27.84\\                    
\bottomrule 
\end{tabular}
\begin{tablenotes}
\scriptsize
\item  \space 
\end{tablenotes}
\end{threeparttable}
\end{center}
\end{table}

\renewcommand{\arraystretch}{1.2}
\begin{table}[h]
\footnotesize
\center
\caption{Results of the proposed approaches, reported on macro-averaged precision, recall and F1 score  (refer to Section \ref{chapter_7_paper_1}).}
\label{table:comparison}
\begin{center}
\begin{threeparttable}
\begin{tabular}{ P{1.0cm} P{2.5cm} P{1.0cm} P{1.0cm} P{1.0cm}}
\toprule
\multirow{2}[2]{*}{\shortstack{Modality}}
&\multirow{2}[2]{*}{\shortstack{Approach}}
&\multicolumn{3}{c}{Metrics}\\ 
\cmidrule(lr){3-5}
& &Precision &Recall &F1\\
\midrule
\midrule
Video  &\textit{Addition} &44.91 &44.97 &44.60\\
fNIRS  &HbO \& \textit{Addition} &44.68 &45.08 &43.60\\
Fusion &\textit{Single Diagram} &46.76 &47.29 &46.70\\
        
\bottomrule 
\end{tabular}
\begin{tablenotes}
\scriptsize
\item 
\end{tablenotes}
\end{threeparttable}
\end{center}
\end{table}

\clearpage

\section*{Supplementary Figures}
\label{appendix_figures}


\begin{figure}[h]
\begin{center}
\includegraphics[scale=0.20]{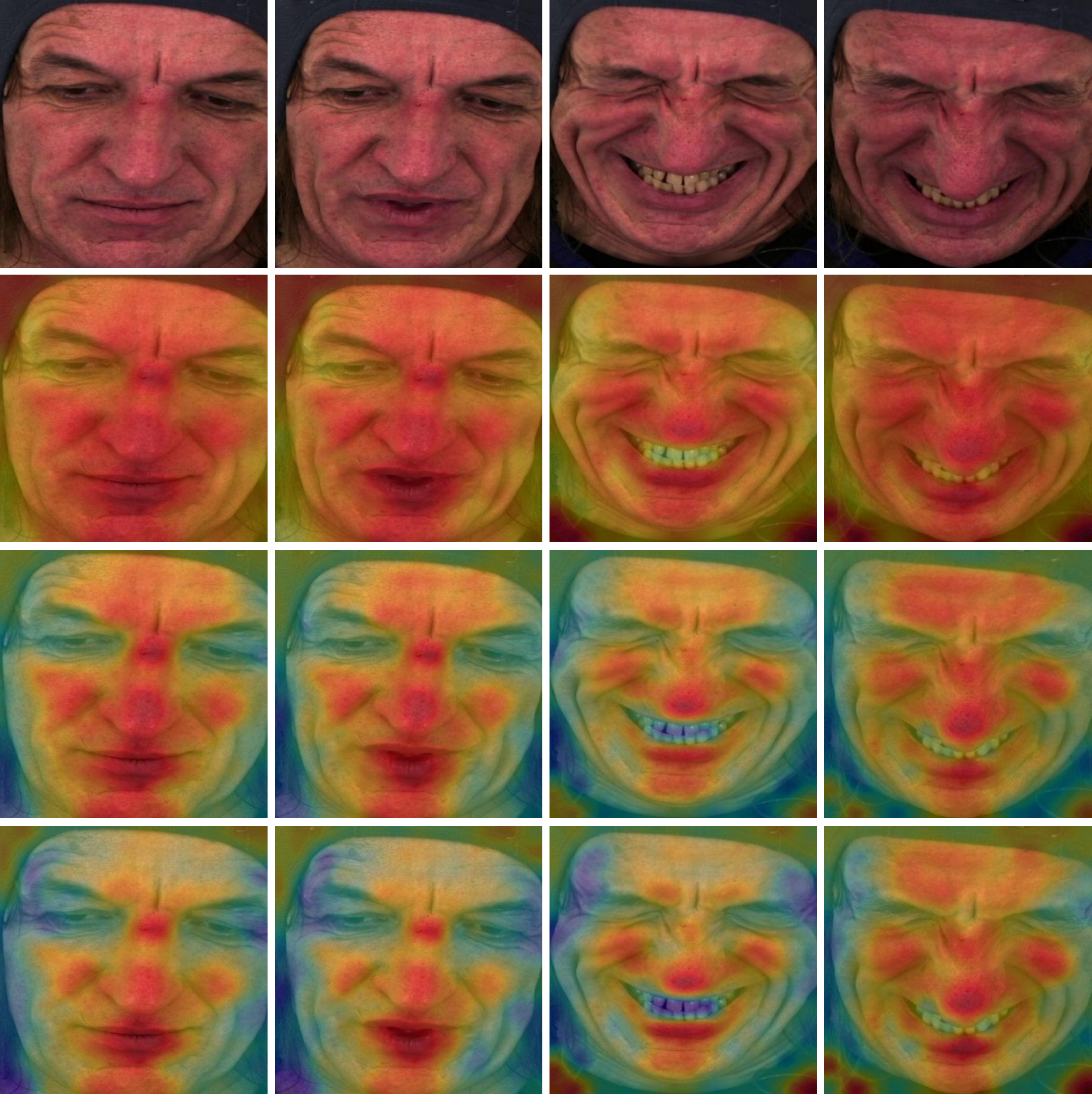}
\caption{Attention maps generated by the \textit{Spatial-Module}. \textit{Yellow and red colors signify intense focus on specific areas.} \textbf{(1\textsuperscript{st} row)} Sequence of original frames. \textbf{(2\textsuperscript{nd} row)} Derived from the \textit{Spatial-Module} after initial stage pretraining. \textbf{(3\textsuperscript{rd} row)} Derived from the \textit{Spatial-Module} post second stage pretraining. \textbf{(4\textsuperscript{th} row)} Derived from the \textit{Spatial-Module} following training on \textit{BioVid} (refer to Section \ref{chapter_5_paper_2}).}
\label{appendic_im_1}
\end{center}
\end{figure}

\begin{figure}[h]
\begin{center}
\includegraphics[scale=0.20]{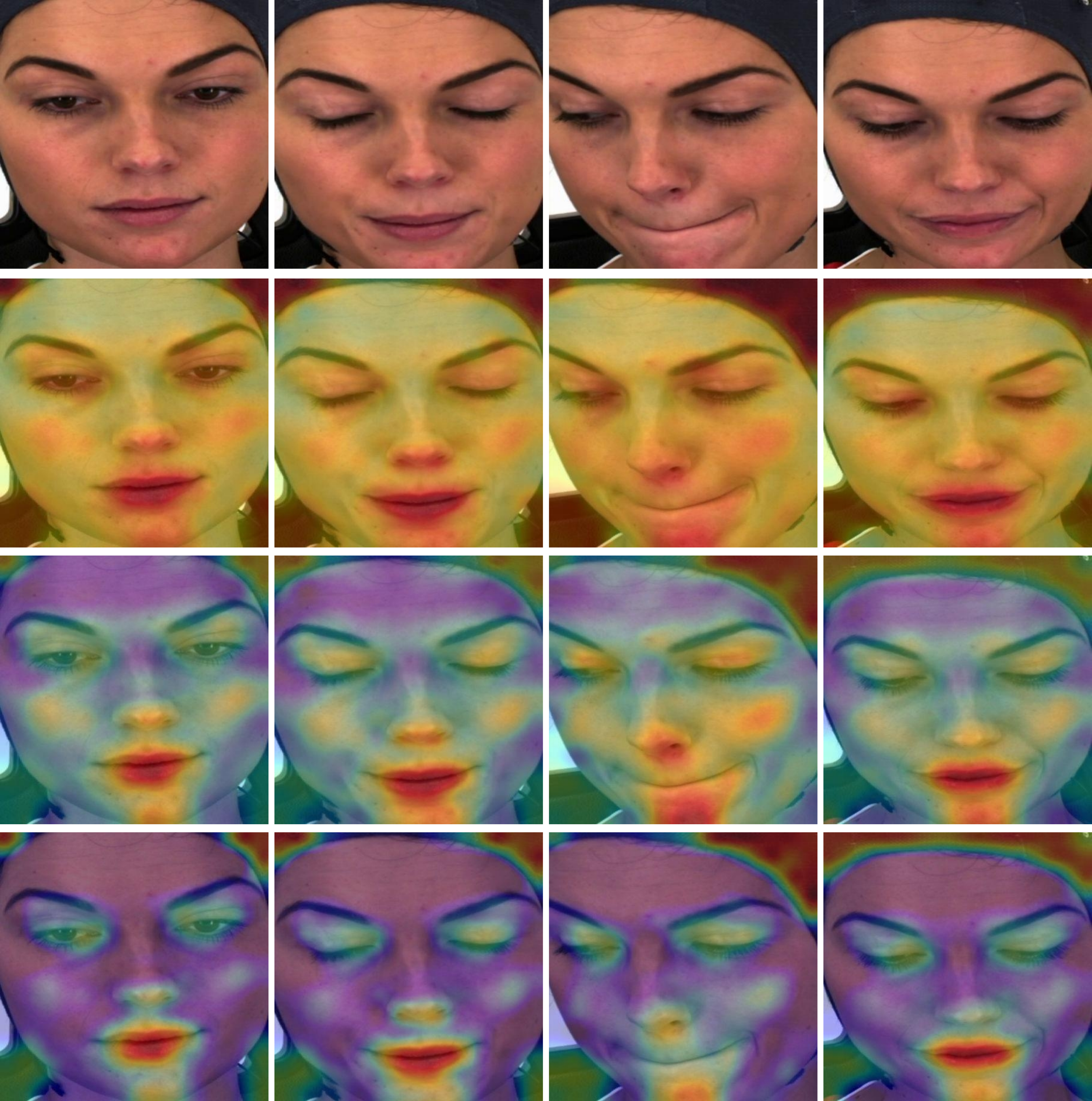}
\caption{Attention maps generated by the \textit{Spatial-Module}. \textit{Yellow and red colors signify intense focus on specific areas.} \textbf{(1\textsuperscript{st} row)} Sequence of original frames. \textbf{(2\textsuperscript{nd} row)} Derived from the \textit{Spatial-Module} after initial stage pretraining. \textbf{(3\textsuperscript{rd} row)} Derived from the \textit{Spatial-Module} post second stage pretraining. \textbf{(4\textsuperscript{th} row)} Derived from the \textit{Spatial-Module} following training on \textit{BioVid} (refer to Section \ref{chapter_5_paper_2}).}
\label{appendix_im_2}
\end{center}
\end{figure}

\begin{figure}[h]
\begin{center}
\includegraphics[scale=0.30]{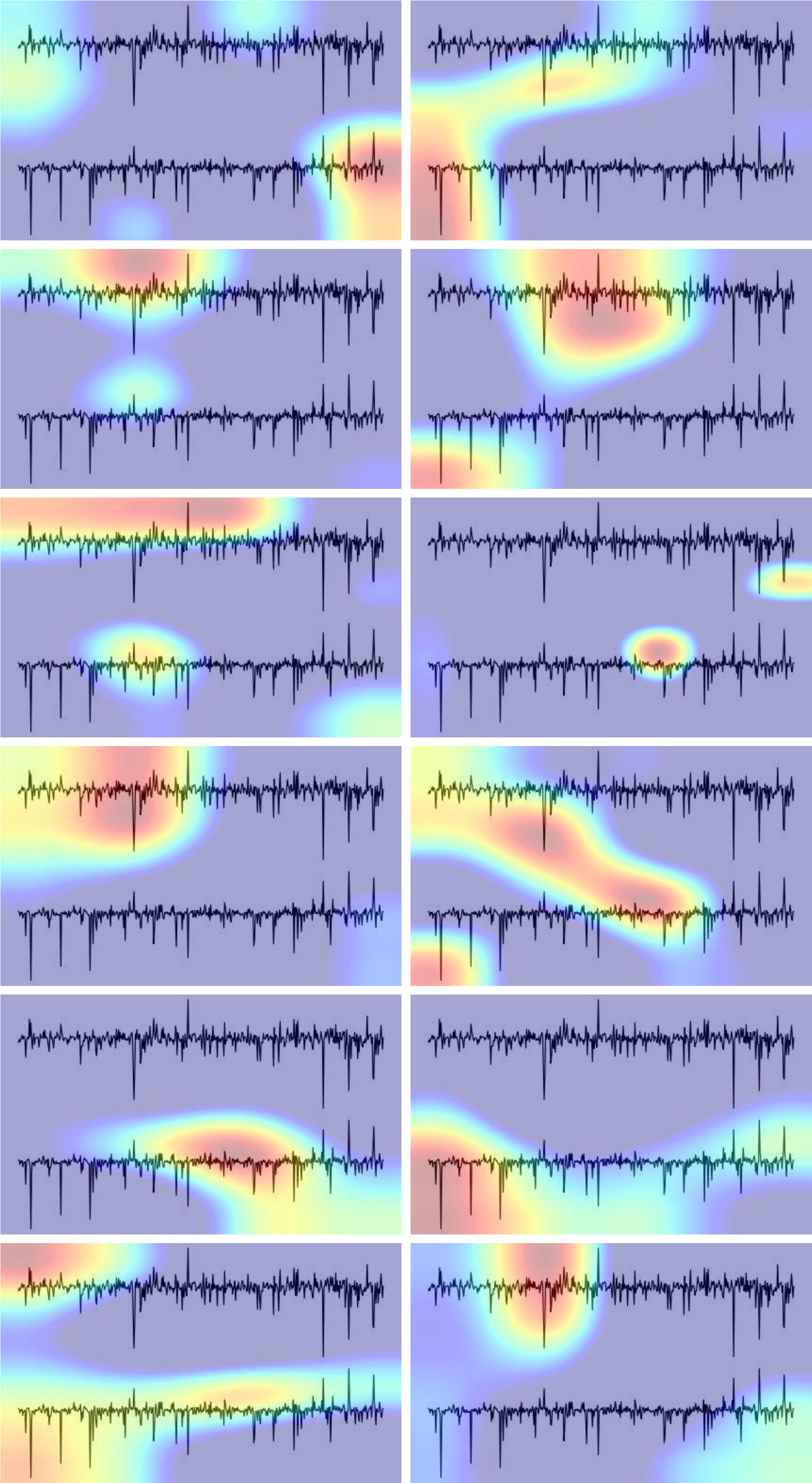}
\end{center}
\caption{Additional attention maps from the \textit{PainViT--2} (refer to Section \ref{chapter_7_paper_1}).}
\label{appendix_im_3}
\end{figure}

\phantomsection
\chapter*{Acronyms}
\label{appendix_acr}

\addcontentsline{toc}{chapter}{Acronyms}
\markboth{Appendix}{Acronyms} 





\begin{table}[h] 
\centering 
\begin{tabular}{ll} 
\textbf{AI}    & \textbf{A}rtificial \textbf{I}ntelligence \\

\textbf{ECG}   & \textbf{E}le\textbf{c}trocardio\textbf{g}raphy \\
\textbf{EDA}   & \textbf{E}lectro\textbf{d}ermal \textbf{A}ctivity\\
\textbf{EEG}   & \textbf{E}lectro\textbf{e}ncephalo\textbf{g}raphy\\
\textbf{EMG}   & \textbf{E}lectro\textbf{m}yo\textbf{g}raphy \\

\textbf{FACS}  & \textbf{F}acial \textbf{A}ction \textbf{C}oding \textbf{S}ystem \\
\textbf{FLOPS} & \textbf{Fl}oating-point \textbf{O}perations \textbf{p}er \textbf{S}econd\\
\textbf{fMRI}  & \textbf{F}unctional \textbf{M}agnetic \textbf{R}esonance \textbf{I}maging \\
\textbf{fNIRS} & \textbf{F}unctional \textbf{N}ear-\textbf{I}nfrared \textbf{S}pectroscopy\\

\textbf{GSR}   & \textbf{G}alvanic \textbf{S}kin \textbf{R}esponse \\

\textbf{ML}    & \textbf{M}achine \textbf{L}earning \\

\textbf{NIPS}  & \textbf{N}eonatal/\textbf{I}nfant \textbf{P}ain \textbf{S}cale\\
\textbf{NIRS}  & \textbf{N}ear-\textbf{I}nfrared \textbf{S}pectroscopy\\

\textbf{PPG}   & \textbf{P}hoto\textbf{p}lethysmo\textbf{g}raphy\\ 
\textbf{PSPI}  & \textbf{P}rkachin and \textbf{S}olomon \textbf{P}ain \textbf{I}ntensity \textbf{S}cale\\

\textbf{RGB}   & \textbf{R}ed \textbf{G}reen \textbf{B}lue \\
\textbf{SpO2}  & \textbf{S}aturation of \textbf{P}eripheral \textbf{O}xygen\\

\textbf{VAS}   & \textbf{V}isual \textbf{A}nalog \textbf{S}cale\\
\textbf{VRS}   & \textbf{V}erbal \textbf{R}ating \textbf{S}cale\\

\end{tabular}
\end{table}

    \clearpage
    \newpage
    \pagestyle{plain} 
    \thispagestyle{empty} 
    \mbox{}


\end{document}